\documentclass[sn-mathphys]{sn-jnl}% Math and Physical Sciences Reference Style
\usepackage{benstyle}
\usepackage{bm}
\usepackage{color}
\usepackage{commath}
\usepackage{lineno}
\usepackage{sistyle}
\SIthousandsep{,}
\usepackage{soul}
\jyear{2021}%

\begin{document} 

%\linenumbers

\title[Christoffel Adaptive Sampling for DL]{CAS4DL: Christoffel Adaptive Sampling for function approximation via Deep Learning}


\author[1]{\fnm{Ben} \sur{Adcock}}\email{ben\_adcock@sfu.ca}

\author*[1]{\fnm{Juan M.} \sur{Cardenas}}\email{jcardena@sfu.ca 
%\FORG{I think this should be ``jcardena@sfu.ca", please check}
}

\author[1]{\fnm{Nick} \sur{Dexter}}\email{ndexter@sfu.ca}

\affil[1]{\orgdiv{Department of Mathematics}, \orgname{Simon Fraser University}, \orgaddress{\street{8888 University Dr}, \city{Burnaby}, \postcode{V5A 1S6}, \state{BC}, \country{Canada}}}


\abstract{The problem of approximating smooth, multivariate functions from sample points arises in many applications in scientific computing, e.g., in computational Uncertainty Quantification (UQ) for science and engineering.
In these applications, the target function may represent a desired quantity of interest of a parameterized Partial Differential Equation (PDE).
Due to the large cost of solving such problems, where each sample is computed by solving a PDE, sample efficiency is a key concern in these applications.
Recently, there has been increasing focus on the use of Deep Neural Networks (DNN) and Deep Learning (DL) for learning such functions from data. In this work, we propose an adaptive sampling strategy, CAS4DL (Christoffel Adaptive Sampling for Deep Learning) to increase the sample efficiency of DL for multivariate function approximation. Our novel approach is based on interpreting the second to last layer of a DNN as a dictionary of functions defined by the nodes on that layer.
With this viewpoint, we then define an adaptive sampling strategy motivated by adaptive sampling schemes recently proposed for linear approximation schemes, wherein samples are drawn randomly with respect to the {\em Christoffel} function of the subspace spanned by this dictionary.
We present numerical experiments comparing CAS4DL with standard Monte Carlo (MC) sampling. Our results demonstrate that CAS4DL often yields substantial savings in the number of samples required to achieve a given accuracy, particularly in the case of smooth activation functions, and it shows a better stability in comparison to MC. These results therefore are a promising step towards fully adapting DL towards scientific computing applications.
}

%%================================%%
%% Sample for structured abstract %%
%%================================%%




\maketitle

\section{Introduction}
\label{sec:introduction} 

In this work, we consider the problem of approximating a smooth, multivariate function $f:\Omega\rightarrow\mathbb{R}$ over a domain $\Omega\subseteq\mathbb{R}^d$ in $d\geq 1$ dimensions from samples using Deep Neural Networks (DNNs). This problem arises in many applications, e.g., in Uncertainty Quantification (UQ) for computational science and engineering, see \cite{Hampton2017,Ghanem2017,Sullivan2015,Hiptmair2018,Tang2013,Tang2014,LeMaitre2010,Smith2013}. The main challenge in such applications is the so-called {\em curse of dimensionality}, which implies computational effort and number of samples required to approximate such functions grows exponentially with the dimension $d$ for many standard approximation schemes. As such problems are often inherently high-dimensional and obtaining samples is expensive, due to the complexity of the underlying model and requirement to solve a Partial Differential Equation (PDE) to obtain new samples, the efficiency of sampling is therefore a key concern. 

In the last decade there has been a great interest in applying Deep Learning (DL) techniques, i.e., machine learning with DNNs, to such problems. This surge of research activity has been driven largely by the impressive results DNNs have achieved on complicated tasks such as speech recognition, image classification, game intelligence, and medical imaging. DL techniques have also recently been applied to the problem of approximating solutions of PDEs from sample data \cite{EWeinan2017DLNM,becker2022learning}. Strategies incorporating physics and domain information into the training procedure, i.e., the so-called Physics-Informed Neural Networks (PINN), have received much research attention \cite{raissi2019physics}. Similarly, DL techniques are now being increasingly applied to various \textit{parametric} PDEs \cite{CyrEtAl2020,dalsanto2020data,geist2021numerical,khoo2021solving,laakmann2020efficient,kutyniok2021theoretical,adcock2021deep,wang2021learning,DehoopEtAL2022,grady2022towards,HerrmannEtAl2020,HerrmannEtAl2022,becker2022learning,CicciEtAl2022,DungEtAl2021,Dung2021,heiss2021neural,khara2021neufenet,kropfl2022operator,lei2022solving,li2020fourier}.

Recent theoretical results suggest DNNs offer many benefits over conventional approximation schemes such as polynomials, due to their ability to approximate a wide class of functions \cite{Yarotsky2017,Bach2017,Petersen2018,Beck2019,Liang2017,Lu2020,Yarotsky2018,Montanelli2017,Unser2019,Schwab2017,Daws2019b,Opschoor2019,Daws2019a}.
Such results establish DNNs achieve best-in-class performance due to their ability to emulate many conventional approximation schemes, and are even capable of mitigating the curse of dimensionality.
However, the problem of {\em sample complexity}, i.e., how many samples are required to train a DNN to achieve a given accuracy, has received less attention from the research community. Furthermore, the question of where to sample when training DNNs has remained largely unaddressed. The former problem has been studied, from both a numerical and theoretical perspective, in the recent work \cite{AdcockDexter2021,adcock2021deep} in the case of Monte Carlo (MC) sampling and fully-connected feedforward $\mathrm{ReLU}$ DNNs (networks where each activation function is the $\mathrm{ReLU}$ function). Theoretically, it has been shown that there exist learning strategies that train DNNs achieving the same sample complexity and error bounds as current best-in-class, polynomial-based schemes for smooth function approximation. However, this work also presented numerous numerical experiments, where such fully-connected $\mathrm{ReLU}$ DNNs were trained using MC samples from smooth functions, demonstrating that current methods of training DNNs with scattered MC sample data rarely achieve theoretical approximation rates, even for problems of moderate dimension.

\subsection{Problem statement and contributions}

The broad objective of this work is to develop new adaptive sampling strategies for function approximation using DNNs.
Throughout this work, we consider a domain $\Omega\subset \mathbb{R}^d$, $d\geq 1$, equipped with a probability measure $\mathrm{d}\tau$, and a smooth unknown target function $f:\Omega\rightarrow\mathbb{R}$. The general setting for our work is the problem of learning a sequence of DNNs $\widehat{\Psi}_1,\widehat{\Psi}_2,\ldots$ approximating $f$ from training data acquired at a \textit{hierarchical} sequences of sample points
\bes{
\{ \bm{y}_i \}^{m_1}_{i=1} \subseteq \{ \bm{y}_i \}^{m_2}_{i=1} \subseteq \cdots .
}
Our focus is on the following question: at step $l$, given the trained DNN $\widehat{\Psi}_l$, how do we choose the subsequent set of sample points $\{ \bm{y}_i \}^{m_{l+1}}_{i = m_l+1}$ to obtain as good an approximation $\widehat{\Psi}_{l+1}$ to $f$ as possible at the next step?

This problem is motivated by the aforementioned tasks in scientific computing and engineering, where generating the data is generally time-consuming. Because of this bottleneck, adaptive sampling techniques, in which the next set of sample points is determined based on the current approximation, can be exploited to enhance the quality of the resulting approximation. Some recent works have considered adaptive sampling schemes for DNNs based on importance sampling and {\em a posteriori} error maps in applications such as solving PDEs and image segmentation \cite{tang2021das,berger2017adaptive}.

Our method is different, and successfully combines two recent, but disparate ideas. The first is the notion of optimal sampling for linear least-squares approximation in a fixed subspace $P \subset L^2(\Omega,\tau)$, typically a polynomial space. This line of work was initiated in \cite{hampton2015coherence,MiglioratiCohenOptimal}. Here it was shown that random sampling according to the \textit{Christoffel} function of the subspace yields a least-squares approximation with provably near-optimal sample complexity: namely, log-linear in the subspace dimension. Unfortunately, constructing this sampling measure requires an orthonormal basis for $P$ and even if one is available, sampling from this measure can be highly nontrivial. In \cite{AdcockCardenas2020,migliorati2021multivariate}, this issue was overcome by first constructing a finite grid over $\Omega$, and then using QR decomposition to obtain an orthonormal basis over the grid. The resulting measure is consequently a discrete measure over the grid, from which it is straightforward to draw samples.

The second key idea stems from the work of \cite{CyrEtAl2020}. A set $\cN$ of DNNs with a fixed architecture does not form a linear subspace. Hence the aforementioned ideas do not apply. However, \cite{CyrEtAl2020} developed the \textit{adaptive basis} viewpoint for DNNs. In short, this work interprets a DNN $\Psi : \bbR^d \rightarrow \bbR$ as a linear combination of the outputs of its penultimate layer -- in other words, an element of the subspace spanned by the functions corresponding to the nodes in this layer.

This idea was used in \cite{CyrEtAl2020} to improve DNN training with a hybrid Least-Squares/Gradient-Descent (LSGD) training procedure. 
The LSGD procedure alternates between gradient descent updates of the weights and biases corresponding to nodes up to the penultimate layer and least-squares solves for the coefficients of the DNN approximation of the target function on the last layer.
In this work, we use the adaptive basis viewpoint in a different way: namely, we combine it with the previously-discussed idea of Christoffel sampling to achieve adaptive sampling for DNNs. Our main contribution is the method termed \textit{Christoffel Adaptive Sampling for Deep Learning (CAS4DL)}.

We present a series of numerical experiments comparing our method with standard MC sampling. Our results show that CAS4DL often presents a considerable saving in the number of samples required to achieve a certain accuracy compared to MC sampling, and it demonstrates better stability compared to MC sampling. Further, we also exploit the adaptive basis viewpoint to provide additional interpretability of the DNN approximations. In particular, we compute the numerical dimension of the learned basis to examine its efficacy. We also compute accuracy, stability and best approximation constants to assess the quality of the learned approximations.

\subsection{Outline} 
The outline of the remainder of this paper is as follows.
In Section \ref{sec:CAS} we describe the motivation and main idea behind adaptive sampling based on the Christoffel function of a basis or dictionary.
Next, in Section \ref{sec:CAS4DL}, we present the derivation of the CAS4DL method for adaptive sampling with DNNs. 
In Section \ref{sec:experiments} we show a series of numerical experiments comparing our method to standard MC sampling. 
In Section \ref{sec:accuracy_stability} we analyze the accuracy and stability of the method. 
Finally, we conclude in Section \ref{sec:conclusion}.

\section{Christoffel Adaptive Sampling}
\label{sec:CAS}

In this section, we first describe Christoffel Adaptive Sampling (CAS), which is a procedure for sampling for weighted least-squares approximation in linear subspaces.

\subsection{Optimal sampling in linear subspaces}\label{ss:opt-samp-LS}

Consider a fixed, $n$-dimensional subspace $P \subset L^2(\Omega,\tau)$ and an unknown function $f \in L^2(\Omega,\tau)$. The works \cite{MiglioratiCohenOptimal,AdcockCardenas2020,migliorati2021multivariate,Migliorati2019} consider the weighted least-squares approximation
\bes{
\tilde{f} \in \argmin{p \in P} \sum^{m}_{i=1} w(\bm{y}_i) \abs{ f(\bm{y}_i) - p(\bm{y}_i) }^2,
}
where $w : \Omega \rightarrow [0,\infty)$ is a weight function. The objective in these works is to choose the sample points $\bm{y}_1,\ldots,\bm{y}_m$ to achieve high \textit{sample efficiency}. Namely, $\tilde{f}$ should be a near-best approximation to $f$ from $P$ and the total number of samples $m$ should be minimized. The proposed approach for doing this involves selecting each sample point $\bm{y}_i$ randomly and independently according to a probability measure $\mu_i$. Choosing the sample points randomly and independently allows one to use techniques from random matrix theory to analyze the resulting weighted least-squares approximation. By doing this, one shows that a log-linear sample complexity bound $m \gtrsim n \cdot \log(n)$ suffices for near-best approximation, provided the measures $\mu_i$ satisfy the following relation:
\be{
\label{opt-samp-cond}
\frac{1}{m} \sum^{m}_{i=1} \D \mu_i(\bm{y}) = \cK(P,H)(\bm{y}) \D \tau(\bm{y}).
}
Here $H = L^2(\Omega,\tau)$ and $\cK(P,H)(\bm{y})$ is the normalized, reciprocal Christoffel function of the subspace $P \subset H$. If $\{ \phi_1,\ldots,\phi_n\}$ is any orthonormal basis for $P$, then this function is given explicitly by 
\be{
\label{Christoffel-fn}
\cK(P,H)(\bm{y}) = \frac1n \sum^{n}_{i=1} \abs{\phi_i(\bm{y}) }^2.
}
Note that in this case the weight function $w$ should be chosen as
\be{
\label{w-def}
w(\bm{y}) = \frac{1}{\cK(P,H)(\bm{y})}.
}
There are various ways one may choose the measures $\mu_i$ to achieve \R{opt-samp-cond}. The approach we employ in this paper is particularly relevant for adaptive schemes, and is based on \cite{Migliorati2019}. Here, we assume that $m = k \cdot n$ for some $k \in \bbN$. Typically, in view of the sample complexity estimate one takes $k = \lceil c \cdot \log(n) \rceil$ for some constant $c$. Then, for every basis function $\phi_j$, we choose $k$ measures as follows:
\be{
\label{induced-distributions}
\D \mu_i(\bm{y}) = \abs{ \phi_j(\bm{y}) }^2 \D \tau (\bm{y}),\quad (j-1) k < i \leq j k,\ j = 1,\ldots,n.
}
It is straightforward to see that this construction satisfies \R{opt-samp-cond}.
Note that these probability measures are sometimes known as the \textit{induced distributions} \cite{gautschi1993set,narayan2018computation}, since they are induced by the orthonormal basis functions $\phi_j$.

\subsection{Finite grids}

Unfortunately, the above construction requires an orthonormal basis $\{ \phi_1,\ldots,\phi_n \}$ for the subspace $P$. This problem was first addressed in \cite{AdcockCardenas2020,migliorati2021multivariate} in the context of polynomial approximation over irregular domains. Here it is straightforward to construct a basis for the polynomial space $P$. But it is generally impossible to explicitly construct orthonormal bases, except for particularly simple domains.

The idea developed in these works is to replace the domain $\Omega$ with a finite, but fine grid $Z = \{ \bm{z}_k \}^{K}_{k=1} \subset \Omega$ on which to perform computations. Mathematically, this involves simply replacing the measure $\tau$ (which is normally assumed to be a continuous measure over $\Omega$) with a discrete measure supported over $Z$. Throughout this work, we employ a MC grid as our fine grid $Z$. That is, we generate $K$ points randomly and independently from $\Omega$, and then define 
\be{
\label{tau-def}
\D\tau(\bm{y}) = \frac1K \sum^{K}_{k=1} \delta(\bm{y} - \bm{z}_k ) \D \bm{y}.
}
One important question is how large to choose $K$ so as to ensure a good approximation off the grid. This has been analyzed theoretically in \cite{AdcockCardenas2020,dolbeault2020optimal}. In this work, we do not consider this question. However, in our numerical experiments we use a distinct, high-order test set to measure the approximation error and therefore assess the performance of the approximation outside of $Z$.

\subsection{Orthogonalization and redundant dictionaries}

Having introduced a finite grid, it is then possible to construct an orthonormal basis over the grid using numerical linear algebra. Previously, this was done via QR factorizations. This procedure is particularly effective when one starts from a desired subspace $P$ and a nonorthogonal, but linearly independent set of functions that span $P$. However, in the context of DNNs we will see later that one does not typically know $P$ explicitly, but instead simply has access to a linearly dependent set of functions that span it, i.e., a redundant dictionary. For this reason, to construct an orthonormal basis (and, in particular, determine the effective dimension of $P$) we use truncated Singular Value Decompositions (SVDs) instead, since these generally have somewhat better stability properties in the face of redundant or near-redundant dictionaries \cite{adcock2019frames,adcock2020frames}.

Specifically, let $\{ \psi_1,\ldots,\psi_N \} \subset L^2(\Omega,\tau)$ be a dictionary and consider the matrix
\bes{
\bm{B}=\left ( \sqrt{1/K} \psi_j(\bm{z}_i) \right )_{i,j=1}^{K,N} \in\mathbb{R}^{K\times N}.
} 
Let $\bm{B}$ have SVD $\bm{U}\bm{\Sigma} \bm{V}^*$ and write $\sigma_1,\ldots,\sigma_N$ for its singular values (note that we assume $K \geq N$ throughout) and $\bm{u}_1,\ldots,\bm{u}_K$ and $\bm{v}_1,\ldots,\bm{v}_N$ for its left and right singular vectors. Then we threshold all singular values below a tolerance $\epsilon_{\mathrm{tol}}$ and set
\bes{
n = \max \{ i : \sigma_i / \sigma_1 > \epsilon_{\mathrm{tol}}  \} \in \{1,\ldots,N\}.
}
This number is the \textit{numerical dimension} of the subspace spanned by the dictionary. In practice we take the tolerance to be either $\epsilon_{\mathrm{tol}} = 10^{-6}$ (single precision) or $\epsilon_{\mathrm{tol}} = 10^{-12}$ (double precision).  Having done this, we then construct an orthonormal basis for the numerical span of this dictionary using the left singular vectors $\bm{u}_1,\ldots,\bm{u}_n$. Specifically, we let $P = \spn \{ \phi_1,\ldots,\phi_n\}$, where
\bes{
\phi_i(\bm{y}) = \frac{1}{\sigma_i} \sum^{N}_{j=1} (\bm{v}_i)_j \psi_j(\bm{y}),\quad i = 1,\ldots,n.
}

\subsection{CAS}

Having constructed an orthonormal basis as above, we may now define the sampling measures. These are given by \R{induced-distributions}, keeping in mind that $\tau$ is given by \R{tau-def}. If $(j-1) k < i \leq j k$, then we get
\bes{
\D \mu_i(\bm{y}) = \abs{\phi_j(\bm{y}) }^2 \D \tau(\bm{y}) = \frac1K \sum^{K}_{l=1} \delta(\bm{y} -\bm{z}_l) \abs{\phi_j(\bm{y}) }^2 \D \bm{y}.
}
Now we observe that
\bes{
\phi_j(\bm{z}_l) = \frac{1}{\sigma_j} \sum^{N}_{i=1}  (\bm{v}_j)_i \psi_i(\bm{z}_l) = \frac{1}{\sigma_j} \left ( \bm{B} \bm{v}_j \right )_l = (\bm{u}_j)_l,\quad l = 1,\ldots,K.
}
Therefore, writing $(\bm{u}_j)_l = u_{lj}$ for convenience, we obtain
\be{
\label{mu_i-disc-SVD}
\D \mu_i(\bm{y}) = \sum^{K}_{l=1} \abs{u_{lj}}^2 \delta(\bm{y} -\bm{z}_l) \D \bm{y}.
}
In particular, $\mu_i$ is a discrete measure, supported over the grid $Z$, and is therefore straightforward to sample from. Specifically, to sample from it, one first draws an integer $j \in \{1,\ldots,K\}$ randomly according to the discrete distribution $\{ \vert u_{lj} \vert^2 \}^{K}_{l=1}$ and then sets $\bm{y} = \bm{z}_j$.

With this in hand, we now have all the ingredients needed to define \textit{Christoffel Adaptive Sampling (CAS)}. This is a procedure that, given a (possibly redundant) dictionary $\{ \psi_1,\ldots,\psi_N\}$ and uniform grid $Z = \{\bm{z}_l\}^{K}_{l=1}$, generates an orthonormal basis for their span, and then draws $m$ samples according to the measures \R{mu_i-disc-SVD}. We summarize this in Algorithm \ref{alg:CAS}. Note that this algorithm allows for the possibility that $m / n \notin \bbN$. In this case, it simply draws one additional sample from the first $m - \lfloor m/n \rfloor n$ measures to ensure a total of $m$ samples.

\renewcommand{\algorithmicrequire}{\textbf{Input:}}
\renewcommand{\algorithmicensure}{\textbf{Output:}}
\begin{algorithm}[t] \caption{Christoffel Adaptive Sampling (CAS)}
\label{alg:CAS} 
\begin{algorithmic}[1]
\Require Finite grid $Z = \{\bm{z}_i \}^{K}_{i=1}$, dictionary $\{\psi_1,\ldots,\psi_N\}$ with $N \leq K$, number of samples $m$, tolerance $\epsilon_{\mathrm{tol}}>0$.
\Ensure Sample points $\bm{y}_1,\ldots,\bm{y}_m$.

\State Construct the matrix $\bm{B}=\{\sqrt{1/K}\psi_j(\bm{z}_i)\}_{i,j=1}^{K,N}$.
\State Compute the SVD $\bm{B}=\bm{U}\bm{\Sigma}\bm{V}^*$, and write  
$\bm{U}=\{u_{ij}\}_{i,j=1}^{K,N}$ and  $\bm{\Sigma}=\mathrm{diag}(\sigma_1,\ldots,\sigma_N)$.
\State Compute the numerical dimension $n =  \max \{ i : \sigma_i / \sigma_1 > \epsilon_{\mathrm{tol}} \}$.
\State Construct the discrete measures $\mu_j=\lbrace \vert u_{ij} \vert^2\rbrace_{i=1}^{K}$ for $j=1,\ldots,n$.
\State Set $k=\lfloor m/n \rfloor$ and $s=m-kn$.
\For{$j=1,\ldots,n$}
    \State Draw $k$ integers $\{j_1,\ldots,j_{k}\}$ randomly and independently from $\{1,\ldots,K\}$, according to the discrete distribution $\mu_j$. Set $\bm{y}_{q}=\bm{z}_{j_i}$, where $q=k(j-1)+i$ for $i=1,\ldots,k$.
\EndFor
\For{$t=1,\ldots,s$}
    \State Draw an integer $j$ randomly and independently from $\{1,\ldots,K\}$, according to the discrete distribution $\mu_t$. Set $\bm{y}_{kn+t}=\bm{z}_j$.
\EndFor 
\end{algorithmic}
\end{algorithm}

\section{Christoffel Adaptive Sampling for Deep Learning}
\label{sec:CAS4DL}

In this section, we describe the method \textit{Christoffel Adaptive Sampling for Deep Learning (CAS4DL)}.

\subsection{Neural networks and deep learning}
\label{ss:NN_DL}

We commence by recapping some standard deep learning concepts.

\begin{definition}{(feedforward neural network)} Let $L\in\mathbb{N}_0$ and $N_0,\ldots,N_{L+2}\in\mathbb{N}$. A map $\Psi:\mathbb{R}^{N_0}\rightarrow\mathbb{R}^{N_{L+2}}$ given by 
\begin{equation}\label{DNN}
\Psi(\bm{x}) = 
\begin{cases} 
      \mathcal{A}_1(\rho(\mathcal{A}(\bm{x}))) & L=0 \\
      \mathcal{A}_{L+1}(\rho(\mathcal{A}_L(\rho(\ldots\rho(\mathcal{A}_0(\bm{x}))\ldots)))), & L\geq 1 \\
\end{cases}
\end{equation}
with affine linear maps $\mathcal{A}_l:\mathbb{R}^{N_l}\rightarrow\mathcal{R}^{N_{l+1}}$, $l=0,\ldots,L+1$, and the activation function $\rho$ acting component-wise (i.e., $\rho(\bm{x})=(\rho(x_1),\ldots,\rho(x_d))$ for $\bm{x}=(x_1,\ldots,x_d)$) is called a Neural Network (NN). The map corresponding to layer $l$ is given by $\mathcal{A}_l(\bm{x})=\bm{W}_l\bm{x}+\bm{b}_l$, where $\bm{W}_l\in\mathbb{R}^{N_{l+1}\times N_l}$ is the $l$th weight matrix and $\bm{b}_l\in\mathbb{R}^{N_{l+1}}$ the $l$th bias vector. We refer to $L$ as the depth of the network and $\max_{1\leq l\leq L+1}N_l$ as its width.
\end{definition}

The layers $l=1,\ldots,L$ are termed to as \textit{hidden} layers. Informally, we term any NN with $L \geq 1$ hidden layers a Deep NN (DNN). Throughout this paper, we consider standard, feedforward neural networks \R{DNN}. There are many other constructions, such as ResNets, to which our adaptive sampling procedure is equally applicable. However, we shall not consider these. 
We also consider so-called \textit{fully connected} networks, i.e., DNNs where the weights and biases can take arbitrary real values.

The \textit{architecture} of a DNN is the particular choice of activation function $\rho$ and parameters $L$ and $N_1,\ldots,N_{L+1}$. We denote the set of neural networks of a given architecture by $\mathcal{N}$. Note that this is a class of functions parametrized by the weight matrices $\bm{W}_l$ and biases $\bm{b}_l$. 

We now describe the DNN architectures used in this work. Throughout this paper, we assume the widths of the hidden layers to be equal, $N_1=\ldots=N_{L+1}=N$. Note that $N_0$ and $N_{L+2}$ are defined by the problem, as its input and output dimensions, respectively. In particular, $N_0=d$ and $N_{L+2}=1$ in the majority of this work. 

In practice, there are numerous options for the activation function $\rho$, and one can also use different activation functions in different layers. In this work, we maintain the same activation function $\rho$ in all the layers. One of the most commonly-used activation functions is the \textit{Rectified Linear Unit} (ReLU), which is defined by $\rho(x) = \max\{0,x\}$, and it is considered in our numerical experiments. However, in this work since we primarily strive to approximate smooth functions, we compare ReLU with smoother activation functions. These typically give better performance. Specifically, we also consider the \textit{hyperbolic tangent} ($\tanh$) and \textit{Exponential Linear Unit} (ELU): 

\begin{alignat*}{2} 
&\text{tanh:} & \quad \rho(x) & = \tanh(x), \\
&\text{ELU:}  & \rho(x) & = \begin{cases} x & x>0 \\ \E^x-1 & x\leq 0 \end{cases}.
\end{alignat*}
In terms of terminology, we refer to a DNN architecture with activation function $\rho$, $L$ hidden layers and $N$ nodes per layer as a $\rho$ $L\times N$ DNN.
This choice of architecture is motivated by the recent work \cite{AdcockDexter2021} which studied the relation between $L$ and $N$ for training fully connected feedforward ReLU networks on function approximation tasks.

Having defined the types of DNNs we consider, we now discuss training. Given samples $f(\bm{y}_1),\ldots,f(\bm{y}_m)$ of an unknown function $f\in L^2(\Omega,\tau)$, \textit{training} is the procedure of computing a neural network $\Psi \in \cN$ that approximates $f$ from the data $\{(\bm{y}_i,f(\bm{y}_i))\}_{i=1}^m$. This is typically accomplished by minimizing a \textit{loss function}. In the context of approximation or regression problems the $\ell^2$-loss (also known as \textit{empirical risk}) is a standard choice. However, in this work, similar to \cite{AdcockCardenas2020,MiglioratiCohenOptimal,migliorati2021multivariate}  
, we consider the \textit{weighted $\ell^2$-loss function}, 
\be{
\label{eq:minimization_prob}
\min_{\Psi\in \mathcal{N}}\sum_{i=1}^m
w(\bm{y}_i)\abs{\Psi(\bm{y}_i)-f(\bm{y}_i)}^2,
} 
where $w$ is a positive function, in this case the weight function introduced in \eqref{w-def}. We notice that minimizing over $\mathcal{N}$ is equivalent to a minimization problem for the weights $\bm{W}_l$ and biases $\bm{b}_l$.  

\subsection{CAS4DL}

We are now ready to introduce our method for adaptive sampling in DL. We first fix a DNN architecture $\cN$ consisting of $L \times N$ DNNs with activation function $\rho$. In a small departure from the above setup, we assume the bias in the final layer is set to zero (this makes little difference either theoretically or practically).
Let $\Psi \in \cN$ be arbitrary. The idea, based on \cite{CyrEtAl2020}, is to view $\Psi$ as a linear combination of functions defined by its penultimate layer. Specifically, we write
\bes{
\Psi = \sum^{N}_{i=1} c_i \psi_i,
}
where $c_i \in \bbR$ and $\psi_i : \bbR^d \rightarrow \bbR$ is the value of the DNN $\Psi$ in the $i$th neuron in its penultimate layer. More compactly, we may write
\bes{
\Psi = \bm{c}^{\top} \bm{\Psi},\quad \bm{c} \in \bbR^N,
}
where
\bes{
\bm{\Psi} : \bbR^d \rightarrow \bbR^N,\ \bm{y} \mapsto (\psi_i(\bm{y}) )^{N}_{i=1},
}
is the DNN obtained from the first $L$ layers of $\Psi$.

CAS4DL is now defined as follows. Suppose an initial DNN $\widehat{\Psi} = \widehat{\Psi}^{(0)}$ has been computed. In our experiments, this is simply constructed via a standard (random) initialization strategy. Then, write
\bes{
\widehat{\Psi}^{(0)} = (\hat{\bm{c}}^{(0)})^{\top} \widehat{\bm{\Psi}}^{(0)},\quad \bm{c}^{(0)} \in \bbR^N,
}
which yields a dictionary $\{ \hat{\psi}^{(0)}_1,\ldots,\hat{\psi}^{(0)}_{N} \}$ for the subspace $P^{(0)}$ within which $\hat{\Psi}^{(0)}$ belongs. We then input this dictionary into the CAS method (Algorithm \ref{alg:CAS}) with the desired number of samples $m = m_1$ to obtain the sample points $\bm{y}_1,\ldots,\bm{y}_{m_1}$. Next, we train a new DNN (of the same architecture) $\widehat{\Psi}^{(1)}$. This is done by applying a standard optimizer to the weighted $\ell^2$-loss
\bes{
\min_{\Psi\in \mathcal{N}}\sum_{i=1}^{m_1}
w(\bm{y}_i)\abs{\Psi(\bm{y}_i)-f(\bm{y}_i)}^2,
} 
using $\widehat{\Psi}^{(0)}$ as an initialization.
We now simply repeat this procedure. Thus, at step $l \geq 2$, we write the current DNN as
\bes{
\widehat{\Psi}^{(l-1)} = (\hat{\bm{c}}^{(l-1)})^{\top} \widehat{\bm{\Psi}}^{(l-1)},
\quad \bm{c}^{(l-1)} \in \bbR^N,
}
so as to obtain the dictionary $\{ \hat{\psi}^{(l-1)}_{1},\ldots,\hat{\psi}^{(l-1)}_{N} \}$ for the subspace $P^{(l-1)} \ni \widehat{\Psi}^{(l-1)}$. Then we input this into Algorithm \ref{alg:CAS} with the number $m = m_l - m_{l-1}$ of new samples desired. We then train a new DNN $\widehat{\Psi}^{(l)}$ using these samples and the previous DNN $\widehat{\Psi}^{(l-1)}$ as initialization. The resulting method, CAS4DL, is formally described in Algorithm \ref{alg:CAS4DL}. Note that the set of sample points in line 4 of Algorithm \ref{alg:CAS4DL} is drawn using Algorithm \ref{alg:CAS}. This algorithm does not reject any samples, meaning that the full set of samples is used to train the DNN $\hat{\Psi}^{(l)}$ at step $l$ of CAS4DL.

\renewcommand{\algorithmicrequire}{\textbf{Input:}}
\renewcommand{\algorithmicensure}{\textbf{Output:}}
\begin{algorithm}[t] \caption{Christoffel Adaptive Sampling for Deep Learning (CAS4DL)}
\label{alg:CAS4DL} 
\begin{algorithmic}[1]
\Require Target function $f$, DNN parameters $L$ and $N$, activation function $\rho$, finite grid $Z = \{ \bm{z}_i \}^{K}_{i=1}$ with $K \geq N$, numbers of samples $0=m_0<m_1<m_2<\ldots$, tolerance $\epsilon_{\mathrm{tol}}>0$.
\Ensure A sequence of DNN approximations $\widehat{\Psi}^{(1)},\widehat{\Psi}^{(2)},\ldots$ to $f$.
\State Initialize the DNN $\widehat{\Psi}^{(0)}$.
\For{$l = 1,2,\ldots,$}
    \State Construct a dictionary $\{ \hat{\psi}^{(l-1)}_1,\ldots,\hat{\psi}^{(l-1)}_{N}\}$ via the relation $\widehat{\Psi}^{(l-1)} = (\hat{\bm{c}}^{(l-1)})^{\top} \widehat{\bm{\Psi}}^{(l-1)}$, where $\widehat{\bm{\Psi}}^{(l-1)} = (\hat{\psi}^{(l-1)}_i)^{N}_{i=1}$.
    \State Draw $m_l-m_{l-1}$ samples points $\bm{y}_{m_{l-1}+1},\ldots,\bm{y}_{m_l}$ using Algorithm \ref{alg:CAS}.
    \State Train a DNN $\widehat{\Psi}^{(l)}$ using the data $\{(\bm{y}_i,f(\bm{y}_i))\}_{i=1}^{m_l}$ by applying a standard optimizer to the $\ell^2$-loss \R{eq:minimization_prob} with initialization $\widehat{\Psi}^{(l-1)}$.
\EndFor
\end{algorithmic}
\end{algorithm}

\subsection{Discussion}

Several remarks are in order. First, we observe that this scheme is adaptive in two key senses. Consider step $l$ of the algorithm. We assume that $\widehat{\Psi}^{(l-1)}$ is a good approximation for $f$, and therefore (by definition), $P^{(l-1)} = \spn \{ \hat{\psi}^{(l-1)}_{1},\ldots,\hat{\psi}^{(l-1)}_N \}$ is a good subspace within which to approximate $f$. Therefore, we choose the next set of $m_l - m_{l-1}$ samples so as to optimize the approximation in this subspace. The samples are therefore adapted to $\widehat{\Psi}^{(l-1)}$, and therefore by extension, to $f$. However, we also retrain the network, using $\widehat{\Psi}^{(l-1)}$ as an initial point, to get a new DNN $\widehat{\Psi}^{(l)}$ and subspace $P^{(l)}$. Hence, the subspace is also adapted to the function.

On the other hand, while the samples $m_l - m_{l-1}$ are fully adapted to the subspace $P^{(l-1)}$, the previous $m_{l-1}$ samples are not. Indeed, they were chosen in terms of the previous subspaces $P^{(0)},\ldots,P^{(l-2)}$. This issue has been considered in previous works on Christoffel adaptive sampling under the assumption that the subspaces are chosen {\em a priori} and are nested, i.e., $P^{(0)} \subseteq P^{(1)} \subseteq \cdots$ \cite{Migliorati2019,arras2019sequential,AdcockCardenas2020}. However, this is not possible in our setting, since the subspaces are constructed adaptively and are generally not nested. Nonetheless, we expect that the samples are `good' for the subspace $P^{(l)}$, even if they are not necessarily optimal.

Finally, it is now possible to see more clearly why a truncated SVD is employed in Algorithm \ref{alg:CAS}. The dictionary $\{ \psi^{(l-1)}_1,\ldots,\psi^{(l-1)}_N \}$ consists of DNNs with the same activation function. It is therefore unsurprising that this dictionary can easily be numerically redundant, depending on the architecture of the DNN and the way it was trained. We see examples of this in our numerical experiments next.
 
\section{Numerical experiments}
\label{sec:experiments}

In this section, we present numerical experiments showing the performance of CAS4DL with different DNN architectures on a variety of smooth function approximation problems.

\subsection{Experimental setup}
\label{sec:experimental_setup}

We now describe the setup for our experiments. Throughout, we consider approximating smooth functions defined on the domain $\Omega=[-1,1]^d$, i.e., a compact hypercube in $d$ dimensions. Let $Z = \{\bm{z}_i\}_{i=1}^K$ be a grid of points of size $K$, drawn uniformly and randomly from $\Omega$, and let $\tau$ be the uniform measure over $Z$. In our experiments, we vary the size $K$ of the grid $Z$ depending on the dimension $d$. When $d=1,2$ we set $K=\num{10000}$, then we use the values $K=\num{20000}$, $K = \num{50000}$ and $K=\num{100000}$ in dimensions $d = 4$, $d = 8$ and $d = 16$, respectively. We consider the following functions, all of which are standard test functions for smooth, multivariate function approximation tasks (see \cite{Genz1984} and, in particular, \cite{adcock2021sparse} for $f_1$, $f_2$, \cite{Migliorati2019} for $f_3$ and \cite{AdcockDexter2021} for $f_4$):
\begin{align*}
&f_1(y) = \exp\left(-\frac{1}{d}\sum_{i=1}^dy_i\right),\\
&f_2(y) = \frac{\prod_{k=\lceil d/2\rceil +1}^d \cos\left(16y_k/2^k\right)}{\prod_{k=1}^{\lceil d/2\rceil}(1-y_k/4^k)},\\
&f_3(y) = \left( 1 + \frac{1}{2d}\sum_{i=1}^d q_iy_i\right)^{-1}, 
\quad 
q_i = 10^{-\frac{3(i-1)}{d-1}}, \quad i=1,\ldots,d. \\
&f_4(y) = \left(
\frac{\prod_{k=1}^{\lceil d/2\rceil} (1+4^k y_k^2) }{\prod_{k=\lceil d/2\rceil +1}^{d} (100+5y_k)}
\right)^{1/d},  
\end{align*}
As mentioned previously, we consider testing DNN performance in function approximation tasks using the MLFA framework introduced in \cite{AdcockDexter2021}. We summarize the key points of our methodology here:\\
\indent\textbf{(i) Implementation.} Our experiments have been implemented in TensorFlow version 2.5, and the code to reproduce these results is available on GitHub at \url{https://github.com/JMcardenas/CAS4DL}. 
\\
\indent\textbf{(ii) Hardware.} 
The majority of our computations were performed in single precision using NVIDIA Tesla P100 GPUs available through Compute Canada's Cedar compute cluster at Simon Fraser University \citep{cedar_SFU}. Some computations were also performed on Google Colab. The experiments in Section \ref{sec:app_parametric_PDEs} were conducted on a single NVIDIA GeForce GTX 1080 Ti.
\\
\indent\textbf{(iii) Choice of architectures and initialization.} We consider fully-connected $\rho$ $L\times N$ DNNs with different values of $L$ and $N$. As discussed, $\rho$ is either the Rectified Linear Unit (ReLU), hyperbolic tangent ($\tanh$) or Exponential Linear Unit (ELU).

As discussed in Section \ref{ss:NN_DL} we consider DNNs with a fixed number of nodes $N$ for each layer. Here we also focus our attention on DNNs with depth $L$ such that the ratio $\beta:= L/N$ is small, motivated by results from \cite{Hanin2018}. In \cite{AdcockDexter2021} it was observed empirically that values of $L$ between 5-10 times smaller than $N$ on average yielded the best results in smooth function approximation problems. Here we consider $5\times 50$ and $10\times 100$ DNNs so that $\beta=0.1$.

For the initialization strategy, we consider weights and biases with entries drawn from the normal distribution with mean 0 and variance 0.01.
For the size of architectures considered, this yields smaller variance than the popular Xavier and He Normal initializations, see e.g. \cite{He2016,HaninRolnick2018}, and was demonstrated to be an effective choice for function approximation in \cite{AdcockDexter2021}. Our networks are also initialized with the same seed to compare the effect of varying the samples.

\indent\textbf{(iv) Optimizers for training and parametrization.}
We use \textit{Adam} as the optimizer for training our network weights and biases, and use an exponentially decaying learning rate with respect to the number of epochs. 
For the CAS4DL strategy, we train the networks for \num{5000} epochs on each new set of samples, using the weights and biases from the previous set as initialization for the next as described in Algorithm \ref{alg:CAS4DL}. We repeat this procedure 10 times, for a total of \num{50000} epochs.
For MC sampling, we train the network for a total of \num{50000} epochs, stopping every \num{5000} epochs to add more data points drawn from the uniform distribution, and continuing to train from the previous point.
\\
\indent\textbf{(v) Training data and design of experiments.} To better understand the performance of our method, we display the average testing error and run statistics over 20 trials of both the CAS4DL and MC sampling approaches for each plot.
We train our DNNs over a set of training data consisting of the values $\{(\bm{y}_i,f(\bm{y}_i))\}_{i=1}^{m_l}$. To generate each data set of size $m_l$, with $0<m_l<\ldots<m_{\max}$, we sample 20 i.i.d.\ sets of points $\{\bm{y}_i\}_{i=1}^{m_k}$ from the Monte Carlo uniform distribution on $(-1,1)^d$ and evaluate our target function $f$ at these points to form training data for Monte Carlo sampling. However, for CAS4DL we consider 20 sets of points $\{\bm{y}_i\}_{i=1}^{m_k}$ from the probability distribution defined for each DNN. To train the networks we used the following number of samples points $m\in\{1000,1400,1900,2300,2800,3200,4100,4600,5000\}$.
\\
\indent\textbf{(vi) Testing data and error metric.} 
We report the average over all trials of the relative $L^2$ error
\begin{equation}
\label{eq:relative_L2_error}
E(f) = \frac{\nmu{f-\widehat{\Psi}}_{L^2}}{\nmu{f}_{L^2}}
\end{equation} 
as the test error, where $\widehat{\Psi}$ is either the CAS4DL or MC approximation. 
As opposed to the training data, we calculate an approximation to the $L^2$ integrals using a high order isotropic Clenshaw-Curtis sparse grid quadrature rule with TASMANIAN \cite{stoyanov2015tasmanian}. The sparse grid method is selected due to the superior convergence over standard Monte Carlo integration in evaluating the global testing error metrics. See \cite{NTW08} for more details. 
In contrast to the training data, we use a large set of testing points, specifically, $\num{16385}$  ($d=1$), $\num{15361}$ ($d=2$), $\num{46721}$ ($d=4$), $\num{190881}$ ($d=8$) and $\num{353729}$, ($d=16$) points, to guarantee accurate approximation of the integral.

\indent\textbf{(vii) Methods comparison.} The methods considered in this work are:
\begin{enumerate}
\item DNNs trained using the CAS4DL sampling strategy, labelled `CAS'.
\item DNNs trained using Monte Carlo sampling, labelled `MC'. 
\end{enumerate}
For CAS4DL we use the threshold parameter $\epsilon_{\mathrm{tol}} = 10^{-6}$, since our computations are performed in single precision.

\subsection{Benefits of CAS4DL over MC sampling} 

\begin{figure}[h]
\centering
{\small 
\begin{tabular}{cccc} 
\hspace{-0.5cm}
\includegraphics[scale=0.11]{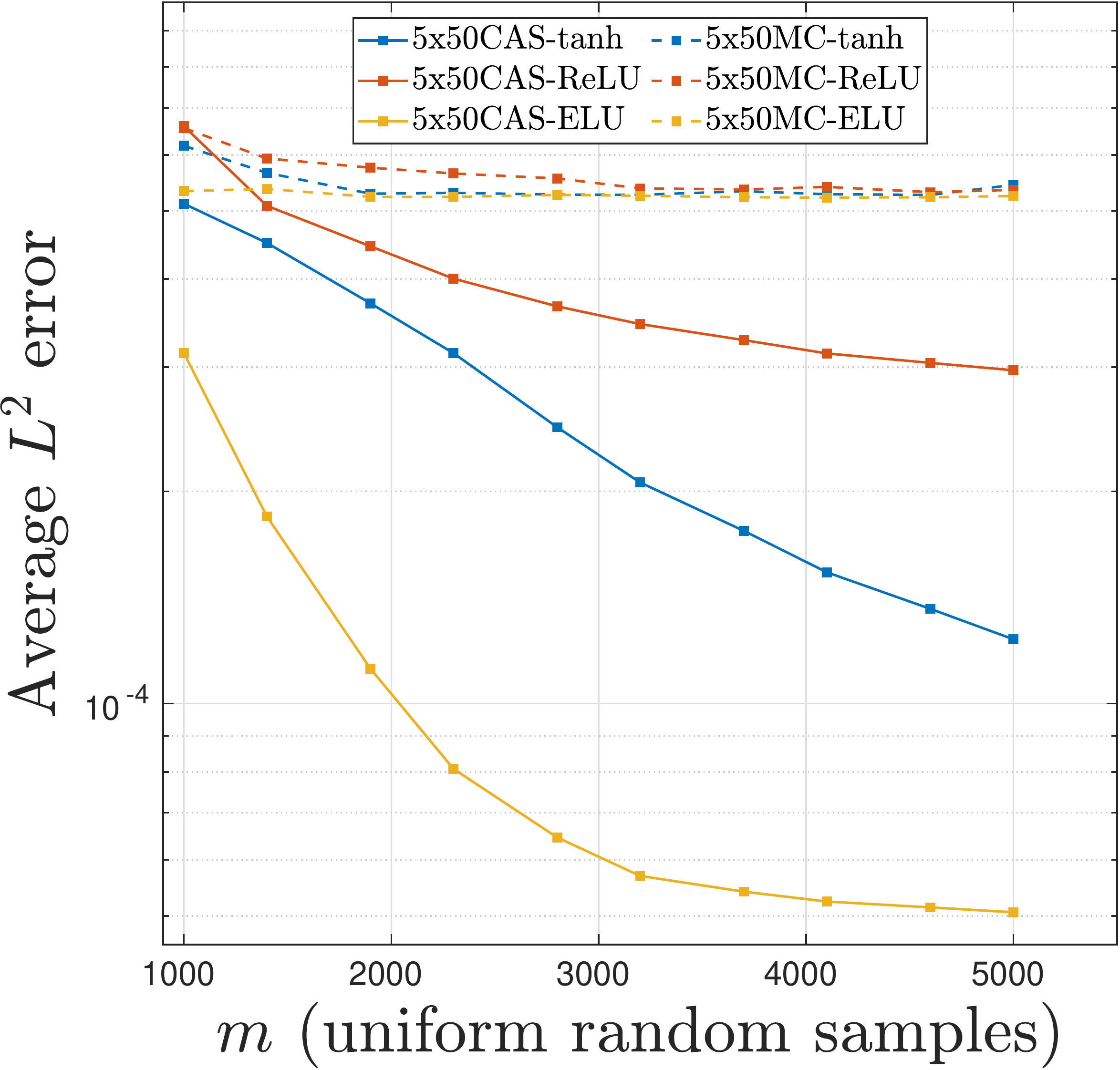} & 
\includegraphics[scale=0.11]{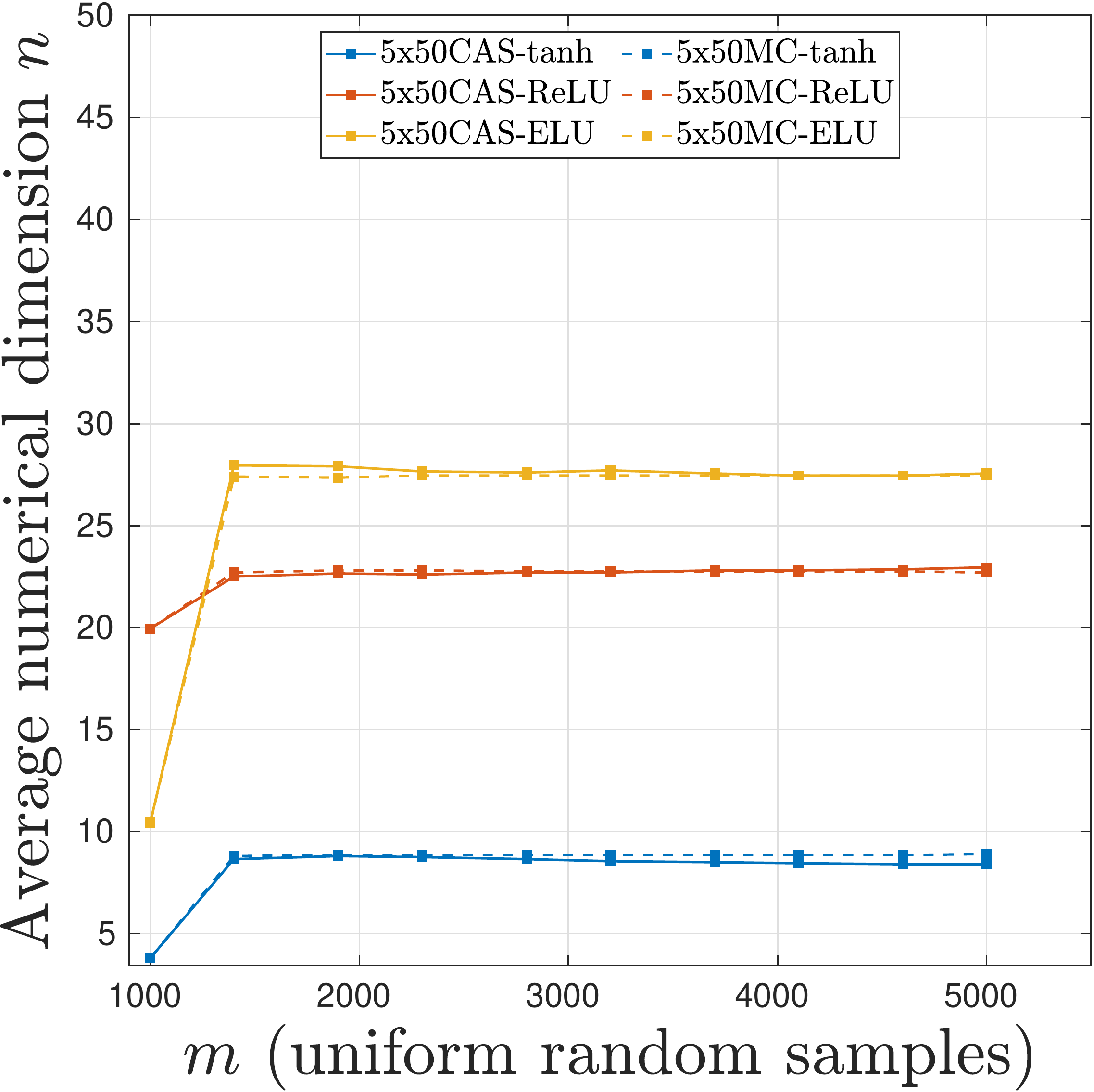} & 
\includegraphics[scale=0.11]{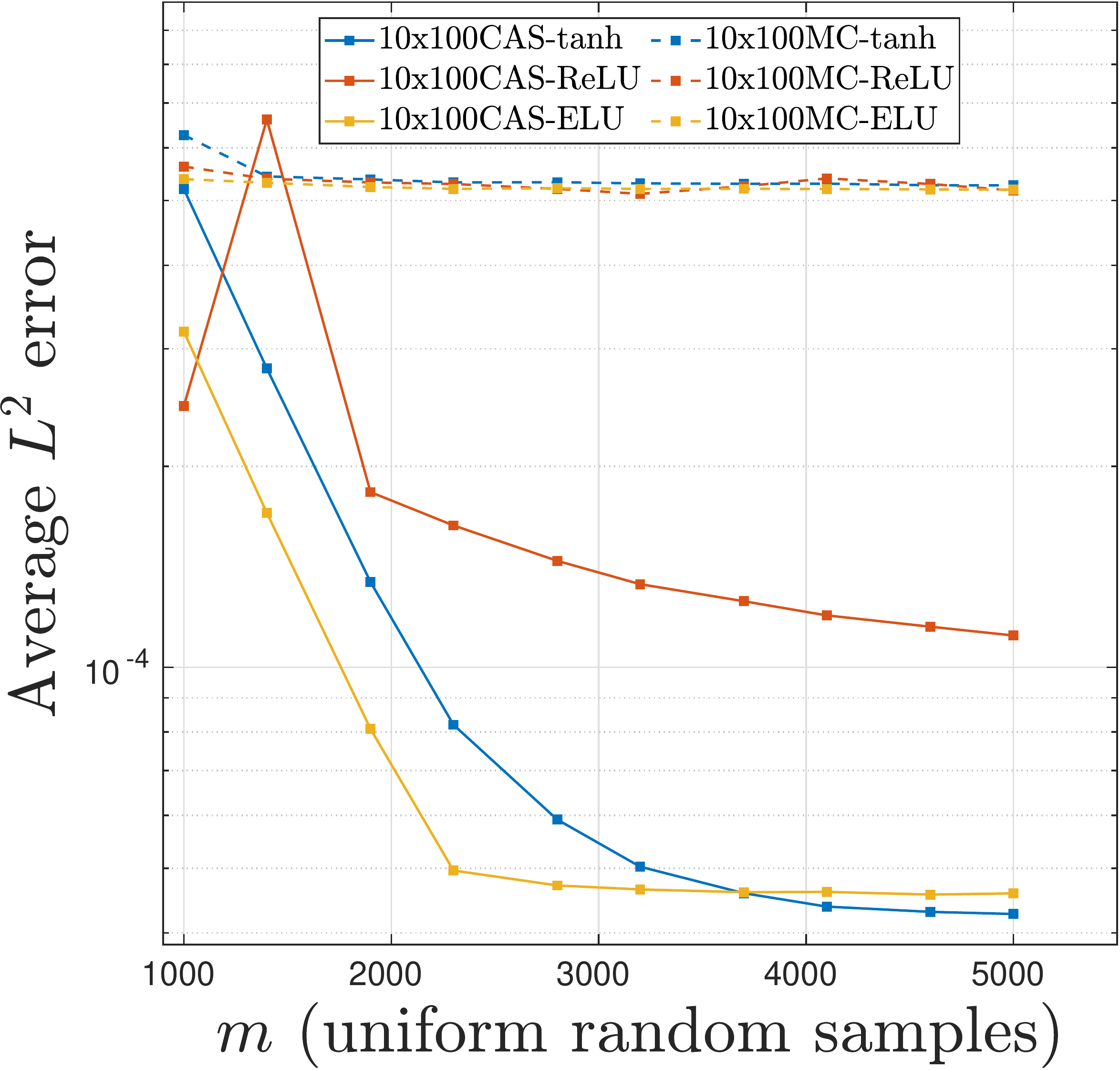} & 
\includegraphics[scale=0.11]{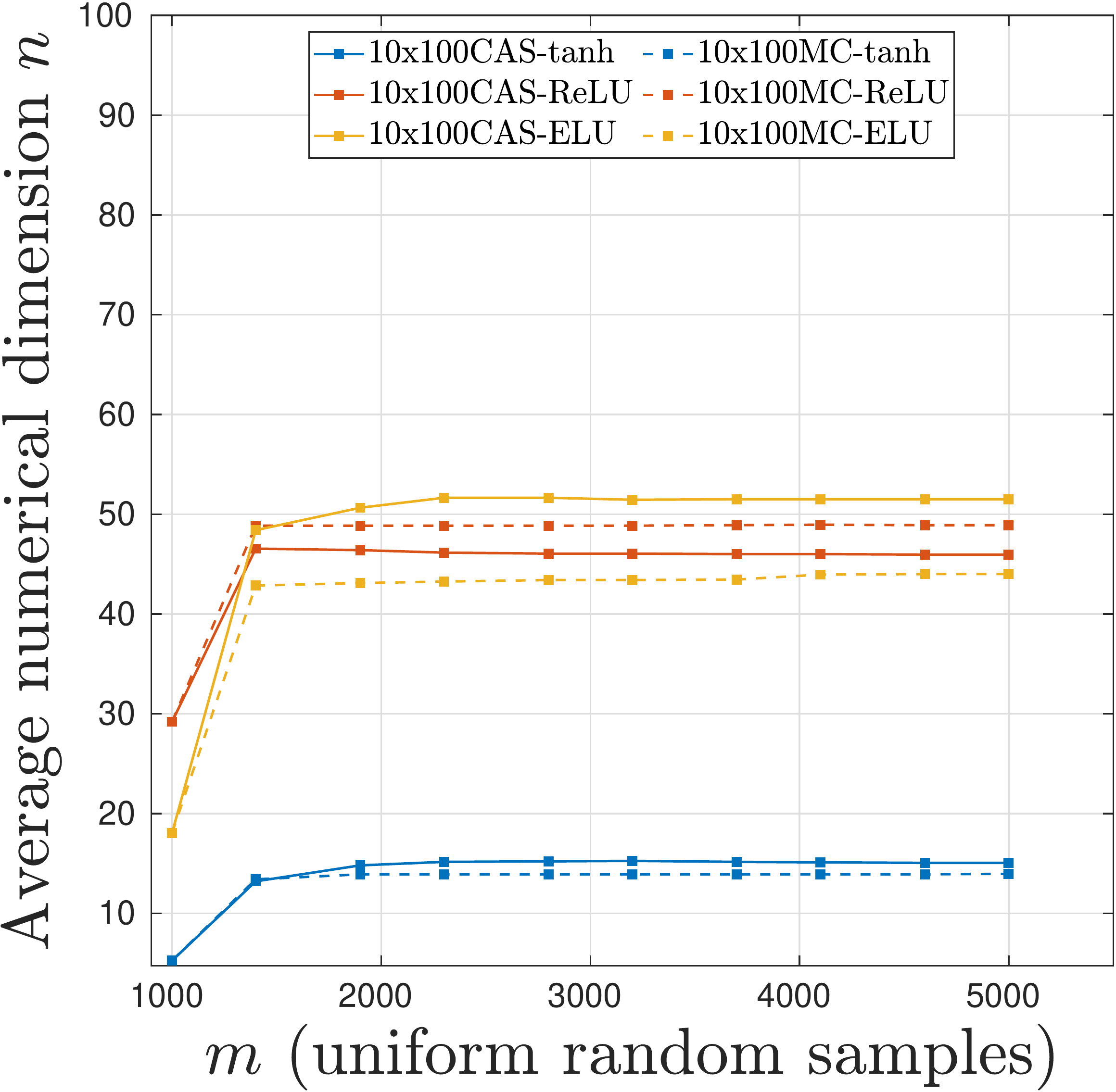} \\  
\hspace{-0.5cm}
$(f,d)=(f_1,1)$ & $(f,d)=(f_1,1)$ & $(f,d)=(f_1,1)$ & $(f,d)=(f_1,1)$ \\
\hspace{-0.5cm}
\includegraphics[scale=0.11]{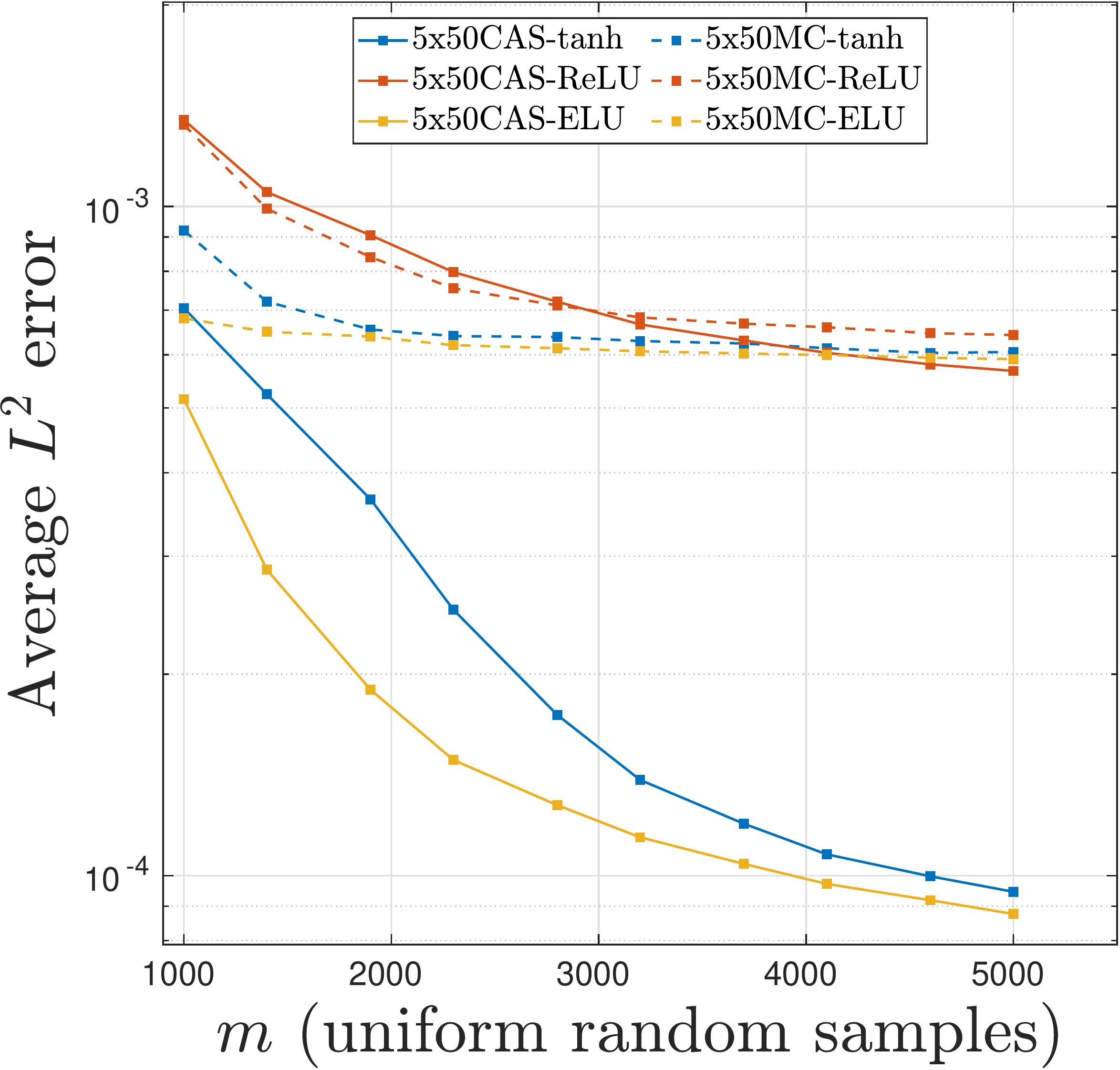} & 
\includegraphics[scale=0.11]{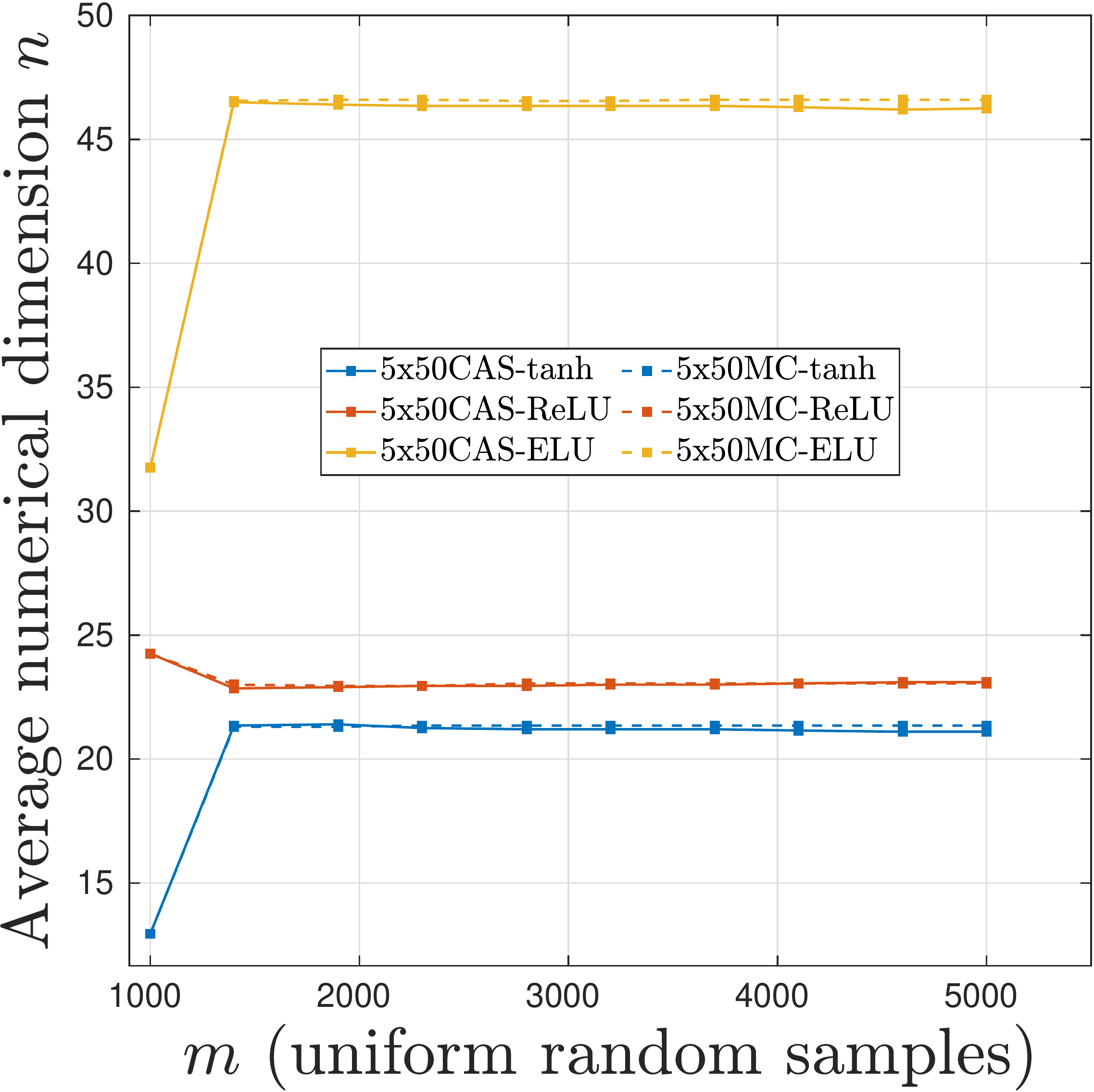} & 
\includegraphics[scale=0.11]{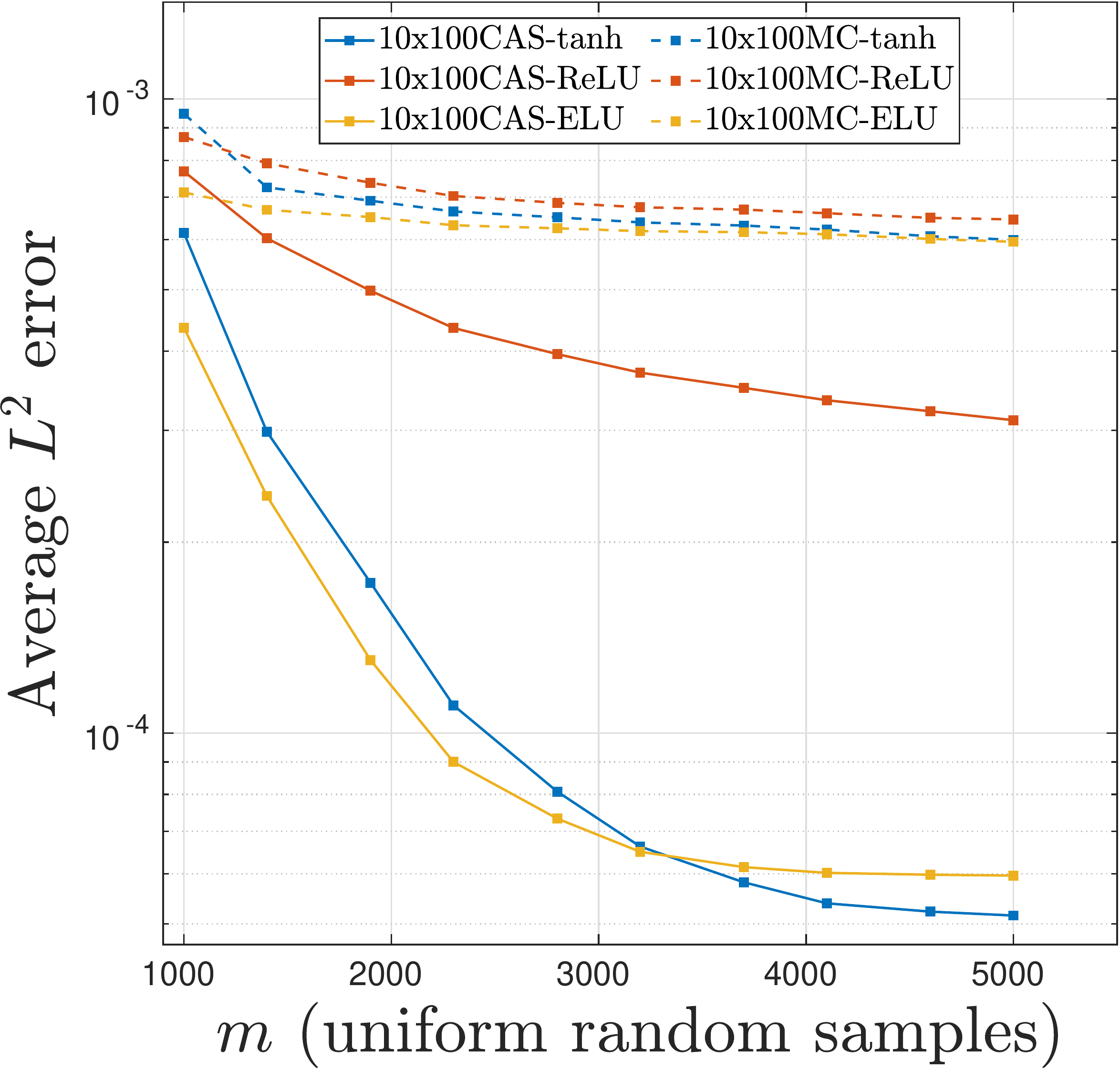} & 
\includegraphics[scale=0.11]{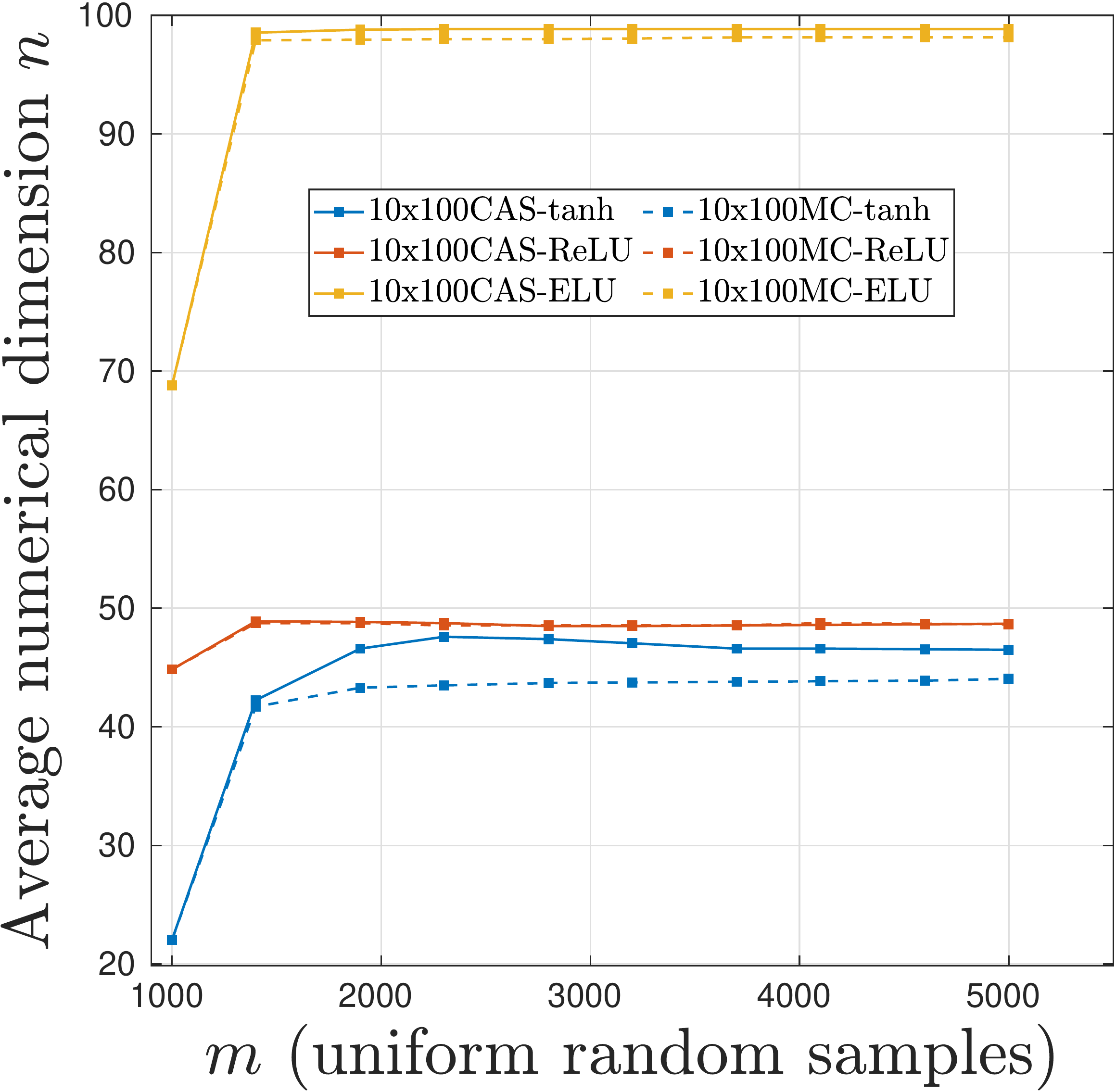} \\ 
\hspace{-0.5cm}
$(f,d)=(f_1,2)$ & $(f,d)=(f_1,2)$ &  $(f,d)=(f_1,2)$ & $(f,d)=(f_1,2)$ \\
\includegraphics[scale=0.11]{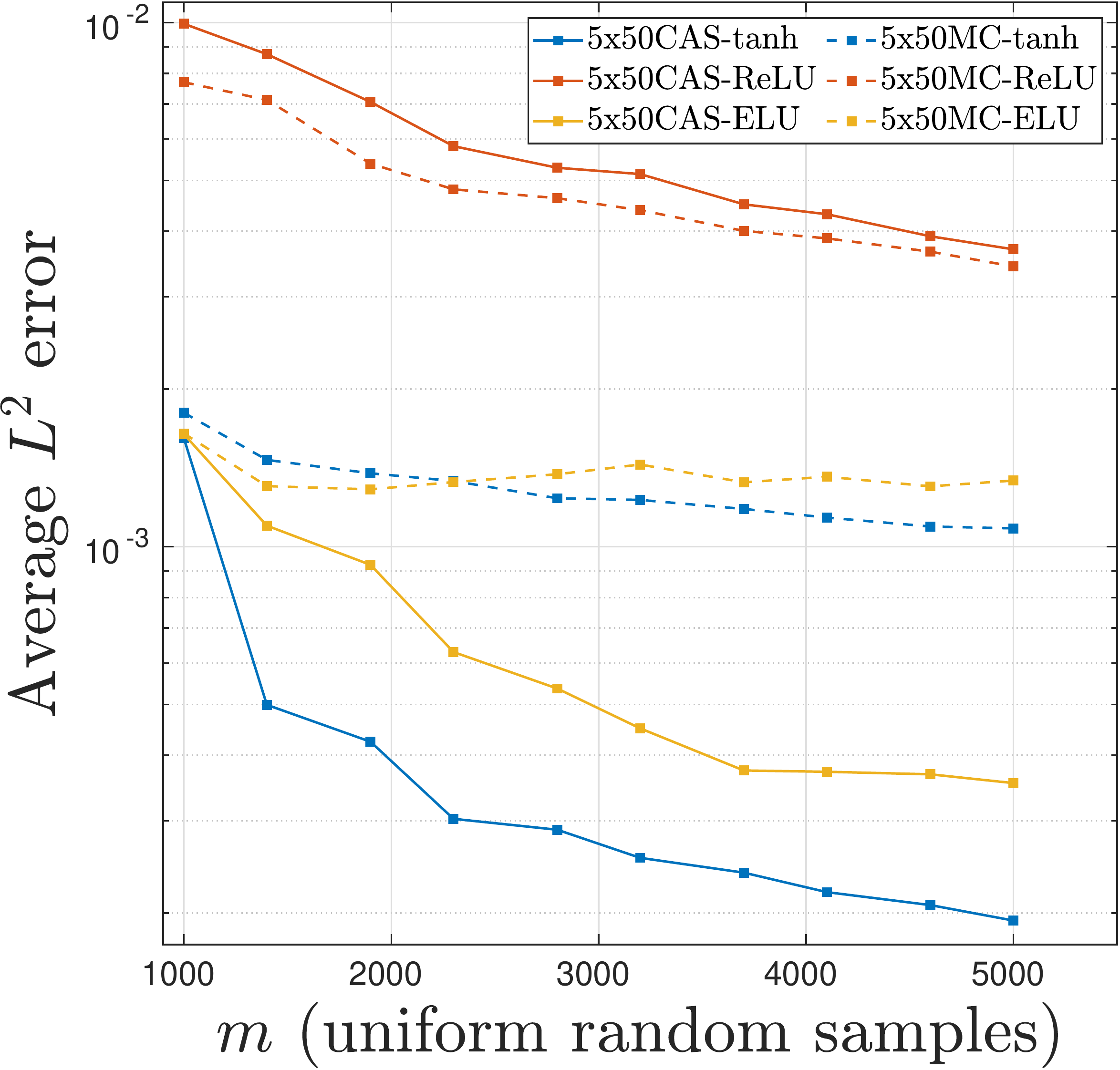} & 
\includegraphics[scale=0.11]{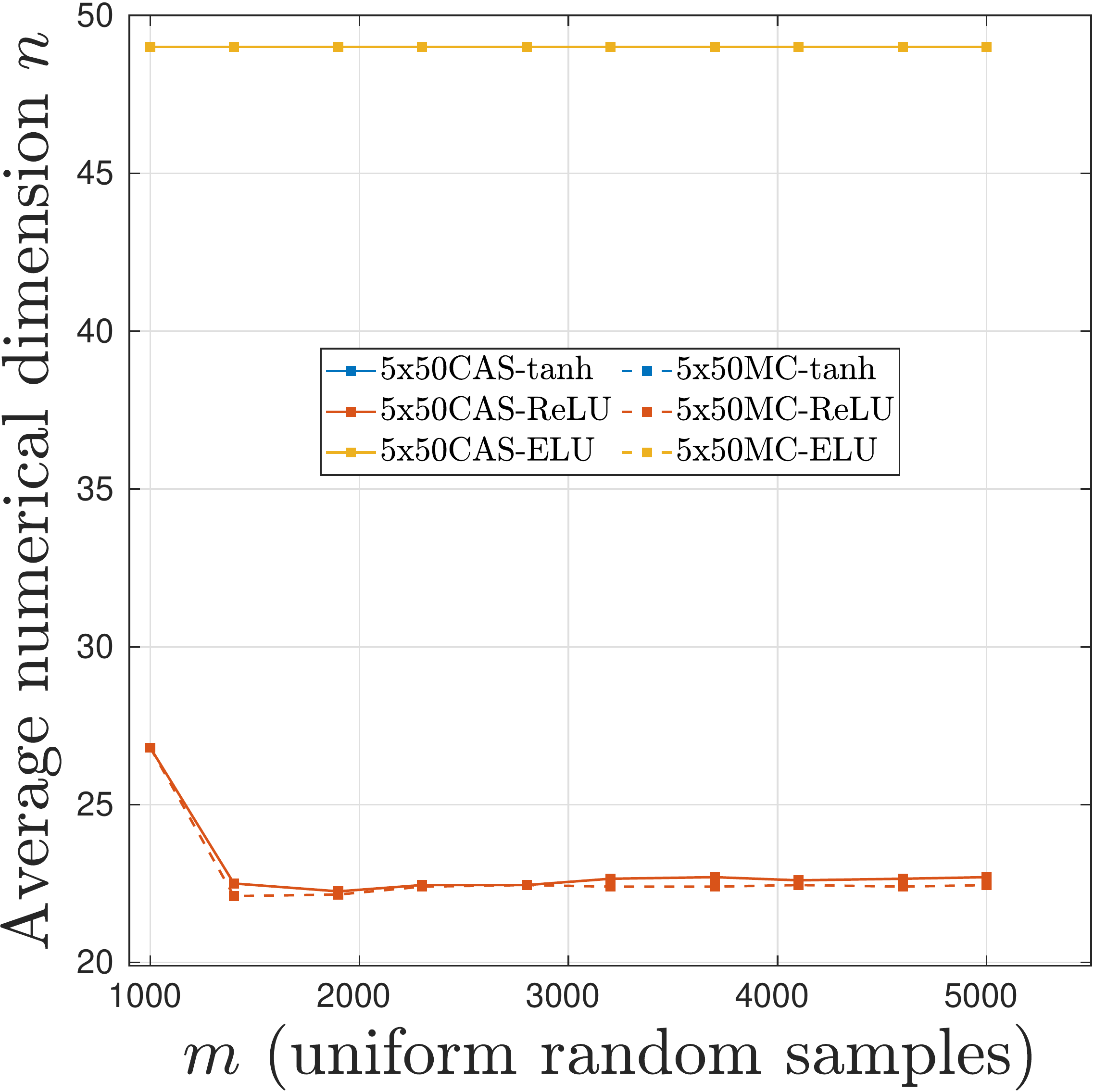} & 
\includegraphics[scale=0.11]{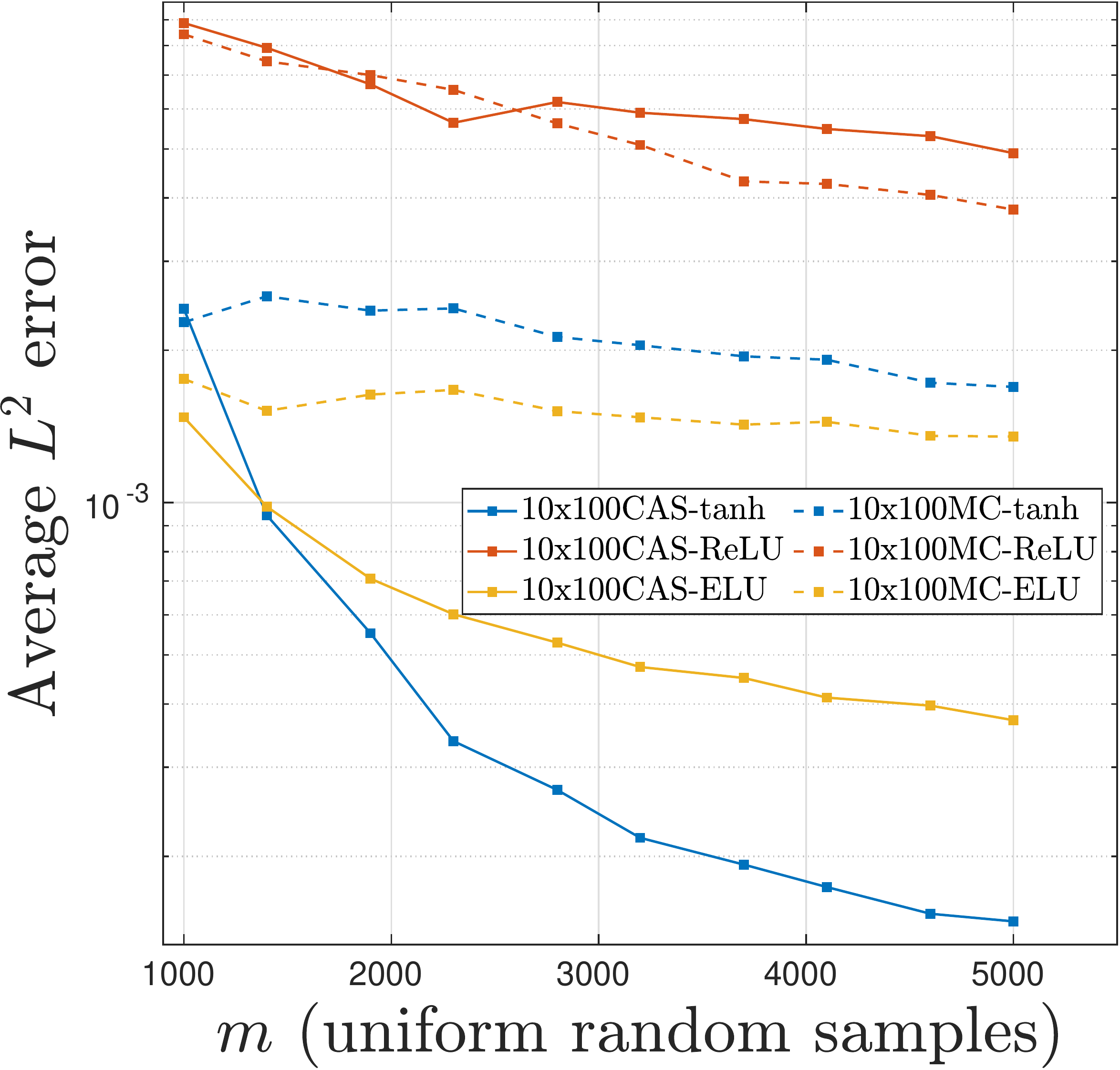} & 
\includegraphics[scale=0.11]{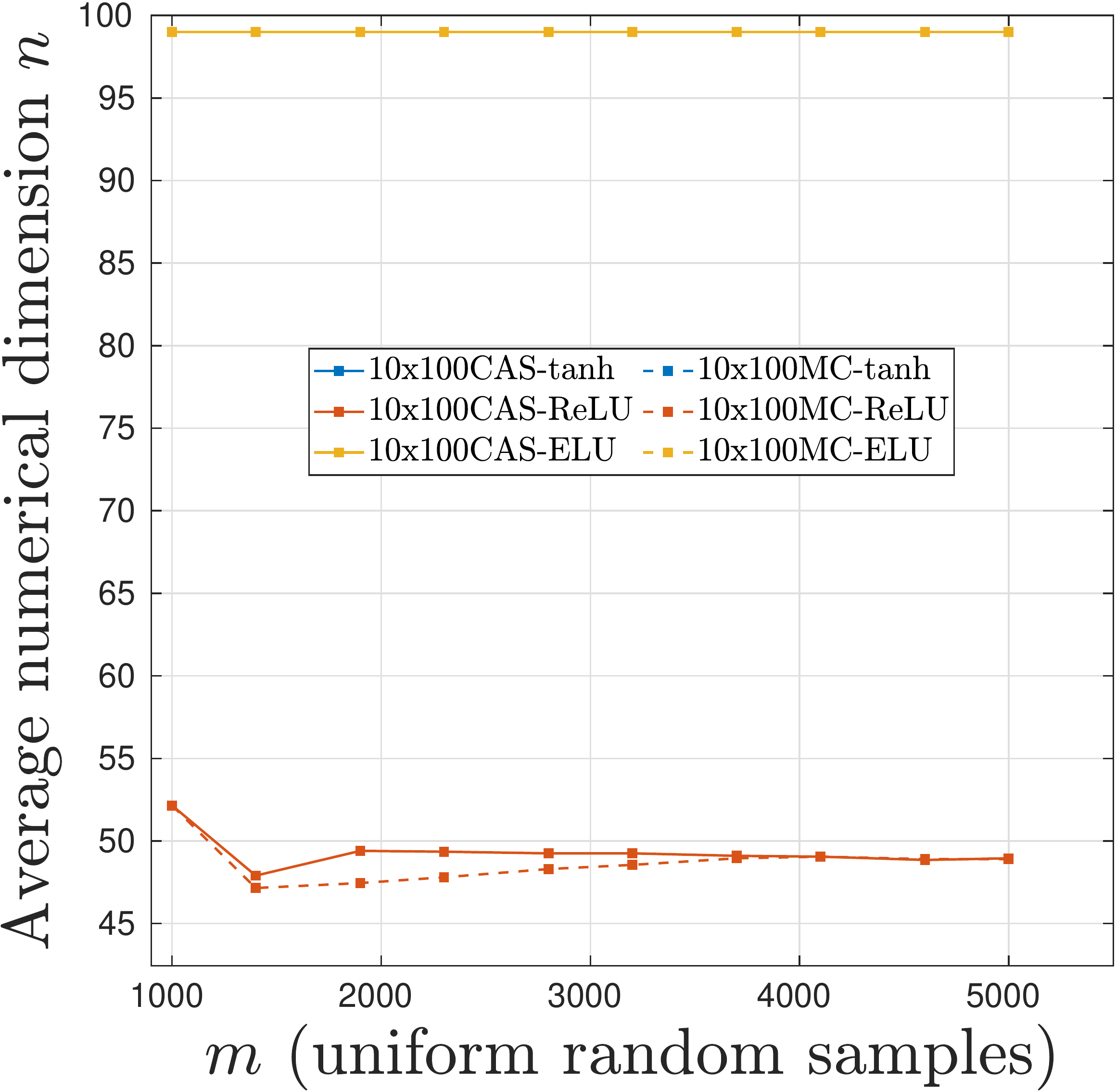} \\ 
\hspace{-0.5cm}
$(f,d)=(f_1,8)$ & $(f,d)=(f_1,8)$ & $(f,d)=(f_1,8)$ & $(f,d)=(f_1,8)$\\
\hspace{-0.5cm}
\includegraphics[scale=0.11]{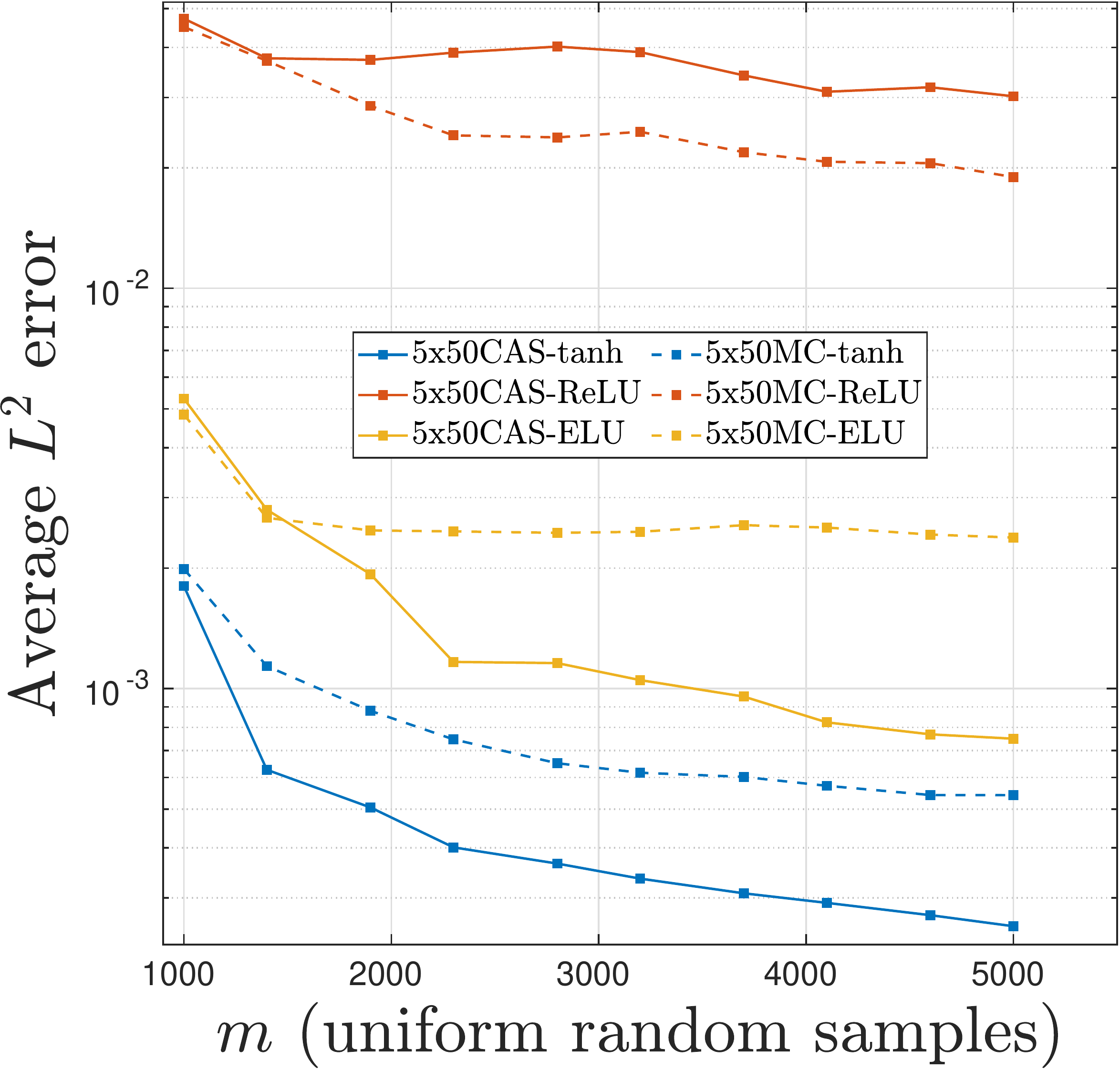} & 
\includegraphics[scale=0.11]{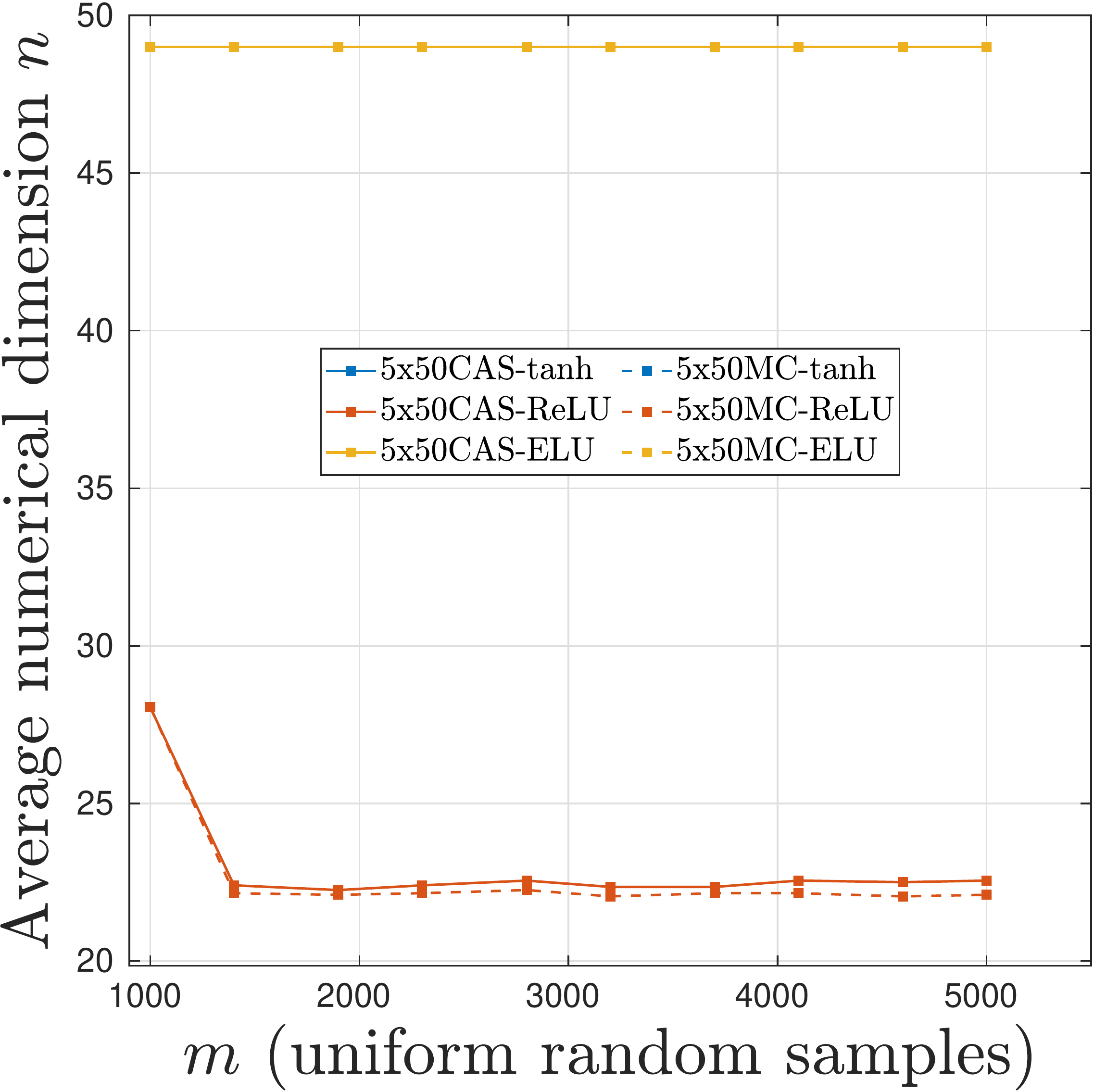} & 
\includegraphics[scale=0.11]{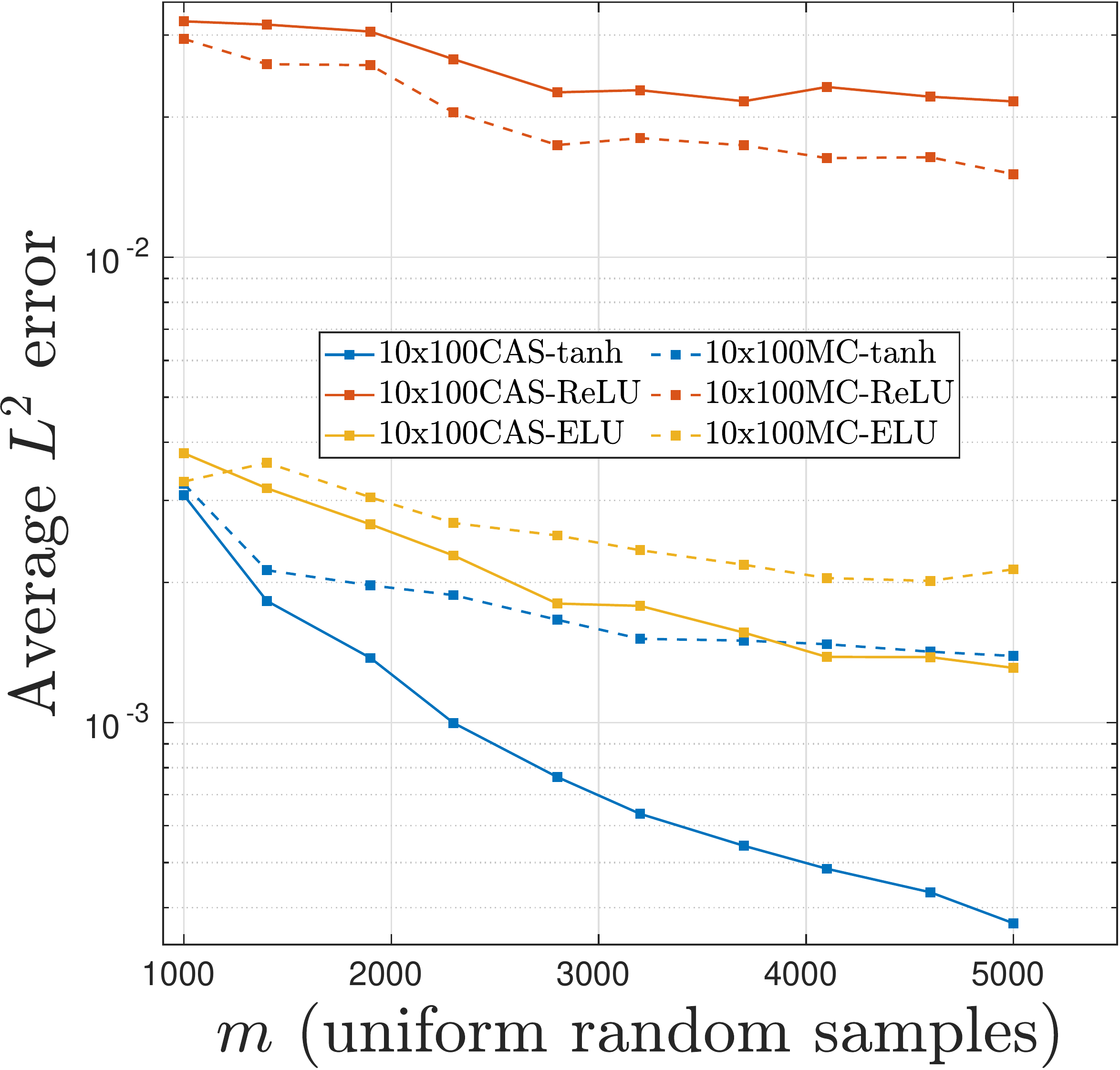} & 
\includegraphics[scale=0.11]{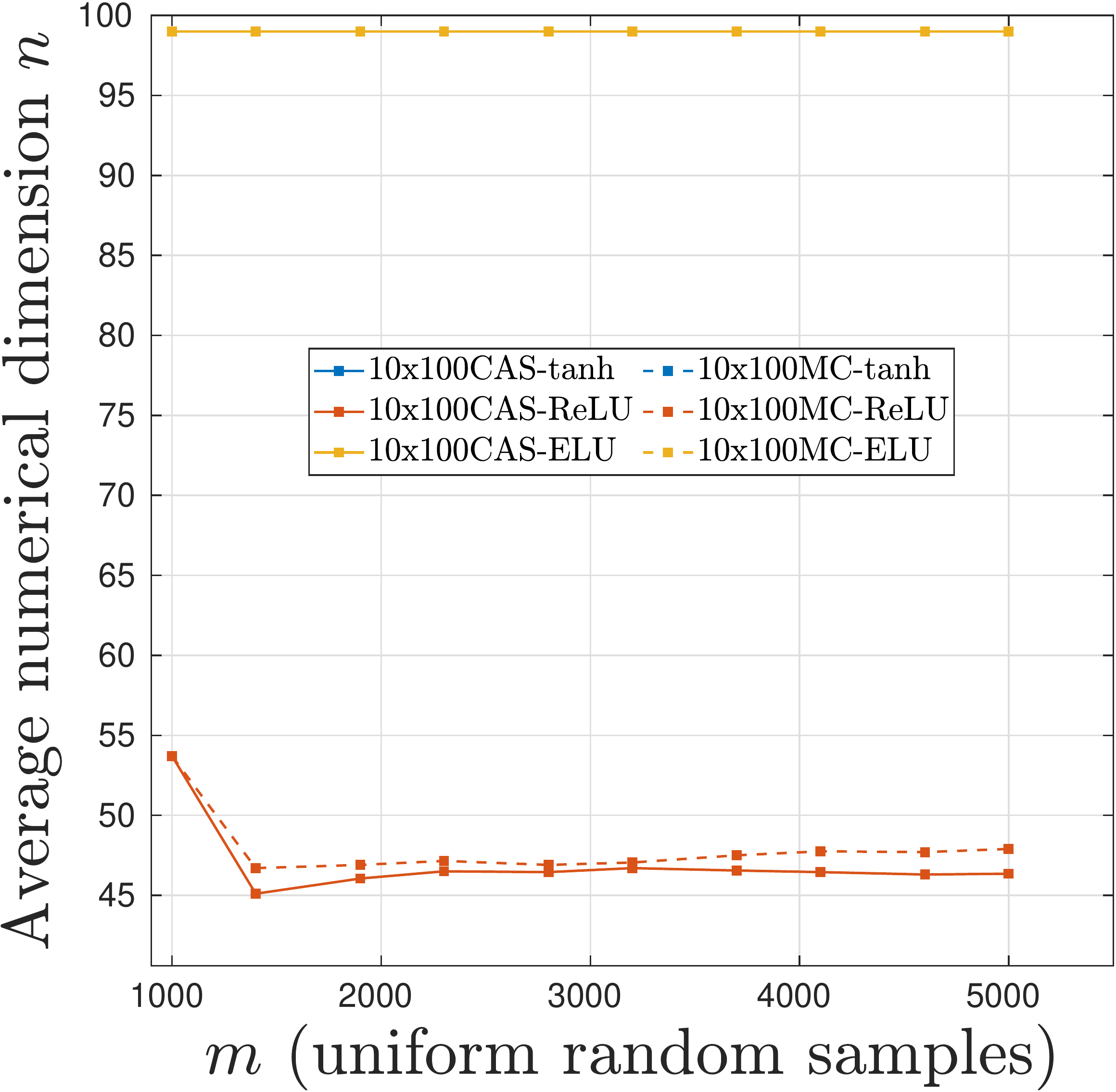} \\ 
\hspace{-0.5cm}
$(f,d)=(f_1,16)$ & $(f,d)=(f_1,16)$ & $(f,d)=(f_1,16)$ & $(f,d)=(f_1,16)$
\end{tabular}
}
\caption{Approximation of the function $f = f_1$ via MC sampling and CAS4DL in \textbf{(1st row)} $d=1$, \textbf{(2nd row)} $d=2$, \textbf{(3rd row)} $d=8$ and \textbf{(4th row)} $d=16$ dimensions. \textbf{First and third columns:} Average relative $L^2$ error vs. number of CAS4DL and MC samples used in training $\tanh$ and ELU $L\times N$ DNNs with \textbf{(first column)} $L=5$ and $N=50$, and \textbf{(third column)} $L=10$ and $N=100$. \textbf{Second and fourth columns:} Average numerical dimension $n$ vs. number of CAS4DL and MC samples used in training $\tanh$ and ELU $L\times N$ DNNs with \textbf{(second column)} $L=5$ and $N=50$, and \textbf{(fourth column)} $L=10$ and $N=100$.}
\label{fig:comp_act_L2_error_example_1}
\end{figure}

\begin{figure}[h]
\centering
{\small
\begin{tabular}{cccc} 
\hspace{-0.5cm}
\includegraphics[scale=0.11]{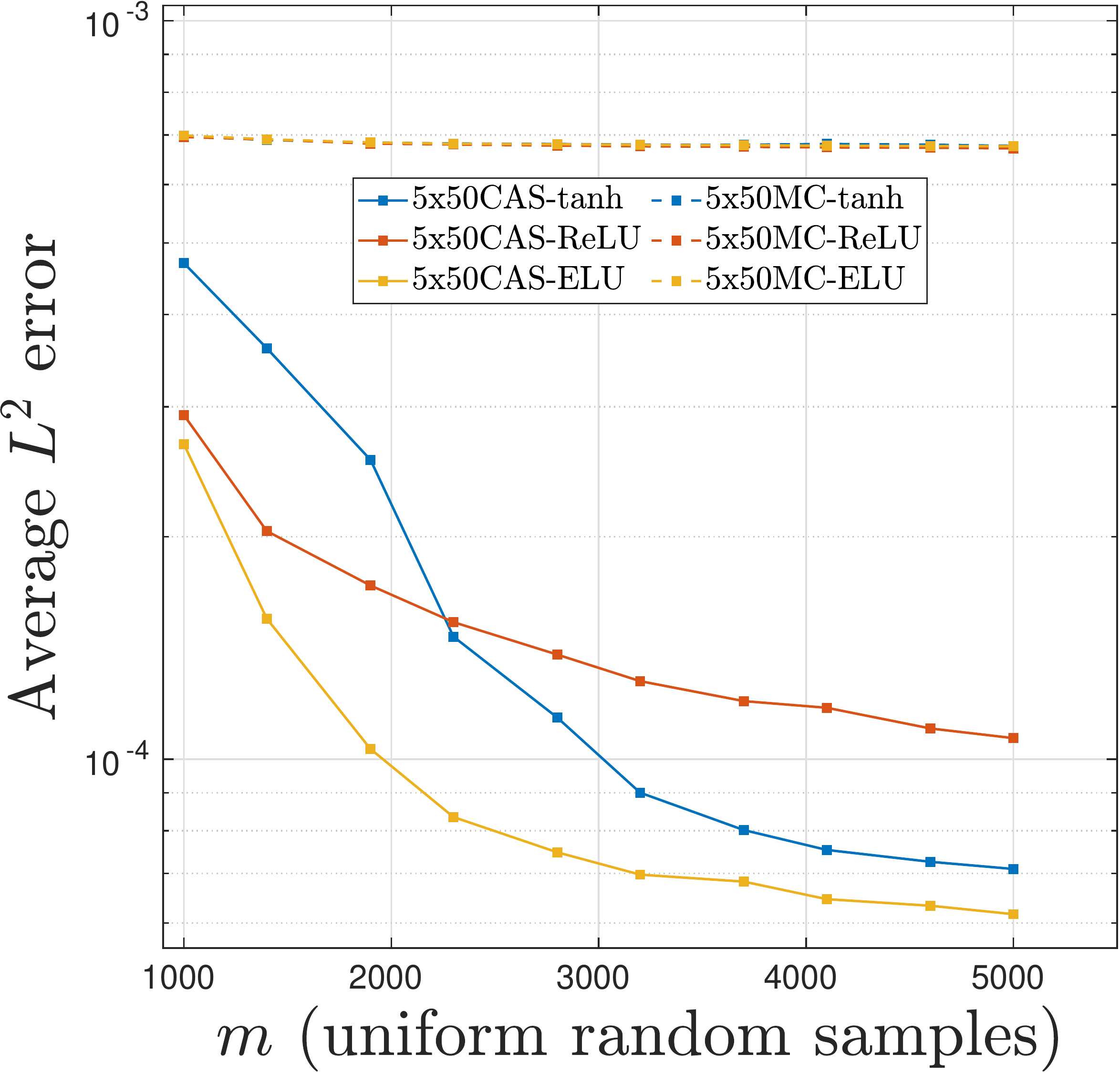} & 
\includegraphics[scale=0.11]{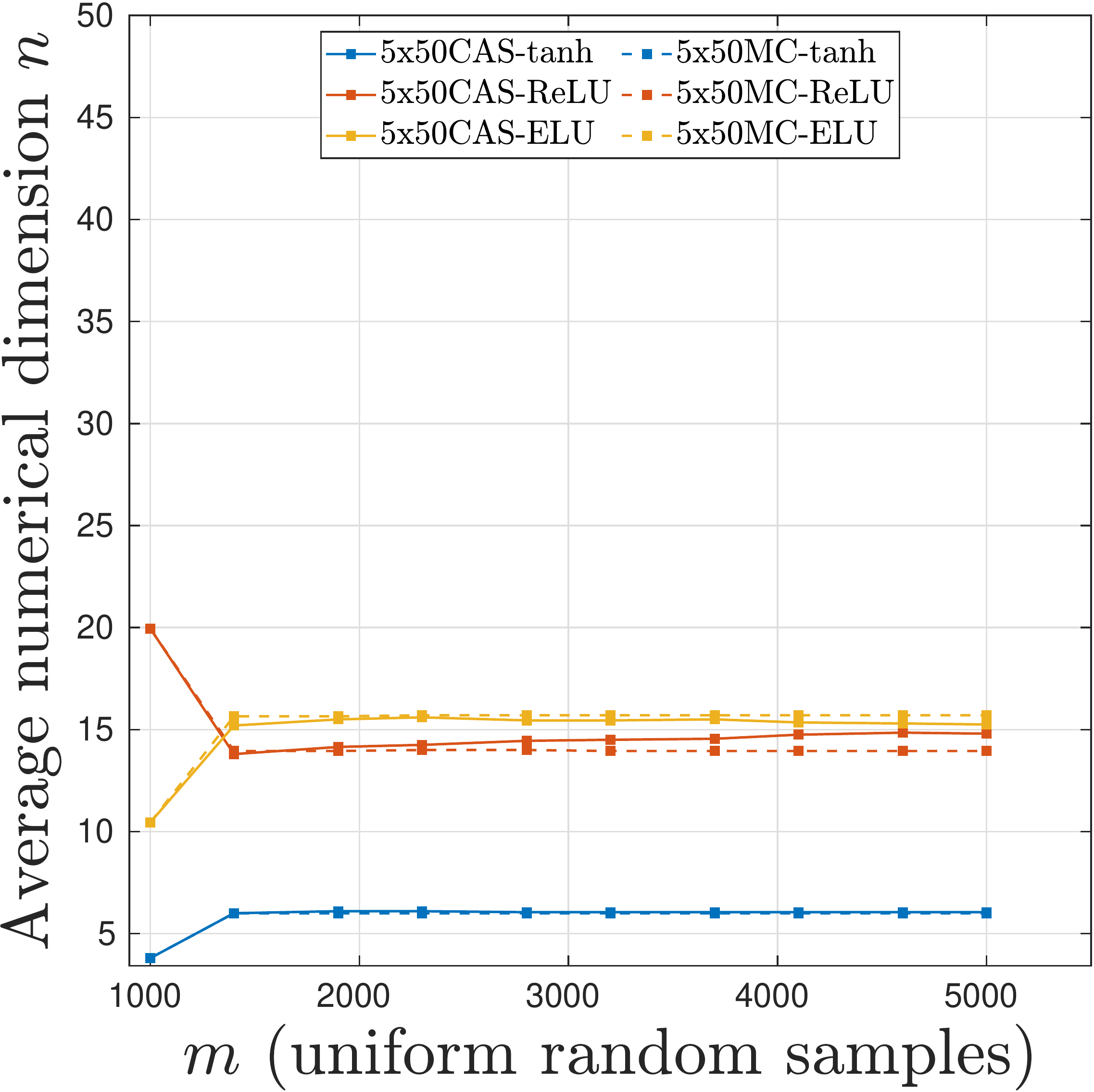} & 
\includegraphics[scale=0.11]{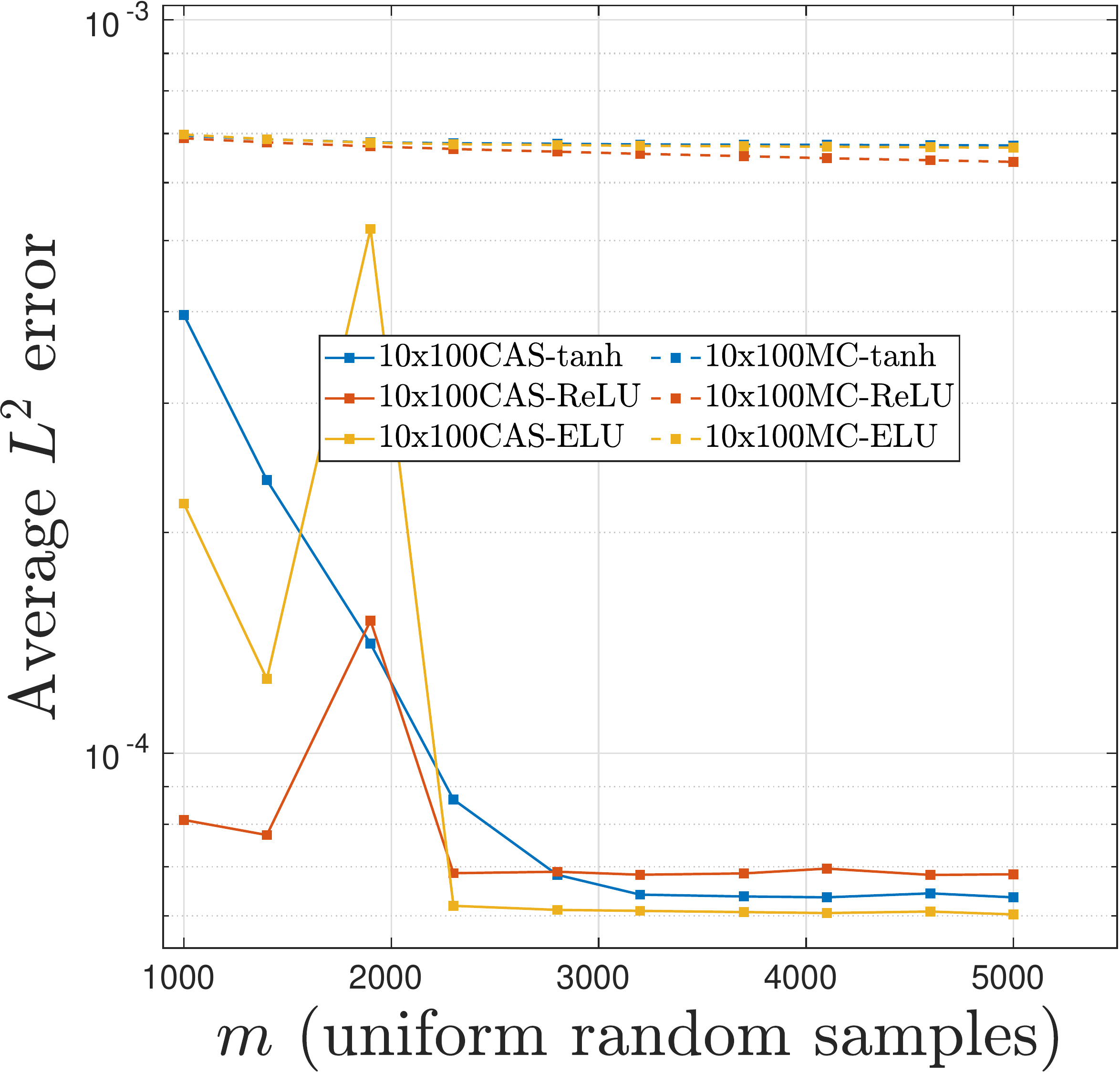} & 
\includegraphics[scale=0.11]{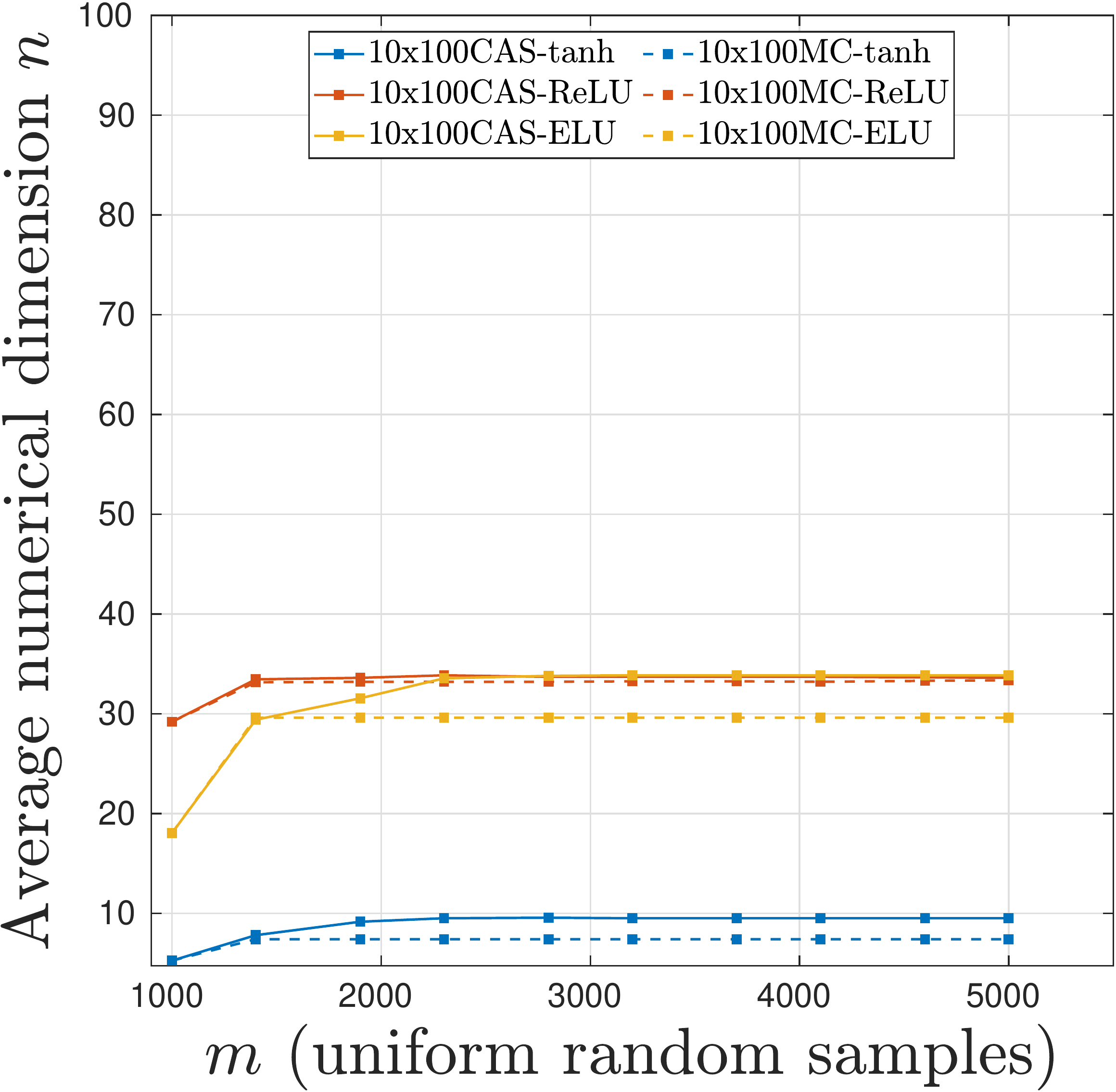} \\  
\hspace{-0.5cm}
$(f,d)=(f_2,1)$ & $(f,d)=(f_2,1)$ & $(f,d)=(f_2,1)$ & $(f,d)=(f_2,1)$ \\
\hspace{-0.5cm}
\includegraphics[scale=0.11]{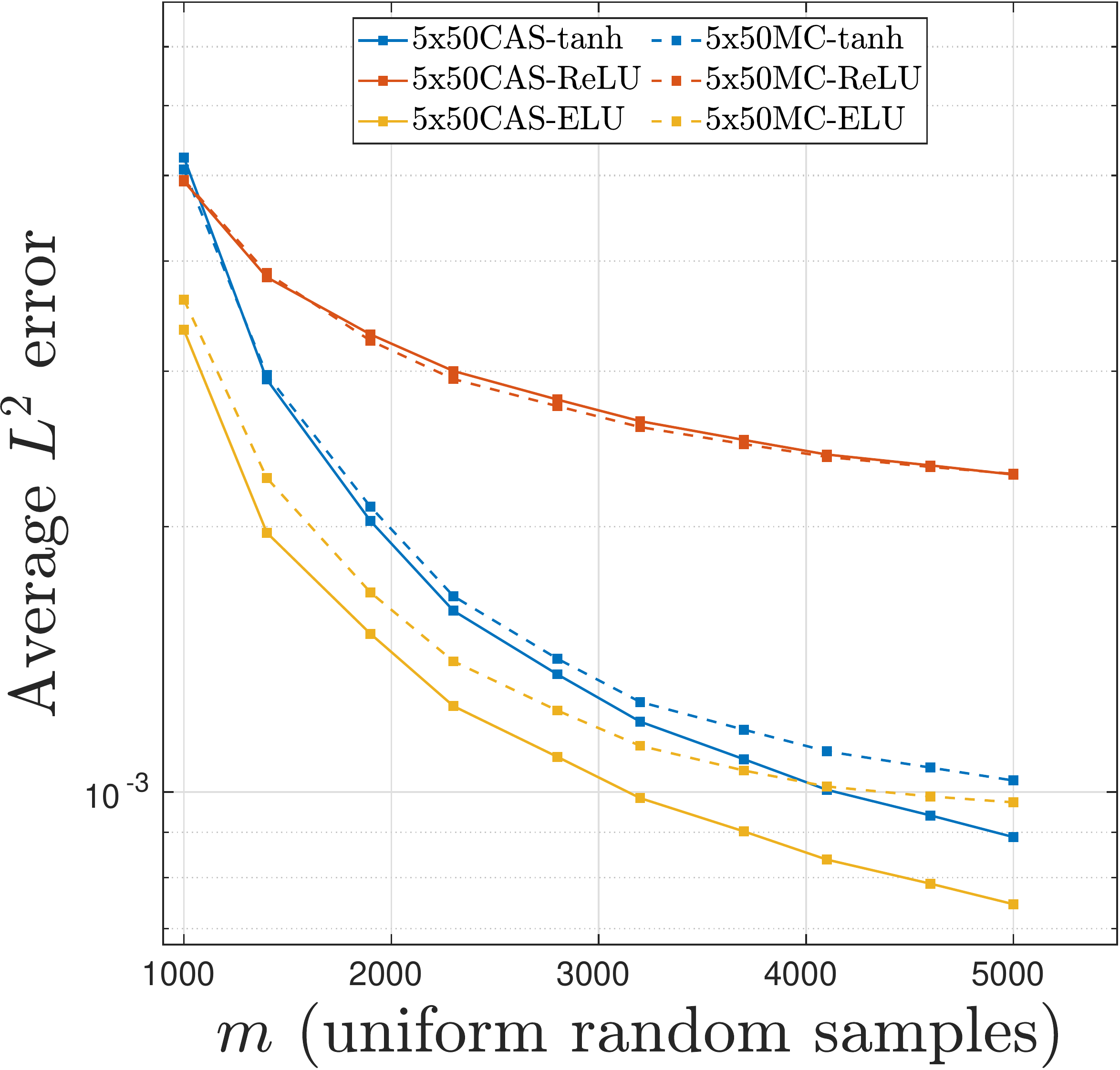} & 
\includegraphics[scale=0.11]{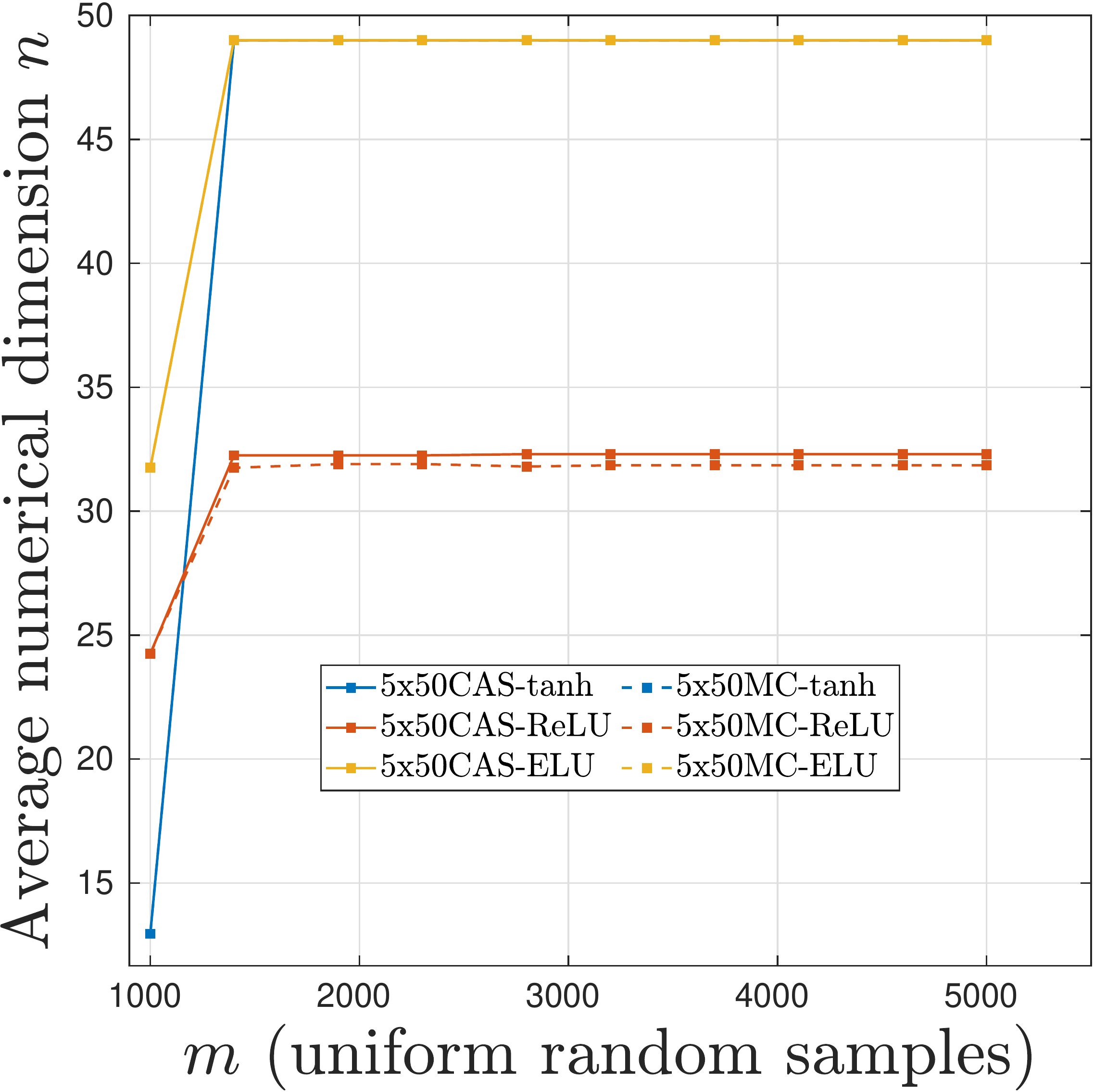} & 
\includegraphics[scale=0.11]{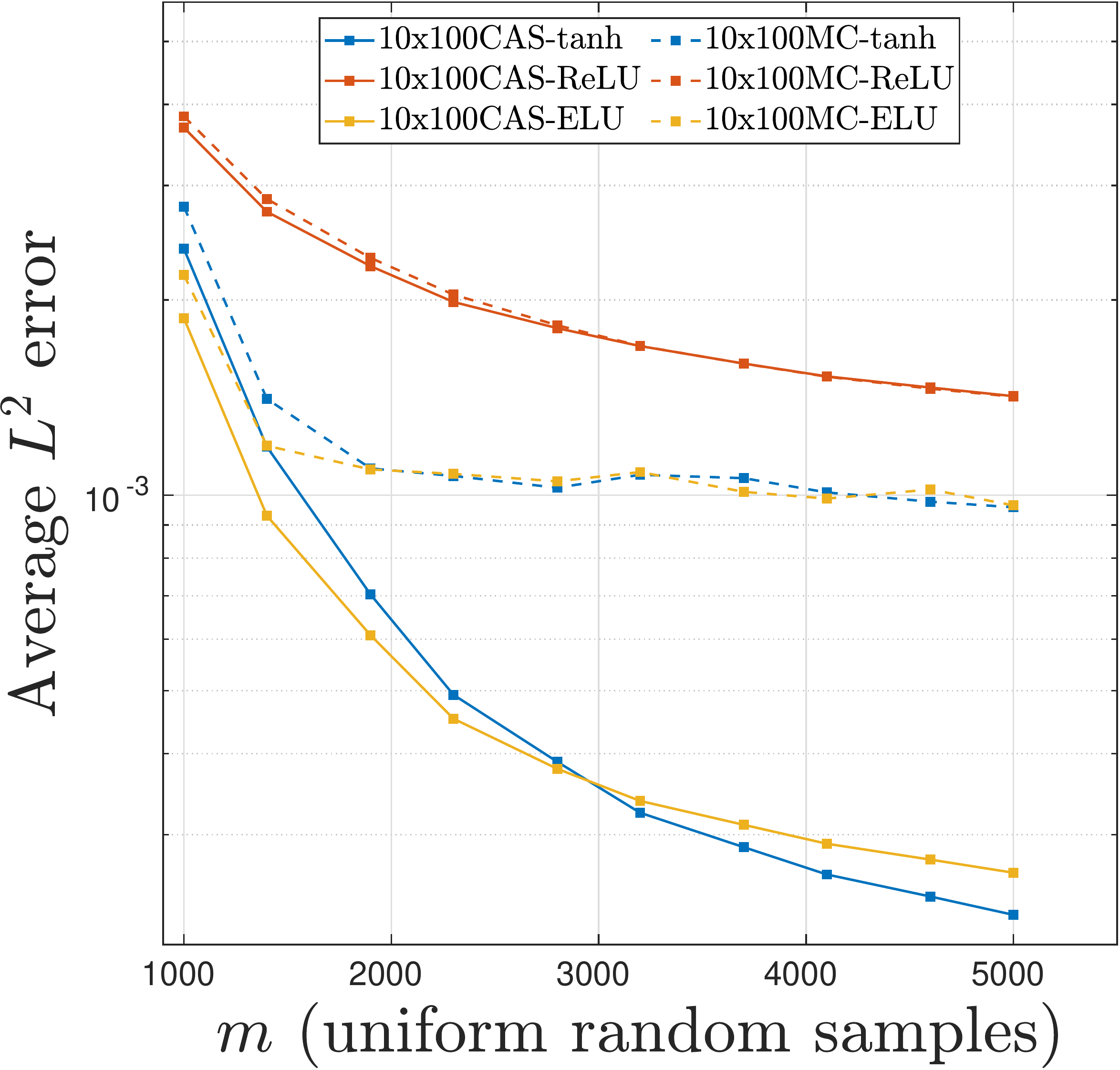} & 
\includegraphics[scale=0.11]{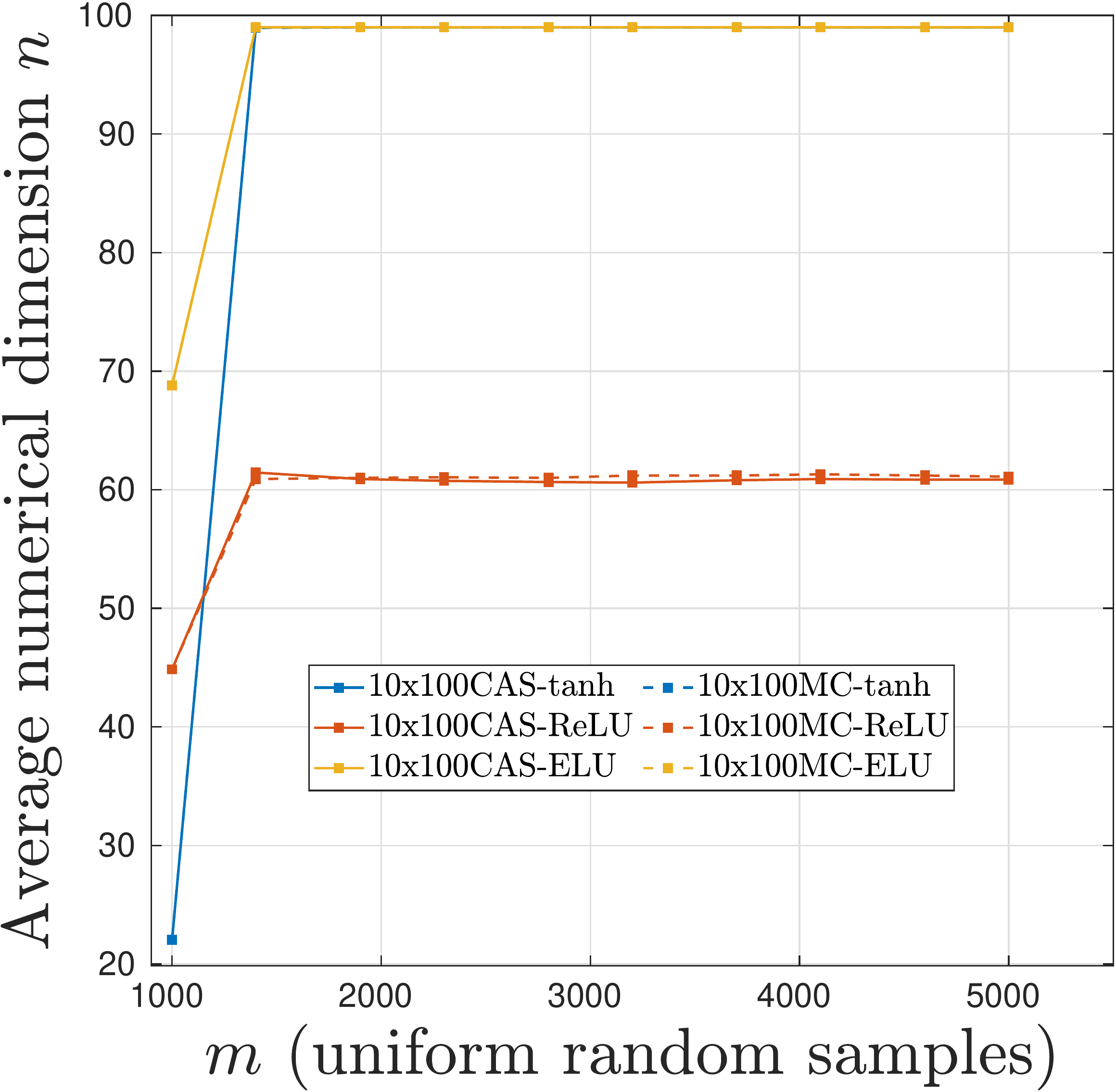} \\ 
\hspace{-0.5cm}
$(f,d)=(f_2,2)$ & $(f,d)=(f_2,2)$ &  $(f,d)=(f_2,2)$ & $(f,d)=(f_2,2)$ \\
\includegraphics[scale=0.11]{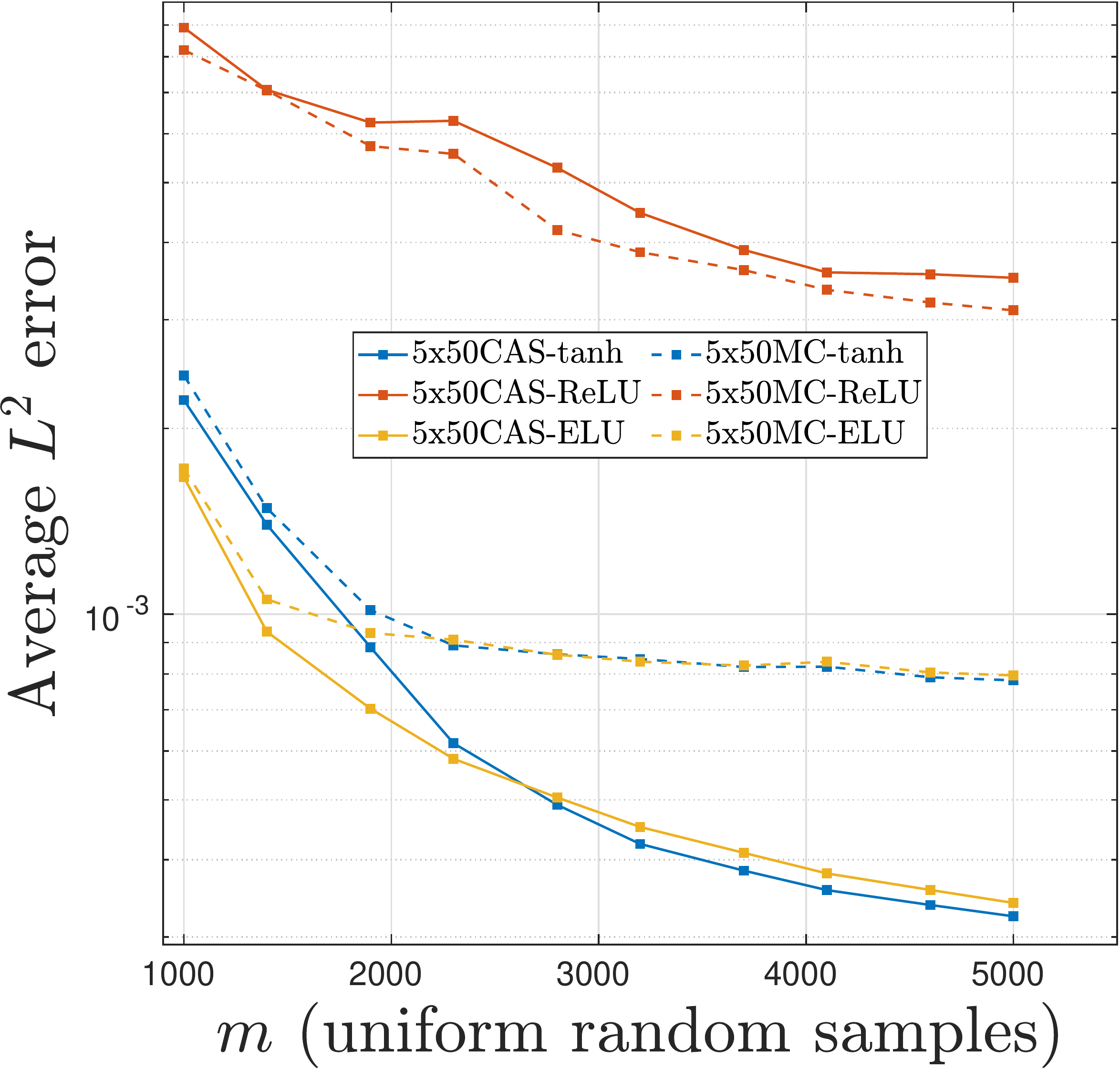} & 
\includegraphics[scale=0.11]{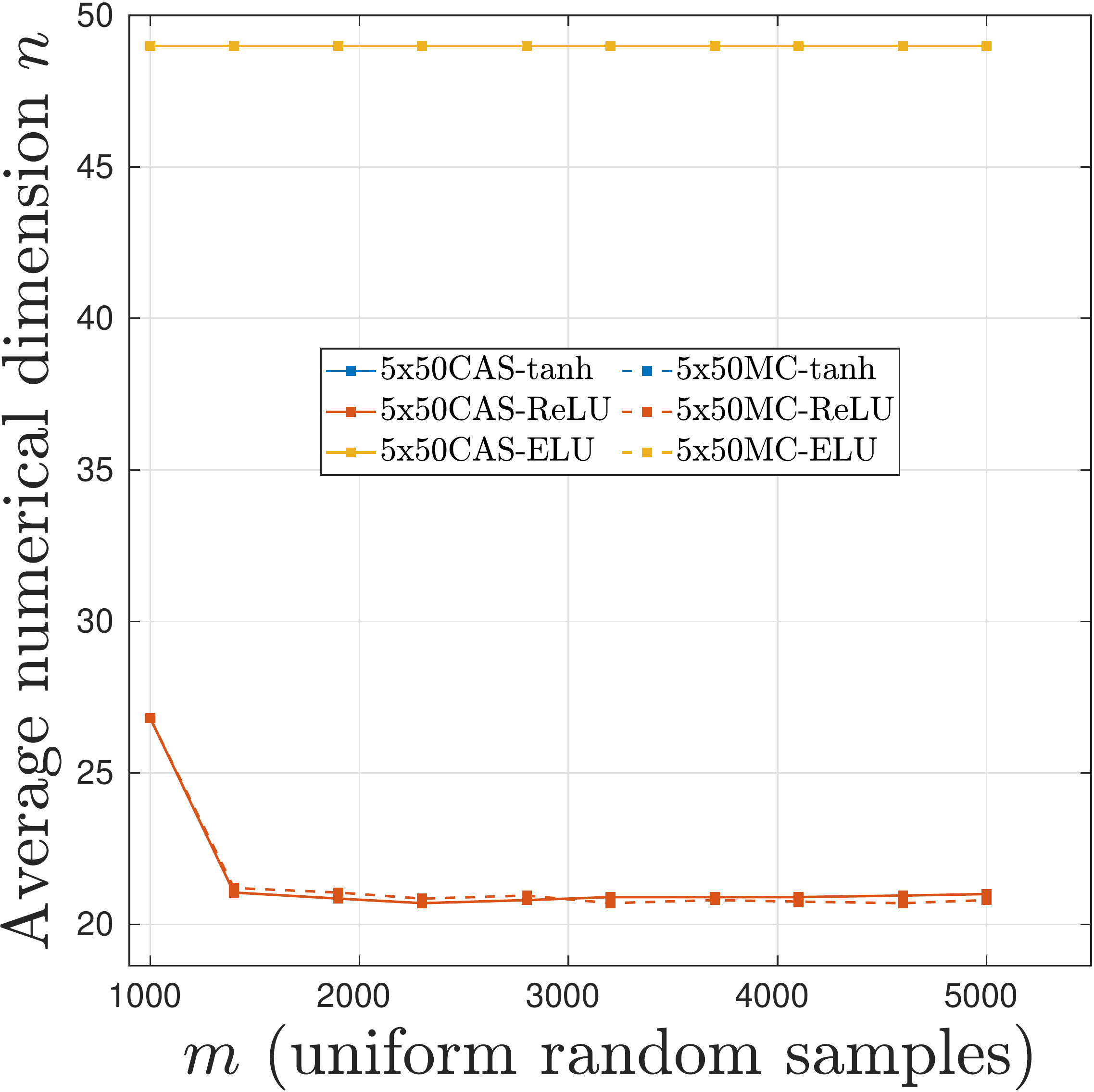} & 
\includegraphics[scale=0.11]{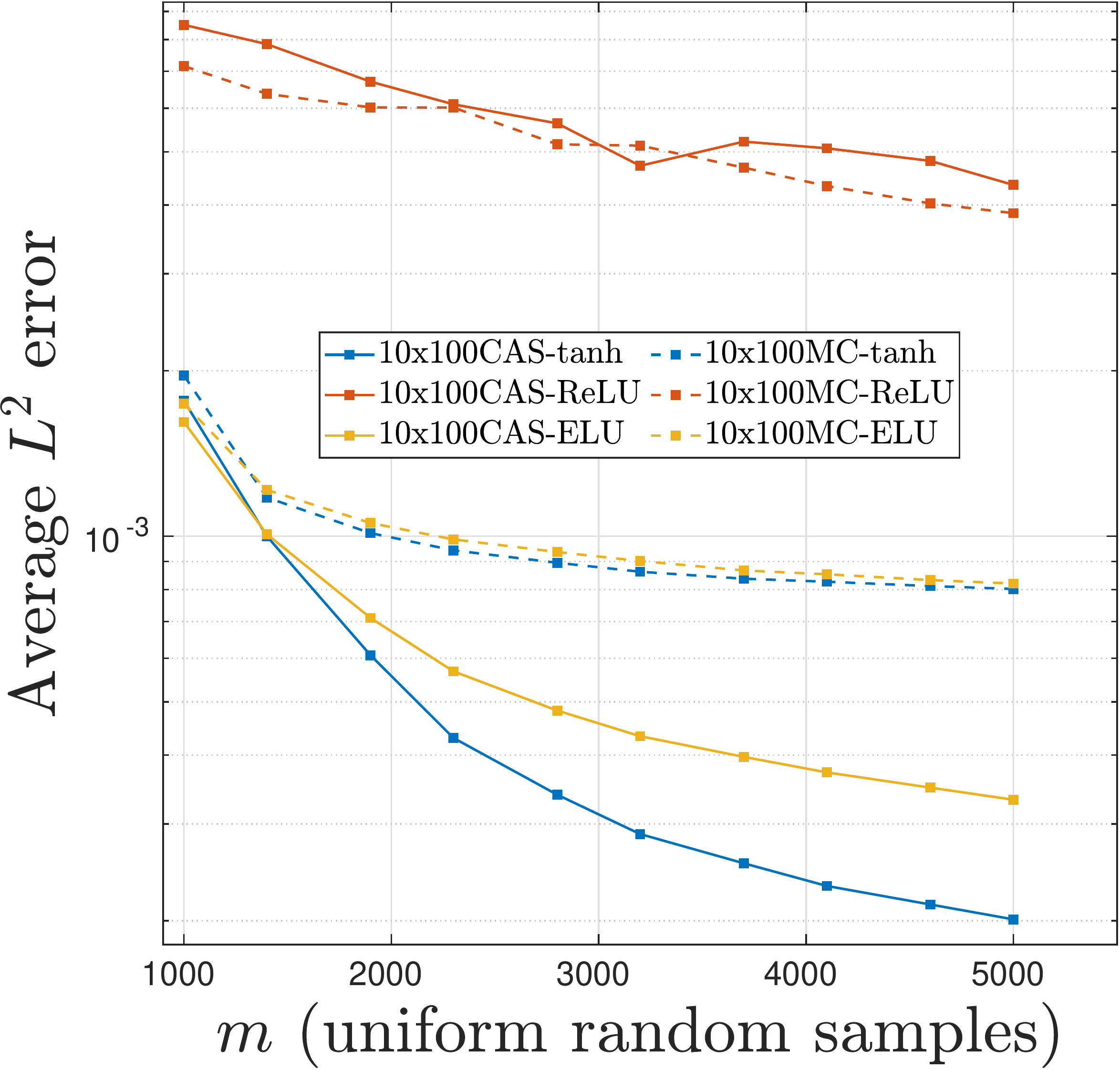} & 
\includegraphics[scale=0.11]{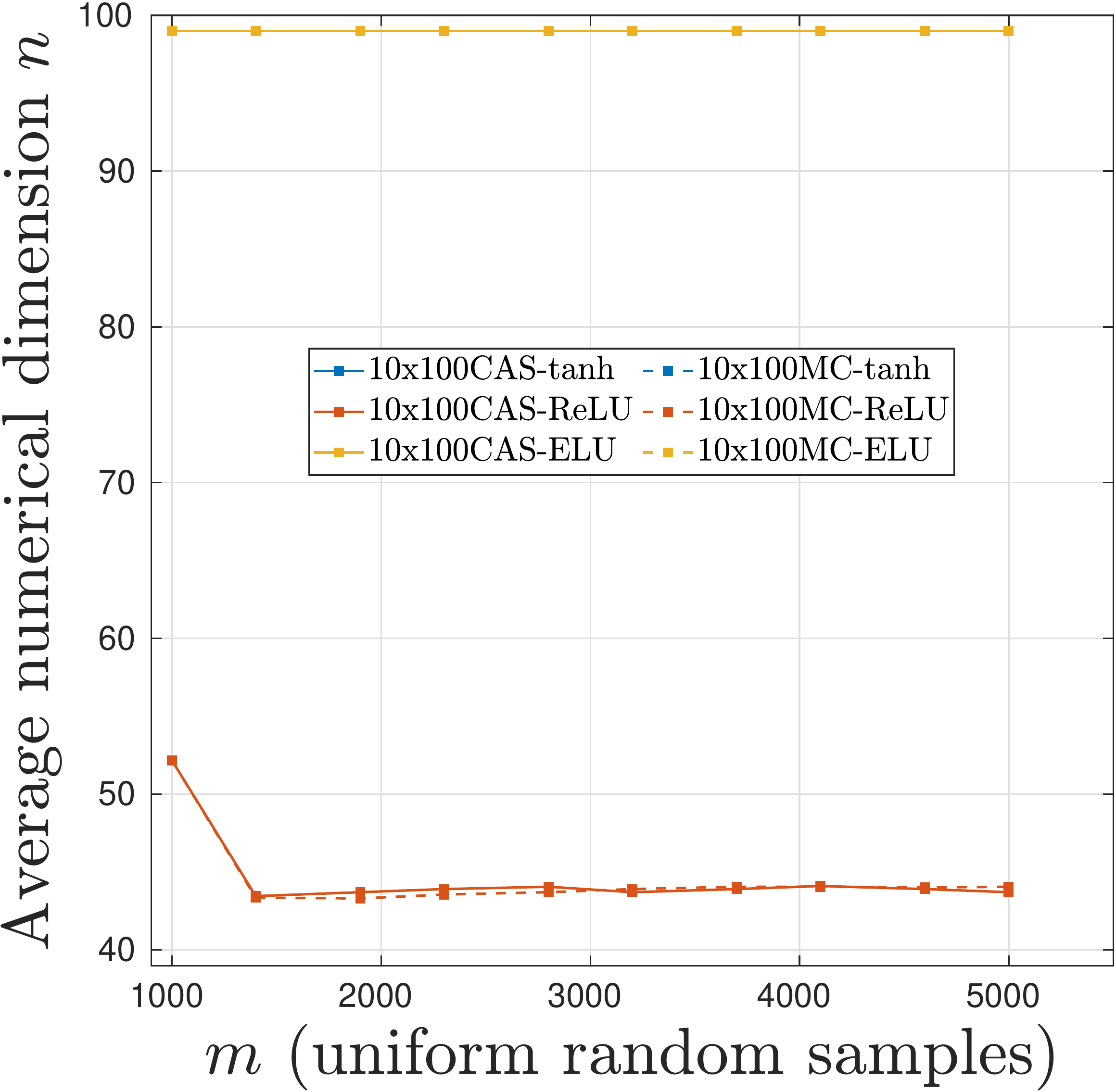} \\ 
\hspace{-0.5cm}
$(f,d)=(f_2,8)$ & $(f,d)=(f_2,8)$ & $(f,d)=(f_2,8)$ & $(f,d)=(f_2,8)$\\
\hspace{-0.5cm}
\includegraphics[scale=0.11]{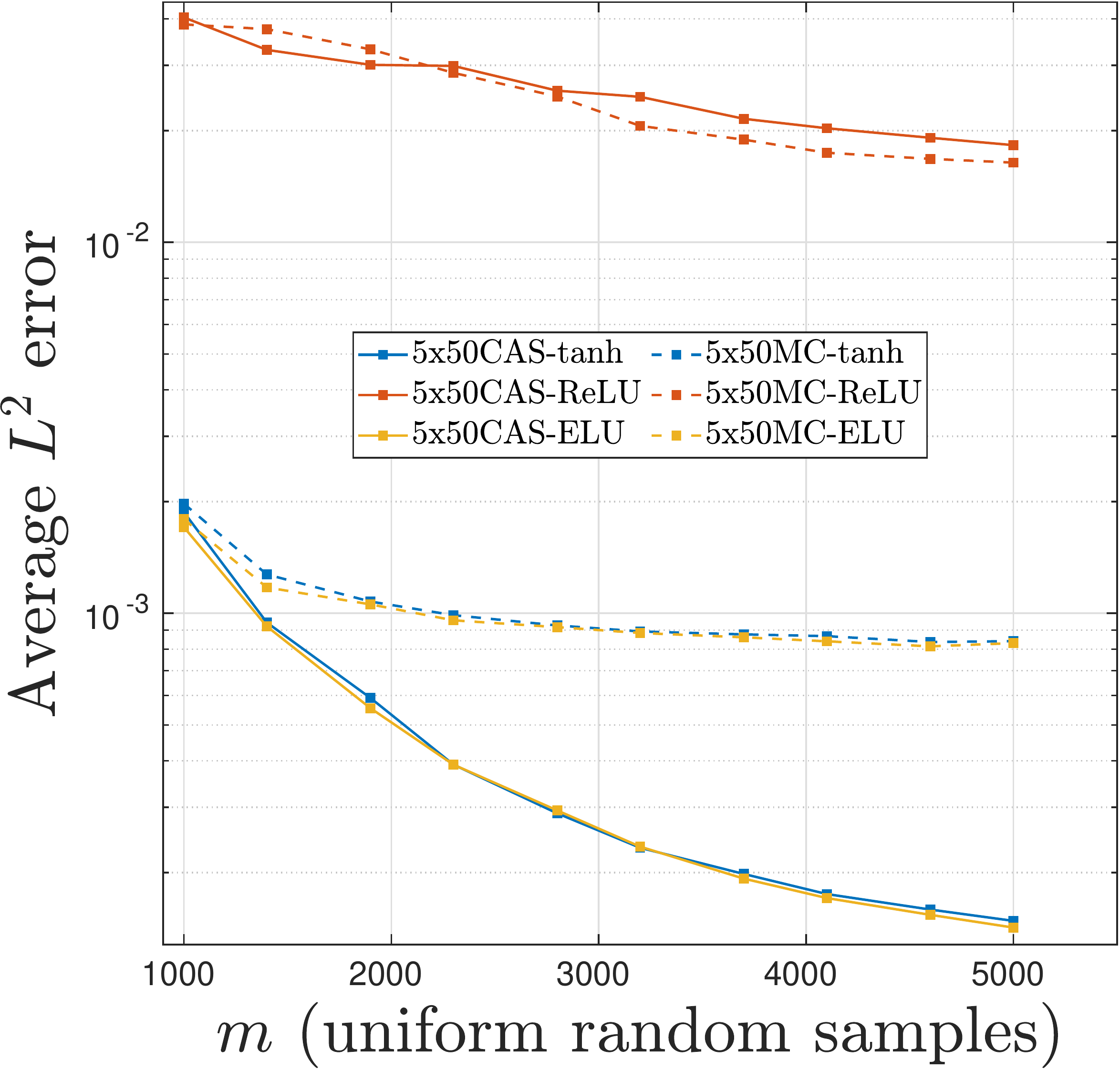} & 
\includegraphics[scale=0.11]{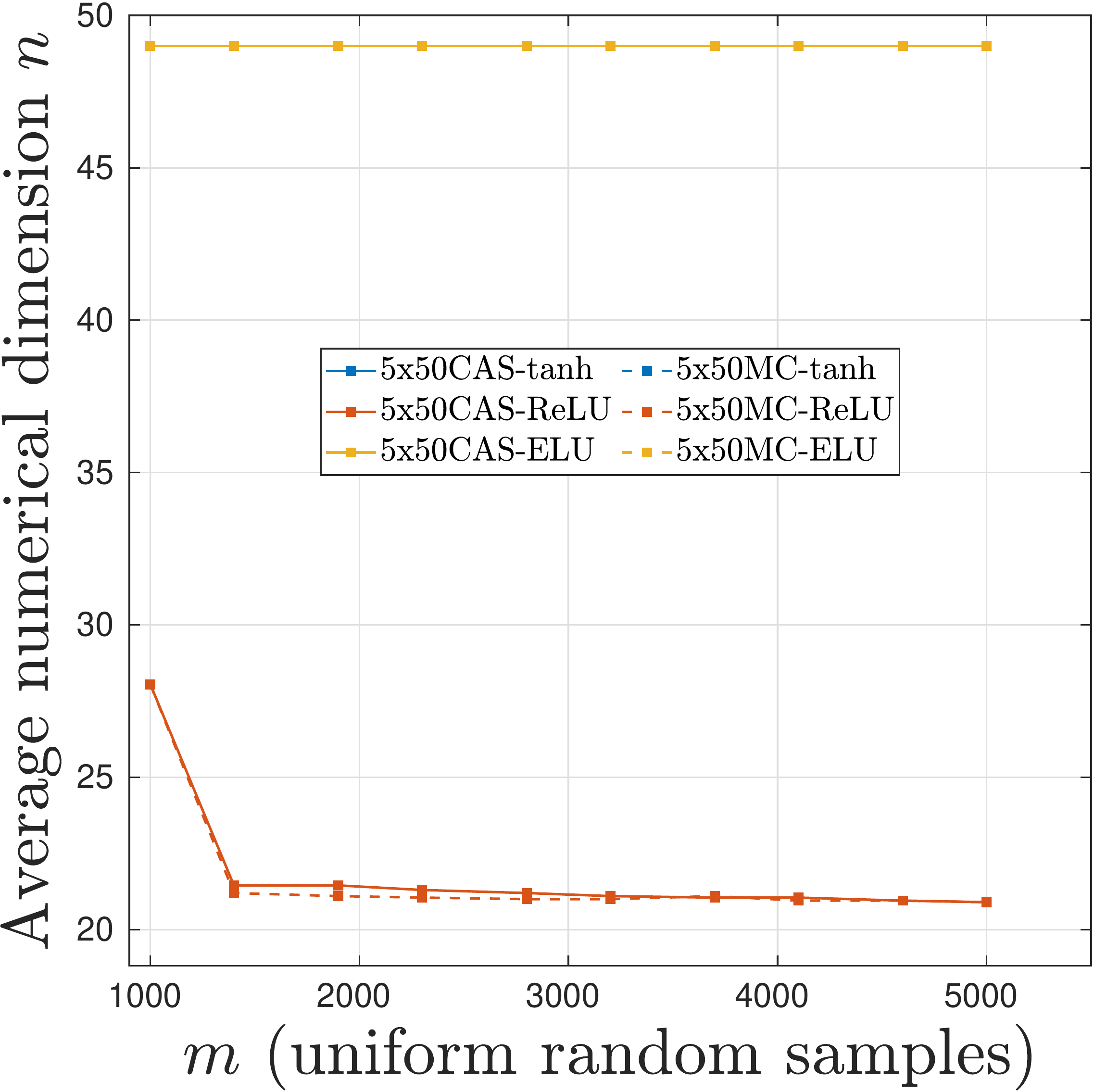} & 
\includegraphics[scale=0.11]{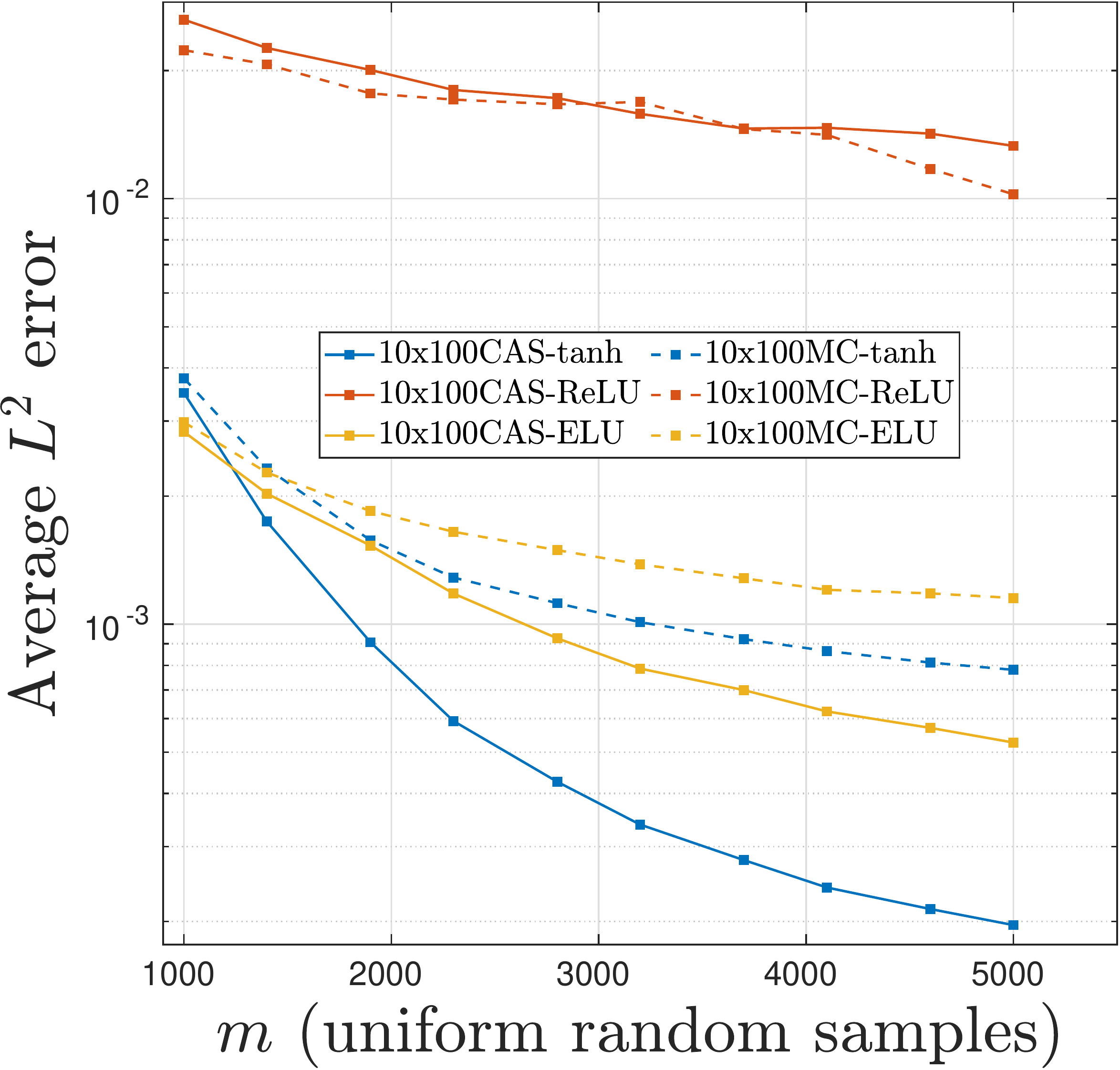} & 
\includegraphics[scale=0.11]{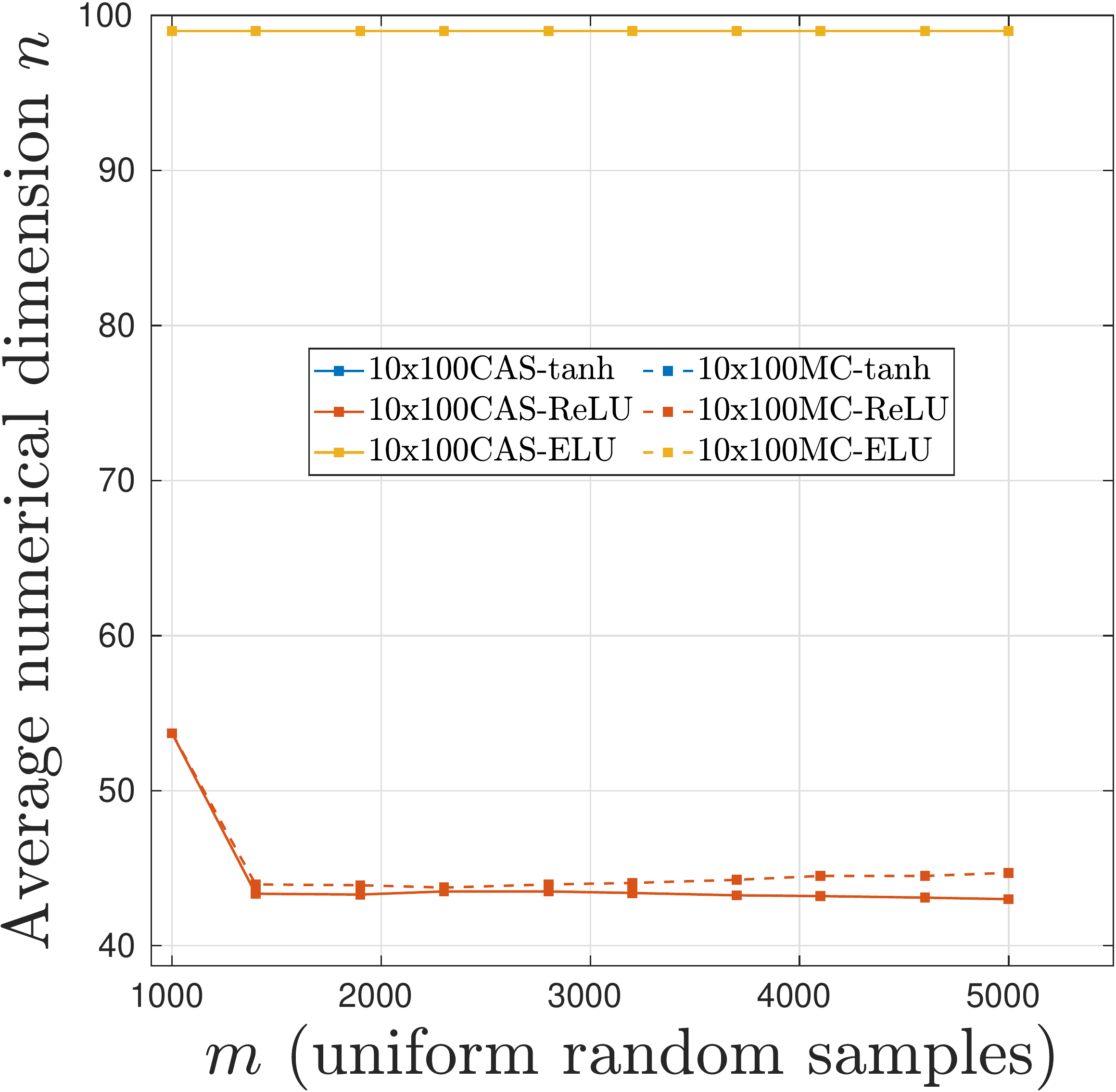} \\ 
\hspace{-0.5cm}
$(f,d)=(f_2,16)$ & $(f,d)=(f_2,16)$ & $(f,d)=(f_2,16)$ & $(f,d)=(f_2,16)$  
\end{tabular}
}
\caption{The same as Figure \ref{fig:comp_act_L2_error_example_1}, except with $f = f_2$.}
\label{fig:comp_act_L2_error_example_2}
\end{figure}

\begin{figure}[h]
\centering
{\small
\begin{tabular}{cccc} 
\hspace{-0.5cm}
\includegraphics[scale=0.11]{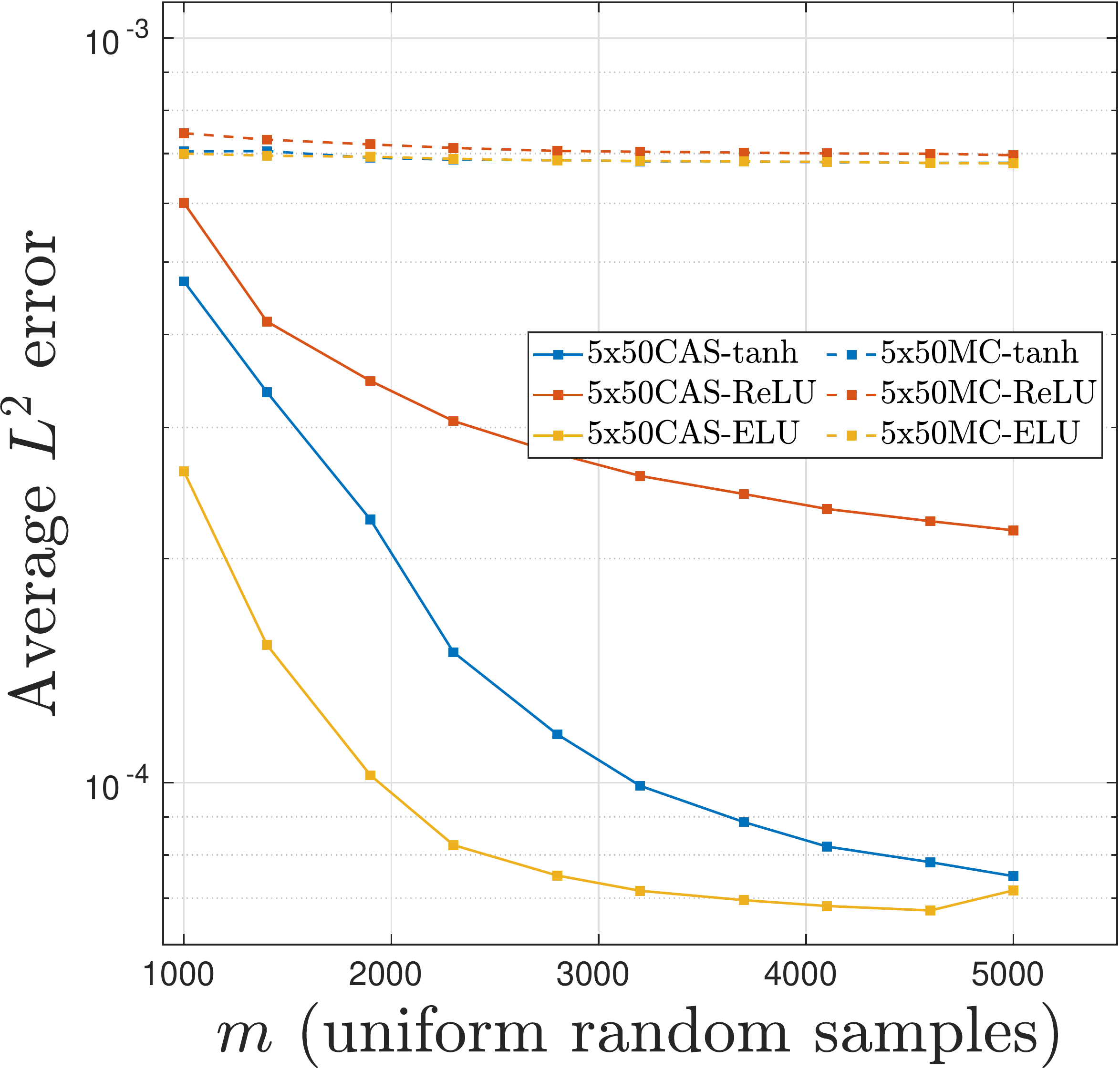} & 
\includegraphics[scale=0.11]{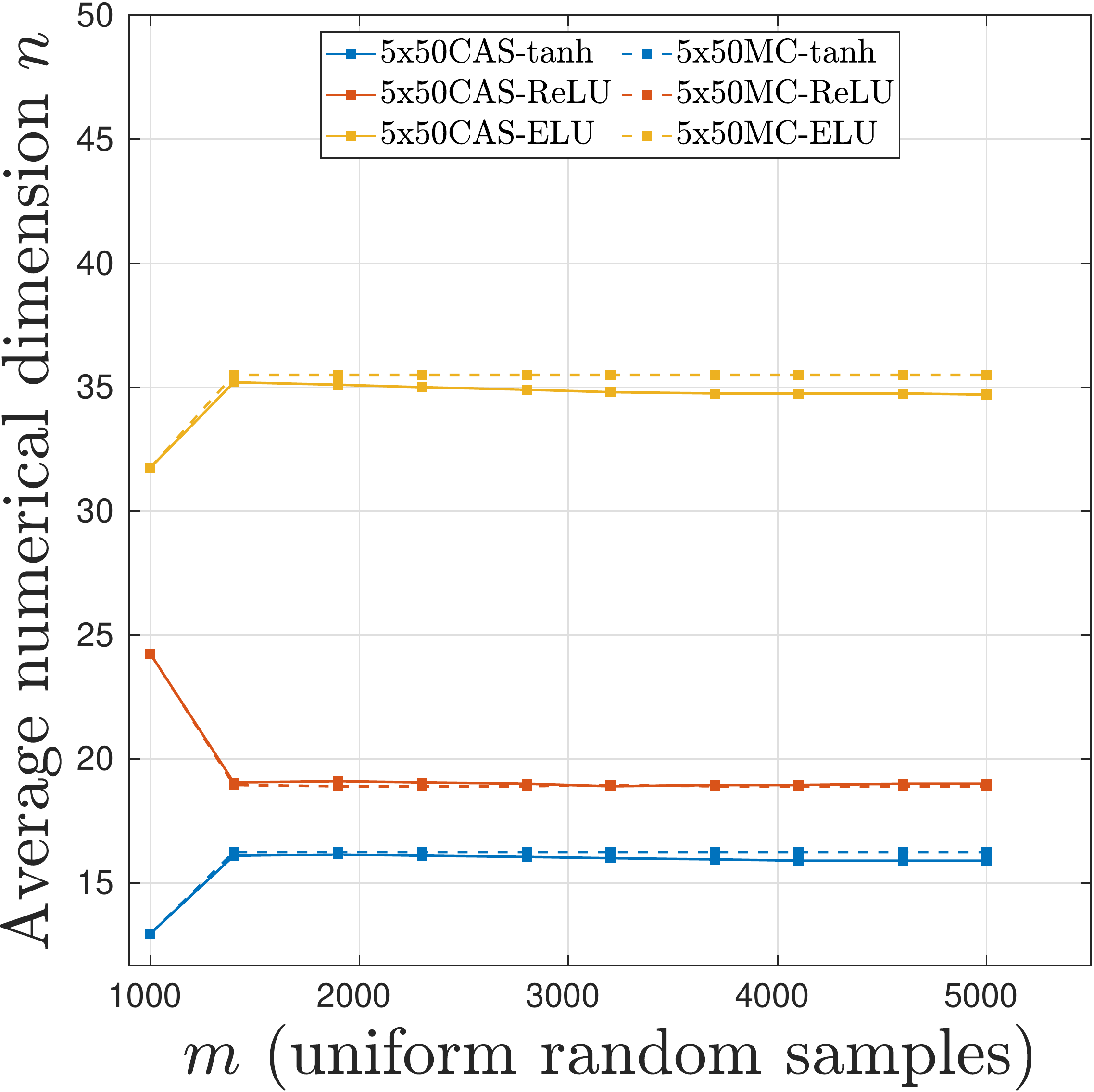} & 
\includegraphics[scale=0.11]{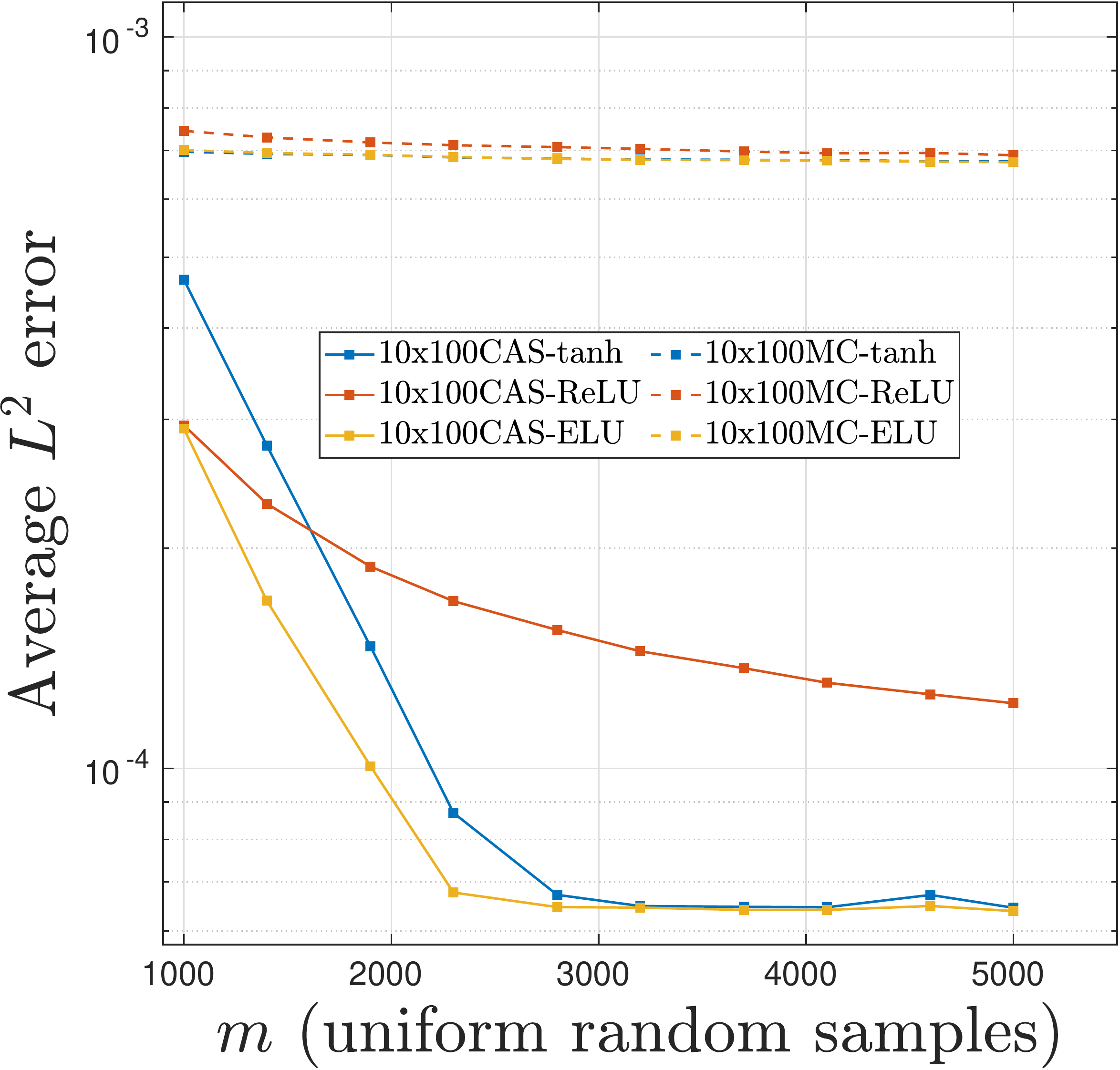} & 
\includegraphics[scale=0.11]{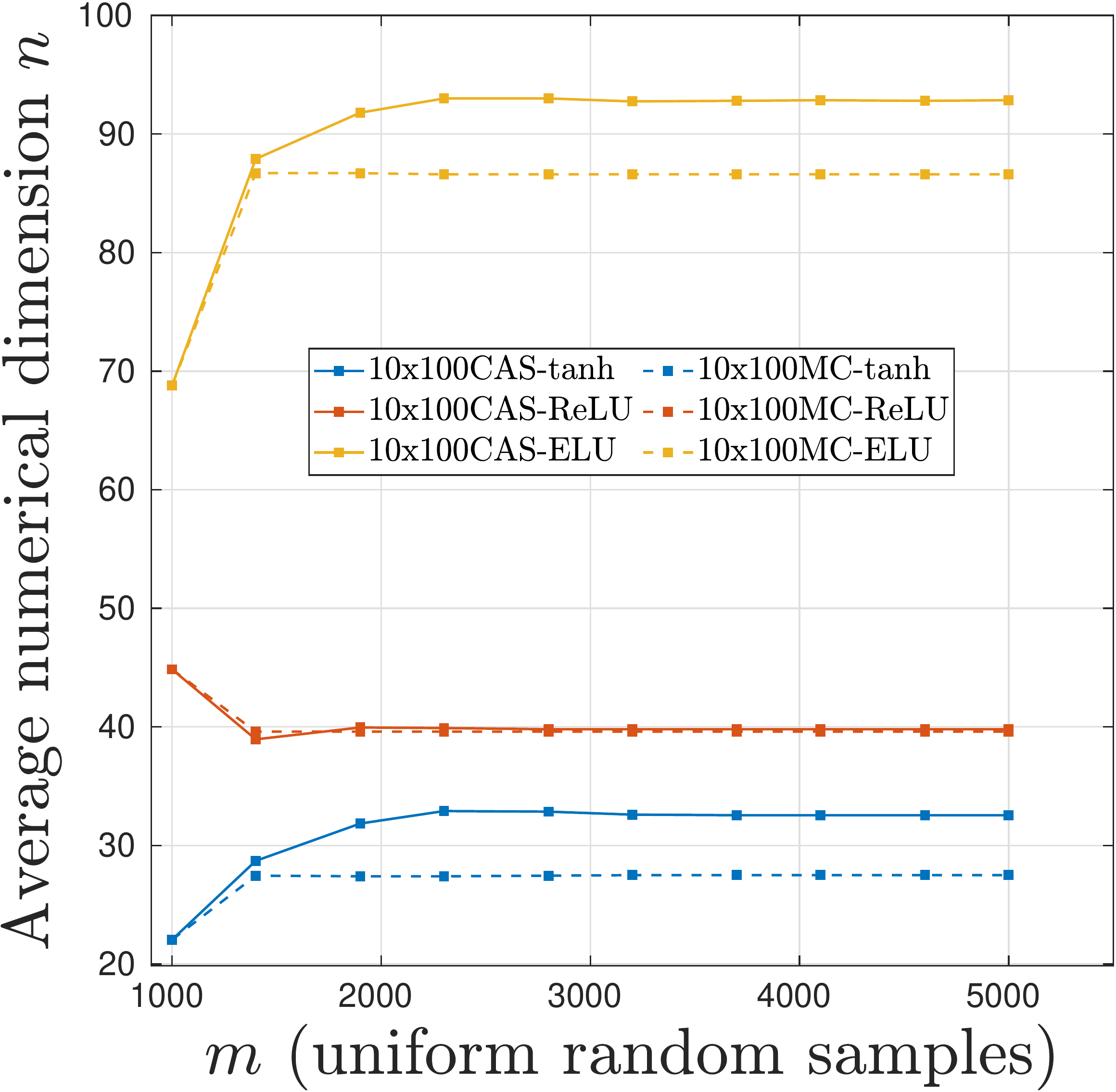} \\ 
\hspace{-0.5cm}
$(f,d)=(f_3,2)$ & $(f,d)=(f_3,2)$ &  $(f,d)=(f_3,2)$ & $(f,d)=(f_3,2)$ \\
\hspace{-0.5cm}
\includegraphics[scale=0.11]{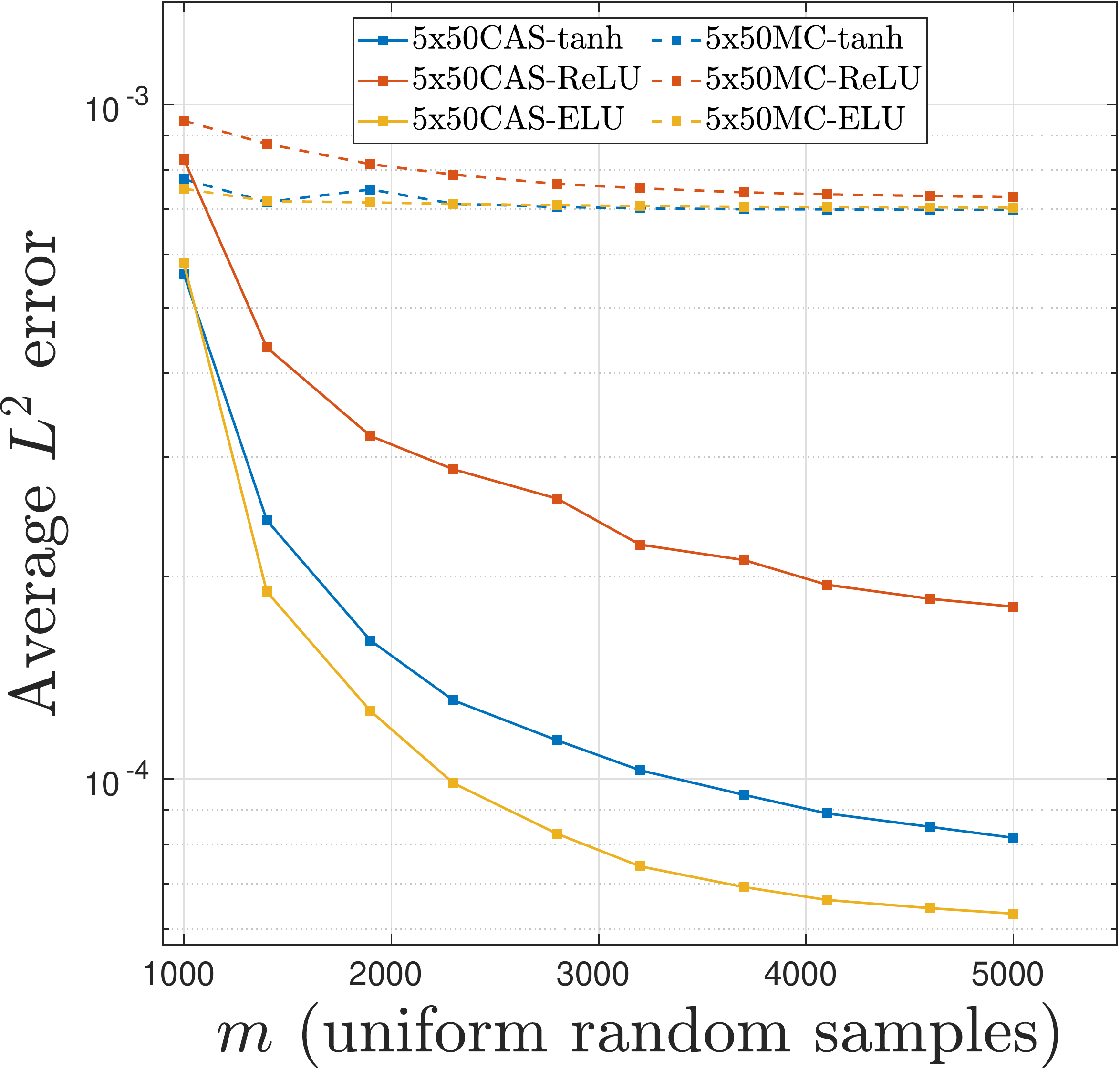} & 
\includegraphics[scale=0.11]{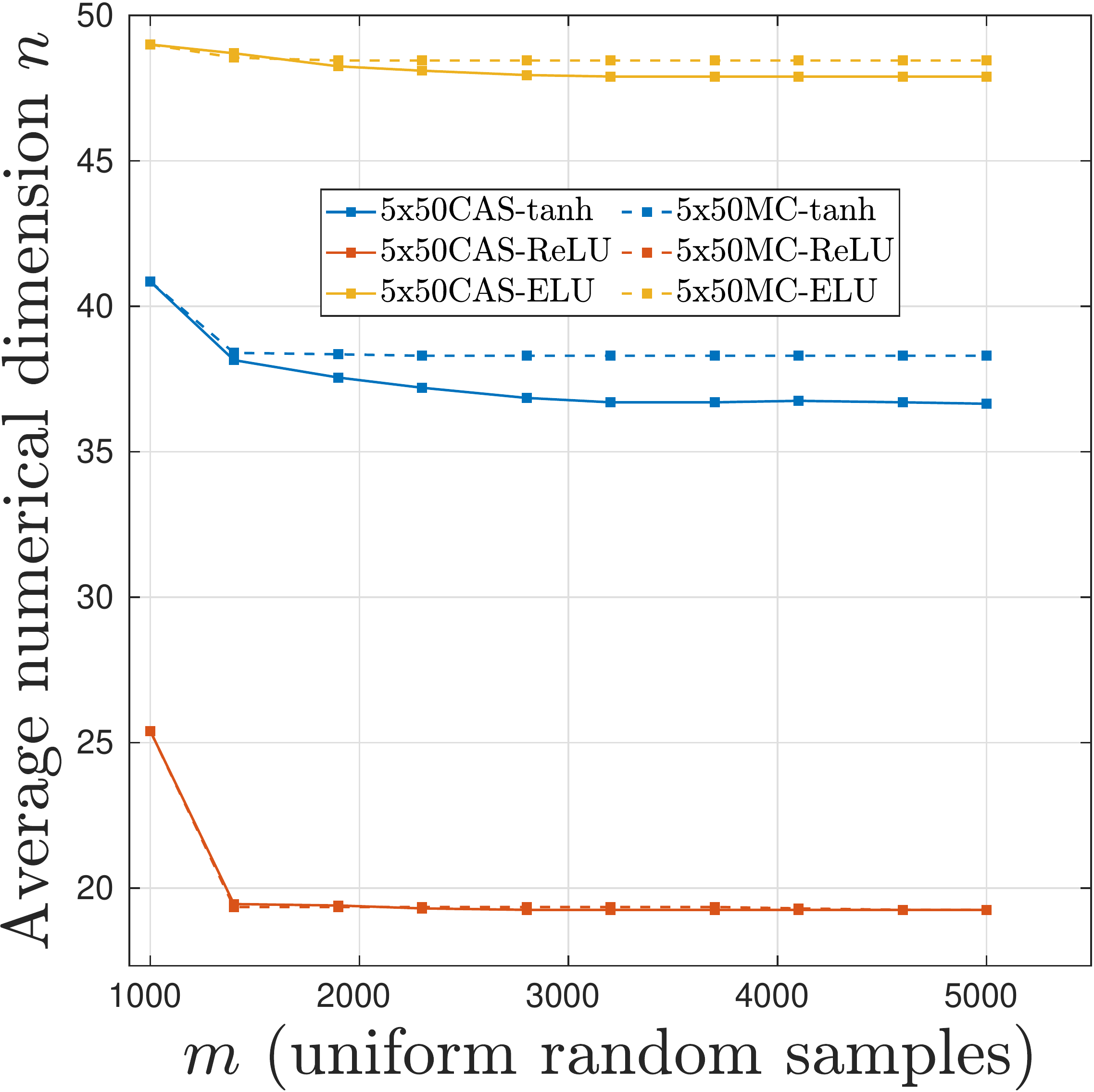} & 
\includegraphics[scale=0.11]{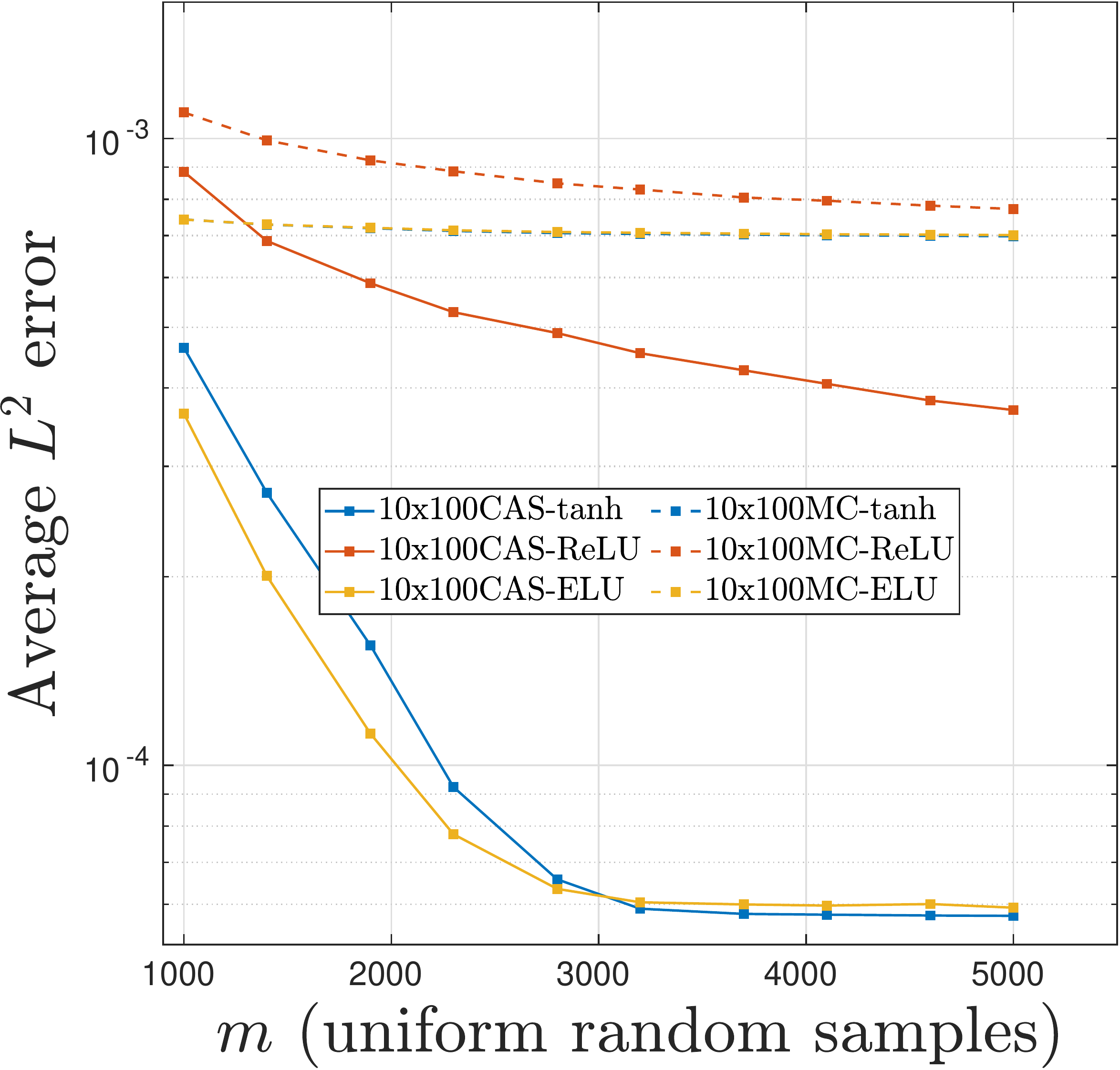} & 
\includegraphics[scale=0.11]{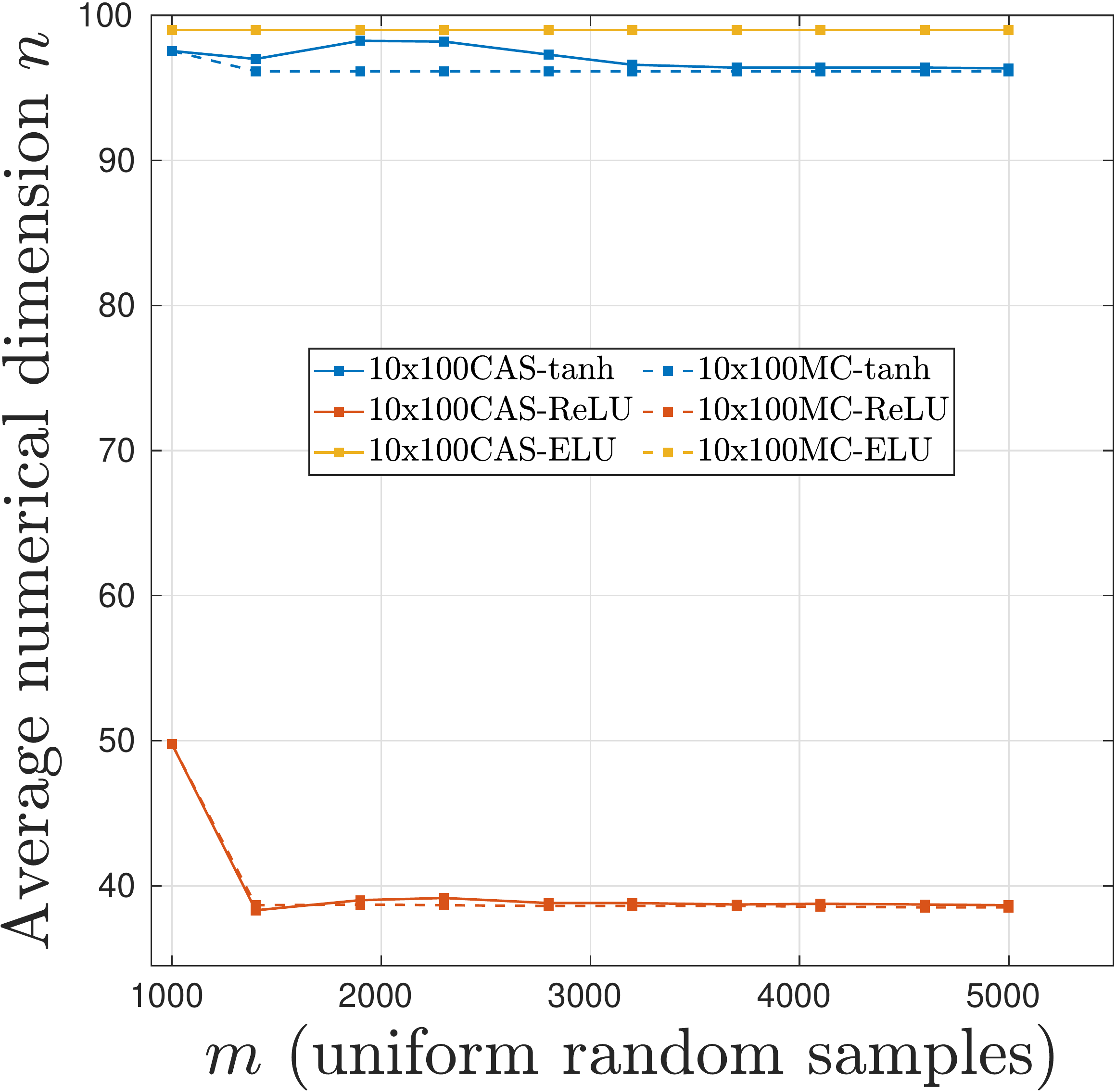} \\ 
\hspace{-0.5cm}
$(f,d)=(f_3,4)$ & $(f,d)=(f_3,4)$ & $(f,d)=(f_3,4)$ & $(f,d)=(f_3,4)$ \\
\hspace{-0.5cm}
\includegraphics[scale=0.11]{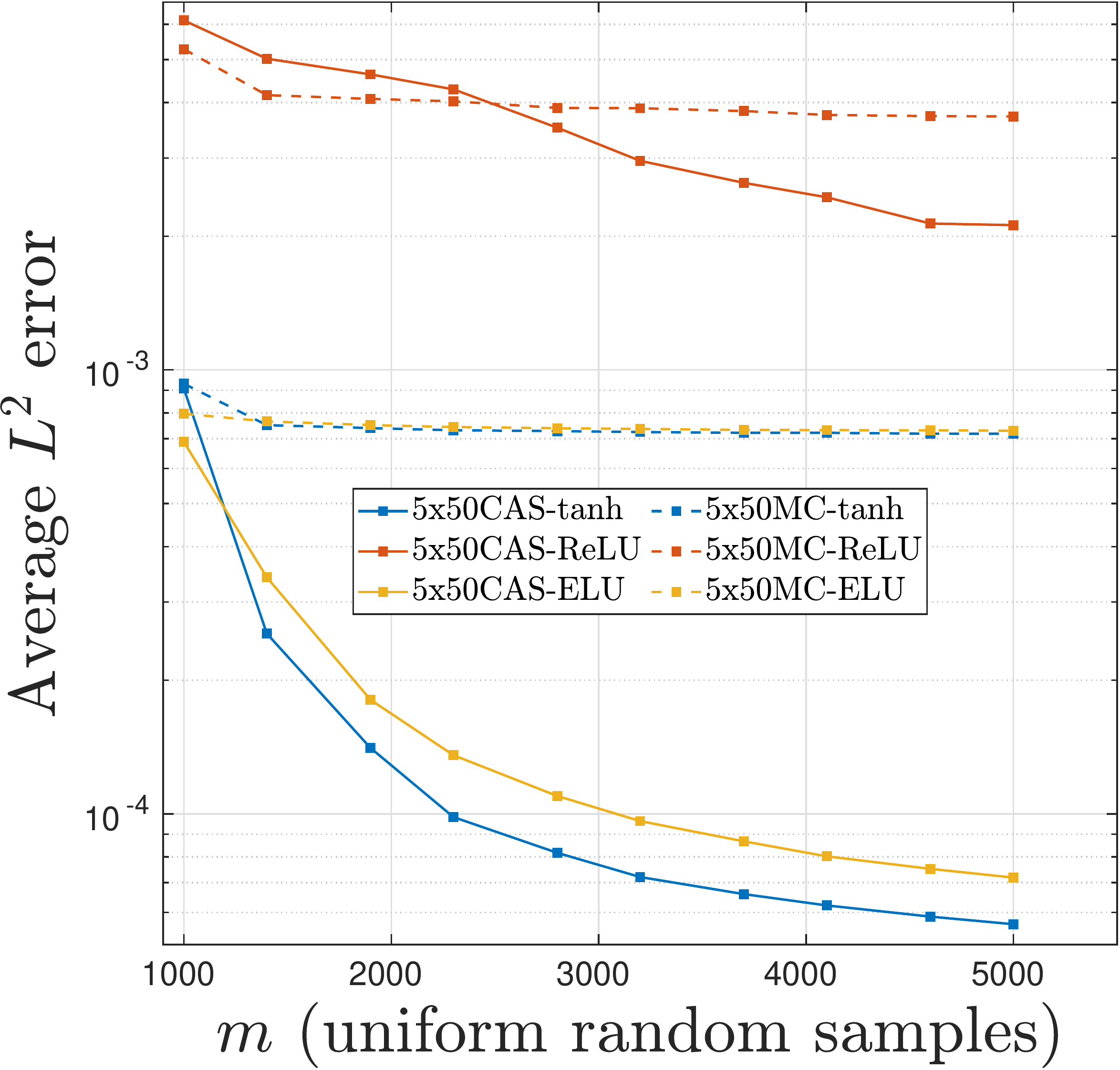} & 
\includegraphics[scale=0.11]{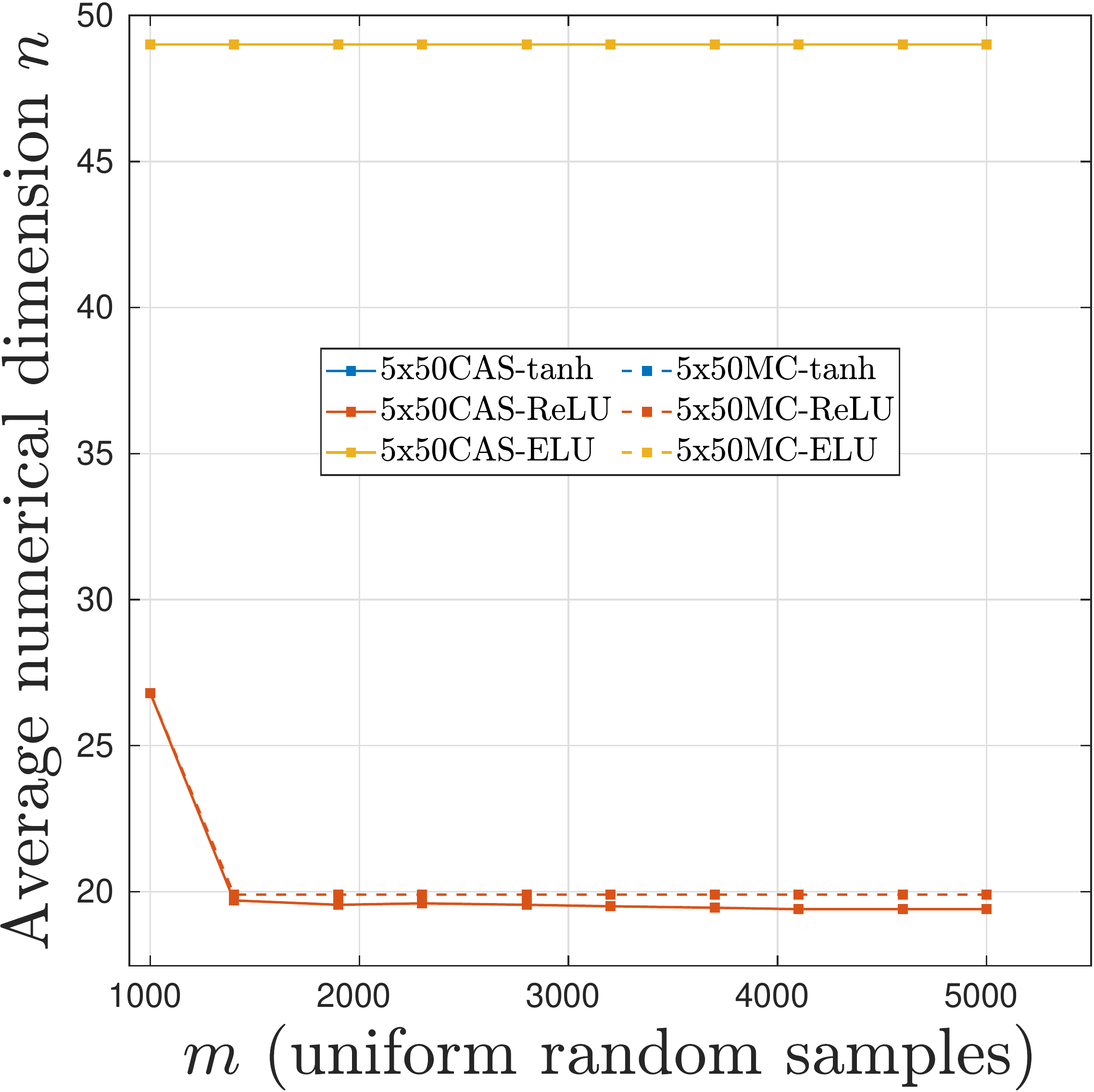} & 
\includegraphics[scale=0.11]{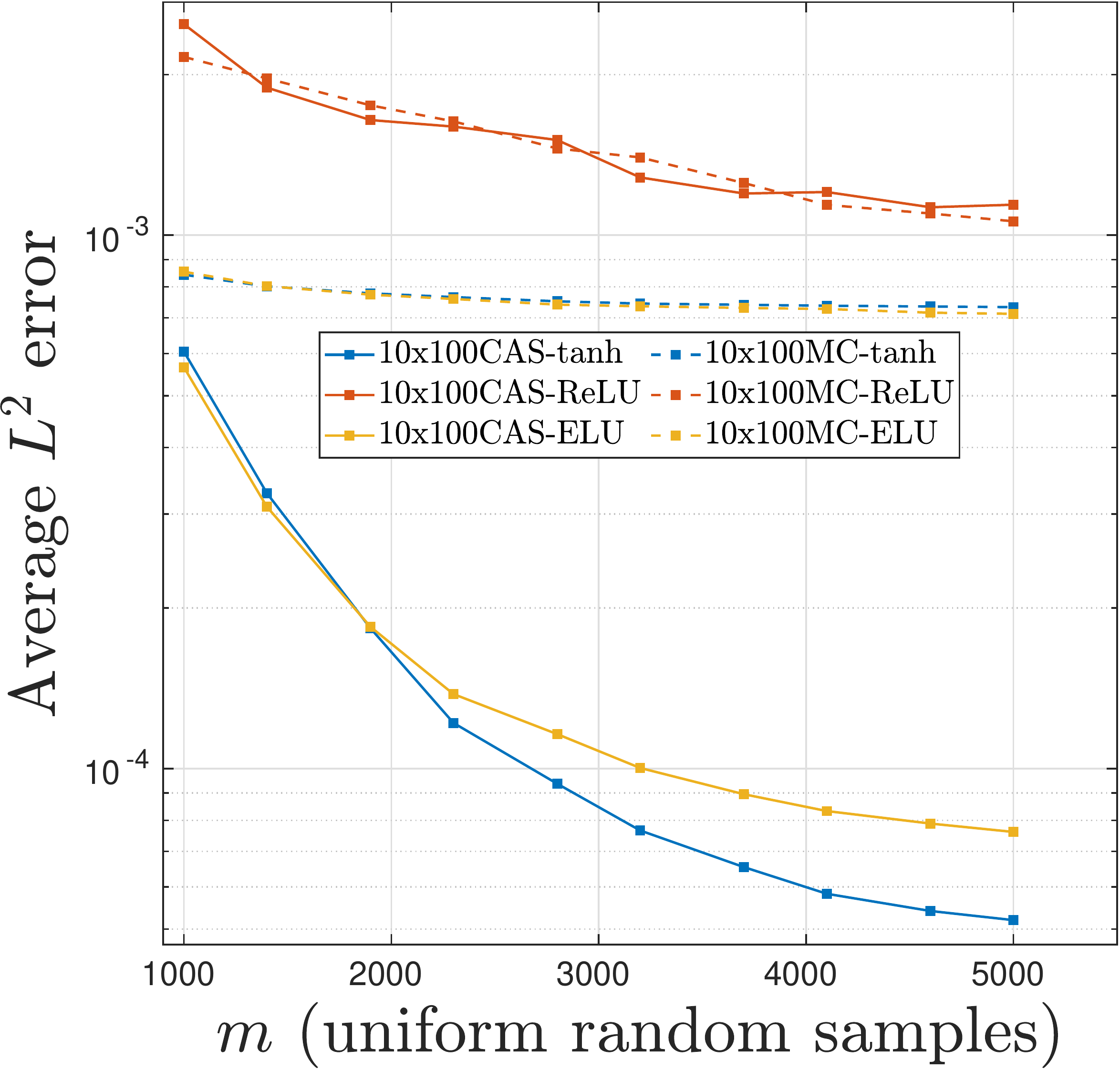} & 
\includegraphics[scale=0.11]{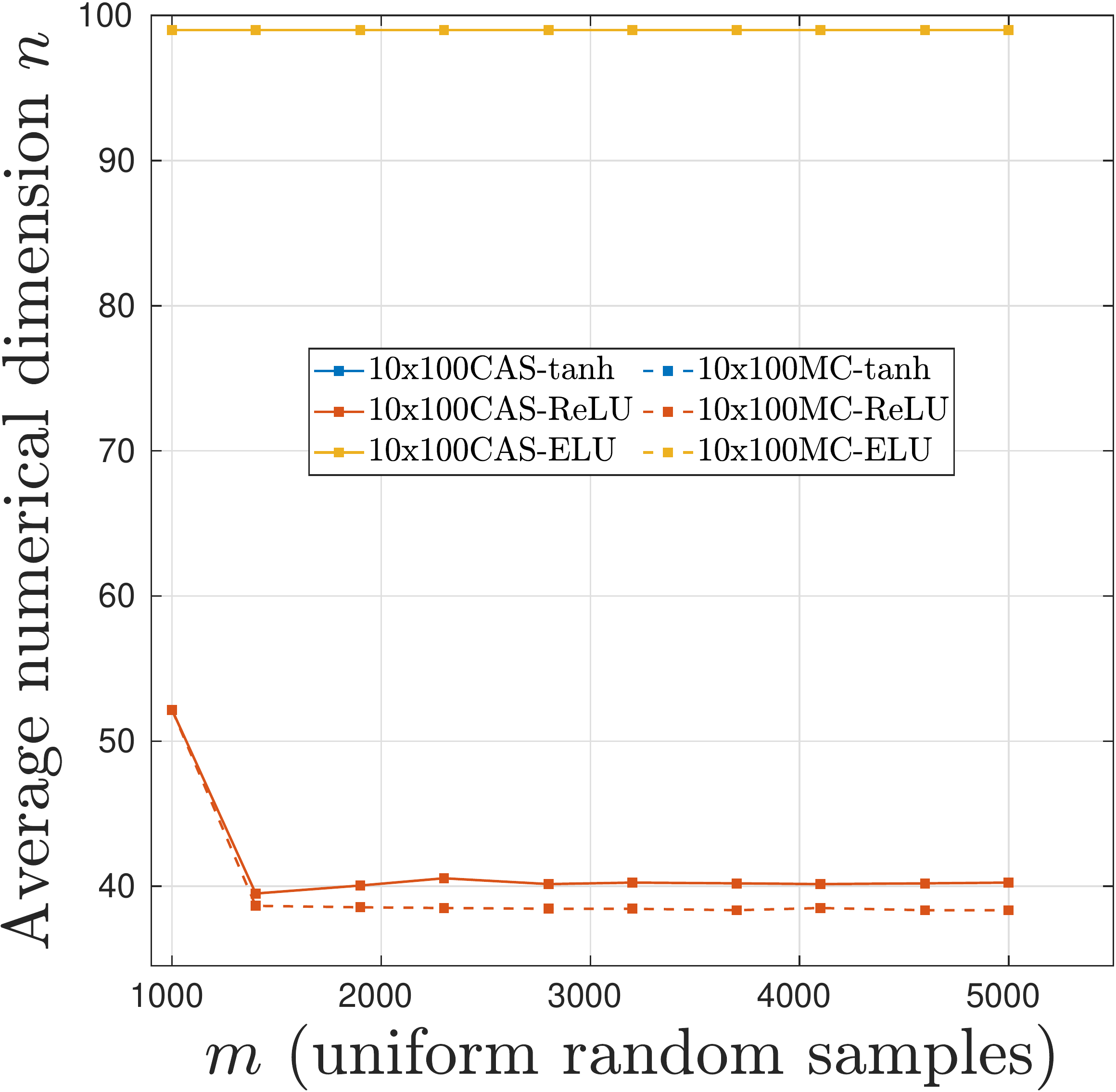} \\ 
\hspace{-0.5cm}
$(f,d)=(f_3,8)$ & $(f,d)=(f_3,8)$ & $(f,d)=(f_3,8)$ & $(f,d)=(f_3,8)$ \\
\hspace{-0.5cm}
\includegraphics[scale=0.11]{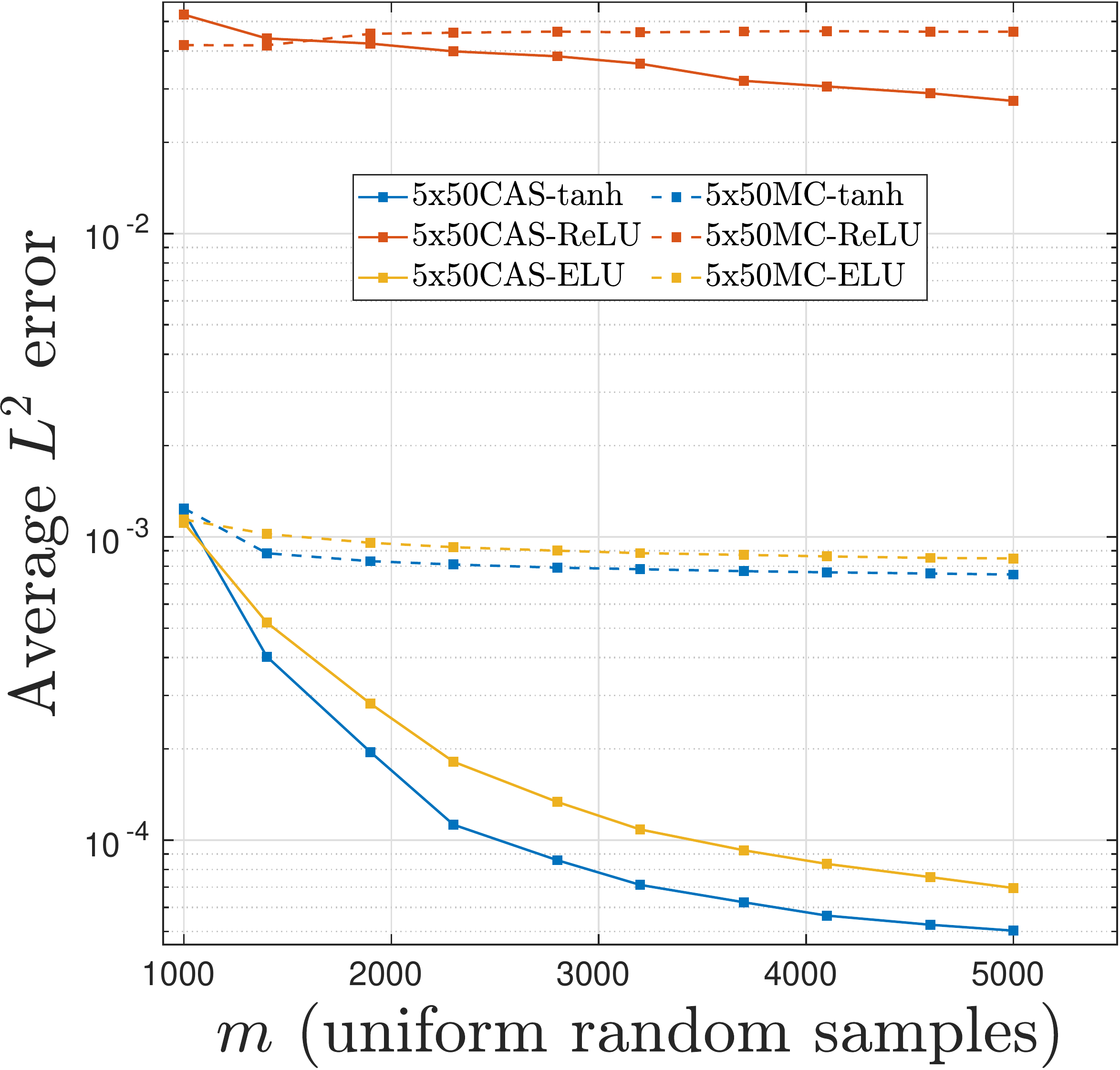} & 
\includegraphics[scale=0.11]{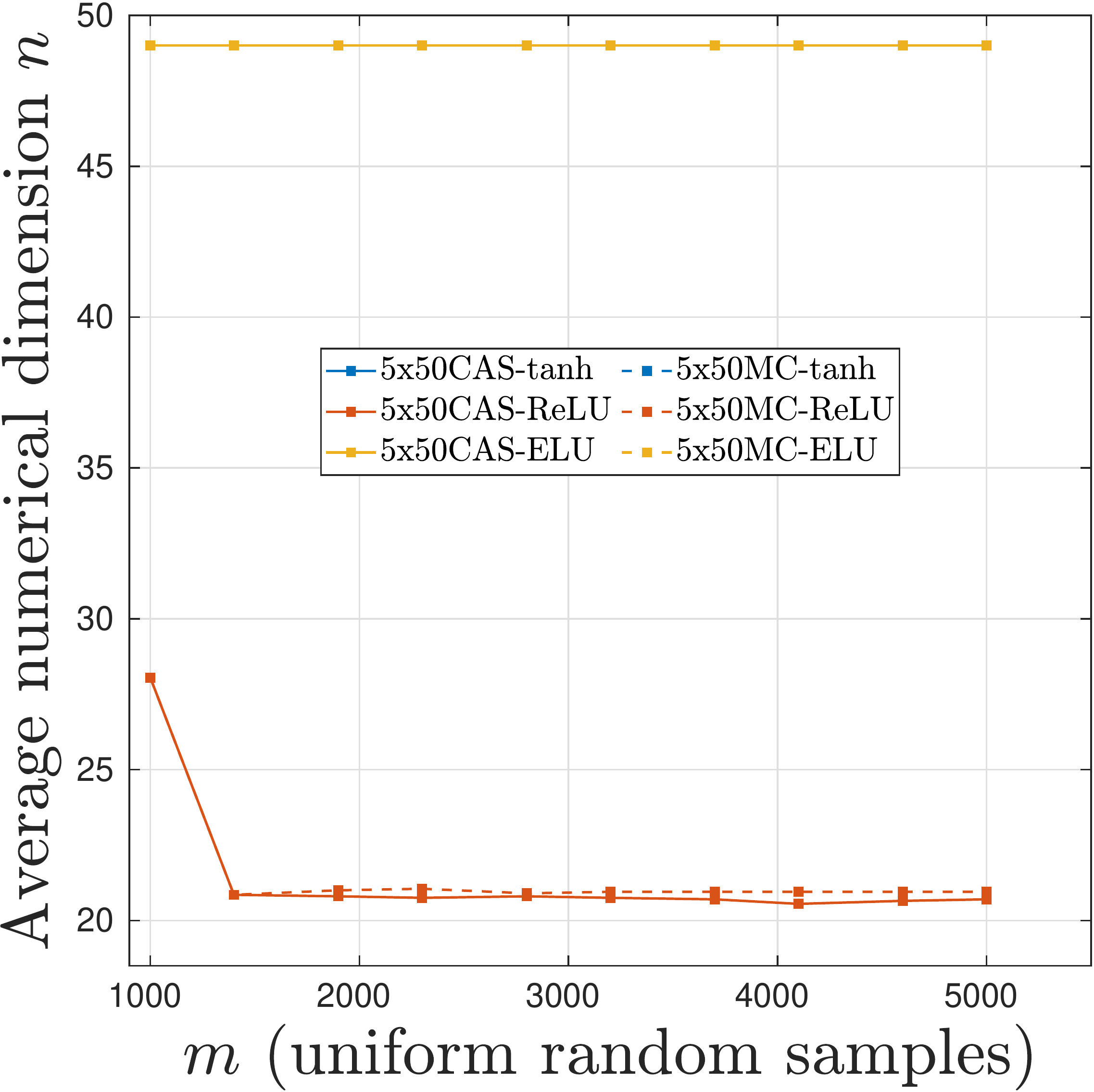} & 
\includegraphics[scale=0.11]{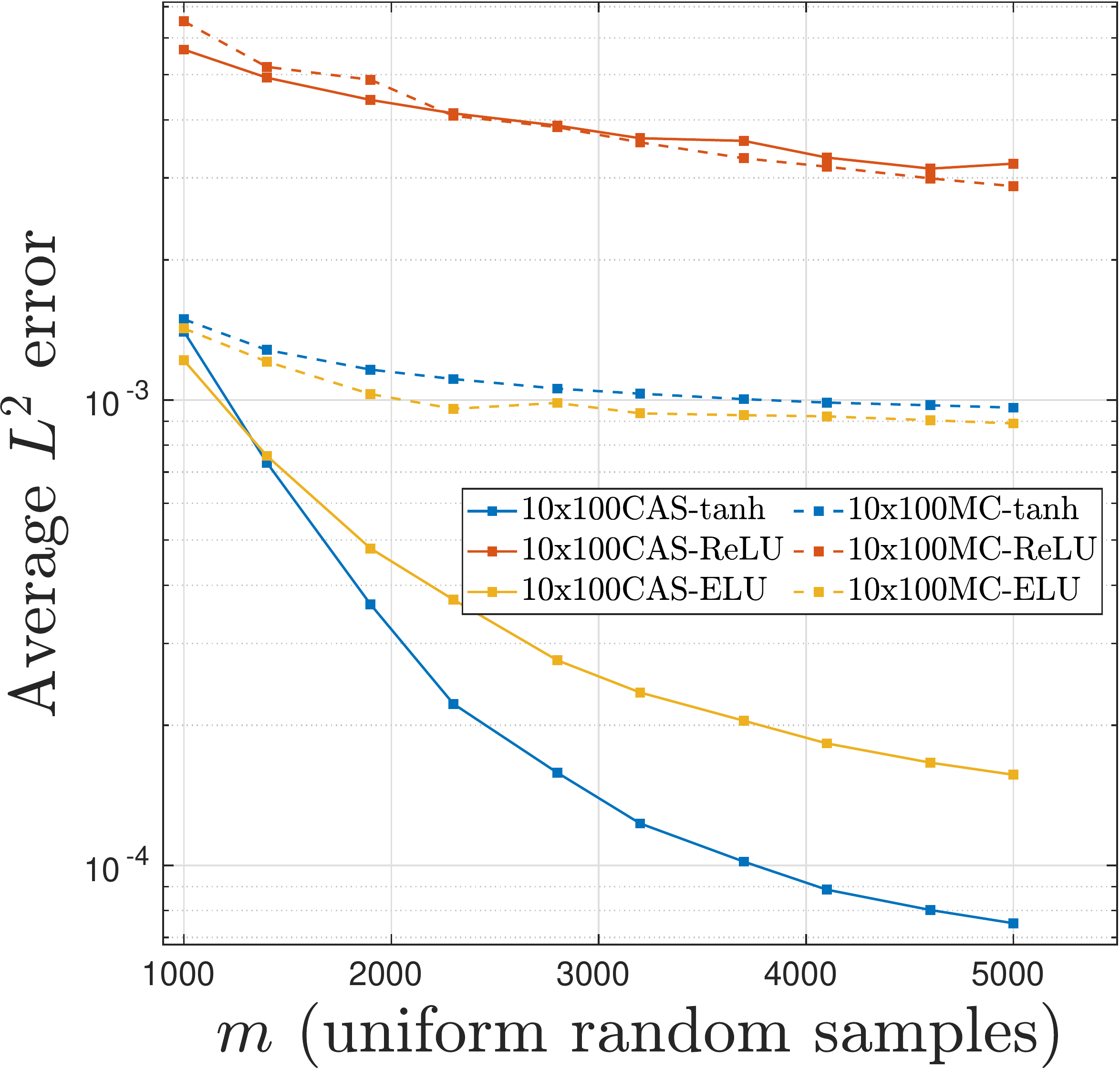} & 
\includegraphics[scale=0.11]{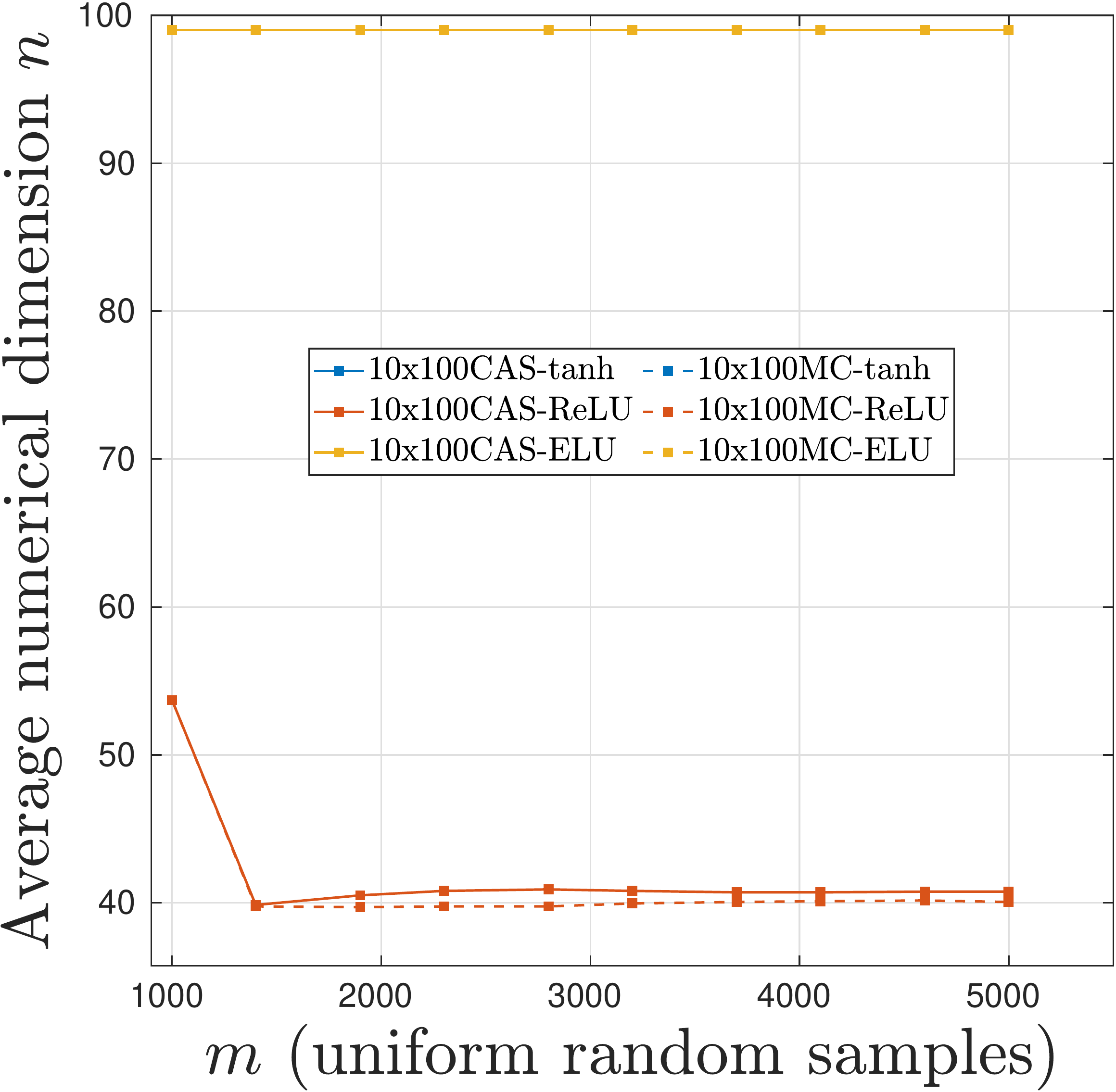} \\ 
\hspace{-0.5cm}
$(f,d)=(f_3 ,16)$  & $(f,d)=(f_3,16)$ & $(f,d)=(f_3,16)$ & $(f,d)=(f_3,16)$\\
\end{tabular}
}
\caption{The same as Figure \ref{fig:comp_act_L2_error_example_1}, except with $f = f_3$.}
\label{fig:comp_act_L2_error_example_6}
\end{figure}

\begin{figure}[h]
\centering
{\small
\begin{tabular}{cccc} 
\hspace{-0.5cm}
\includegraphics[scale=0.11]{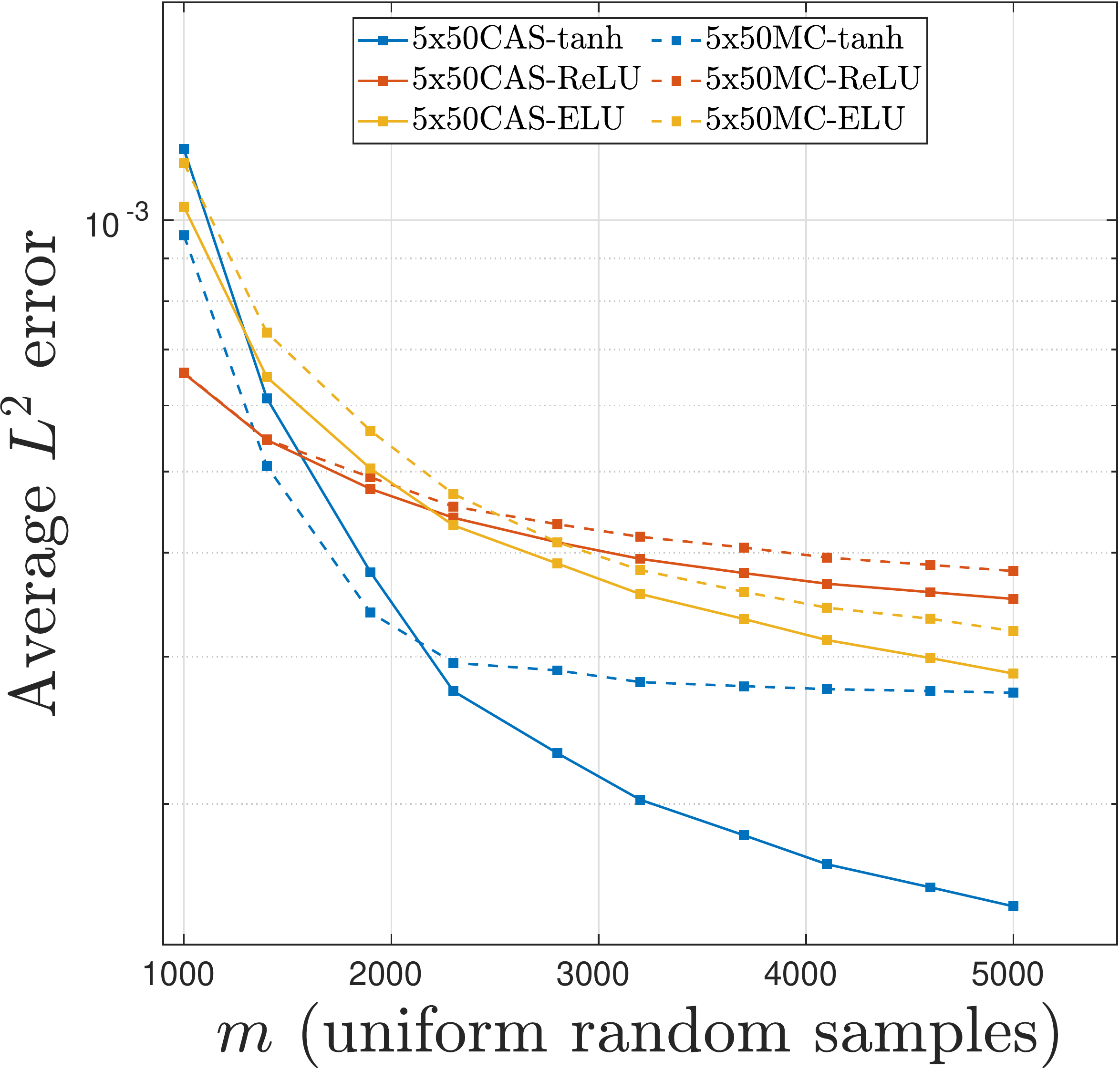} & 
\includegraphics[scale=0.11]{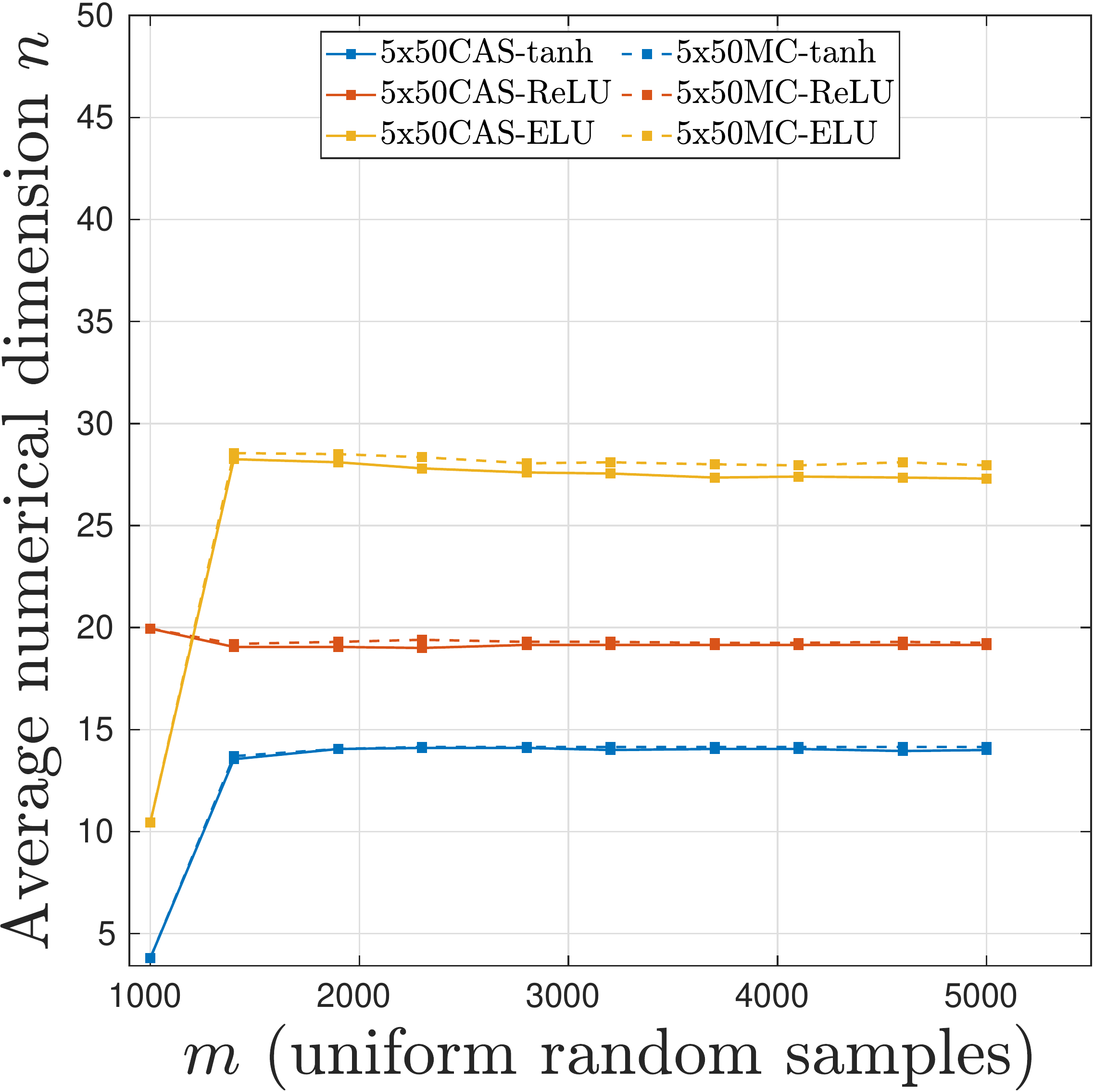} & 
\includegraphics[scale=0.11]{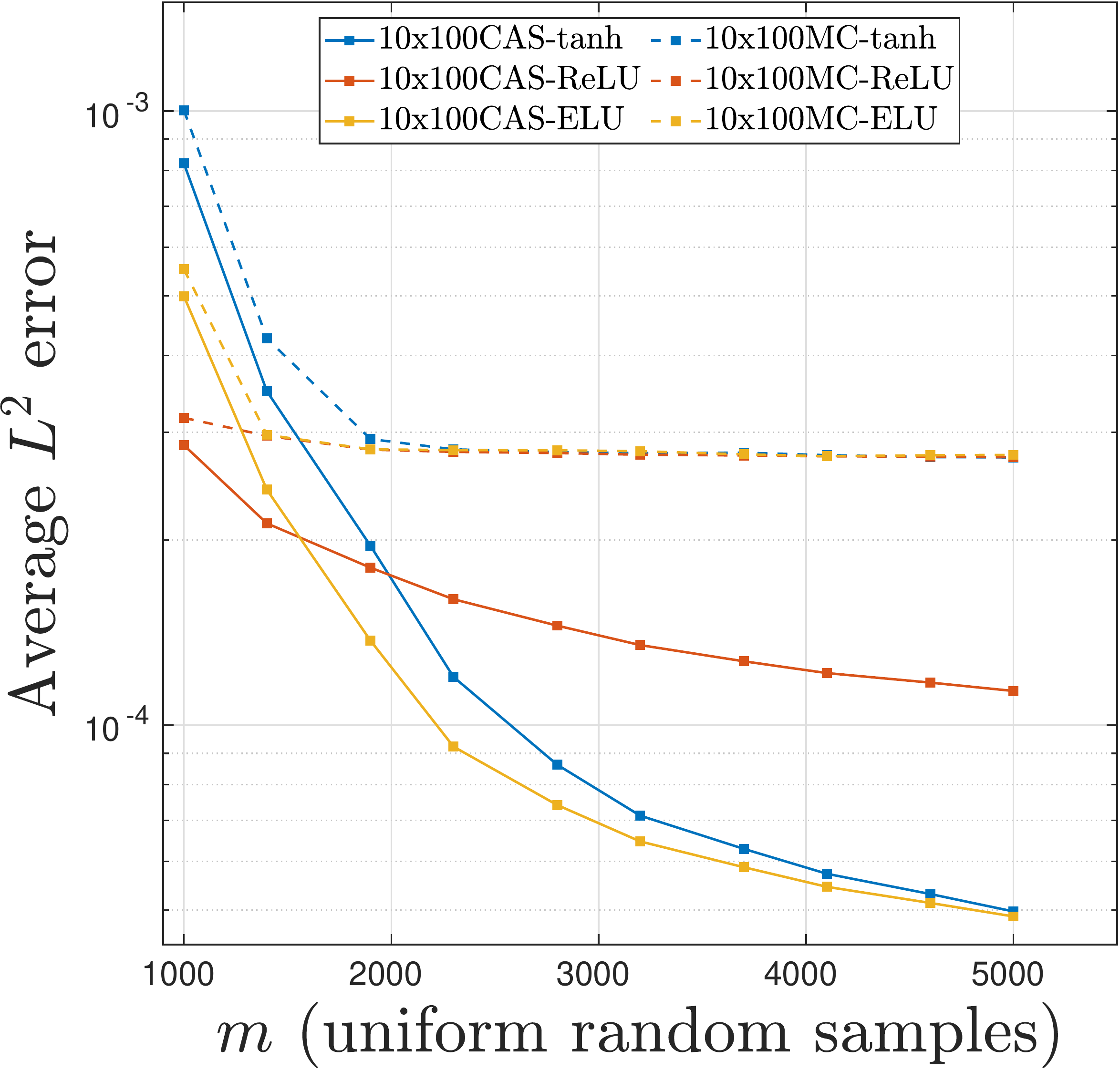} & 
\includegraphics[scale=0.11]{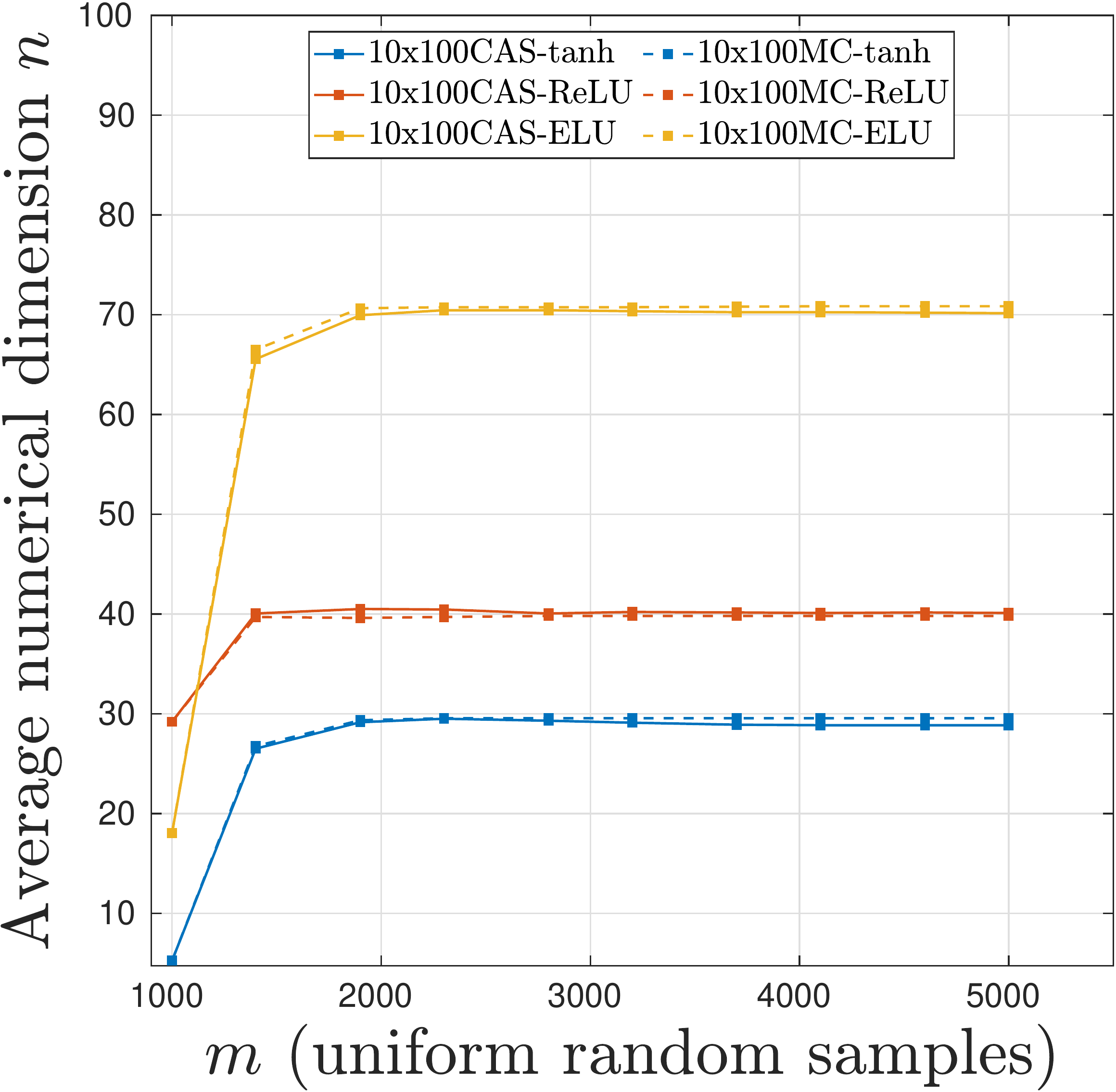} \\ 
\hspace{-0.5cm}
$(f,d)=(f_4,1)$ & $(f,d)=(f_4,1)$ & $(f,d)=(f_4,1)$ & $(f,d)=(f_4,1)$ \\
\hspace{-0.5cm}
\includegraphics[scale=0.11]{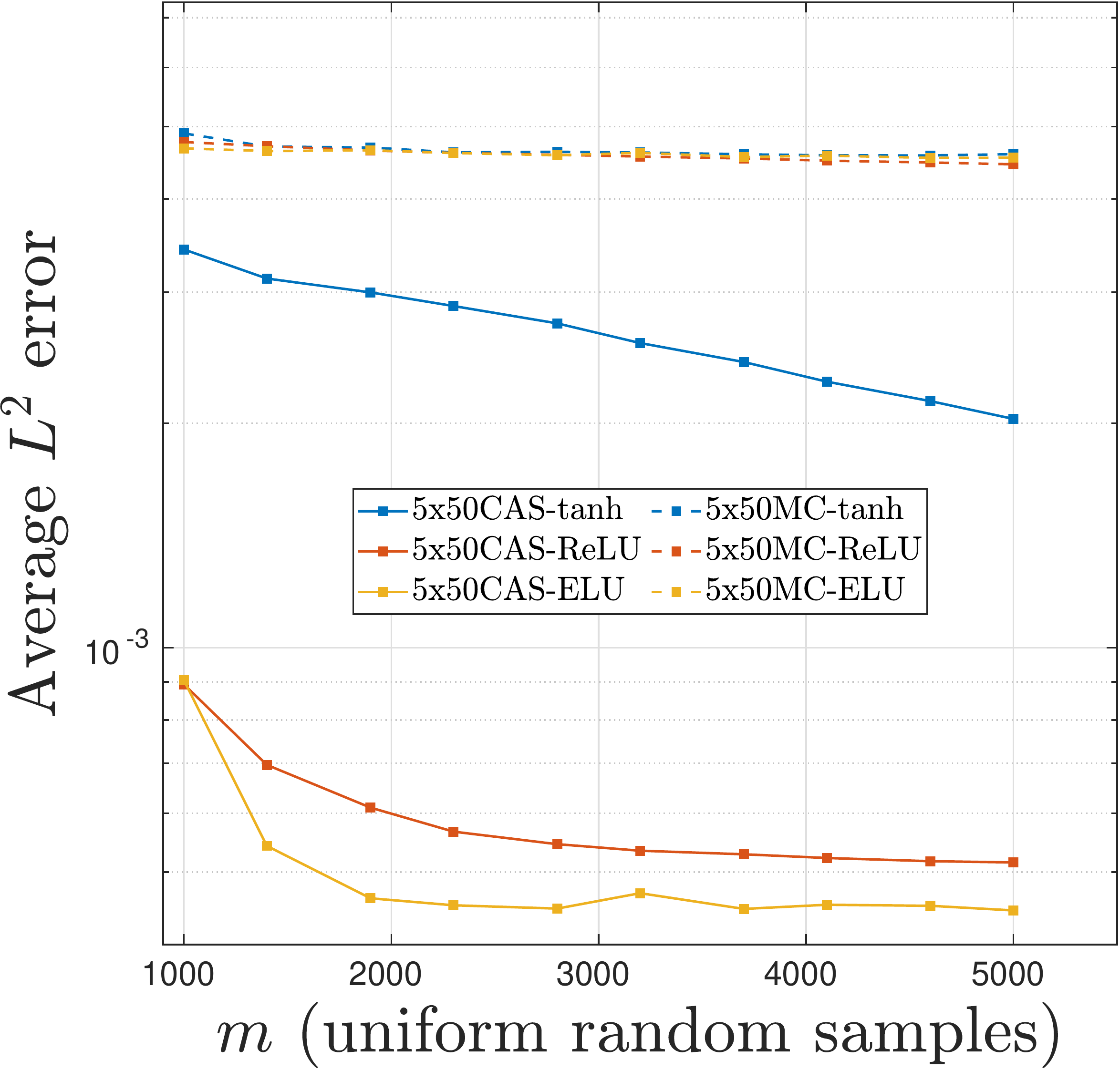} & 
\includegraphics[scale=0.11]{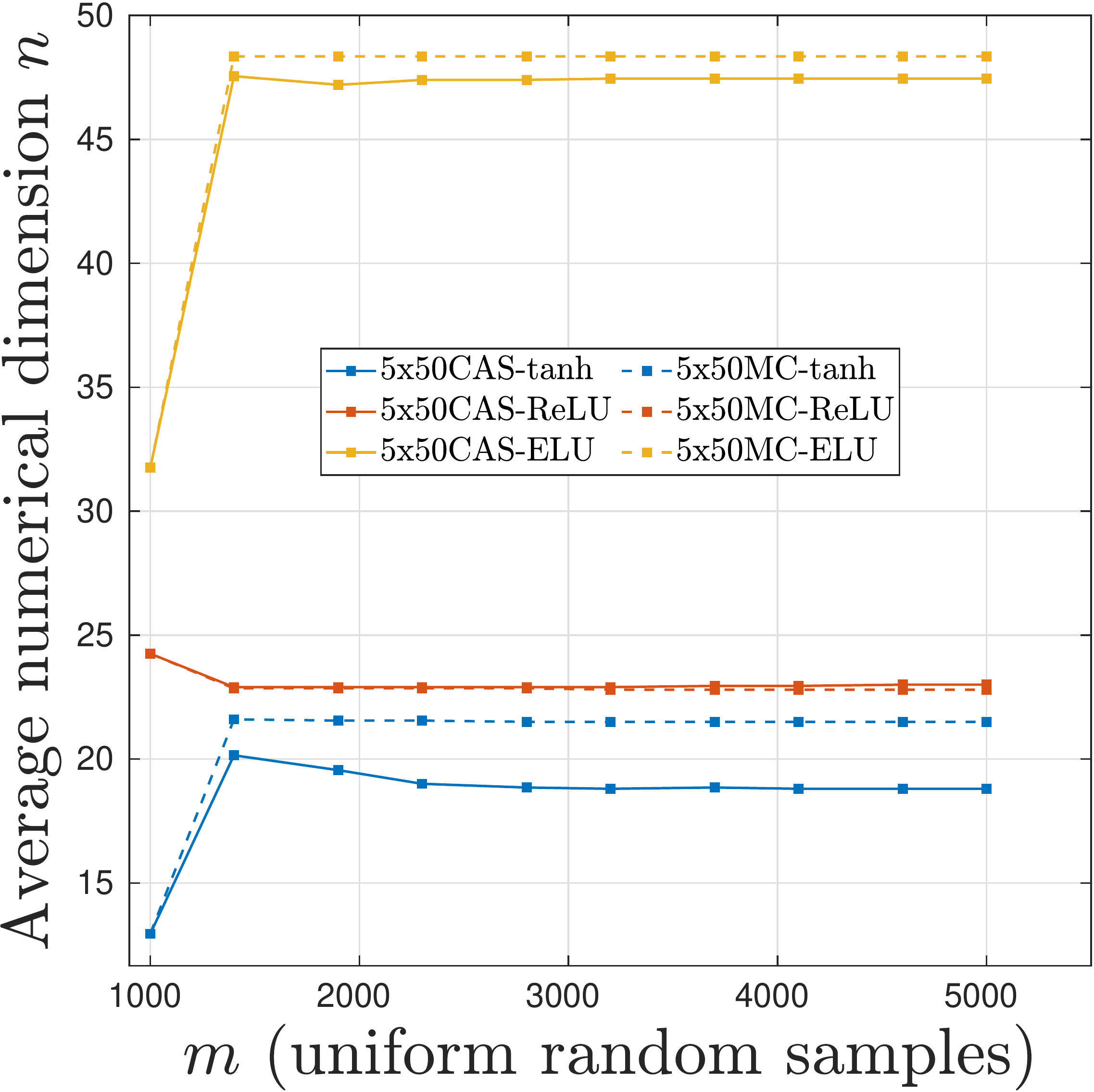} & 
\includegraphics[scale=0.11]{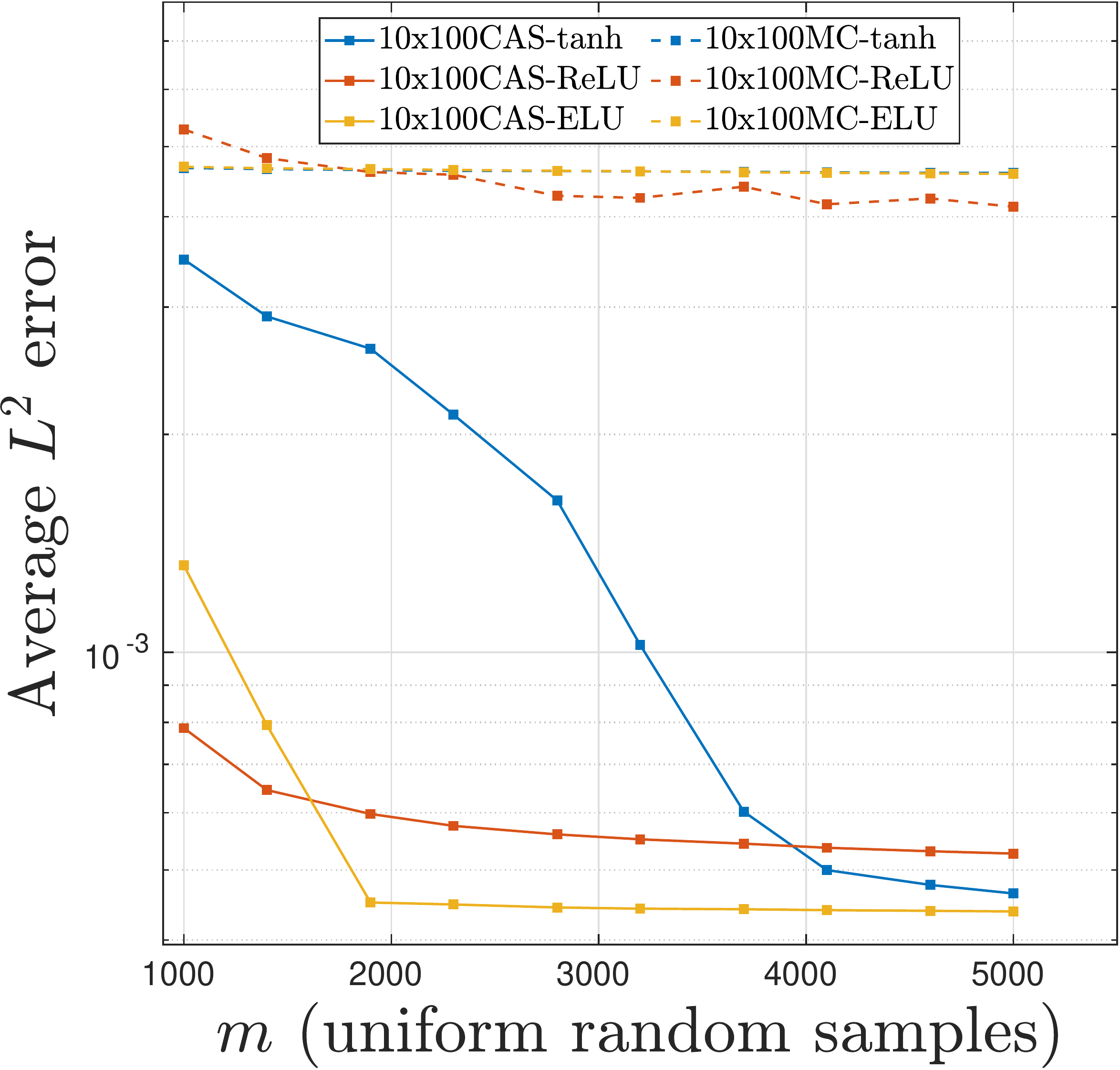} & 
\includegraphics[scale=0.11]{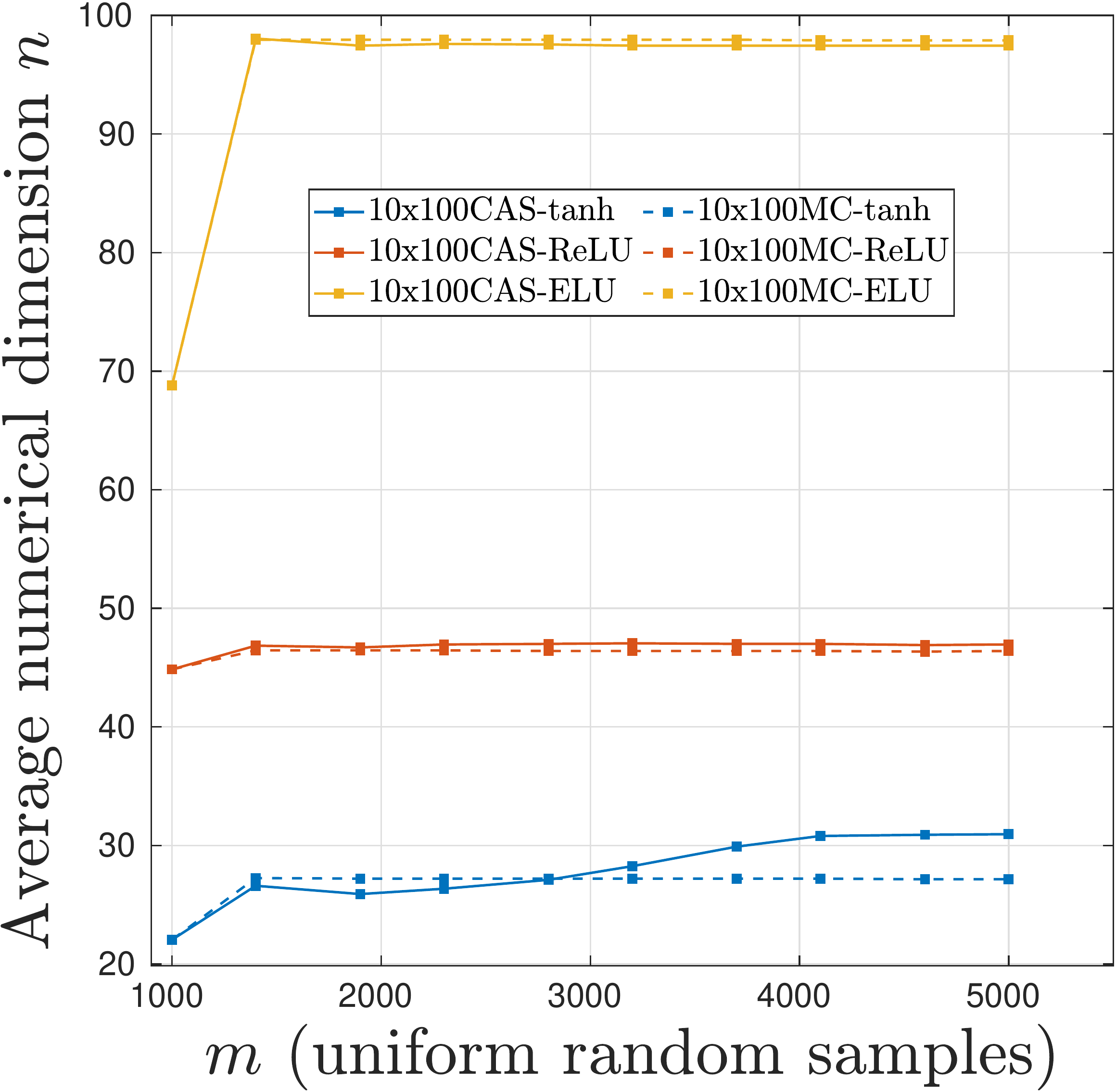} \\ 
\hspace{-0.5cm}
$(f,d)=(f_4,2)$ & $(f,d)=(f_4,2)$ &  $(f,d)=(f_4,2)$ & $(f,d)=(f_4,2)$ \\
\hspace{-0.5cm}
\includegraphics[scale=0.11]{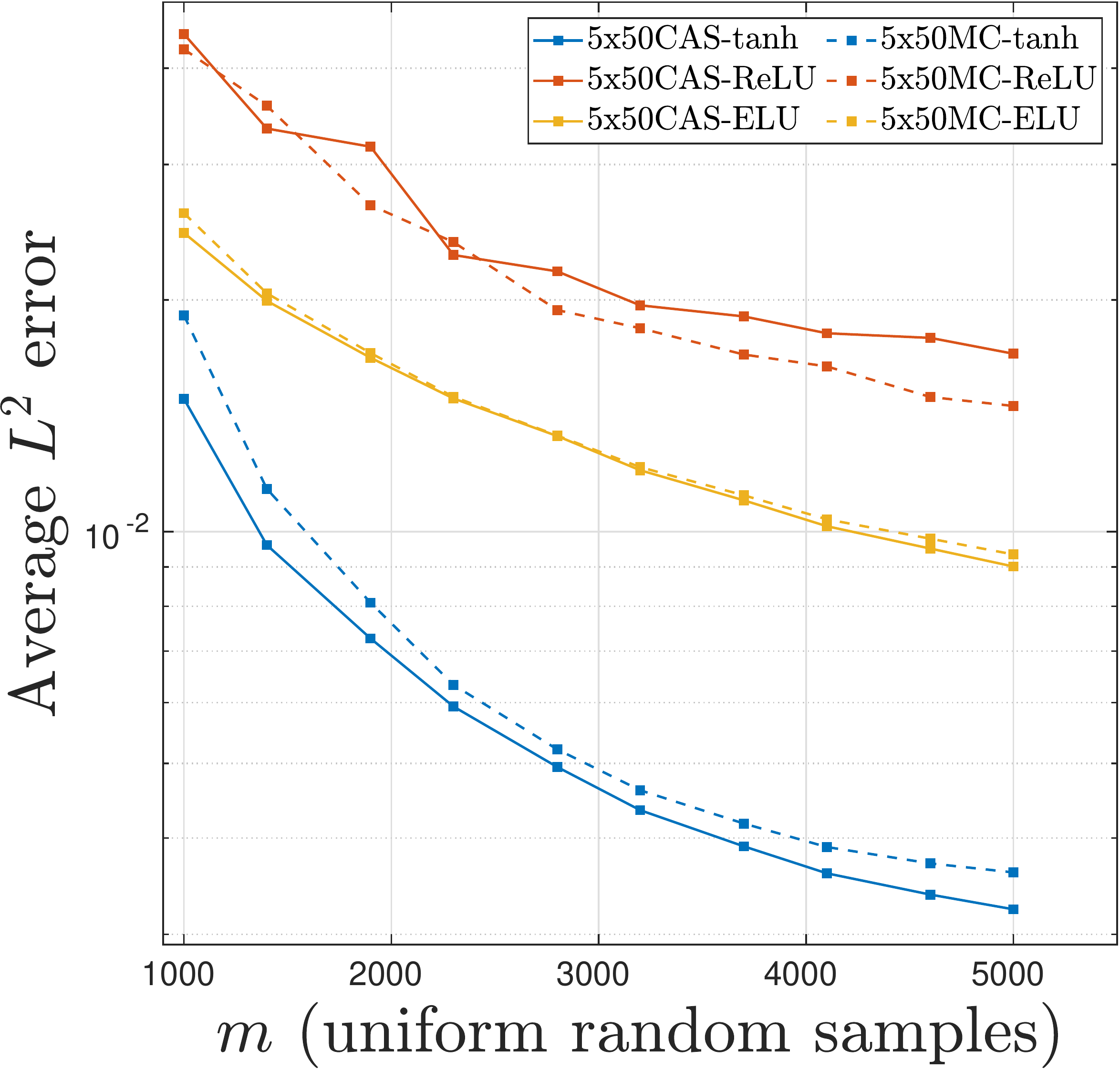} & 
\includegraphics[scale=0.11]{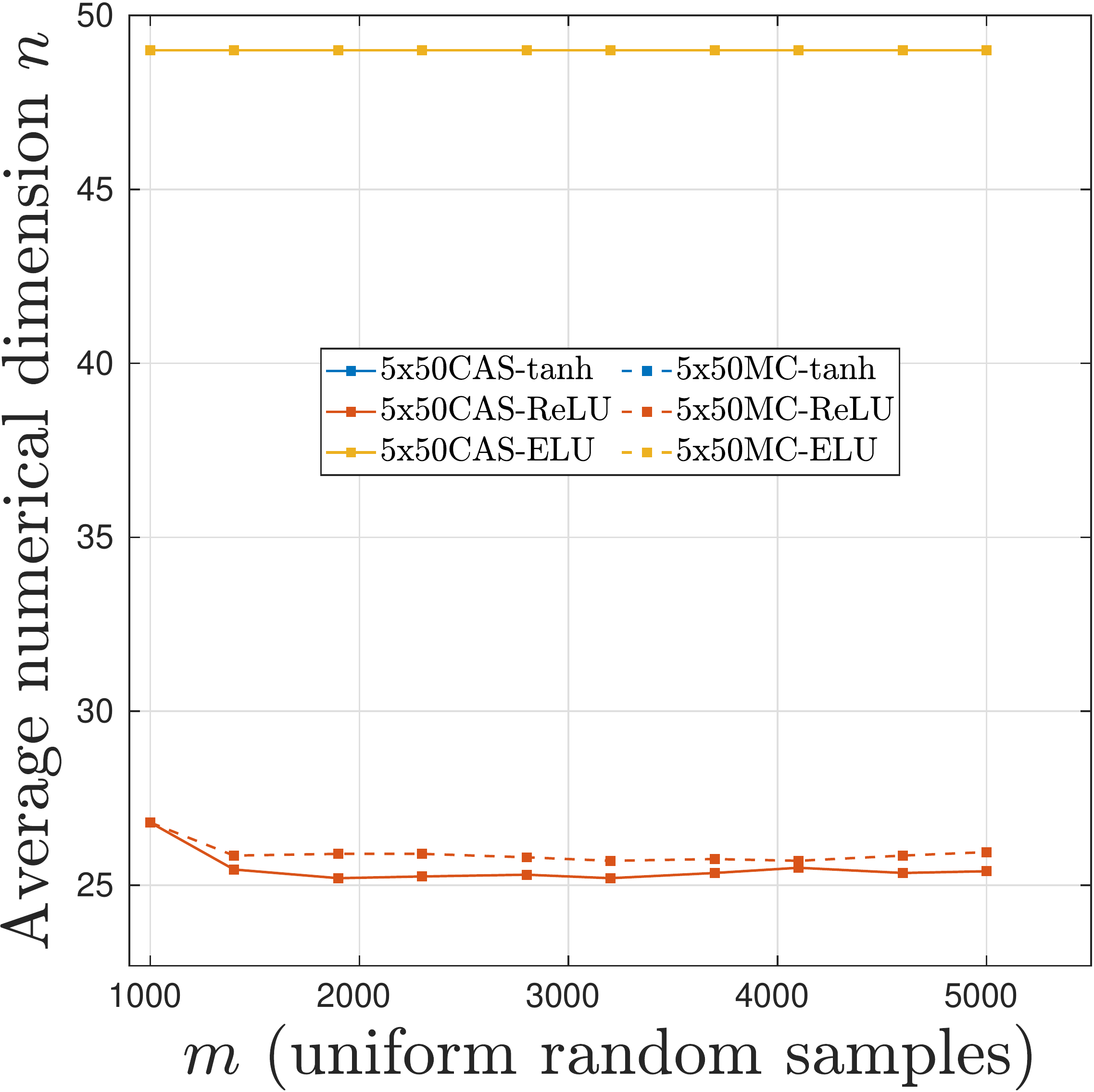} & 
\includegraphics[scale=0.11]{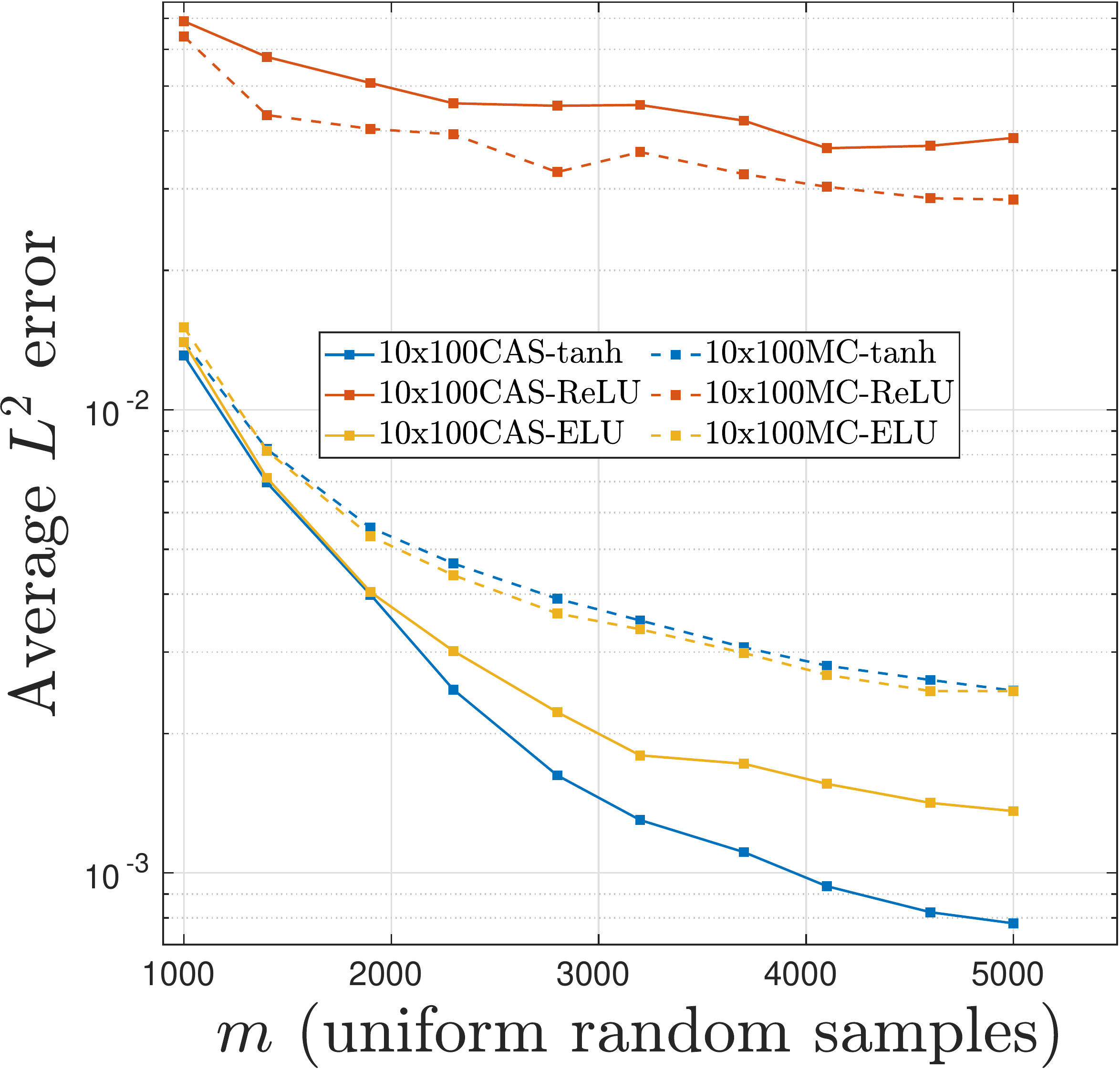} & 
\includegraphics[scale=0.11]{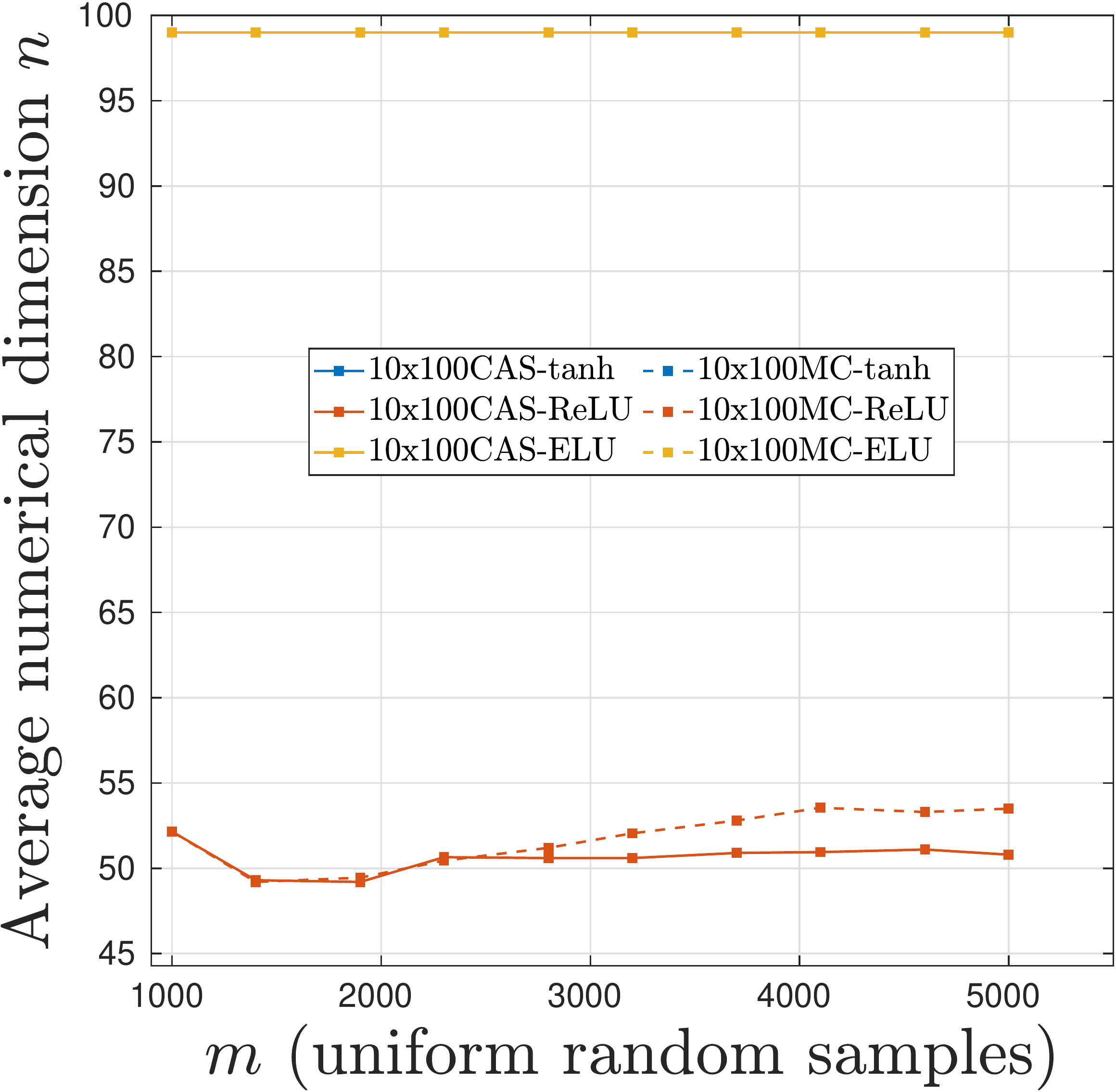} \\ 
\hspace{-0.5cm}
$(f,d)=(f_4,8)$  & $(f,d)=(f_4,8)$ & $(f,d)=(f_4,8)$ & $(f,d)=(f_4,8)$\\
\hspace{-0.5cm}
\includegraphics[scale=0.11]{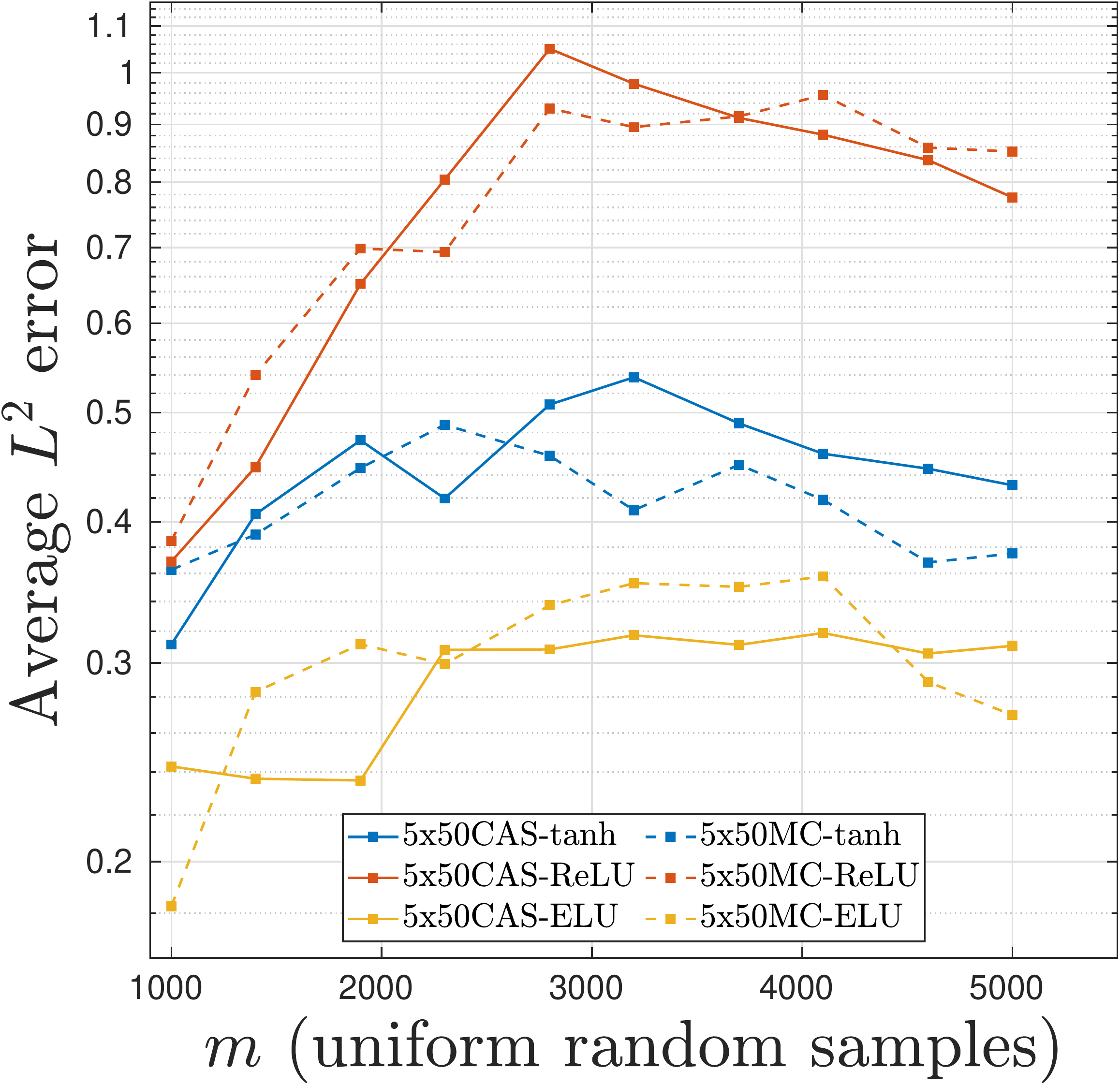} & 
\includegraphics[scale=0.11]{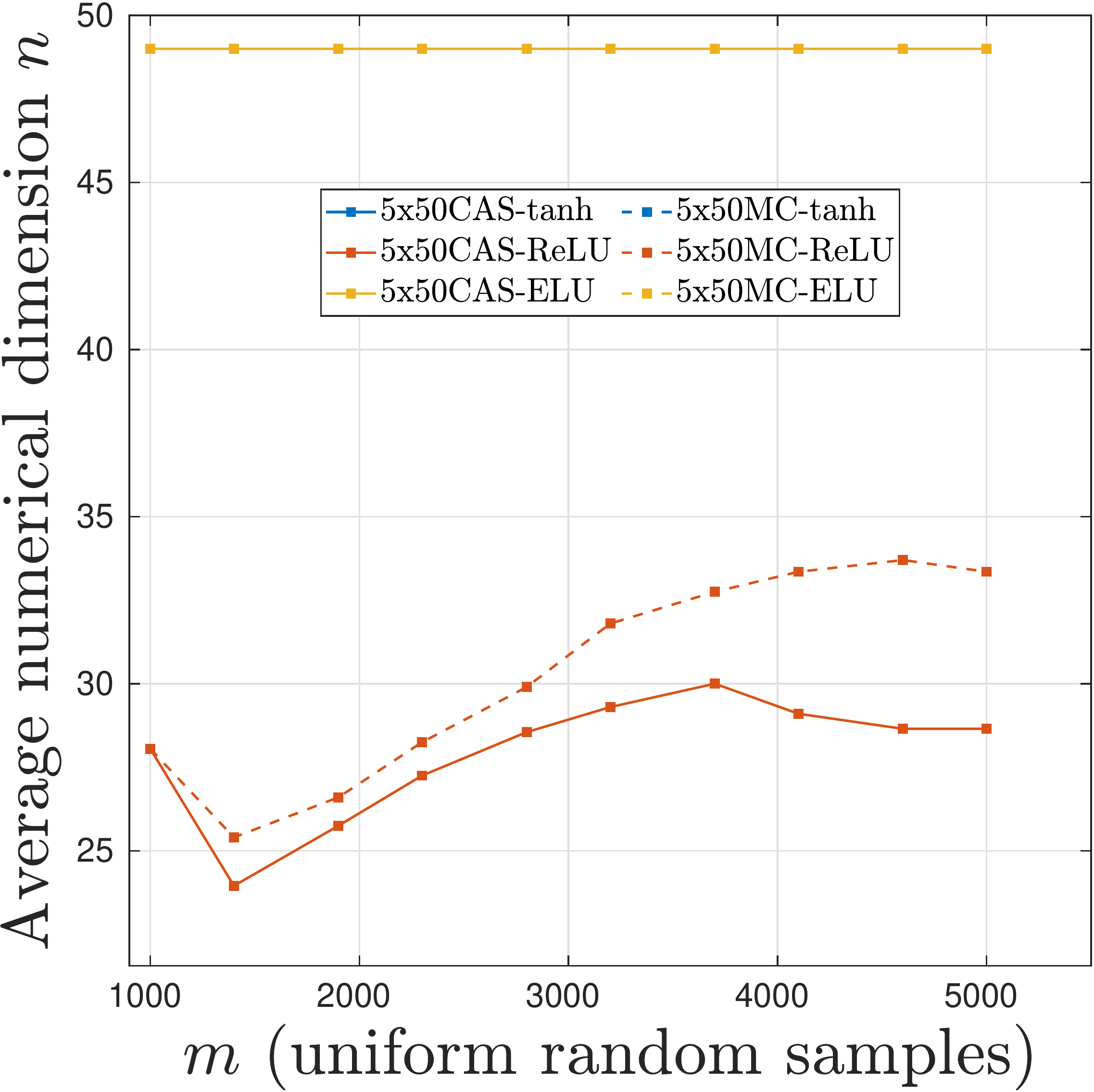} & 
\includegraphics[scale=0.11]{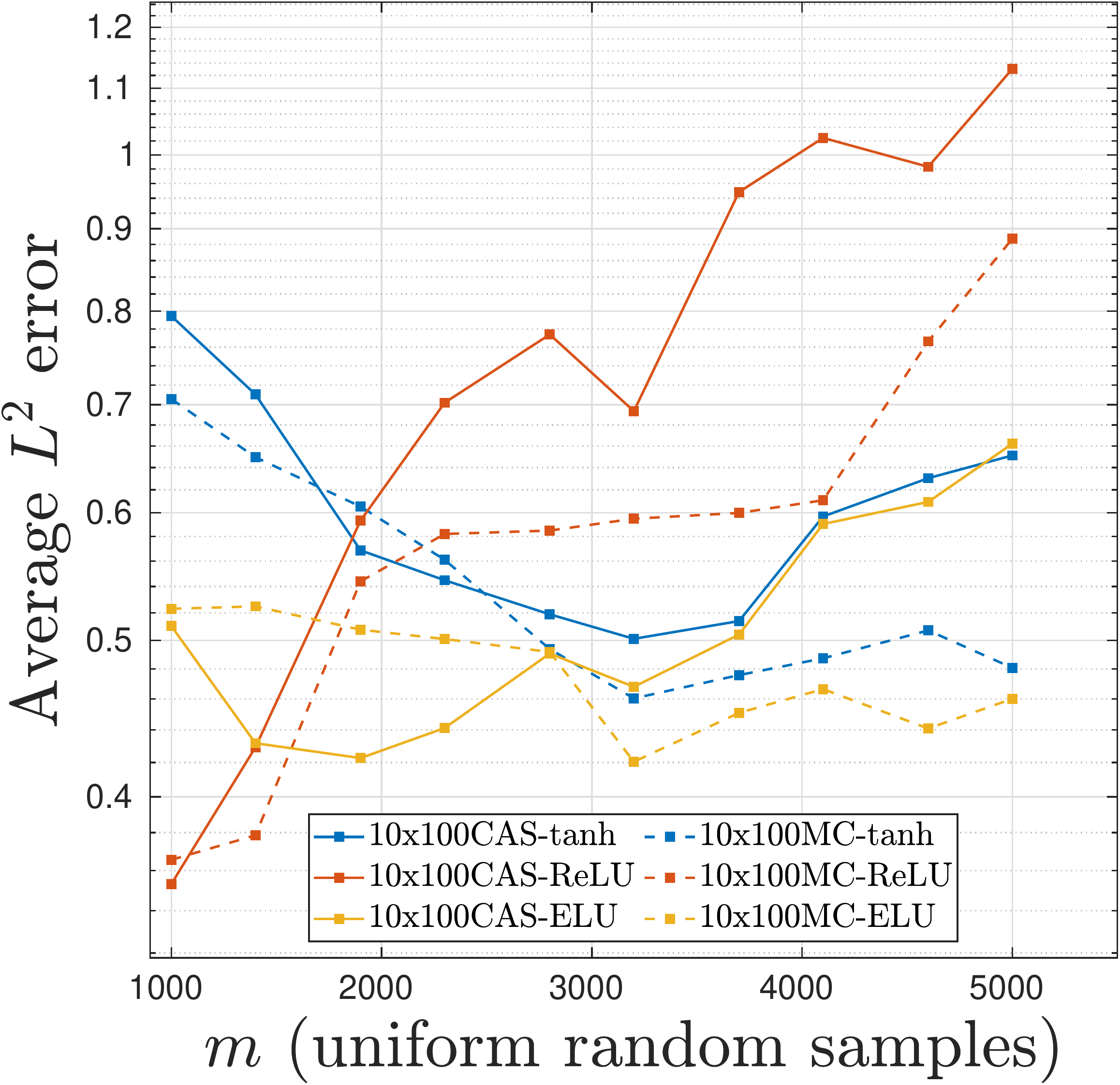} & 
\includegraphics[scale=0.11]{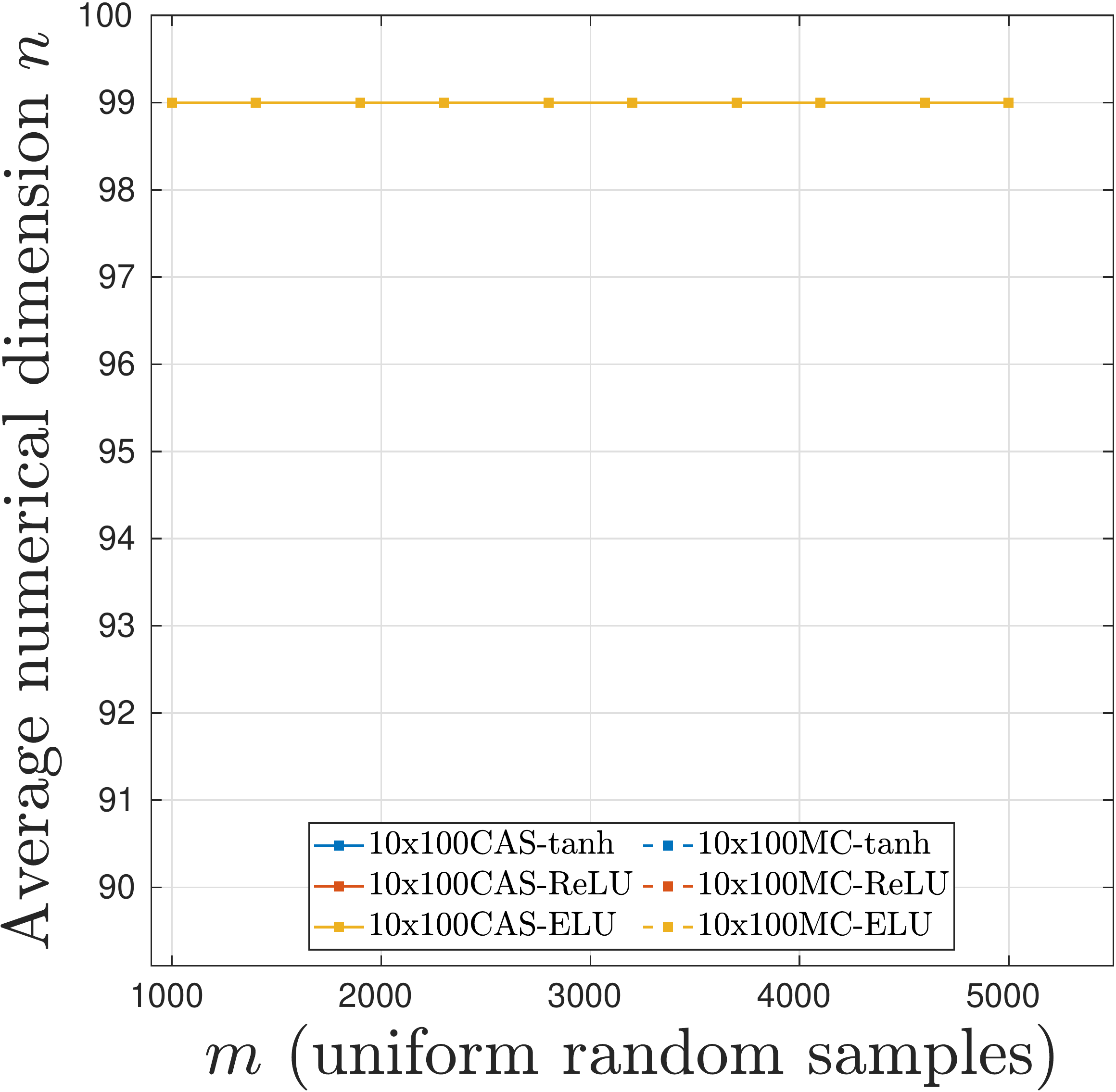} \\ 
\hspace{-0.5cm}
$(f,d)=(f_4,16)$  & $(f,d)=(f_4,16)$ & $(f,d)=(f_4,16)$ & $(f,d)=(f_4,16)$\\
\end{tabular}
}
\caption{The same as Figure \ref{fig:comp_act_L2_error_example_1}, except with $f = f_4$.}
\label{fig:comp_act_L2_error_example_5}
\end{figure}

In Figures \ref{fig:comp_act_L2_error_example_1}--\ref{fig:comp_act_L2_error_example_5} we present results comparing the performance of both strategies. In virtually all cases, we see that CAS4DL leads to a noticeable reduction in the number of samples required by CAS4DL compared to MC sampling to obtain a given accuracy.

In \figref{fig:comp_act_L2_error_example_1}, we present results for the function $f = f_1$. One can see a clear reduction in the number of samples required by CAS4DL compared to MC sampling to obtain a given accuracy in both low and high dimensions. 

We observe an improvement of CAS4DL over MC for the smoother activation functions ELU and $\tanh$ in all dimensions. In fact, the CAS4DL method often achieves two to ten times smaller errors for the same number of samples. There is also an improvement of CAS4DL over MC when using ReLU in dimensions $d=1$ and $d=2$. However, this is not the case in dimensions $d = 8$ and $d = 16$. Here CAS4DL shows an identical performance to MC in dimension $d=8$, but it is marginally worse than MC in dimension $d=16$. 

Similar improvements in performance for CAS4DL over MC using ELU and $\tanh$ are also observed for the functions $f=f_2, f_3, f_4$ in Figures \ref{fig:comp_act_L2_error_example_2}--\ref{fig:comp_act_L2_error_example_5}. In addition, the benefit of CAS4DL over MC when using ReLU is noticeable in dimensions $d=1$ and $d=2$, but their performance is almost identical in dimensions $d=8$ and $d=16$. Finally, we mention for $f=f_4$ in dimension $d= 16$ there is no improvement decreasing the error for all DNNs.

\newpage
\vspace*{\fill}
\vspace*{\fill}

\newpage
\vspace*{\fill}
\vspace*{\fill}

\newpage
\vspace*{\fill}
\vspace*{\fill}

\newpage
\vspace*{\fill}
\vspace*{\fill}

\newpage
\subsection{The effect of the activation function on the sampling distribution and learned dictionary} 

\begin{figure}[h!]
\centering
{\small
\begin{tabular}{ccc}
\hspace{-1cm}
\includegraphics[scale=0.2]{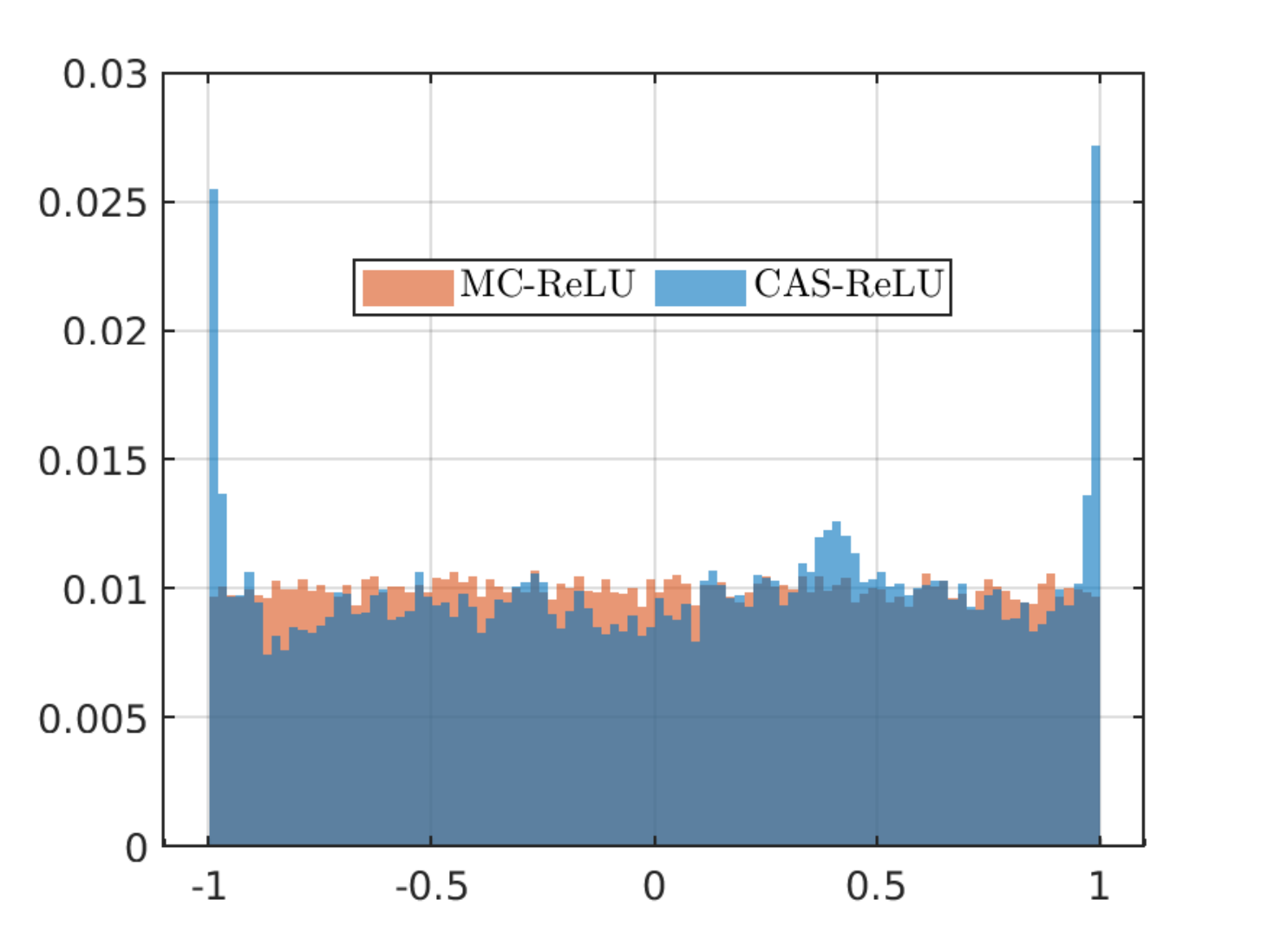} & 
\includegraphics[scale=0.2]{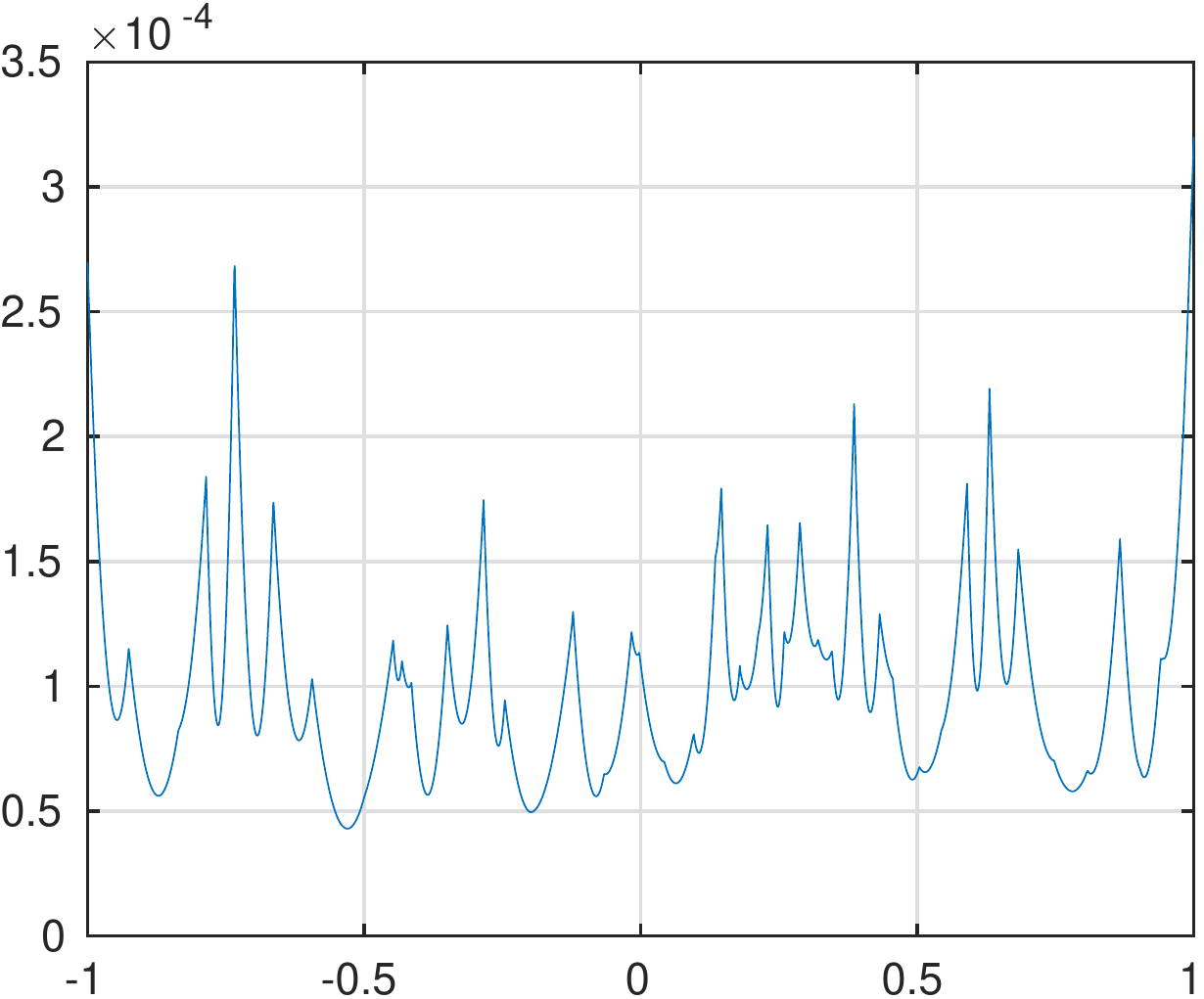} &
\includegraphics[scale=0.2]{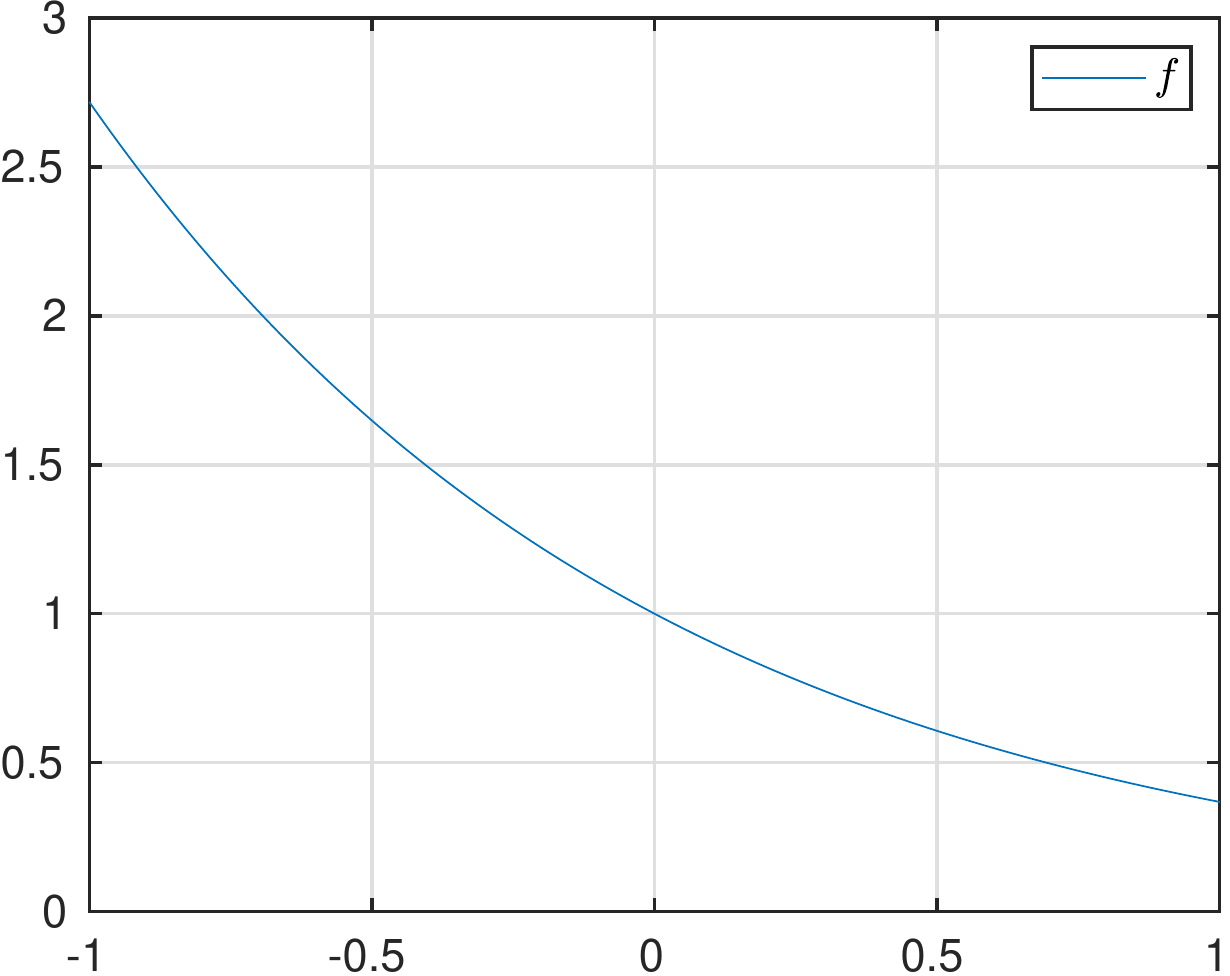}\\
\hspace{-1cm}
$(f,\rho) = (f_1,\mathrm{ReLU})$ & $(f,\rho) = (f_1,\mathrm{ReLU})$&  $(f,d) = (f_1,1)$ \\
\hspace{-1cm}
\includegraphics[scale=0.2]{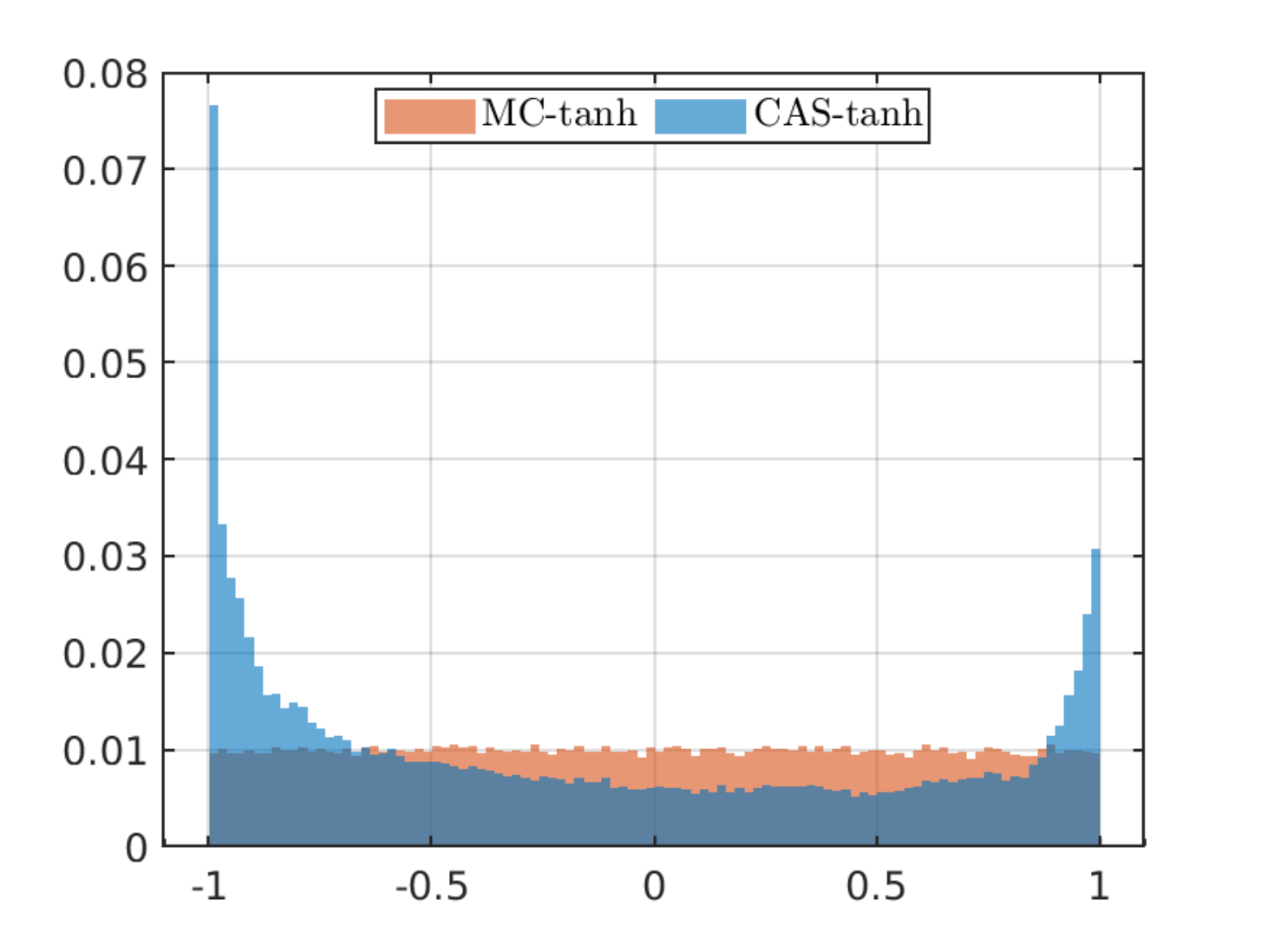} & 
\includegraphics[scale=0.2]{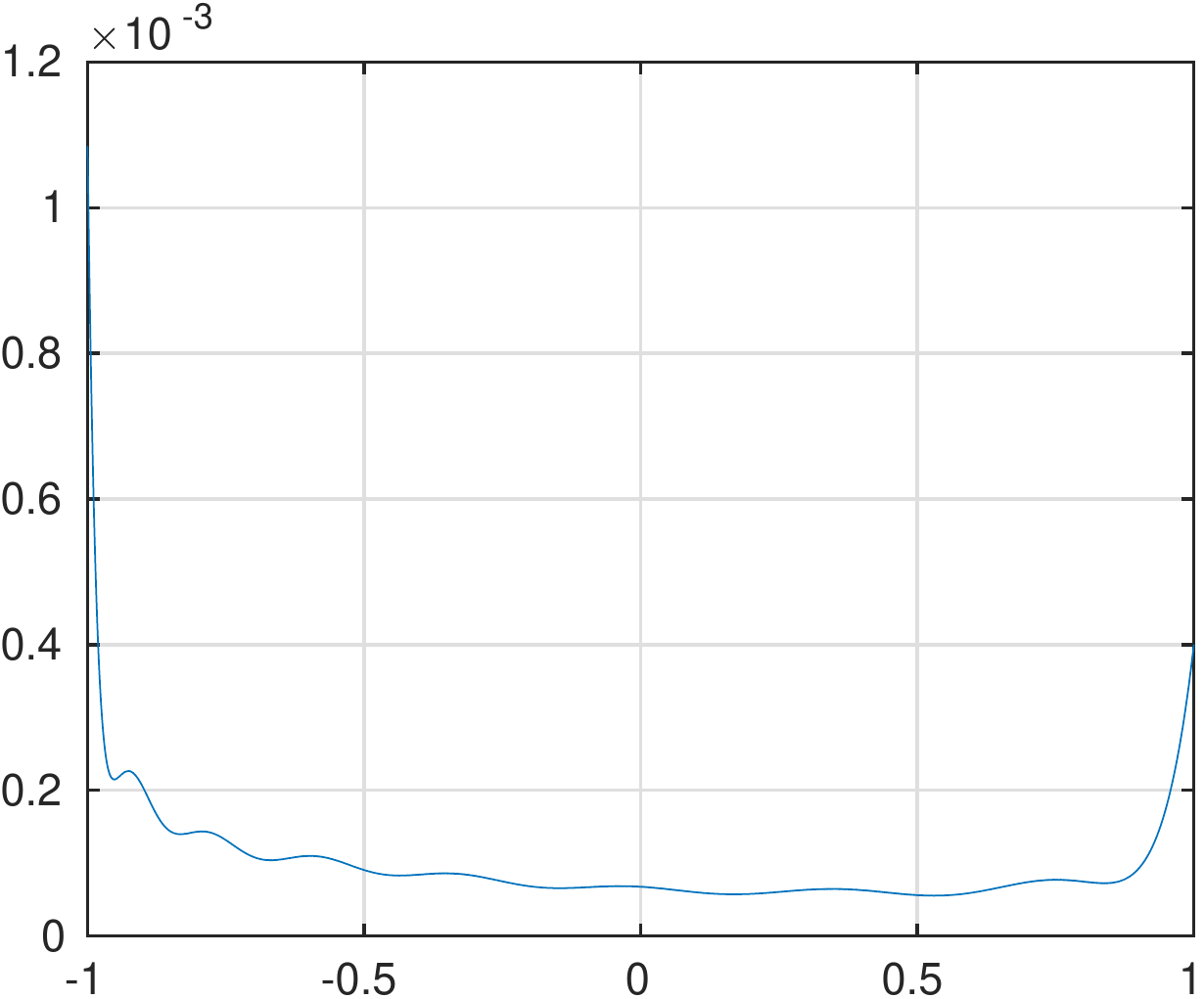} &
\includegraphics[scale=0.2]{train_data_example_1_dim_1-eps-converted-to.pdf}\\
\hspace{-1cm}
$(f,\rho) = (f_1, \tanh)$ & $(f,\rho) = (f_1, \tanh)$  &  $(f,d) = (f_1,1)$ \\
\hspace{-1cm}
\includegraphics[scale=0.2]{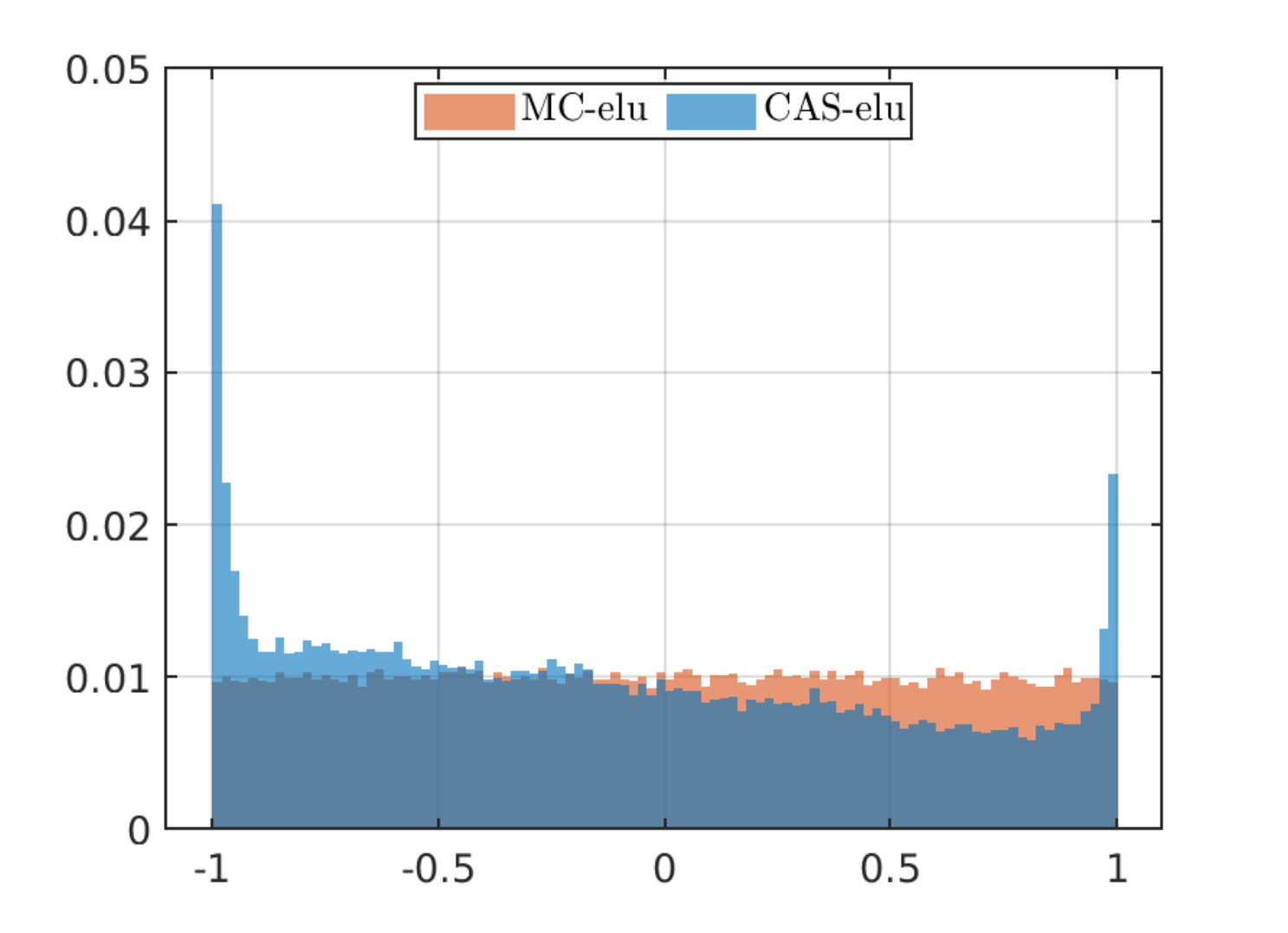} & 
\includegraphics[scale=0.2]{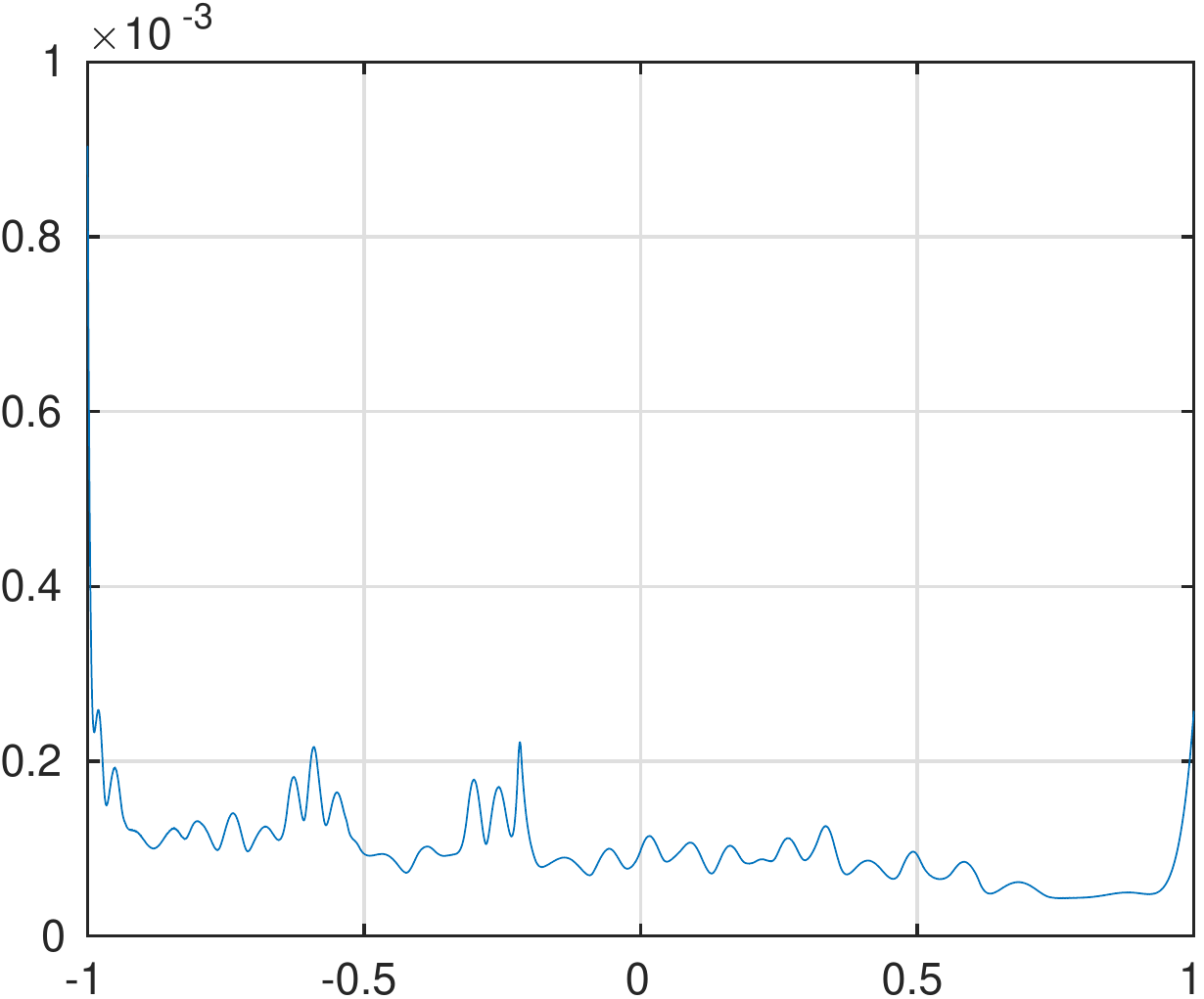} &
\includegraphics[scale=0.2]{train_data_example_1_dim_1-eps-converted-to.pdf}\\
\hspace{-1cm}
$(f,\rho) = (f_1, $eLU$)$ & $(f,\rho) = (f_1, $ELU$)$ &   $(f,d) = (f_1,1)$ \\
\hspace{-1cm}
\includegraphics[scale=0.2]{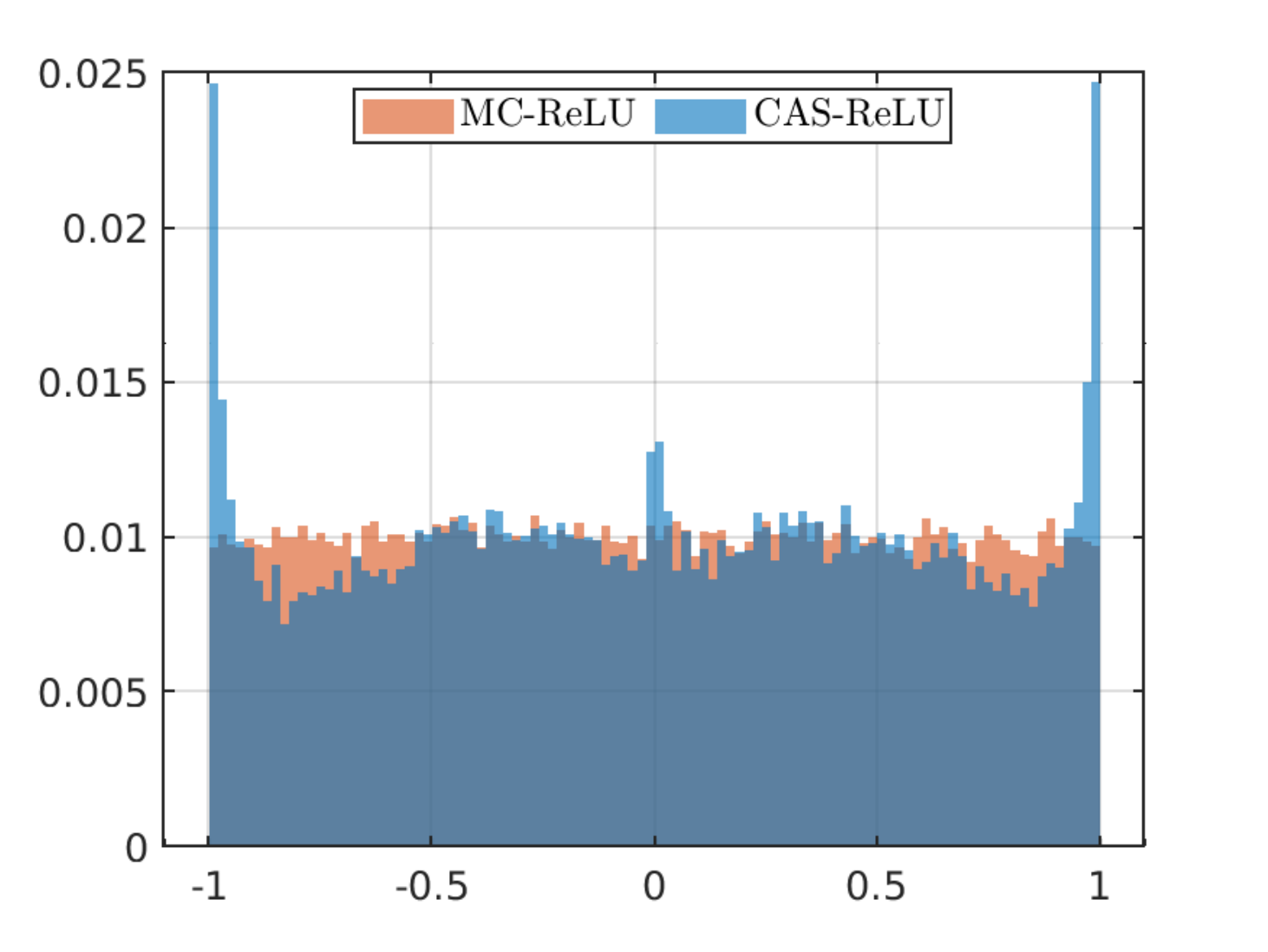} & 
\includegraphics[scale=0.2]{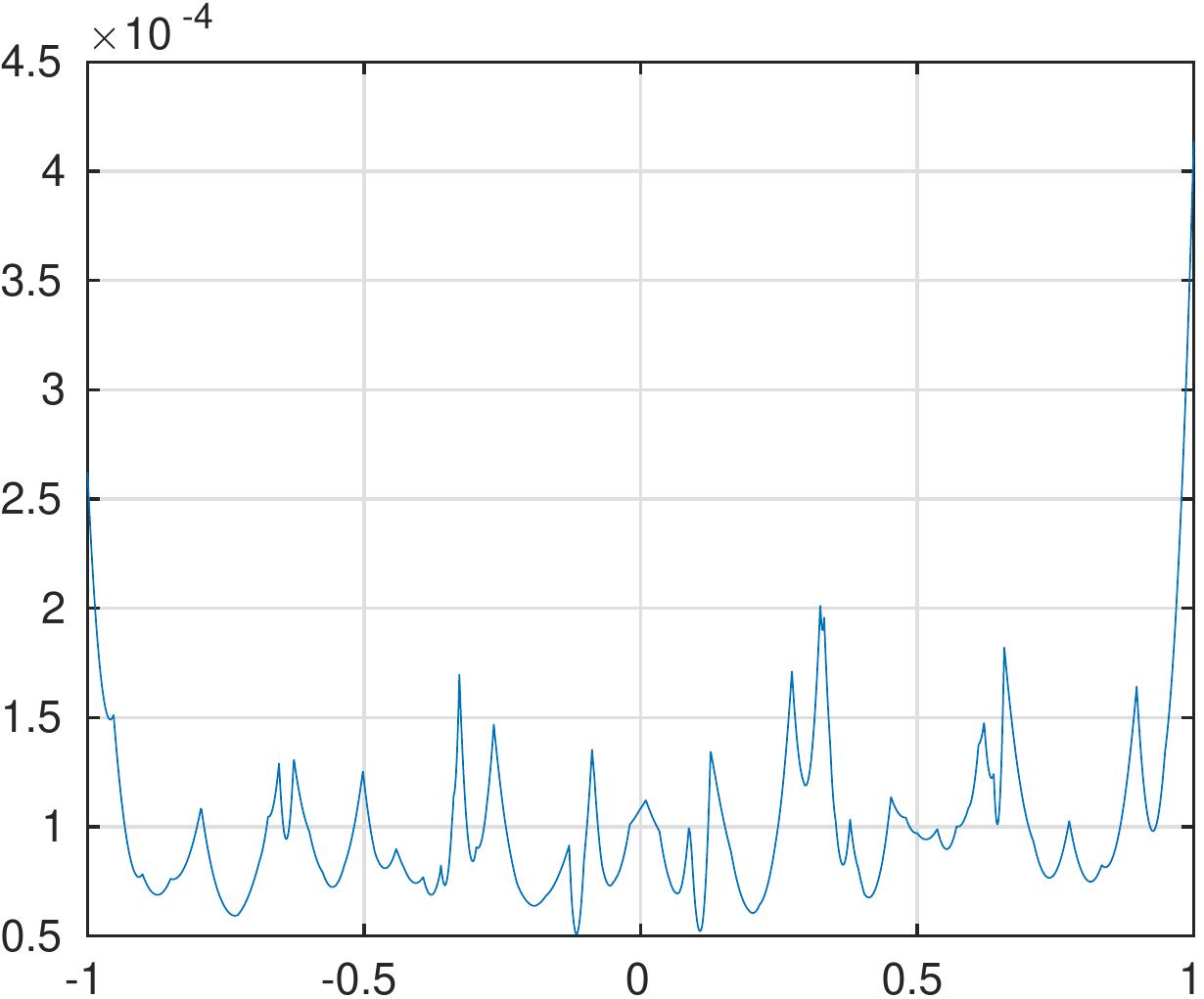} &
\includegraphics[scale=0.2]{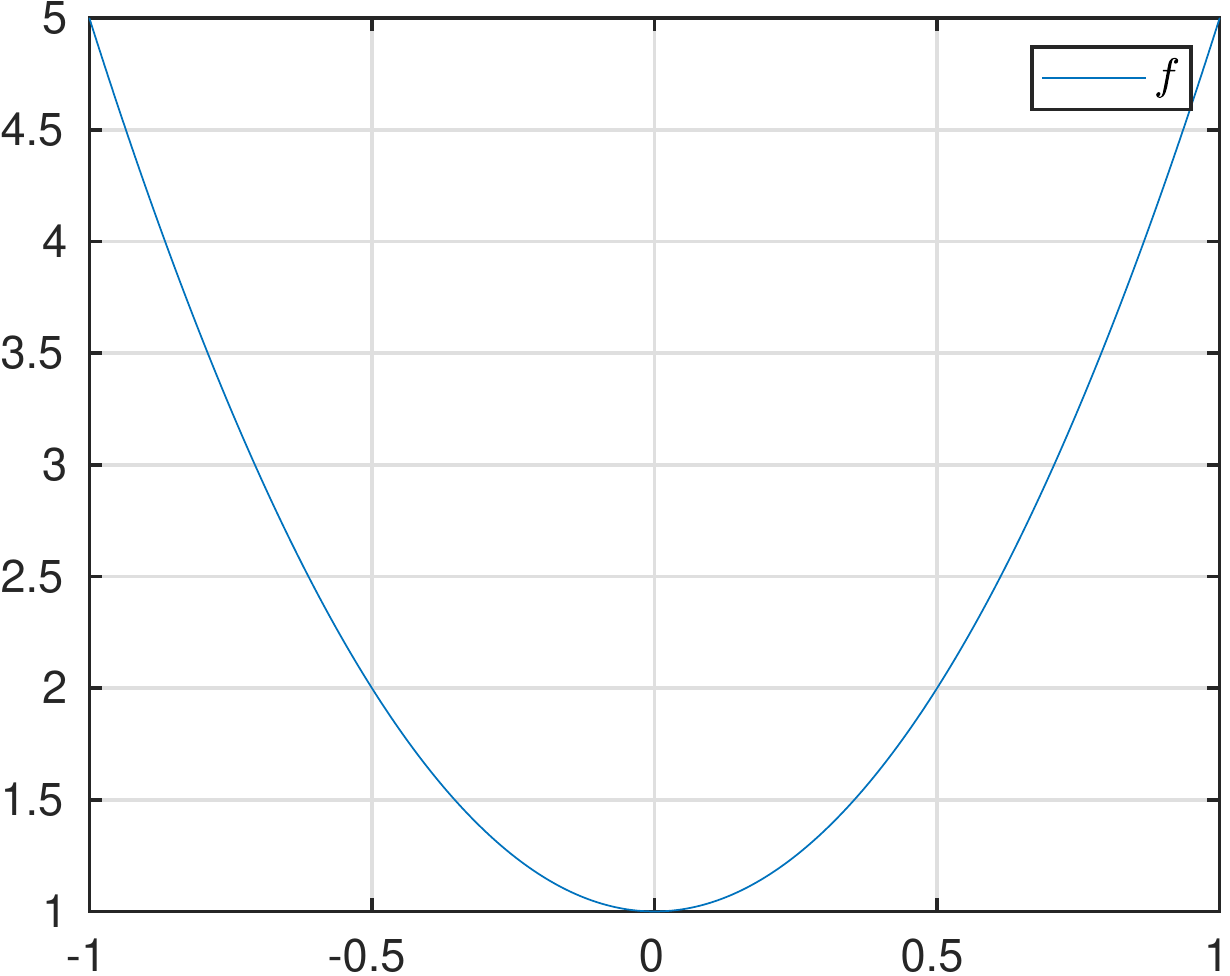}\\
\hspace{-1cm}
$(f,\rho) = (f_4,\mathrm{ReLU})$ & $(f,\rho) = (f_4,\mathrm{ReLU})$&  $(f,d) = (f_4,1)$ \\ 
\hspace{-1cm}
\includegraphics[scale=0.2]{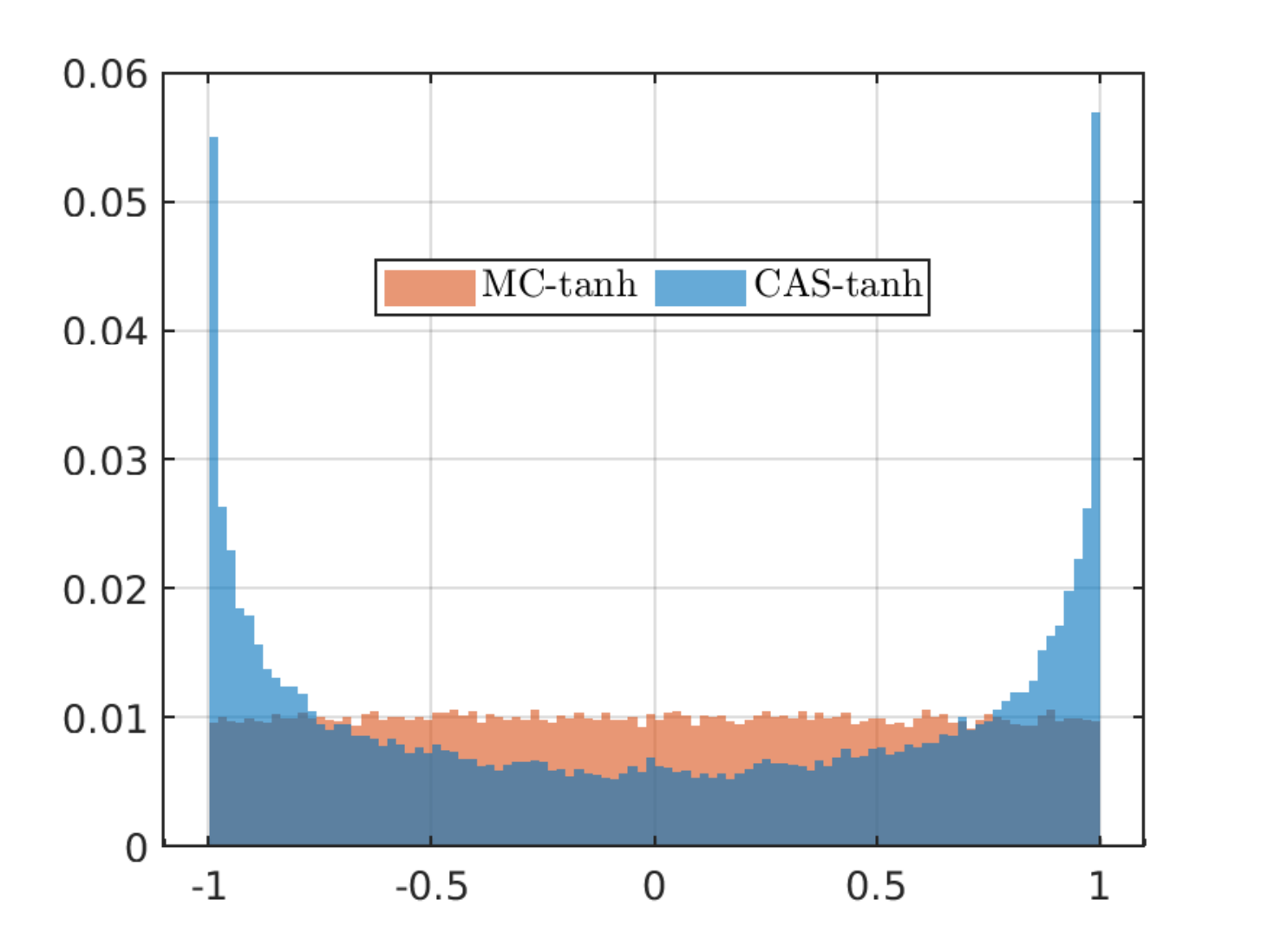} & 
\includegraphics[scale=0.2]{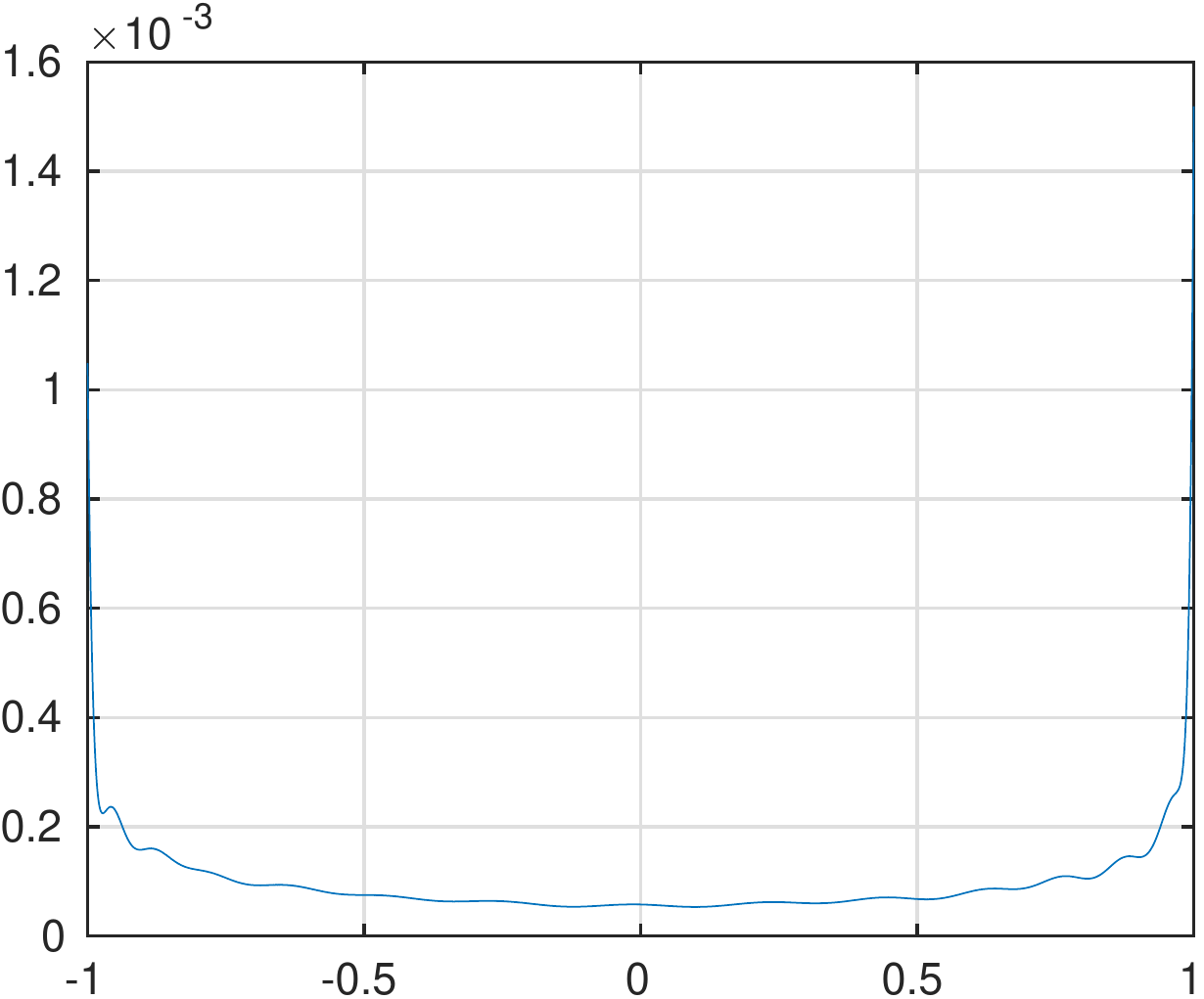} &
\includegraphics[scale=0.2]{train_data_example_5_dim_1-eps-converted-to.pdf}\\
\hspace{-1cm}
$(f,\rho) = (f_4,\tanh)$ & $(f,\rho) = (f_4,\tanh)$  & $(f,d) = (f_4,1)$ \\
\hspace{-1cm}
\includegraphics[scale=0.2]{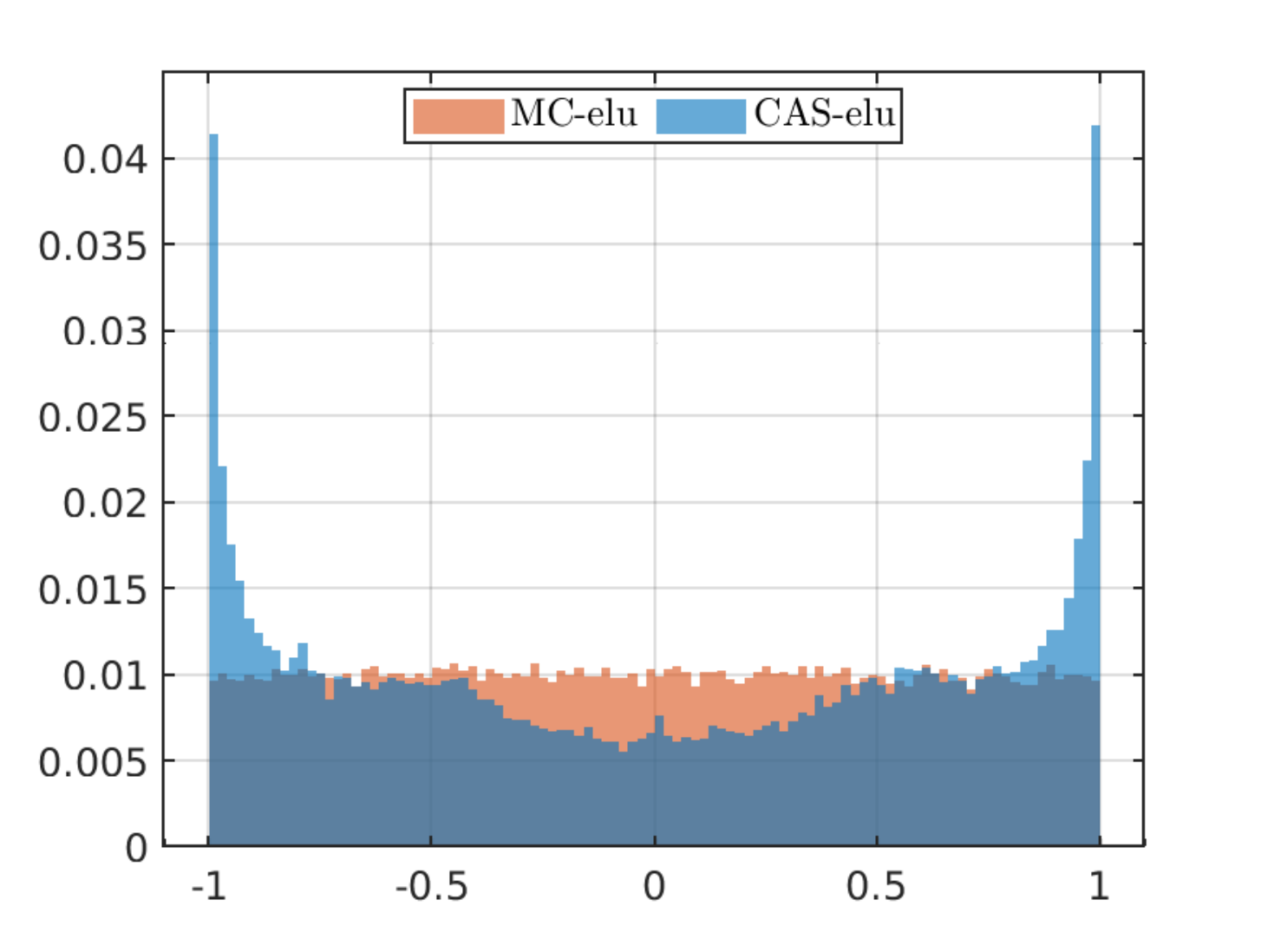} & 
\includegraphics[scale=0.2]{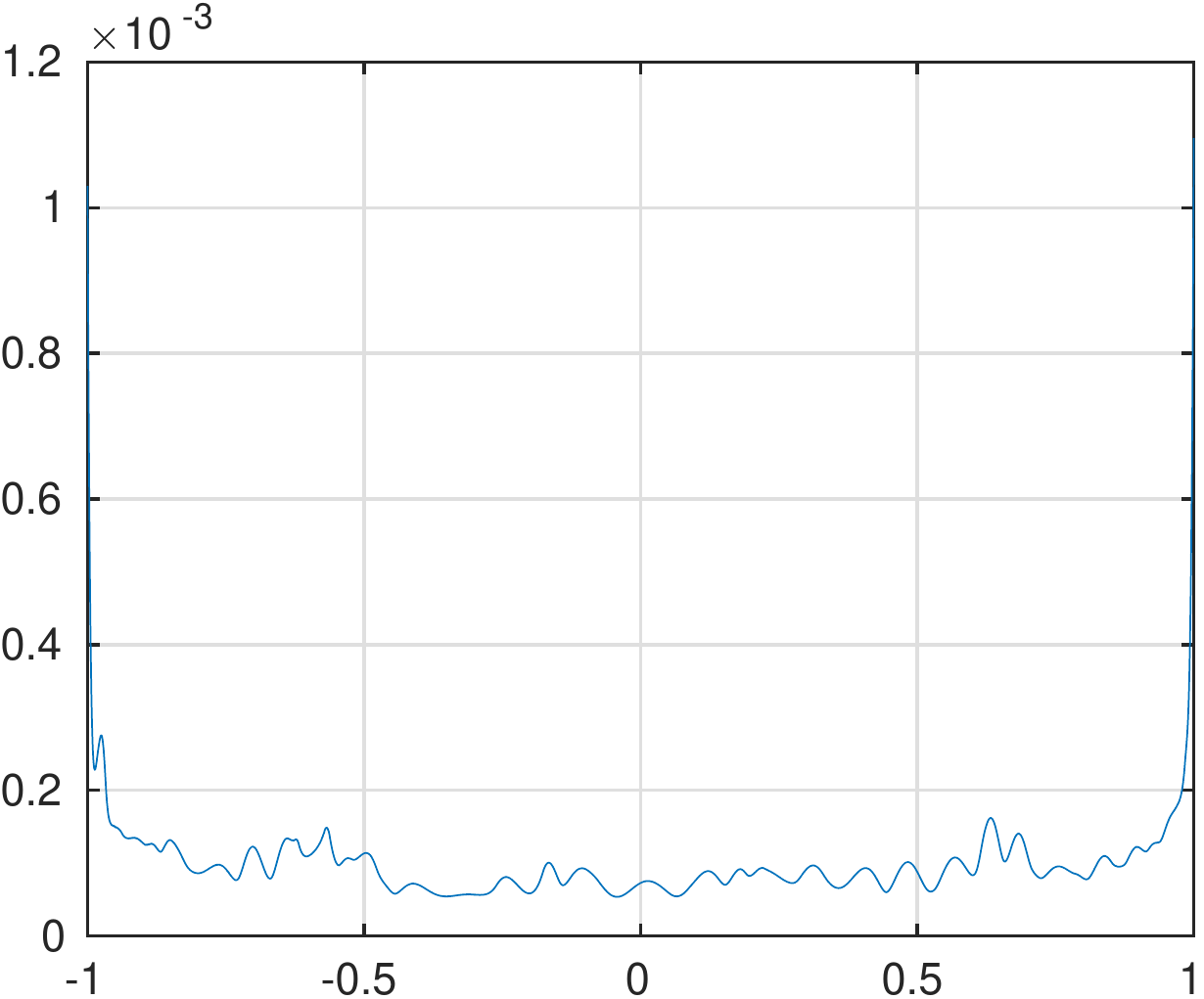} &
\includegraphics[scale=0.2]{train_data_example_5_dim_1-eps-converted-to.pdf}\\
\hspace{-1cm}
$(f,\rho) = (f_4, $ELU$)$ & $(f,\rho) = (f_4, $ELU$)$ &   $(f,d) = (f_4,1)$ \\
\end{tabular}
}
\caption{The learned sampling distributions for the functions $f = f_1$ and $f = f_4$ in $d = 1$ dimensions using ReLU, $\tanh$ and ELU $5 \times 50$ DNNs. \textbf{First column:} Histograms of the samples used in MC and CAS4DL, respectively, at the final iteration. \textbf{Second column:} The probability distribution function obtained at the final iteration of CAS4DL. \textbf{Third column:} Plots of the functions.} 
\label{fig:hist_prob_chris_1d_ex1_ex4}
\end{figure}

In \figref{fig:hist_prob_chris_1d_ex1_ex4}, we examine the sampling distributions learned by CAS. This figure considers the functions $f = f_1$ and $f = f_4$ in one dimension. It plots the histogram of the full set of $m = 5000$ samples chosen by CAS, along with the Christoffel function of subspace $P^{(l)}$ learned at the final step. When compared to MC sampling, the adaptivity of the CAS procedure is clearly evident. For both functions, there is a higher density of samples near the endpoints. Additionally, for $f = f_1$ we see a higher density around the point $y = -1$, where the function is varying more rapidly, and for $f = f_3$ we see a higher density around the points $y = -1$ and $y = 1$, where the function is peaked. Note that a higher sampling at the endpoints (boundary) is generally expected, when approximating smooth functions on compact intervals (or hypercubes) \cite{APS18}.

These effects are also reflected in the reciprocal Christoffel function \R{Christoffel-fn}. This function is obtained from the orthonormal system $\{\phi_j\}_{j=1}^{n_{\mathrm{final}}}$, which is computed by the SVD decompostion with thresholding, where $n_{\mathrm{final}}$ is the rank of the matrix $\bm{U}$ at the last iteration. As explained in Algorithm \ref{alg:CAS}, this is the maximum index $i$ such that $\sigma_i/\sigma_1>10^{-6}$, where $\sigma_i$ are the singular values of the dictionary $\{\psi_j^{(l_{\mathrm{final}})}\}_{j=1}^{N}$  
and $l_{\mathrm{final}}$ is the last iteration. As expected, the reciprocal Christoffel function is large near regions of high sampling density. This is indicative of the fact that the learned subspaces $P^{(l)}$ are adapted to the function being approximated.

\begin{figure}[h!]
\centering
{\small
\begin{tabular}{cccc}
\includegraphics[scale=0.2]{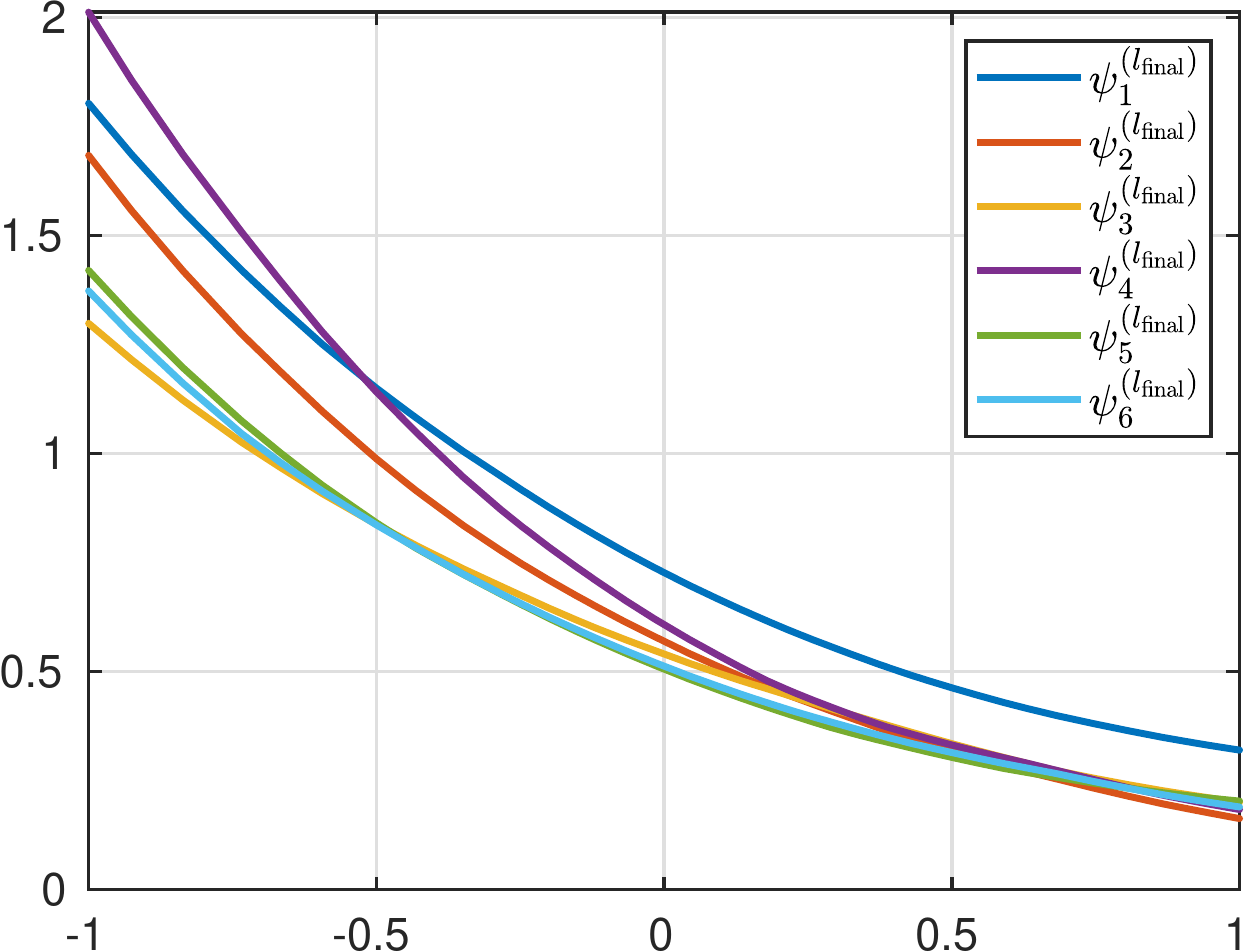} & 
\includegraphics[scale=0.2]{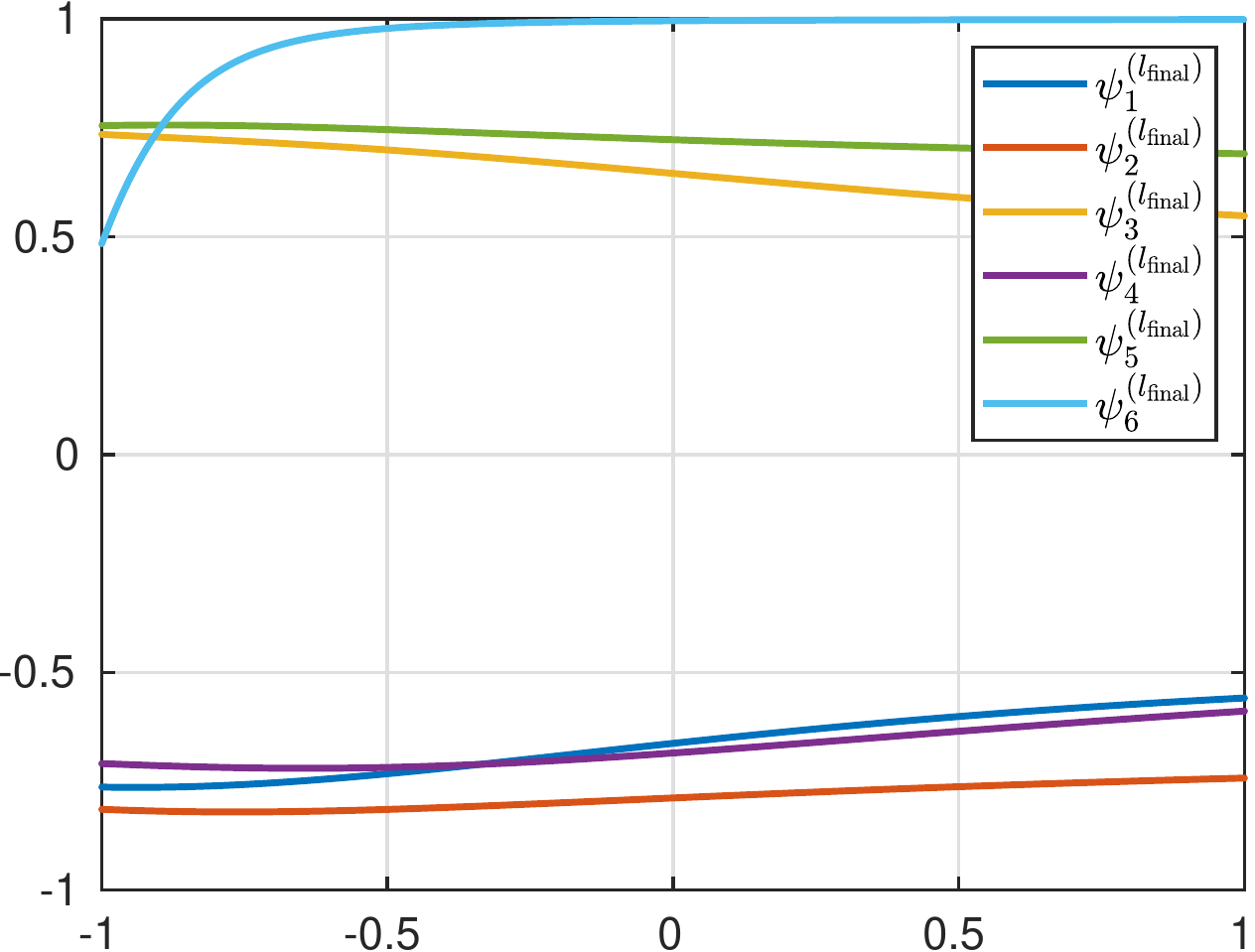} &
\includegraphics[scale=0.2]{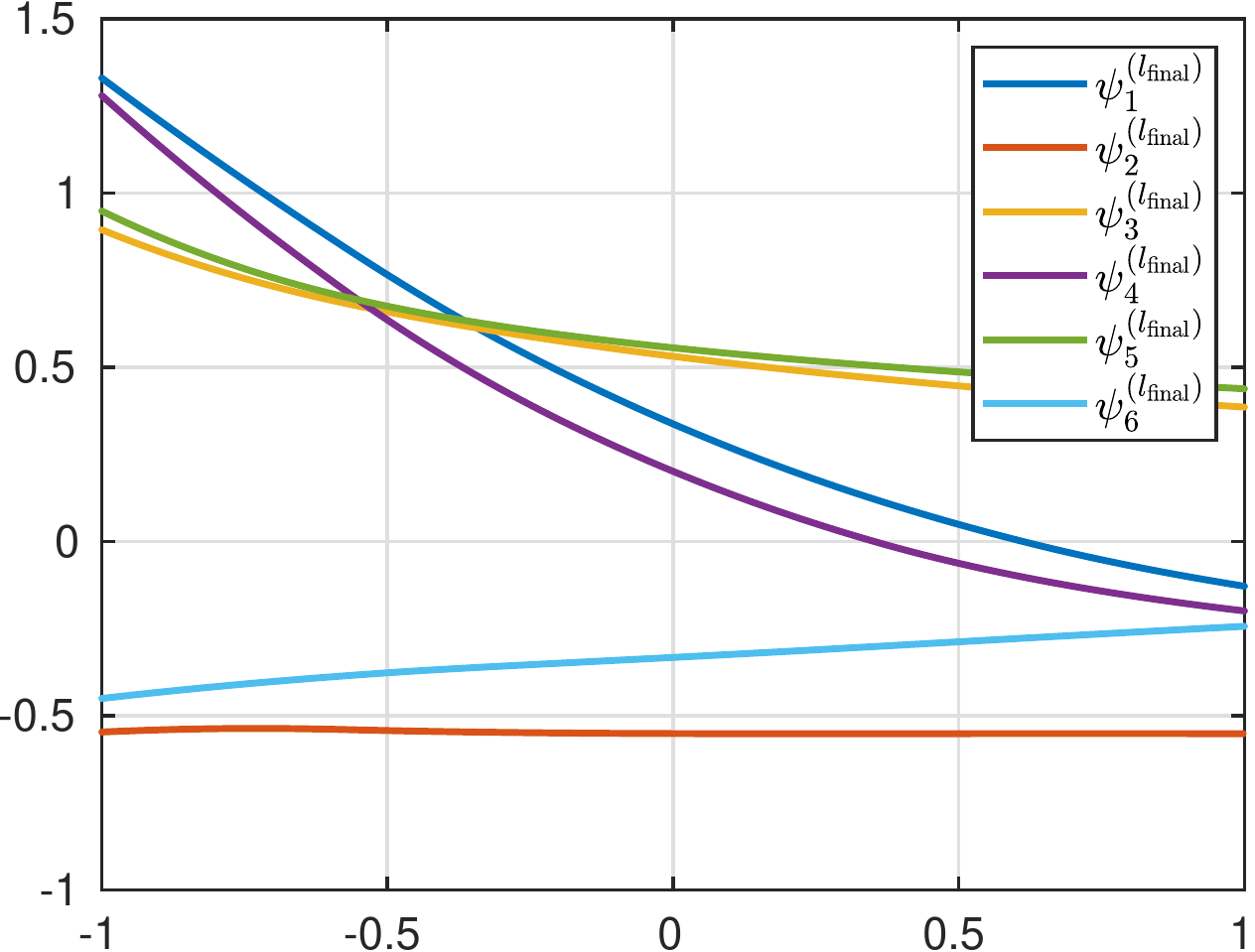} & 
\includegraphics[scale=0.2]{train_data_example_1_dim_1-eps-converted-to.pdf}\\
$(f,\rho) = (f_1,\mathrm{ReLU})$ & $(f,\rho)=(f_1,\tanh)$ & $(f,\rho)=(f_1,$ELU$)$ & $(f,d)=(f_1,1)$\\
\includegraphics[scale=0.2]{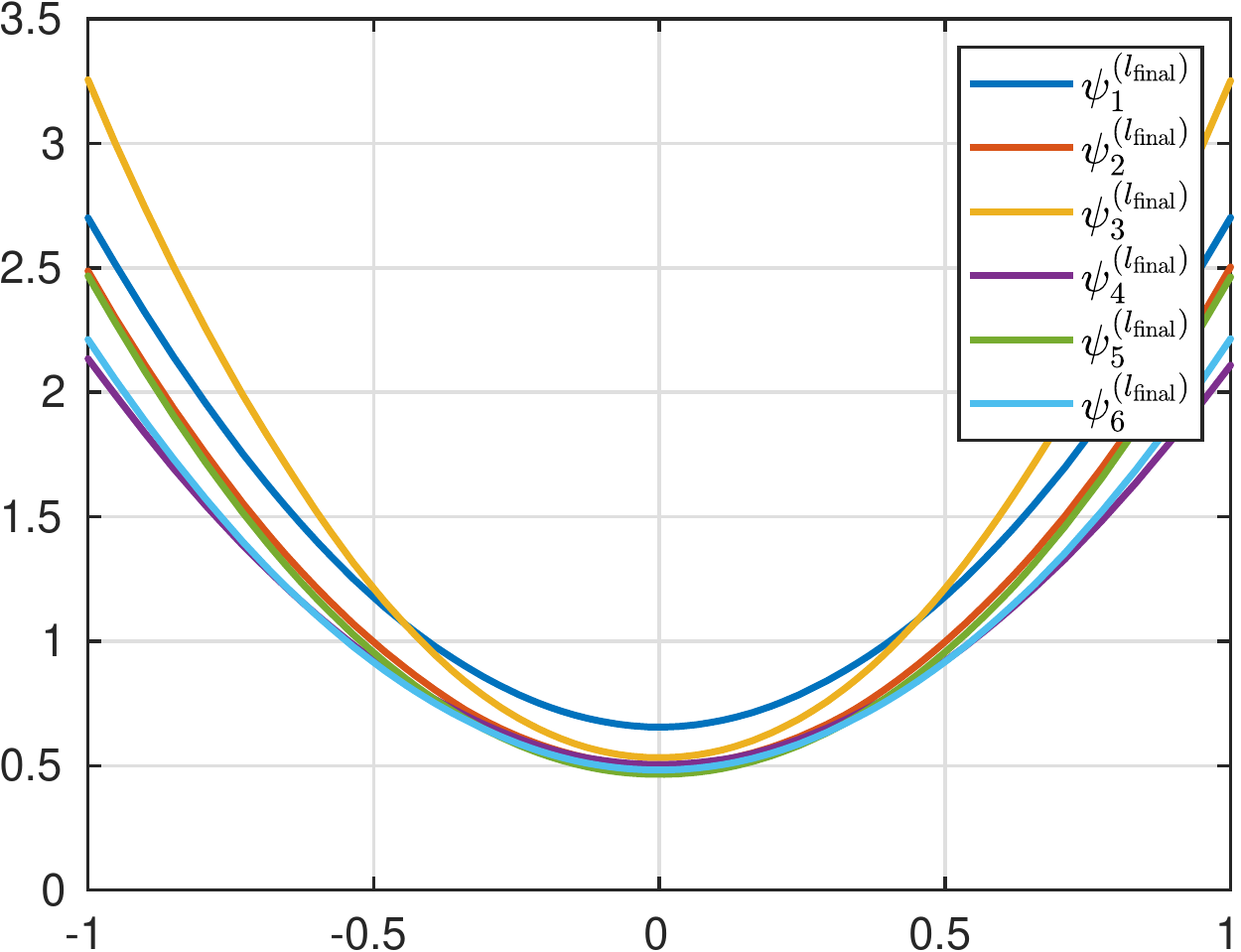} & 
\includegraphics[scale=0.2]{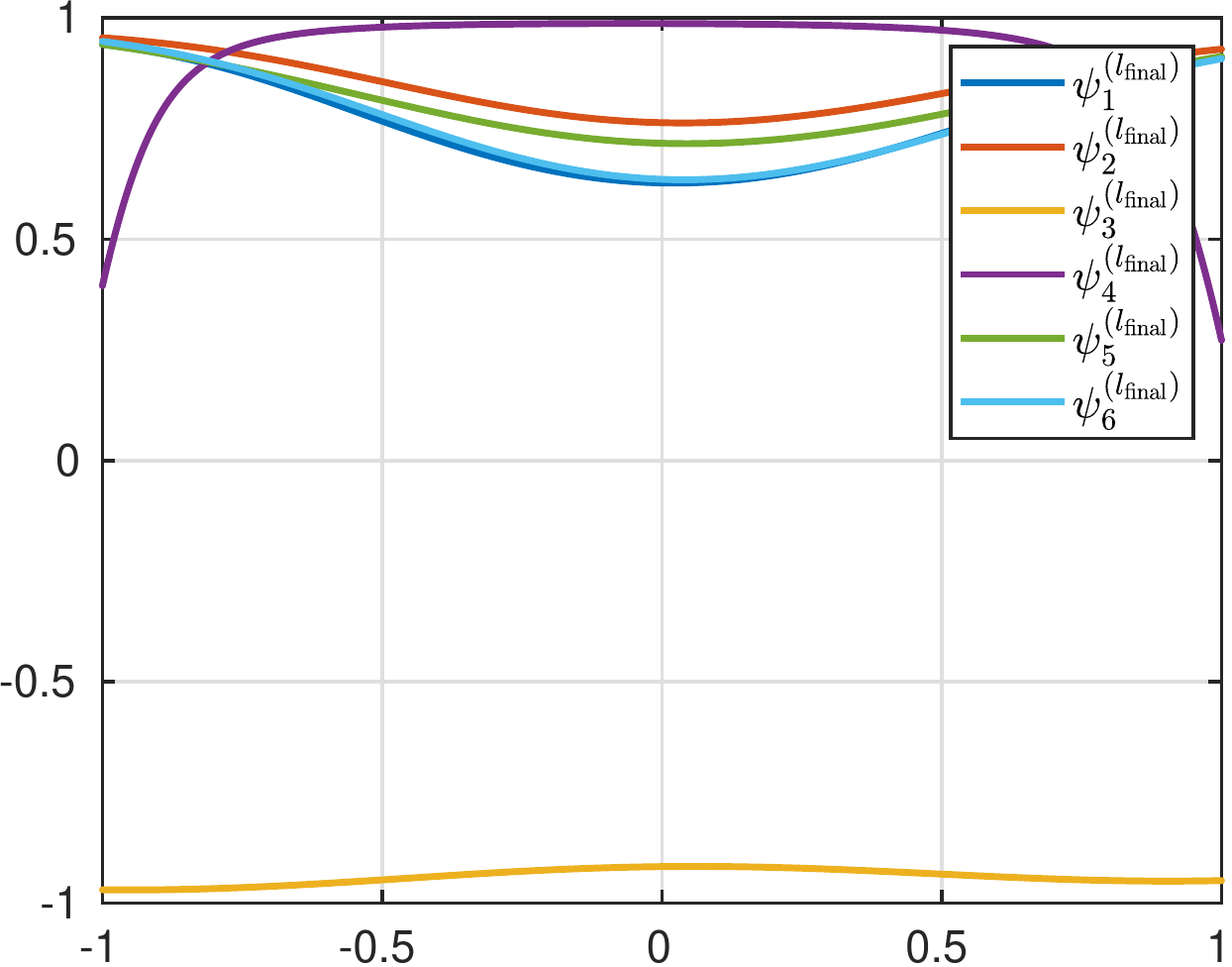} &
\includegraphics[scale=0.2]{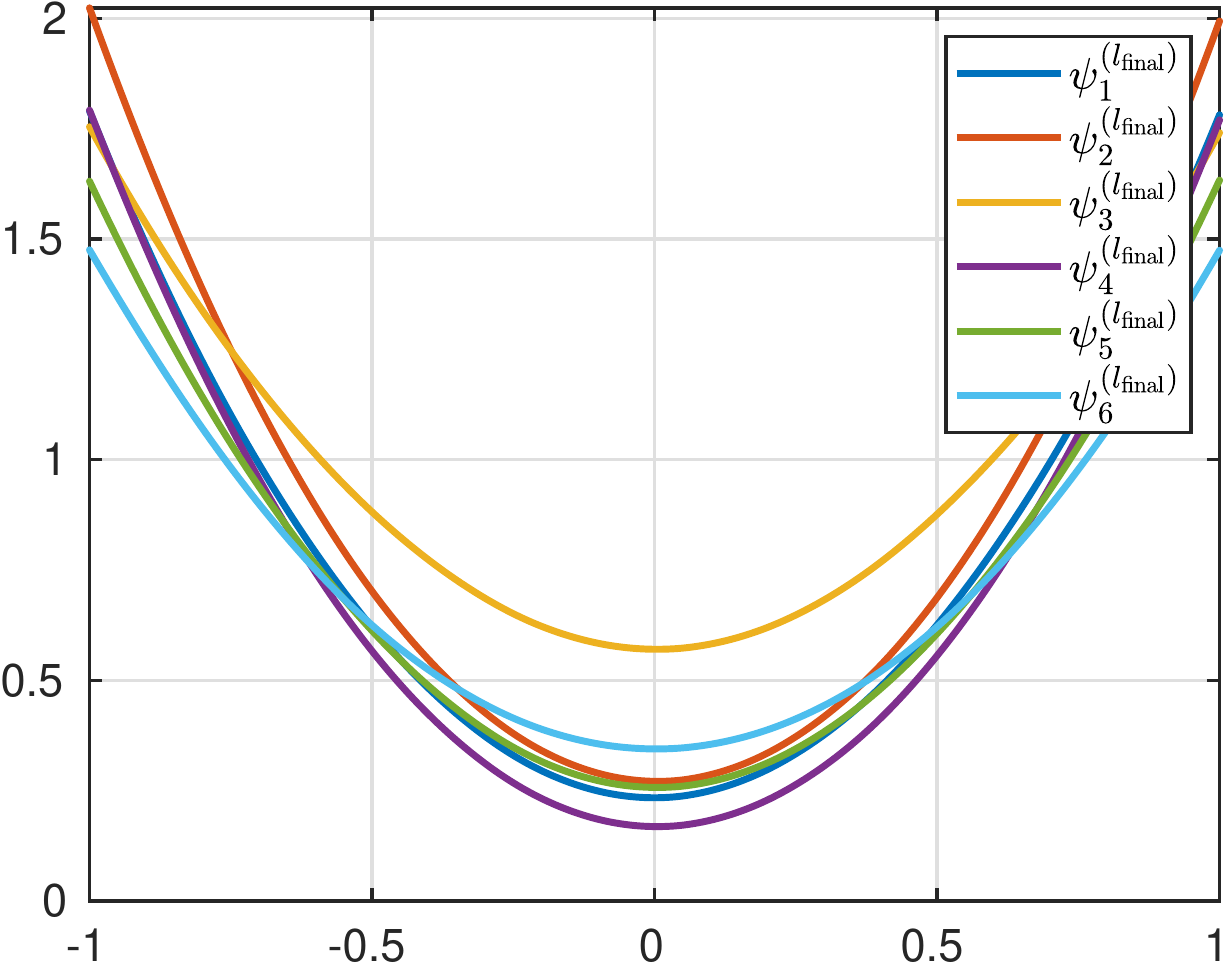} & 
\includegraphics[scale=0.2]{train_data_example_5_dim_1-eps-converted-to.pdf}\\
$(f,\rho) = (f_4,\mathrm{ReLU})$ & $(f,\rho)=(f_4,\tanh)$ & $(f,\rho)=(f_4,$ELU$)$ & $(f,d)=(f_4,1)$
\end{tabular}
}
\caption{The first six learned dictionary functions $\{\psi_j^{(l_{\mathrm{final}})}\}_{j=1}^{6}$ obtained at the final iteration for the functions $f = f_1$ and $f = f_4$ in $d = 1$ dimensions using ReLU, $\tanh$ and ELU $5 \times 50$ DNNs.}
\label{fig:Basis_1D_f1_f4}
\end{figure}

This adaptivity is also evident in the learned dictionary functions $\{\psi_j^{(l_{\mathrm{final}})}\}_{j=1}^{N}$. In \figref{fig:Basis_1D_f1_f4}, we show the first six such functions learned by CAS4DL after the training procedure. 
In this figure, we present the first six dictionary functions obtained by the DNN using ReLU, $\tanh$ and ELU as activation functions for the functions $f_1$ and $f_4$. Here the order is decreasing in magnitude of the product of the corresponding coefficients $c_i$ with the reciprocal of the $L^2$ norm of the dictionary elements. It is noticeable that for these architectures the dictionary elements learned by the CAS4DL method resemble or share features with the target function.

\begin{figure}[h!]
\centering 
{\small
\begin{tabular}{cccc}
\hspace{-0.4cm}  
\includegraphics[scale=0.2]{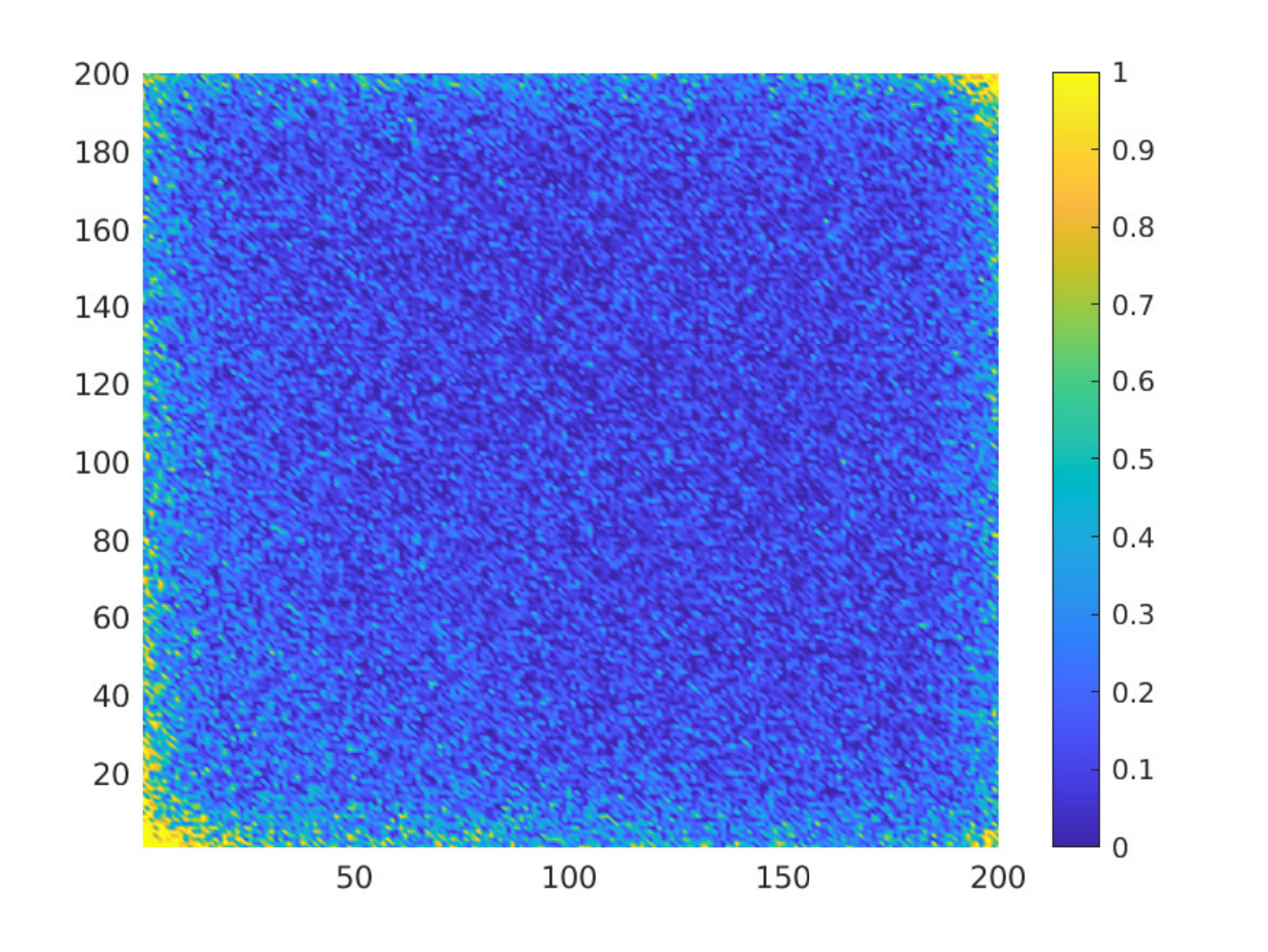} & 
\hspace{-0.6cm}
\includegraphics[scale=0.2]{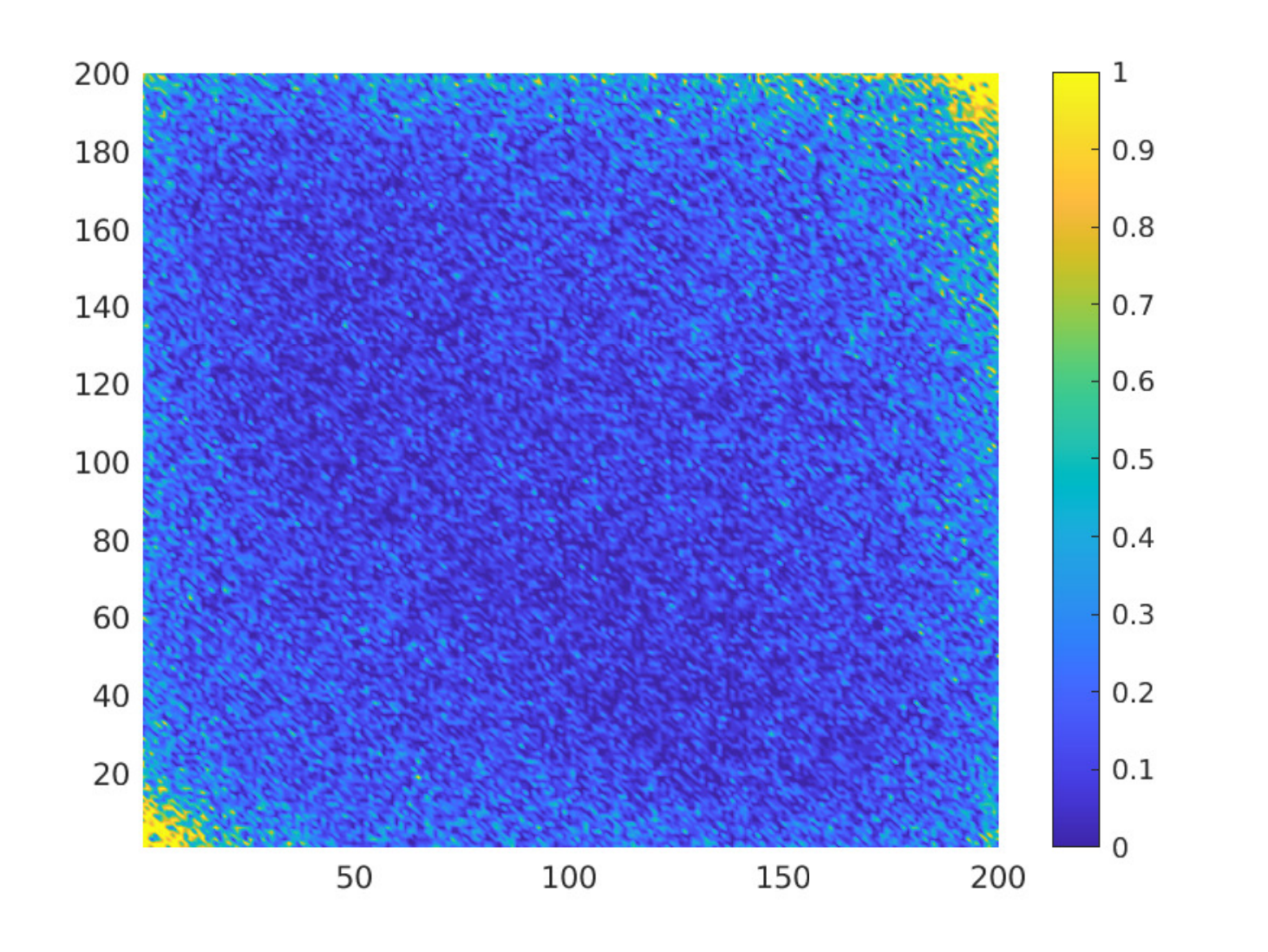} & 
\hspace{-0.6cm}
\includegraphics[scale=0.2]{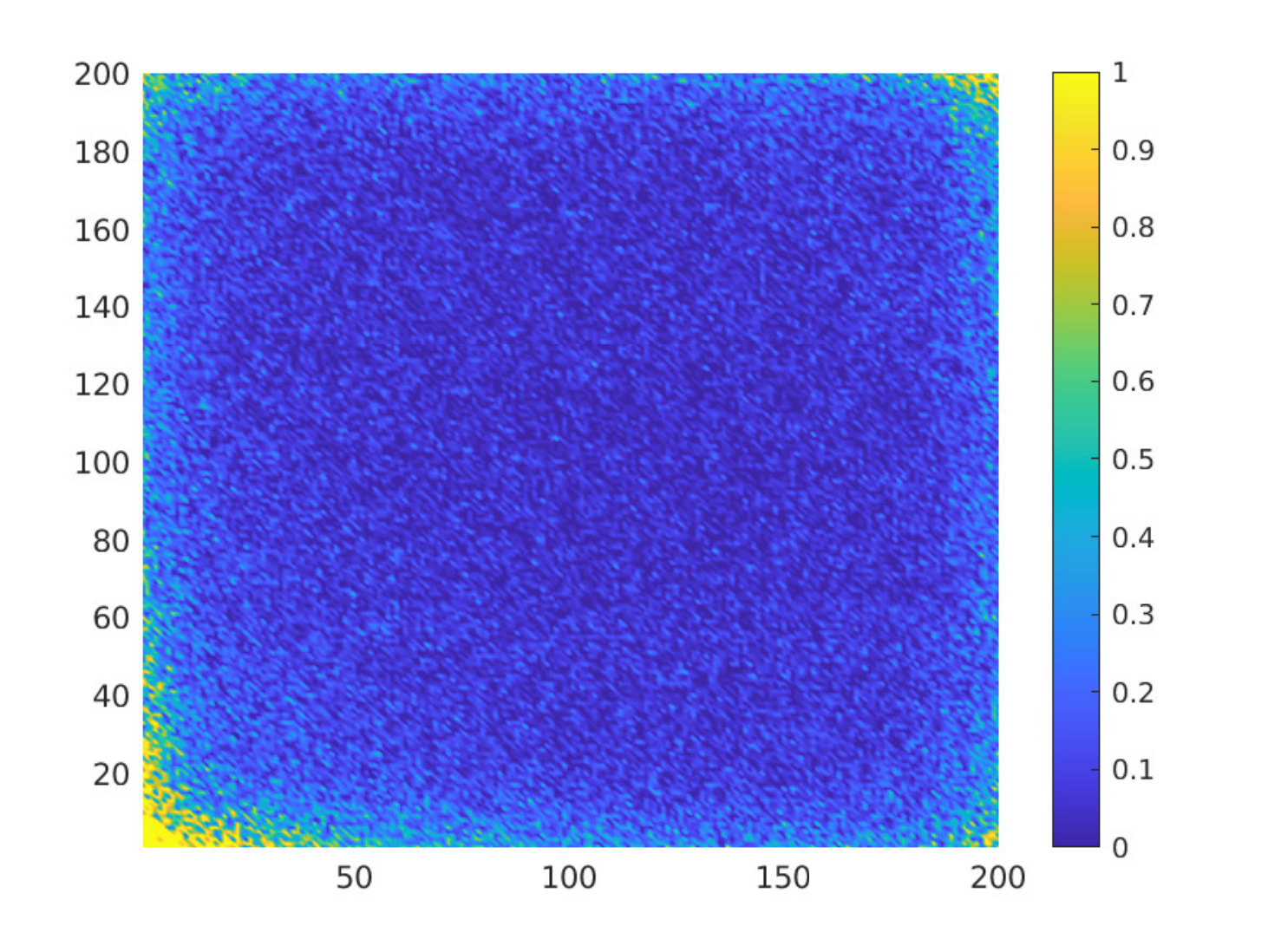} & 
\hspace{-0.6cm}
\includegraphics[scale=0.2]{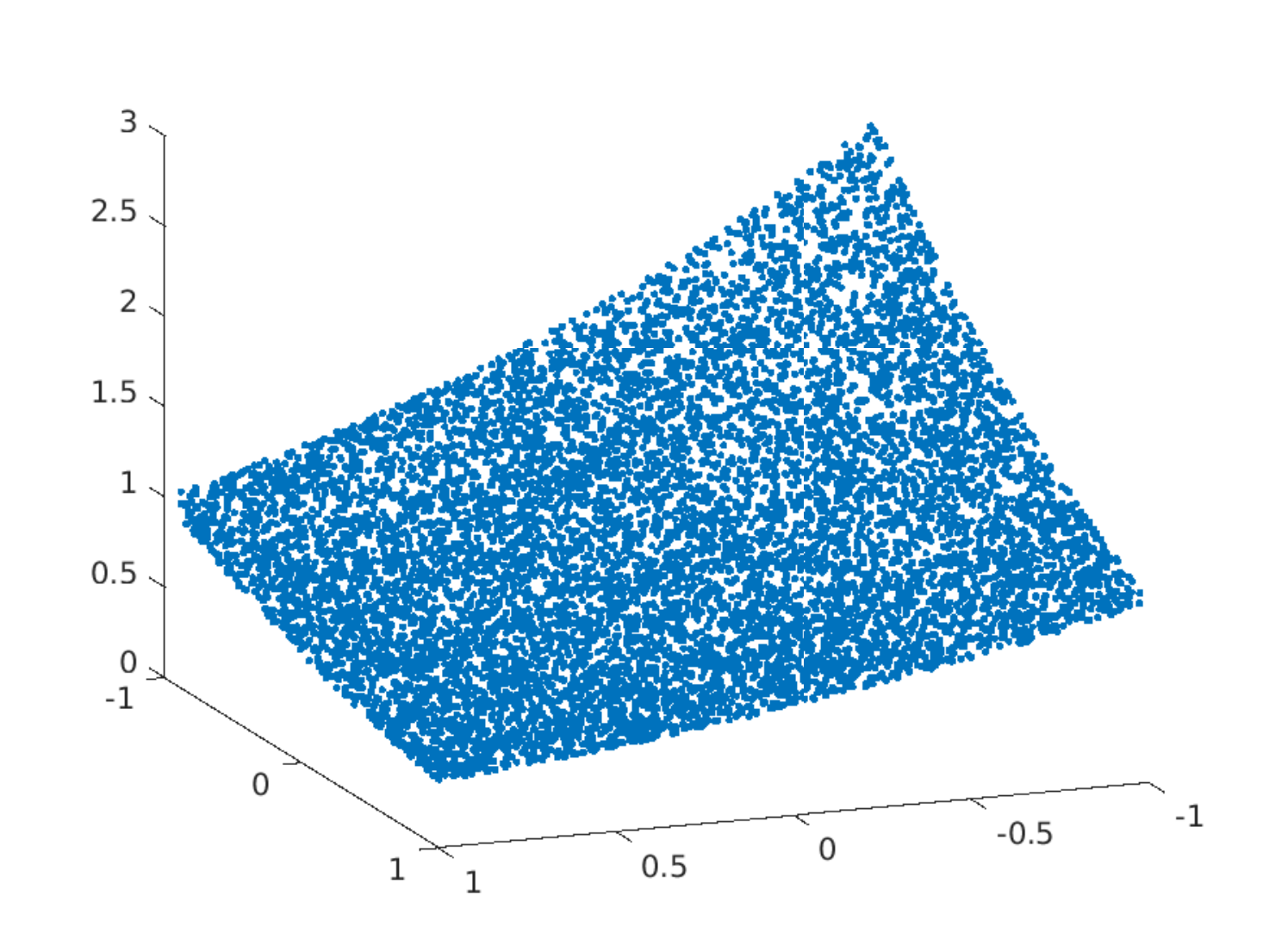}\\
\hspace{-0.4cm} 
$(f,\rho) = (f_1,\mathrm{ReLU})$ & 
\hspace{-0.6cm}
$(f,\rho)=(f_1,\tanh)$ & 
\hspace{-0.6cm}
$(f,\rho)=(f_1,$ELU$)$ & 
\hspace{-0.6cm}
$(f,d) = (f_1,2)$ \\
\hspace{-0.4cm} 
\includegraphics[scale=0.2]{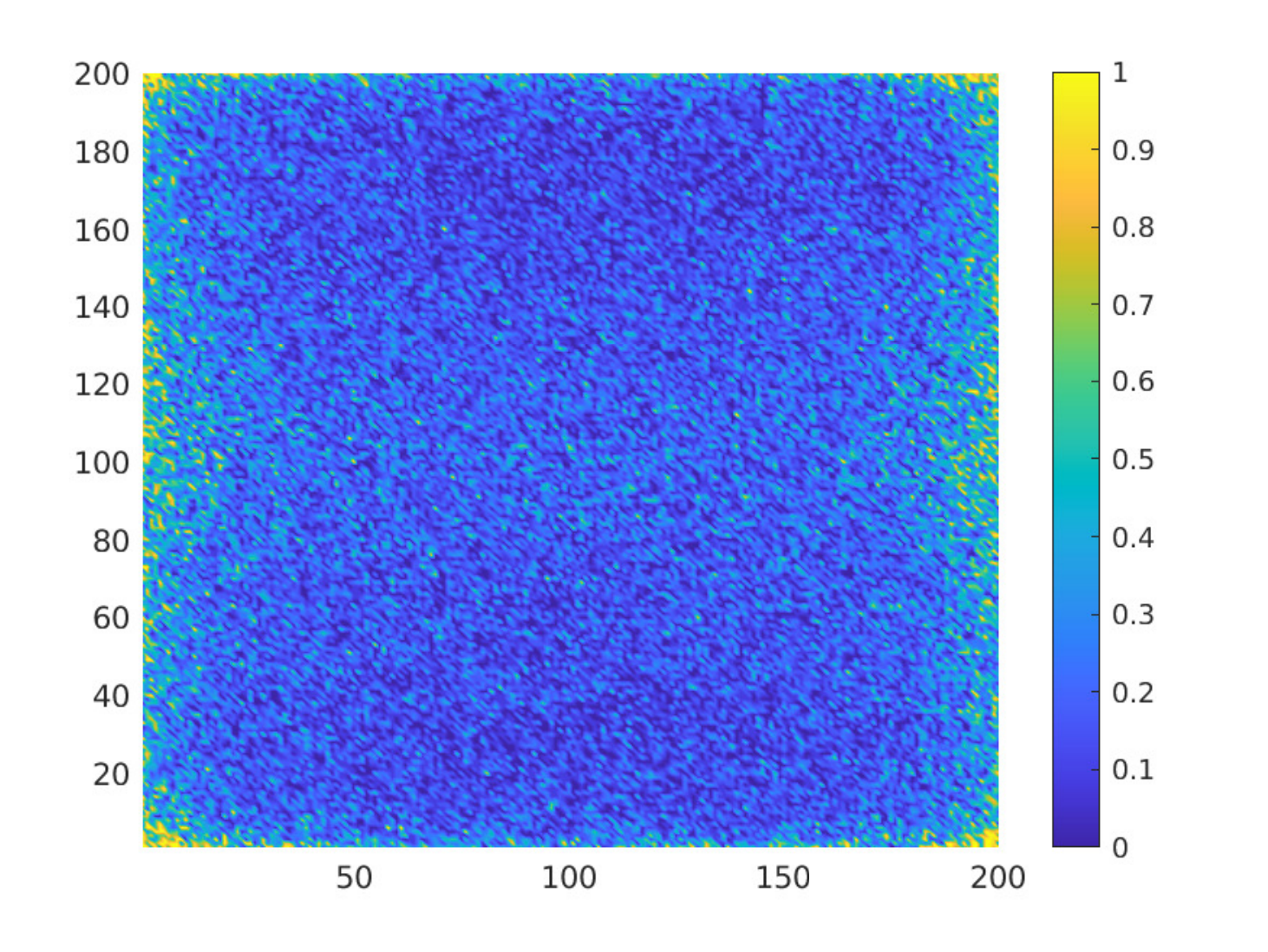} & 
\hspace{-0.6cm}
\includegraphics[scale=0.2]{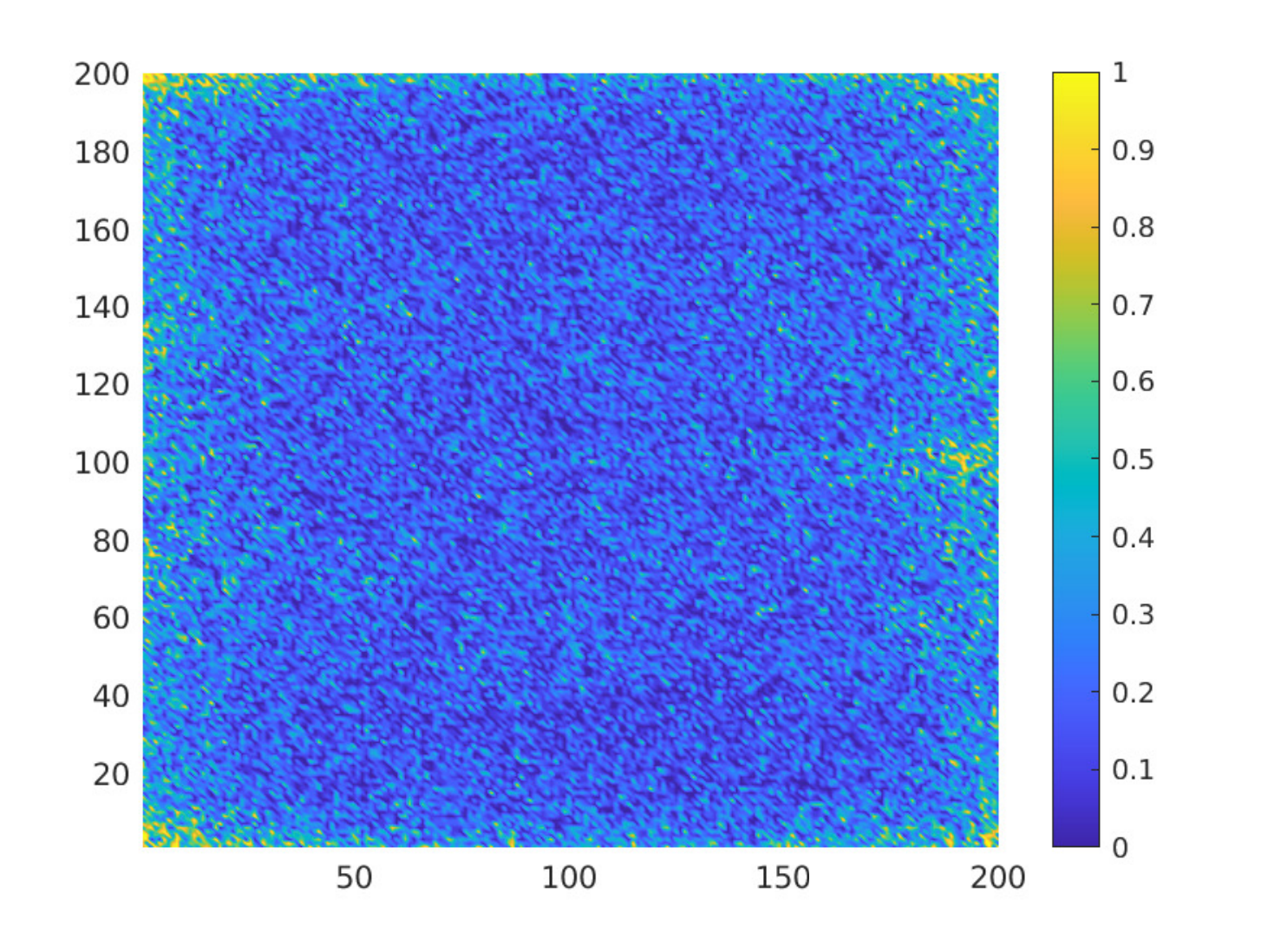} & 
\hspace{-0.6cm}
\includegraphics[scale=0.2]{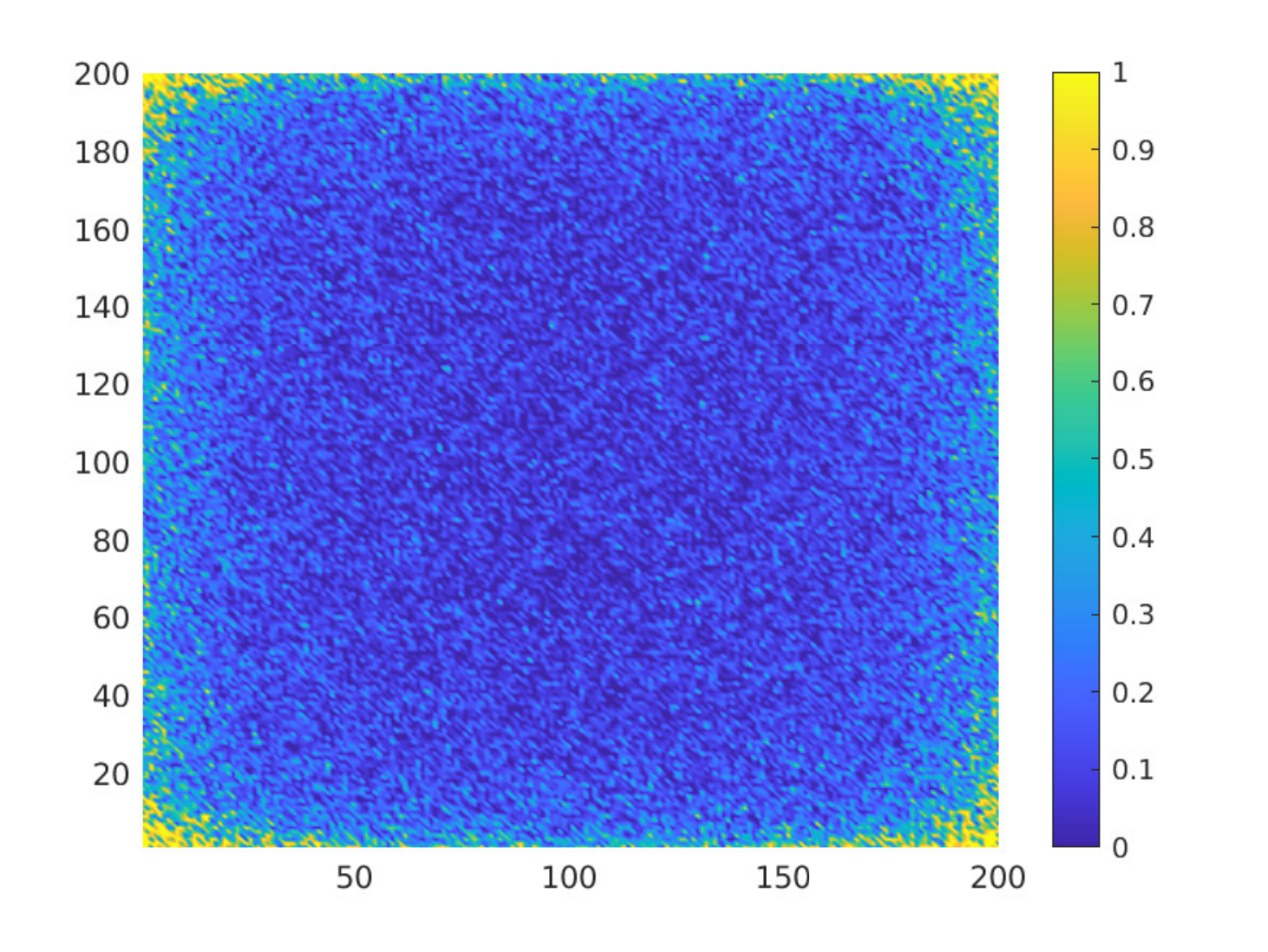} & 
\hspace{-0.6cm}
\includegraphics[scale=0.2]{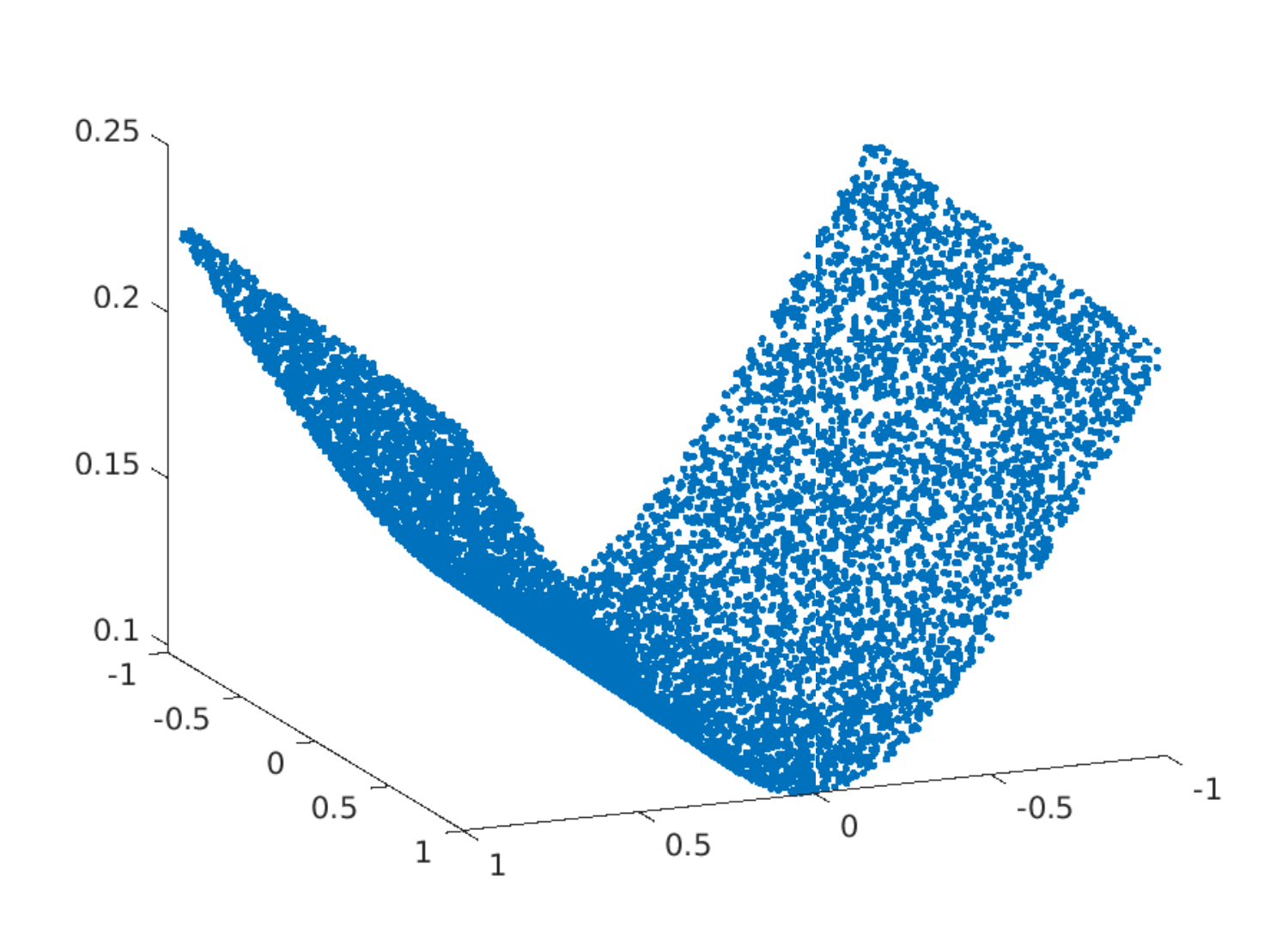}\\
\hspace{-0.4cm} 
$(f,\rho) = (f_4,\mathrm{ReLU})$ & 
\hspace{-0.6cm}
$(f,\rho)=(f_4,\tanh)$ & 
\hspace{-0.6cm}
$(f,\rho)=(f_4,$ELU$)$ & 
\hspace{-0.6cm}
$(f,d) = (f_4,2)$ \\
\end{tabular}
}
\caption{The learned sampling distributions for the functions $f = f_1$ and $f = f_4$ in $d = 2$ dimensions using ReLU, $\tanh$ and ELU $5 \times 50$ DNNs. \textbf{First, second and third columns:} Histograms of the samples used in CAS4DL at the final iteration. \textbf{Third column:} plots of the functions.} 
\label{fig:Histogram_2D}
\end{figure}

Finally, we also plot the histograms of the sampling distributions for CAS4DL in $d=2$ dimensions in \figref{fig:Histogram_2D}. The function $f=f_1$ exhibits a drastic change near the corners of the domain, and we observe that CAS4DL samples more densely over these areas. For $f=f_4$ we notice that the points tend to accumulate on the boundaries, helping to learn the function where its values change most quickly. 

\vspace*{\fill}
\vspace*{\fill}

\newpage
\vspace*{\fill}

\vspace*{\fill}

\newpage
\subsection{The effect of the numerical dimension $n$ on the error}
\label{ss:rank_error}

In Figures \ref{fig:comp_act_L2_error_example_1}--\ref{fig:comp_act_L2_error_example_5} we also show the average numerical dimensions $n_1,n_2,\ldots$ for the learned subspaces. 
There we observe a significant difference between the numerical dimension of ReLU, $\tanh$ and ELU DNNs in the lower-dimensional examples ($d= 1,2$), with the ReLU activation resulting in 40-50\% lower numerical dimension than ELU, and the $\tanh$ activation resulting in 45-70\% lower numerical dimension than ELU. Note that the maximum dimension here is $N$, the number of nodes. We also note that when $d = 1$ neither of the DNNs achieves full numerical dimension, suggesting that their corresponding dictionaries are highly redundant. For $d \ge 2$ the ELU activation function gives nearly full numerical dimension, and in $d=2$ (for $f=f_2$), $d=8$ and $d=16$ this is true for $\tanh$ as well.  
In high dimensions, the numerical dimensions obtained by these architectures and sampling procedures are close to the number of nodes $N$. Therefore, the dictionaries $\{\phi^{(l)}_j\}_{j=1}^{N}$ learned for each $l\geq 1$ via both CAS and MC are well conditioned. This is not the case, however, for ReLU, where the numerical dimension is consistently around 60\% lower than ELU or tanh in higher dimensions. However, we note an exception for $f = f_4$ in $d = 16$ dimensions. 

Interestingly, there is also a close correspondence between the numerical dimension and the comparative accuracies of the different schemes. ELU DNNs typically outperform $\tanh$ DNNs when their numerical dimensions are higher. However, when ELU and $\tanh$ DNNs yield full-numerical dimension dictionaries, $\tanh$ DNNs typically outperform ELU and ReLU DNNs.  

\subsection{Application to parametric PDEs}
\label{sec:app_parametric_PDEs}

In this section, we present a parametric PDE example motivated by UQ for applications in computational science and engineering. We consider the parametric elliptic diffusion equation, defined on the physical domain $D = (0,1)^2$ and the parametric domain $\Omega = [-1,1]^d$. In particular, given $g\in L^2(D)$, we seek a function $u:\Omega\times D\rightarrow \cV$ that satisfies 
\ea{
\label{PDE}
-\nabla \cdot (a(\bm{x},\bm{y}) \nabla u(\bm{x},\bm{y})) &=g(\bm{x}),\  \forall \bm{x} \in D,\, \bm{y} \in \cU,\\
\label{PDE:init_cond}
u(\bm{x},\bm{y})&= 0,\ \forall \bm{x} \in \partial D,\, \bm{y} \in \cU.
}
This is a standard benchmark parametric PDE problem in UQ, see \cite{dalsanto2020data,Dung2021,DungEtAl2021,heiss2021neural,lei2022solving,li2020fourier}. Also, for simplicity, we take $g(\bm{x}) = 10$, as a constant function. As for the diffusion coefficient, we consider the coefficient from \cite[Eqn.\ (24)]{adcock2021deep}, modified from an earlier example from \cite[Eqn.\ (5.2)]{NTW08}, with  one-dimensional (layered) spatial dependence defined by 
 
\ea{
\label{eq:diff_coeff}
a(\bm{x},\bm{y}) 
& = \exp
\left(
1+ y_1 \left(\frac{\sqrt{\pi}\beta}{2}\right)^{1/2} + \sum_{i=2}^d \; \zeta_i \; \vartheta_i(\bm{x}) y_i
\right)  \\
\zeta_i 
& := (\sqrt{\pi} \beta)^{1/2} \exp{\left((-\lfloor i/2 \rfloor\pi \beta)^2/8\right)}
\nonumber\\
\vartheta_i(\bm{x}) 
& := \begin{cases} 
\sin\left( \left\lfloor i/2 \right\rfloor \pi x_1/\beta_p \right) & \mbox{$i$ even} \\
\cos\left(\left\lfloor i/2 \right\rfloor \pi x_1/\beta_p \right) &  \mbox{$i$ odd} 
\end{cases}
\nonumber\\
\beta_c &= 1/8 ,\ \beta_p = \max\{1, 2 \beta_c\},\ \beta = \beta_c/\beta_p.
\nonumber
} 
We consider this problem with parameter dimensions $d=1,2,4,8$ and $16$. In this case, the solution map $f=u(\cdot,)$ is a Hilbert-valued function defined as follows 
\bes{
f:\Omega \rightarrow \cV,\ \bm{y}\in\Omega \mapsto u(\cdot,\bm{y})\in\cV,
}
where $\cV$ is the Sobolev space $H^1_0(D)$. 
We can rewrite this problem as a approximating the unknown function $u:\Omega\rightarrow\cV$ from \textit{sample values}
\be{
\label{u-snapshots}
u(\cdot,\bm{y}_1),\ldots,u(\cdot,\bm{y}_m).
}
Notice that since $\cV$ is infinite dimensional, we cannot work explicitly with this space. Then, we consider a finite-dimensional discretization  
\begin{equation}
\label{eq:conforming}
\cV_h \subseteq \cV.
\end{equation}
Here $h > 0$ denotes a discretization parameter, e.g., the mesh size in the case of a finite element discretization. In the finite elements community, the assumption of \eqref{eq:conforming} means that we consider the so-called \textit{conforming} discretizations. We let $\{ \varphi_k \}^{J}_{k=1}$ be a (not necessarily orthonormal) basis of $\cV_h$, where $J = J(h) = \dim(\cV_h)$.  
Hence, we write the approximation of $u$ depending on the basis $\{\varphi_k\}_{k=1}^{J}$ as 
\be{
\label{eq:u_Phi_FE}
u\approx u_h(\bm{y}) = \sum_{k=1}^{J} c_k(\bm{y})\varphi_k.
}
 
In the previous section, our objective was to approximate the function $f$ from sample values. However, in this case our goal is to approximate the coefficients $c_k(\cdot)$ with a DNN.

Let us consider the standard feedforward DNN architectures (defined in Section \ref{sec:CAS4DL}) of the form  $\Psi : \bbR^d \rightarrow \bbR^J$. Specifically, we write 
\be{
\label{eq:Psi_DNN_PDE}
\Psi(\bm{y}) = \mathcal{A}_{L+1}\left(\sum_{i=1}^N \alpha_i\psi_i(\bm{y})\right),
}
where $\cA_{L+1} : \bbR^{N} \rightarrow \bbR^{J}$ is an affine map, $\alpha_i\in\mathbb{R}$ and $\psi_i:\mathbb{R}^d\rightarrow\mathbb{R}$ is the value of the DNN $\Psi$ in the $i$th neuron in its penultimate layer. For simplicity, we may write 
\bes{
\Psi = \bm{\alpha}^\top\bm{\Psi}, 
\quad \bm{\alpha}\in\mathbb{R}^N, 
\quad \bm{\Psi} : \bbR^d \rightarrow \bbR^N,\ \bm{y} \mapsto (\psi_i(\bm{y}) )^{N}_{i=1},
}
Now, given \eqref{eq:Psi_DNN_PDE}, we write 
\be{
\label{eq:u_Phi_DNN}
u_{\Psi,h}(\bm{y}) = \sum^{J}_{k=1} (\Psi(\bm{y}))_k \varphi_k,
} 
for the resulting approximation to $u$.

We use a finite element method for spatial discretization, based on a triangulation with $J = 1089$ nodes and meshsize $h = \sqrt{2}/32$. In this example, we consider only the error in approximating the parametric PDE, hence we use the same finite element discretization in generating the samples for the training and testing data. To train the DNN we consider the following \textit{weighted loss} function 
\begin{equation}
\label{eq:PDE_weighted_MSE_loss}
\cL(\bm{y}) := {\dfrac{1}{m}\sum_{i=1}^m} \sum^{J}_{k=1} {w(\bm{y}_i)(c_{k}(\bm{y}_i)  - (\Psi(\bm{y}_i))_k )^2}.
\end{equation} 
The training data is generated by solving \eqref{PDE} using \eqref{PDE:init_cond} at a set of uniform random points $\{\bm{y}_i\}_{i=1}^{m_{\max}}\in\Omega$, yielding a set of solutions $\{u_h(\cdot,\bm{y}_i)\}_{i=1}^{m_{\max}}$, where $u_{h}(\cdot,\bm{y})\in\cV_h$ is the computed solution of \eqref{PDE}-\eqref{PDE:init_cond}.

The testing data is generated in a similar manner by evaluating \eqref{PDE} at a set of points $\{\bm{y}_i\}_{i=1}^{m_{\textnormal{test}}}$. As we did earlier in the paper, instead of using random points to test, we use a deterministic high-order sparse grid stochastic collocation method to generate the set of testing points and data  $\{u_h(\cdot,\bm{y}_i)\}_{i=1}^{m_{\textnormal{test}}}$ (see \cite{NTW08}). Then, we compute the average relative error using the Bochner norms $L^2_\varrho(\Omega;L^2(D))$ and $L^2_\varrho(\Omega;H_0^1(D))$ (see for more details \cite[Equation(4)]{adcock2021deep}) 
\be{
E(u)=\frac{\nm{u_h-u_{\Psi,h}}_{L^2_\varrho(\Omega;\cV)}}{\nm{u_h}_{L^2_\varrho(\Omega;\cV)}}.
}

\begin{figure}[h]
\centering
{\small
\begin{tabular}{ccc} 
\includegraphics[scale=0.11]{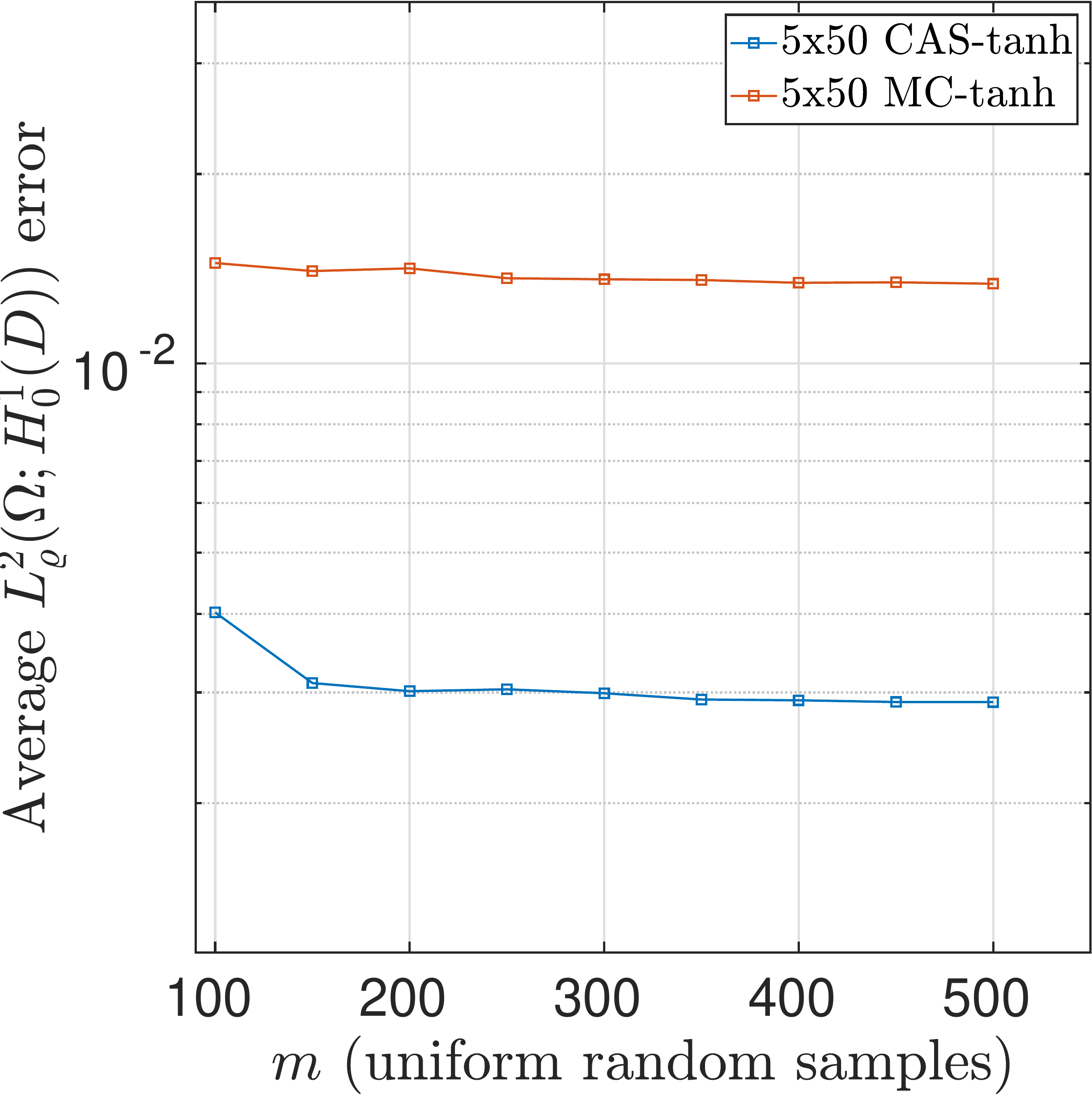} & 
\includegraphics[scale=0.11]{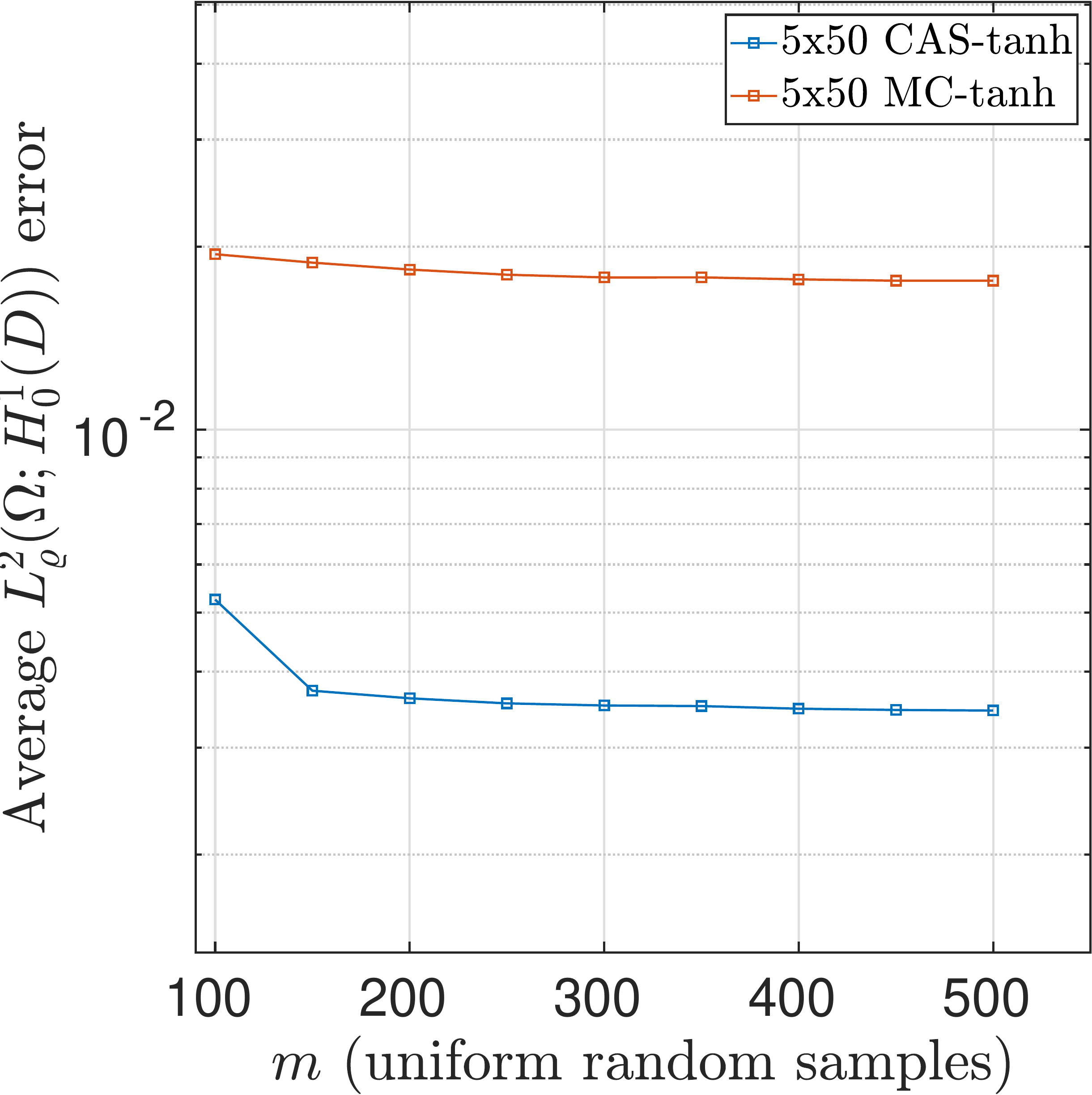} &
\includegraphics[scale=0.11]{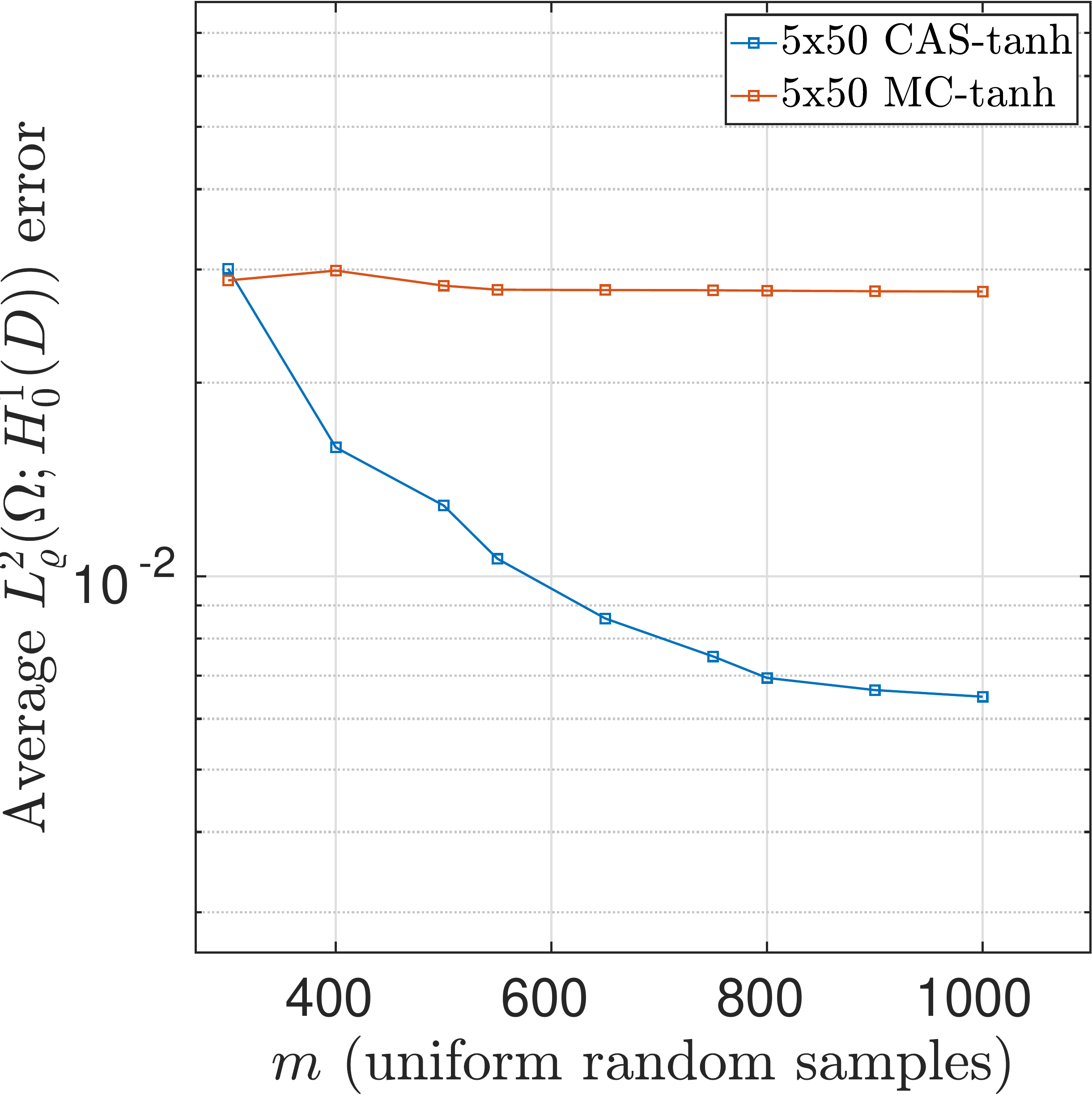}  \\ 
 \hspace{5pt} $d =1$ &  \hspace{5pt} $ d = 2 $ &  \hspace{5pt} $d = 4$ \\
\includegraphics[scale=0.11]{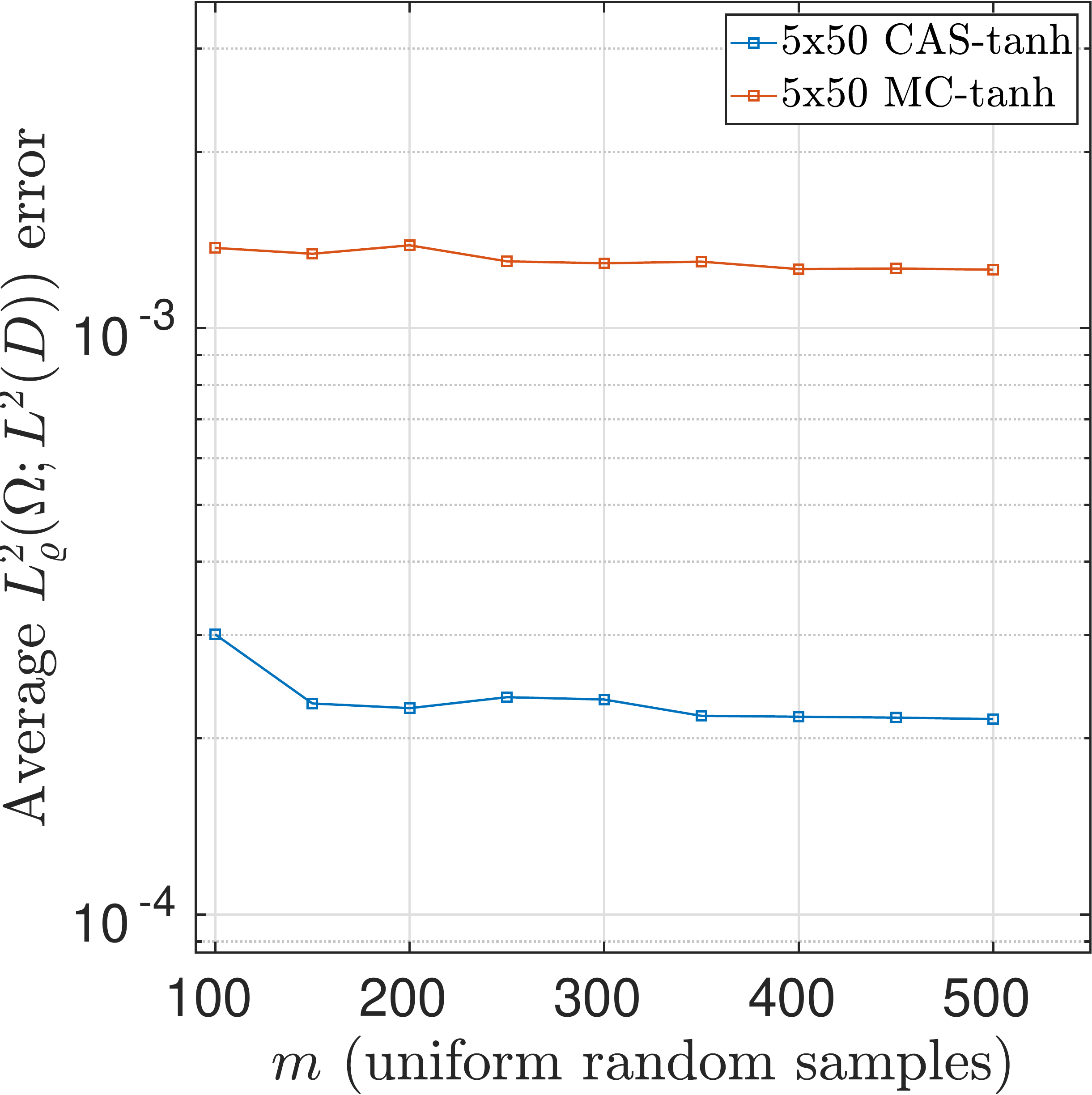} &
\includegraphics[scale=0.11]{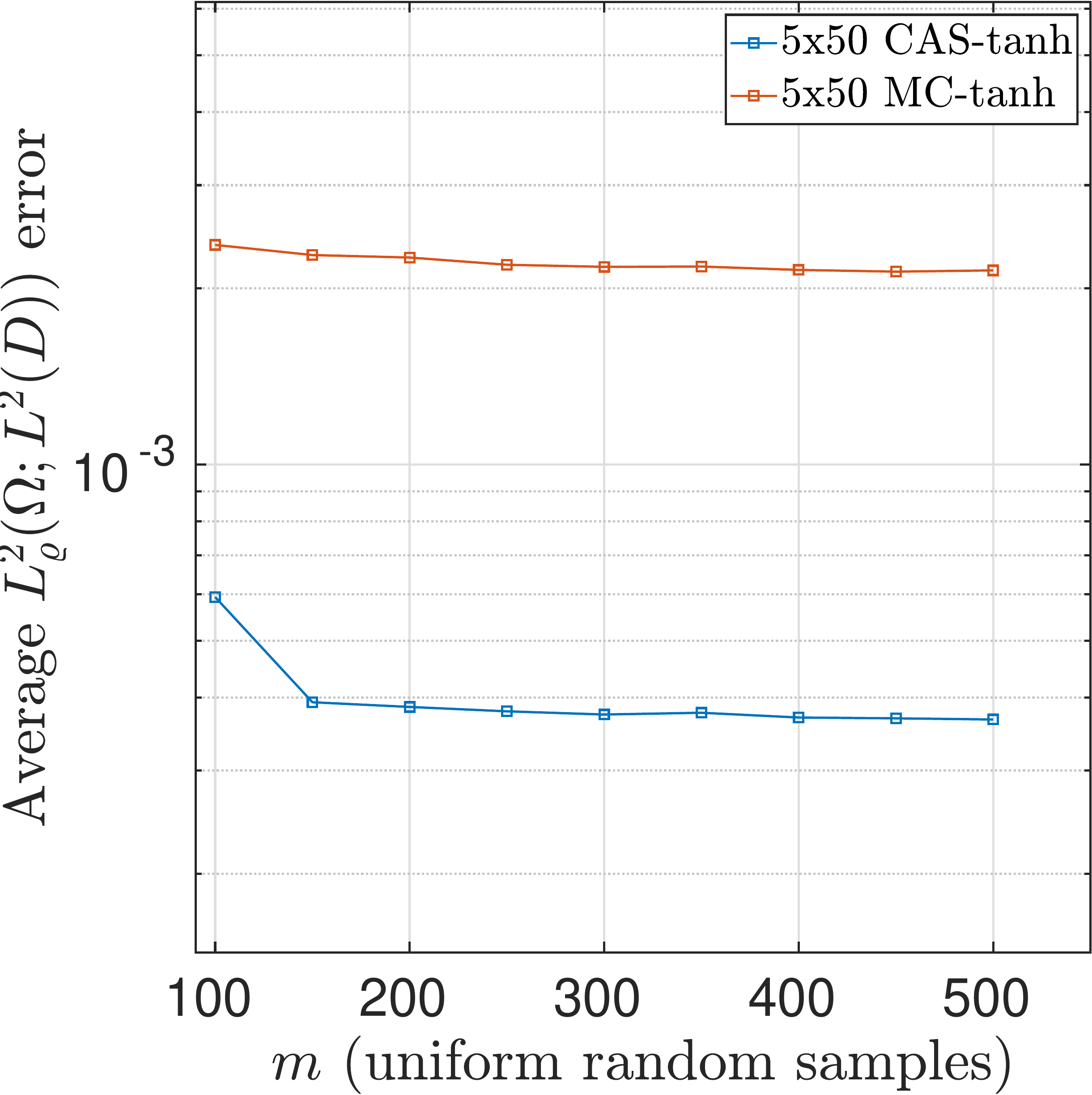} & 
\includegraphics[scale=0.11]{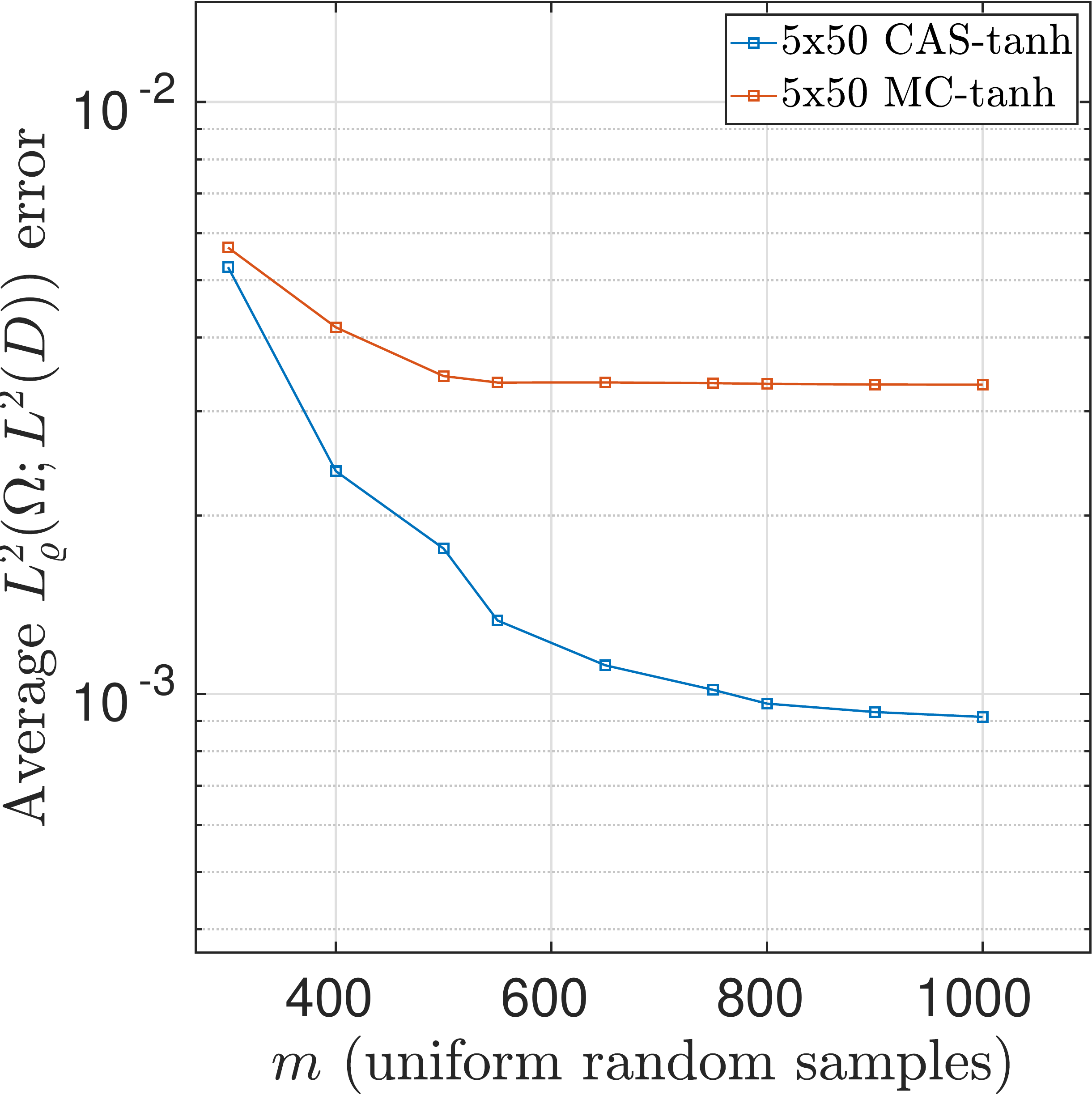} \\  
 \hspace{5pt} $d =1$ &  \hspace{5pt} $ d = 2 $ &  \hspace{5pt} $d = 4$
\end{tabular}
\begin{tabular}{cc}
\includegraphics[scale=0.11]{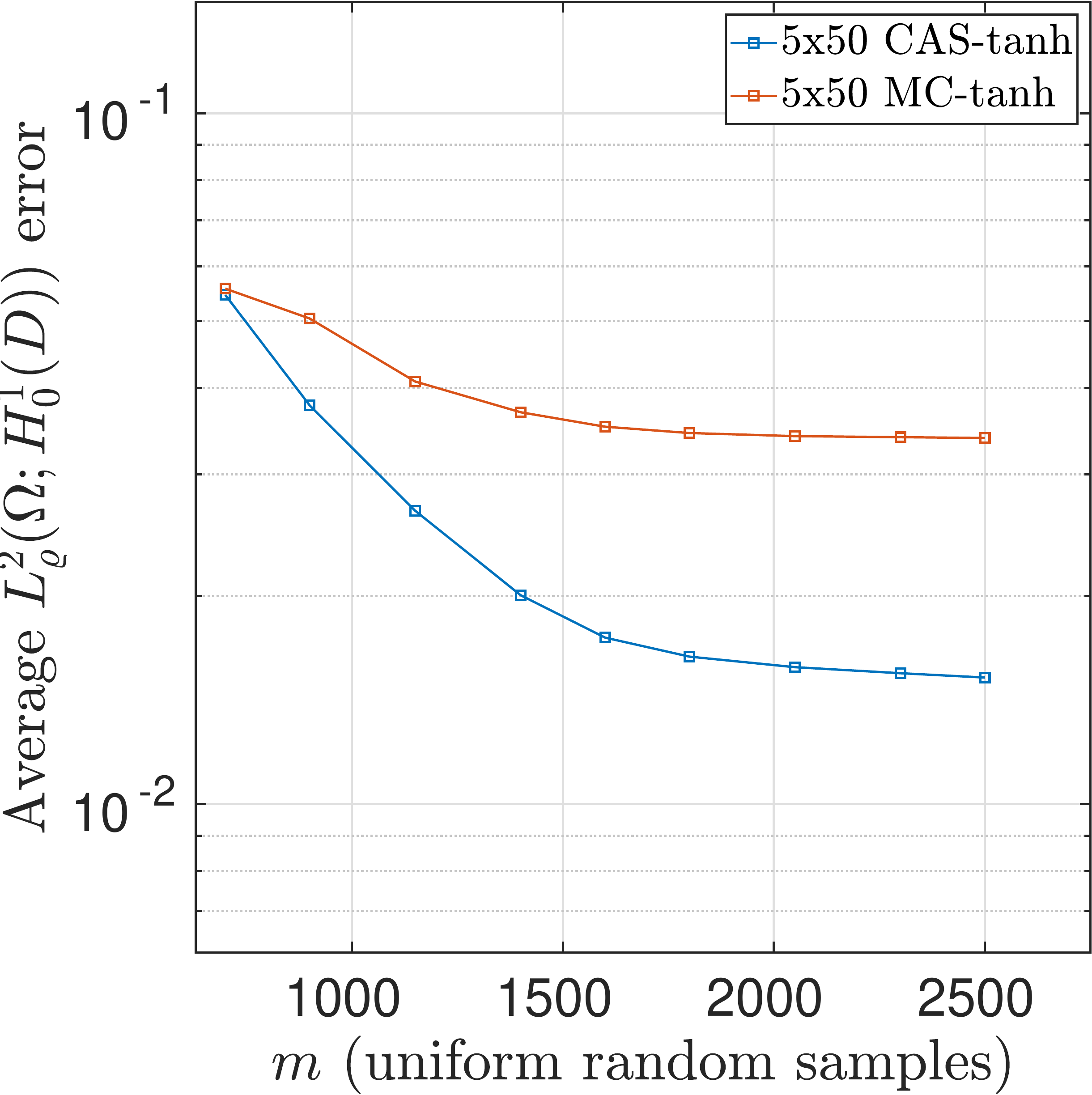} & 
\includegraphics[scale=0.11]{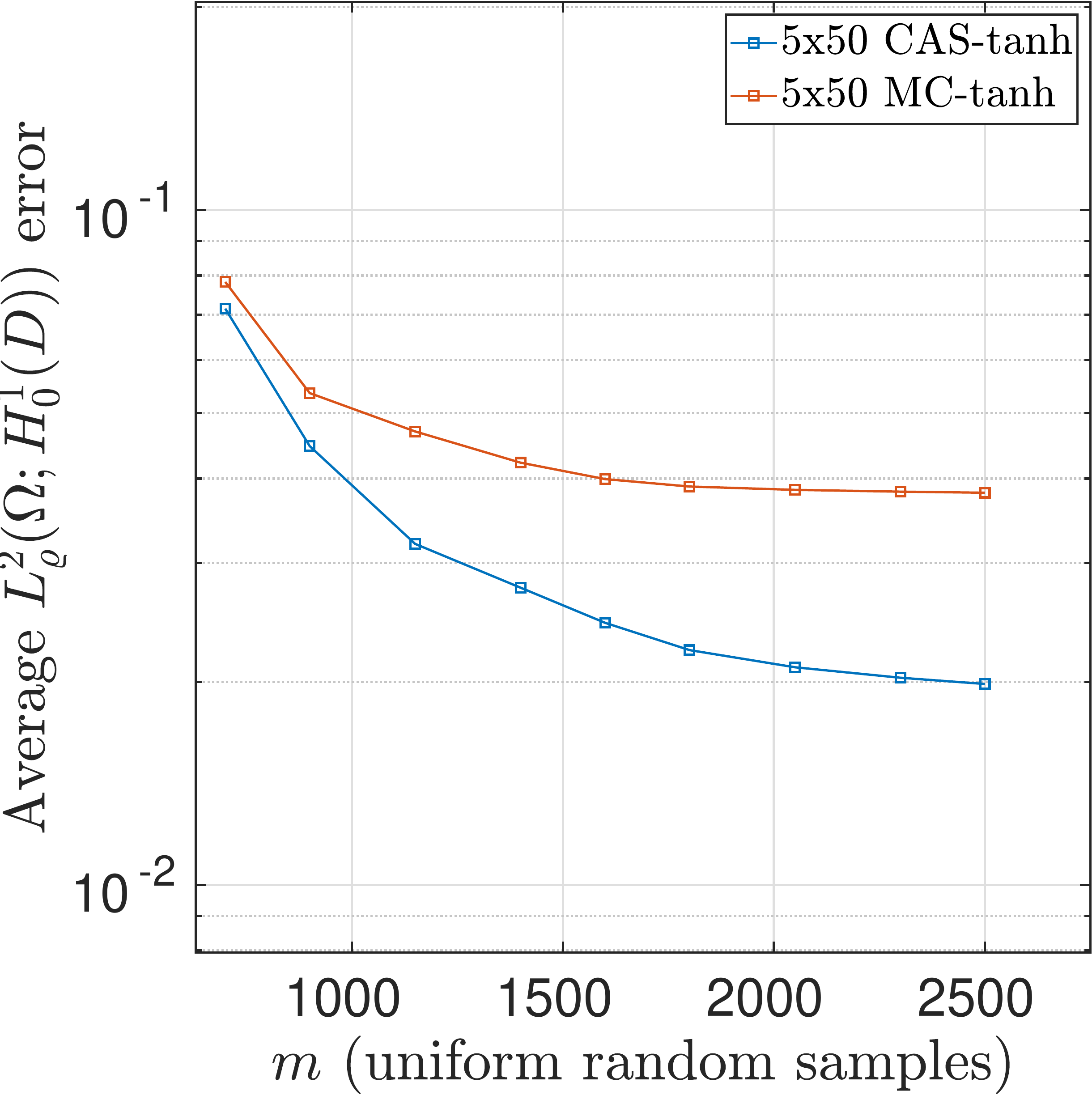} 
\\ 
 \hspace{5pt} $d =8$ &  \hspace{5pt} $ d = 16 $ 
\\
\includegraphics[scale=0.11]{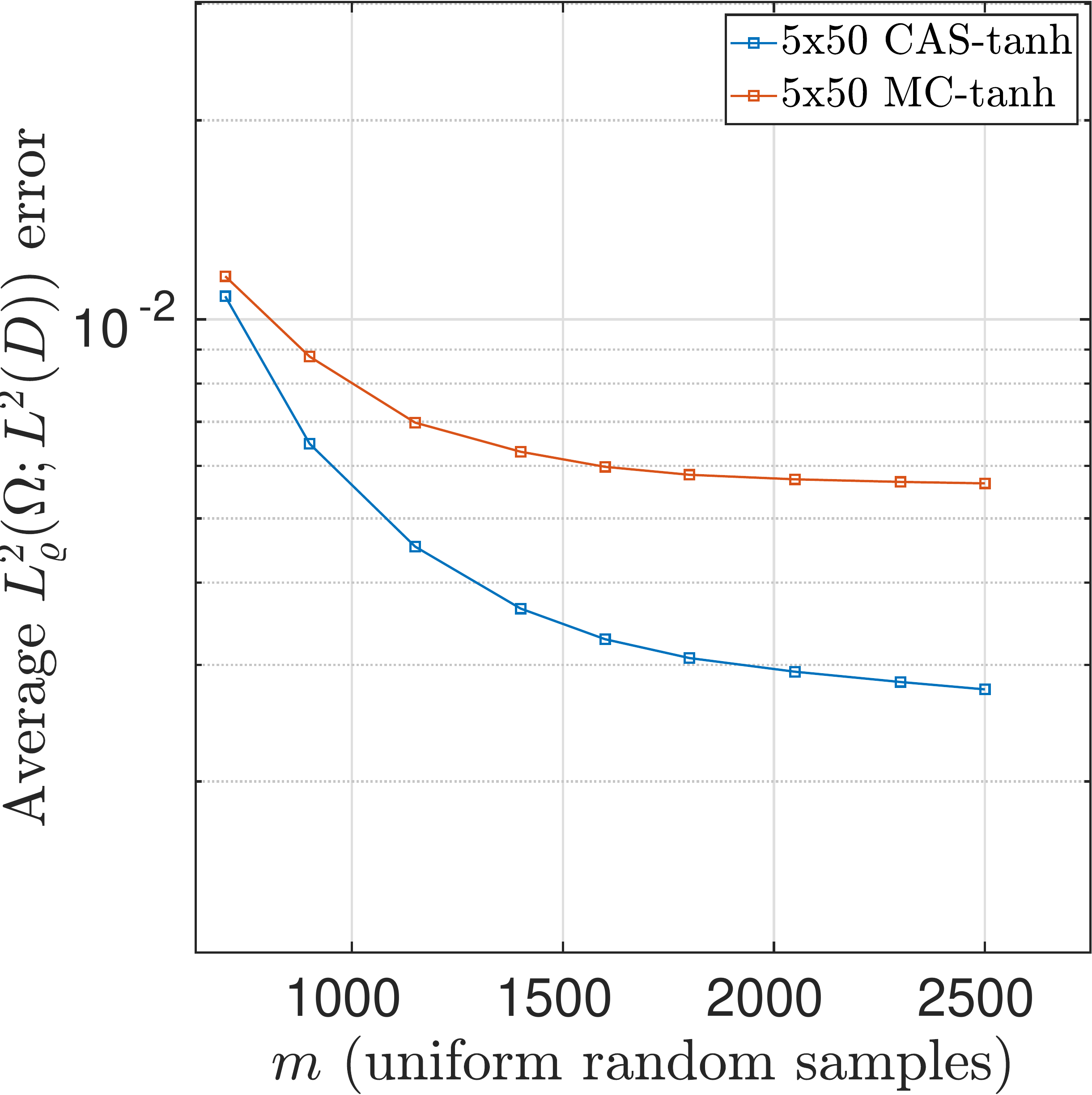} &
\includegraphics[scale=0.11]{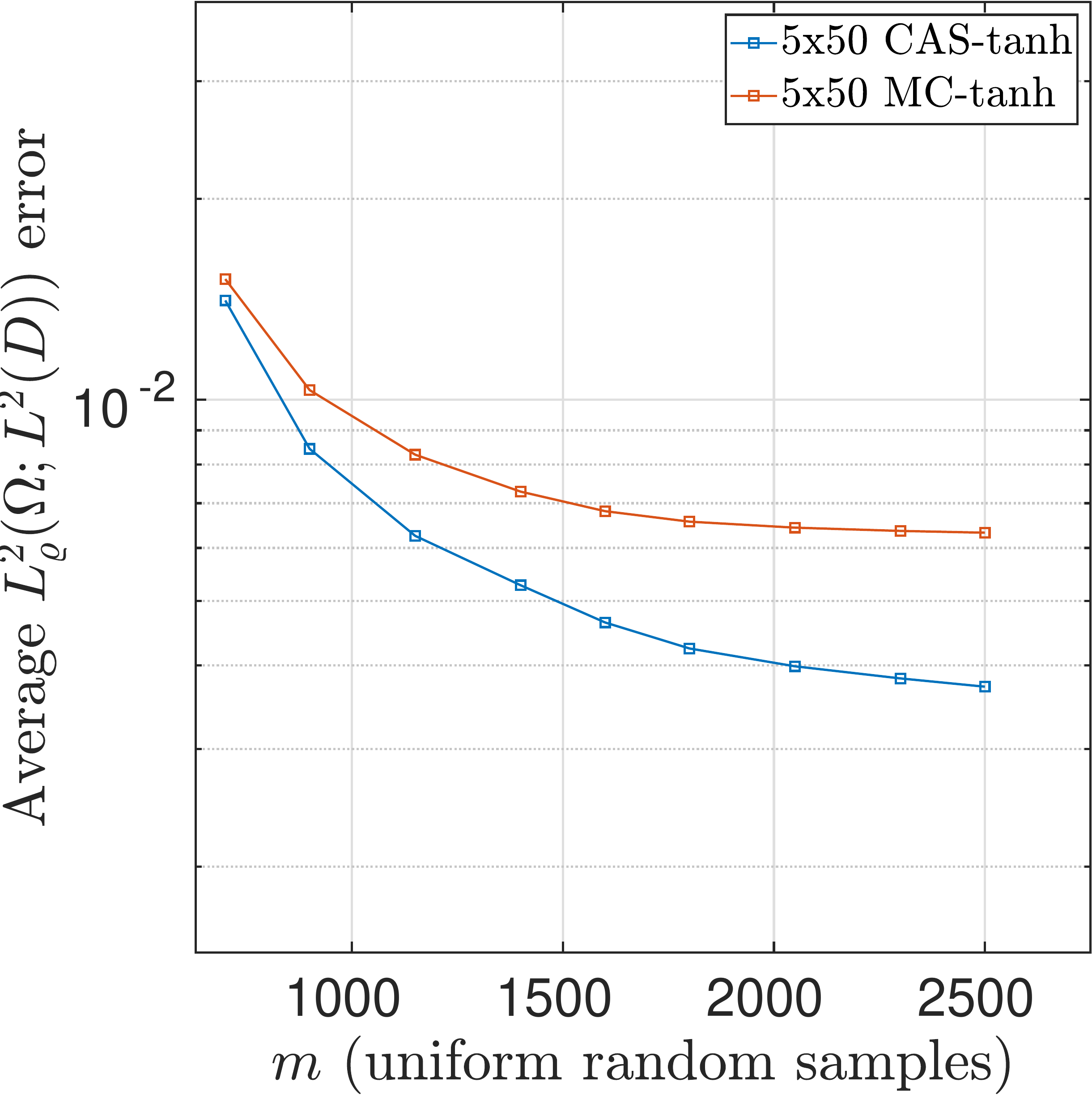} 
\\ 
 \hspace{5pt} $d =8$ &  \hspace{5pt} $ d = 16 $ 
\end{tabular}
}
\caption{ Approximation of the function $f=u(\cdot,)$ via MC sampling and CAS4DL in \textbf{(1st and 2nd row, 1st column)} $d=1$, \textbf{(1st and 2nd row, 2nd column)} $d=2$, \textbf{(1st and 2nd row, 3rd column)} $d=4$, \textbf{(3rd and 4th row, 1st column)} $d=8$ and \textbf{(3rd and 4th row, 2nd column)} $d=16$. \textbf{First and second row:} Average relative $L_\varrho^2(\Omega;H_0^1(D))$ error vs number of CAS4DL and MC samples used in training $\tanh$ $5\times 50$ DNNs. \textbf{Second and fourth column:} Average relative $L_\varrho^2(\Omega;L^2(D))$ error vs number of CAS4DL and MC samples used in training $\tanh$ $5\times 50$ DNNs.}   
\label{fig:parametric_PDE_example_dim_1_to_16}
\end{figure}

In Figure \ref{fig:parametric_PDE_example_dim_1_to_16} we present the average relative $L^2_\varrho(\cU;\cV)$ error, where $\cV$ is either $L^2(D)$ or $H_0^1(D)$. In general, we observe a clear benefit of CAS4DL over MC in all dimensions with respect to the number of samples needed to obtain a given accuracy. In one and two dimensions, the error achieved by the former is between five up to ten times smaller than MC. In higher dimensions the improvement is less, but we still observe that CAS4DL outperforms MC.

\newpage

\vspace*{\fill}
\vspace*{\fill}

\newpage
\section{Accuracy and stability of CAS4DL}
\label{sec:accuracy_stability}

Besides facilitating the method CAS4DL, the adaptive basis viewpoint also provides a way to examine the accuracy and stability of the trained DNN. In this penultimate section, we consider this matter.

\subsection{DL as an optimized least-squares fit}

Recall that in this viewpoint, we consider a class $\cN$ of DNNs of the form
\bes{
\Psi(\bm{y}) = \bm{c}^{\top} \bm{\Psi}(\bm{y}),\qquad \bm{c} \in \bbR^N,
}
where $\bm{\Psi} : \bbR^d \rightarrow \bbR^N$ is the DNN obtained from excluding the final layer of $\Psi$. Write $\bm{\cN}$ for the class of such DNNs. Now recall the CAS4DL strives to compute a DNN as a minimizer of the weighted $\ell^2$-loss:
\bes{
\widehat{\Psi} \in \argmin{\Psi \in \cN} \sum^{m}_{i=1} w(\bm{y}_i)\abs{\Psi(\bm{y}_i) - d_i }^2,\qquad d_i = f(\bm{y}_i) + n_i.
}
Here, for the sake of simplicity we simply write $m$ instead of $m_l$ and $\hat{\Psi}$ instead of $\hat{\Psi}^{(l)}$. We have also added noise $n_i$ to our measurements so as to assess the stability of the method. Then this is equivalent to
\bes{
\widehat{\Psi} = \hat{\bm{c}}^{\top} \widehat{\bm{\Psi}},
}
where the pair $(\hat{\bm{c}},\widehat{\bm{\Psi}})$ is a minimizer of the problem
\bes{
\min_{\bm{c} \in \bbR^N, \bm{\Psi} \in \bm{\cN}} \sum^{m}_{i=1} w(\bm{y}_i)\abs{\bm{c}^{\top}\bm{\Psi}(\bm{y}_i) - d_i }^2 
=  \min_{\bm{\Psi} \in \bm{\cN}} \min_{\bm{c} \in \bbR^N} \sum^{m}_{i=1} w(\bm{y}_i)\abs{\bm{c}^{\top}\bm{\Psi}(\bm{y}_i) - d_i }^2.
}
Given $\bm{\Psi} = (\psi_i)^{N}_{i=1} \in \bm{\cN}$, let
\bes{
P_{\bm{\Psi}} = \spn \{ \psi_1,\ldots,\psi_N \}.
}
The inner minimization is precisely a least-squares fit in $P_{\bm{\Psi}}$. It now follows that $\widehat{\Psi} $ is a least-squares fit in $P_{\widehat{\bm{\Psi}}}$ corresponding to the subspace spanned functions that define the output layer of the learned DNN $\widehat{\bm{\Psi}}$. In this sense, it is an \textit{optimized} least-squares fit over all possible subspaces $P_{\bm{\Psi}}$ with $\bm{\Psi} \in \bm{\cN}$.

\subsection{Stability and accuracy analysis}
With this in hand, standard least-squares analysis (see, e.g., \cite[Thm.\ 5.3]{adcock2021sparse}) now yields the error bound
\be{
\label{bound-1}
\nmu{f - \widehat{\Psi}}_{L^2} \leq \inf_{\Phi \in P_{\widehat{\bm{\Psi}}} } \left \{ \nmu{f - \Phi}_{L^2} + \frac{1}{\alpha(\widehat{\bm{\Psi}})} \nmu{f - \Phi}_{\mathrm{disc}} + \frac{1}{\alpha(\widehat{\bm{\Psi}})}\frac{\nm{\bm{n}}_{\ell^2}}{m}\right \},
}
where $\nm{g}_{\mathrm{disc}} = \sqrt{m^{-1} \sum^{m}_{i=1} w(\bm{y}_i)\abs{g(\bm{y}_i)}^2 }$,
\bes{
\alpha(\widehat{\bm{\Psi}}) = 
\inf \{ \nm{\Phi}_{\mathrm{disc}} / \nm{\Phi}_{L^2} : \Phi \in P_{\widehat{\bm{\Psi}}},\ \Phi \neq 0 \}
}
is a \textit{discrete stability constant} and $\bm{n}=(n_i)^{m}_{i=1}$ is the noise.

This analysis has several benefits. First, we can numerically assess the stability of the learned approximation $\hat{\Psi}$ by computing the discrete stability constant $\alpha(\widehat{\bm{\Psi}})$. Indeed, given an orthonormal basis $\{ \hat{\phi}_1,\ldots,\hat{\phi}_N \} \subset L^2(\Omega,\tau)$ for $P_{\widehat{\bm{\Psi}}}$, this can be computed by finding the minimum singular value of the matrix
\be{
\label{matrix-for-alpha}
\left ( \sqrt{(1/m)w(\bm{y}_i)} \hat{\phi}_j(\bm{y}_i) \right )^{m,N}_{i,j=1}.
}
Since $\tau$ is a discrete measure supported on a finite grid, this basis can be easily generated via SVD. See the previous discussion. In particular, if $1/\alpha(\widehat{\bm{\Psi}})$ is not too large, then $\widehat{\Psi}$ is stable to noise in the measurements.

The bound \R{bound-1} also provides some insight into the accuracy of the approximation. Again, if $1/\alpha(\widehat{\bm{\Psi}})$ is not too large, then we see that $\widehat{\Psi}$ is a quasi-best approximation to $f$ from the learned subspace $P_{\widehat{\bm{\Psi}}}$. Unfortunately, this does not provide a direct insight, since it is generally unknown how well $f$ can be approximated in $P_{\widehat{\bm{\Psi}}}$. Nonetheless, it supports the fact that one key way to obtain better accuracy from DNNs involves focusing on learning a good subspace in which the target function is approximated. See Section \ref{sec:conclusion} for some further discussion on this issue.

\subsection{Numerical examples}

To conclude this section, we present several examples showing the computed values of the constant $\alpha(\widehat{\bm{\Psi}})$. Recall that, in practice, the trained DNN yields a system of functions that is numerically redundant. Hence, in this computation, we use the same singular value thresholding strategy to determine the numerical dimension $n$ of the learned subspace $P_{\widehat{\bm{\Psi}}}$ and to compute an orthonormal basis $\{ \hat{\phi}_1,\ldots,\hat{\phi}_n \}$ for it. Thus, in practice, the matrix \R{matrix-for-alpha} is modified by replacing $N$ by $n$.

\begin{figure}[t]
\centering
{\small
\begin{tabular}{cccc}
\hspace{-1cm} 
\includegraphics[scale=0.11]{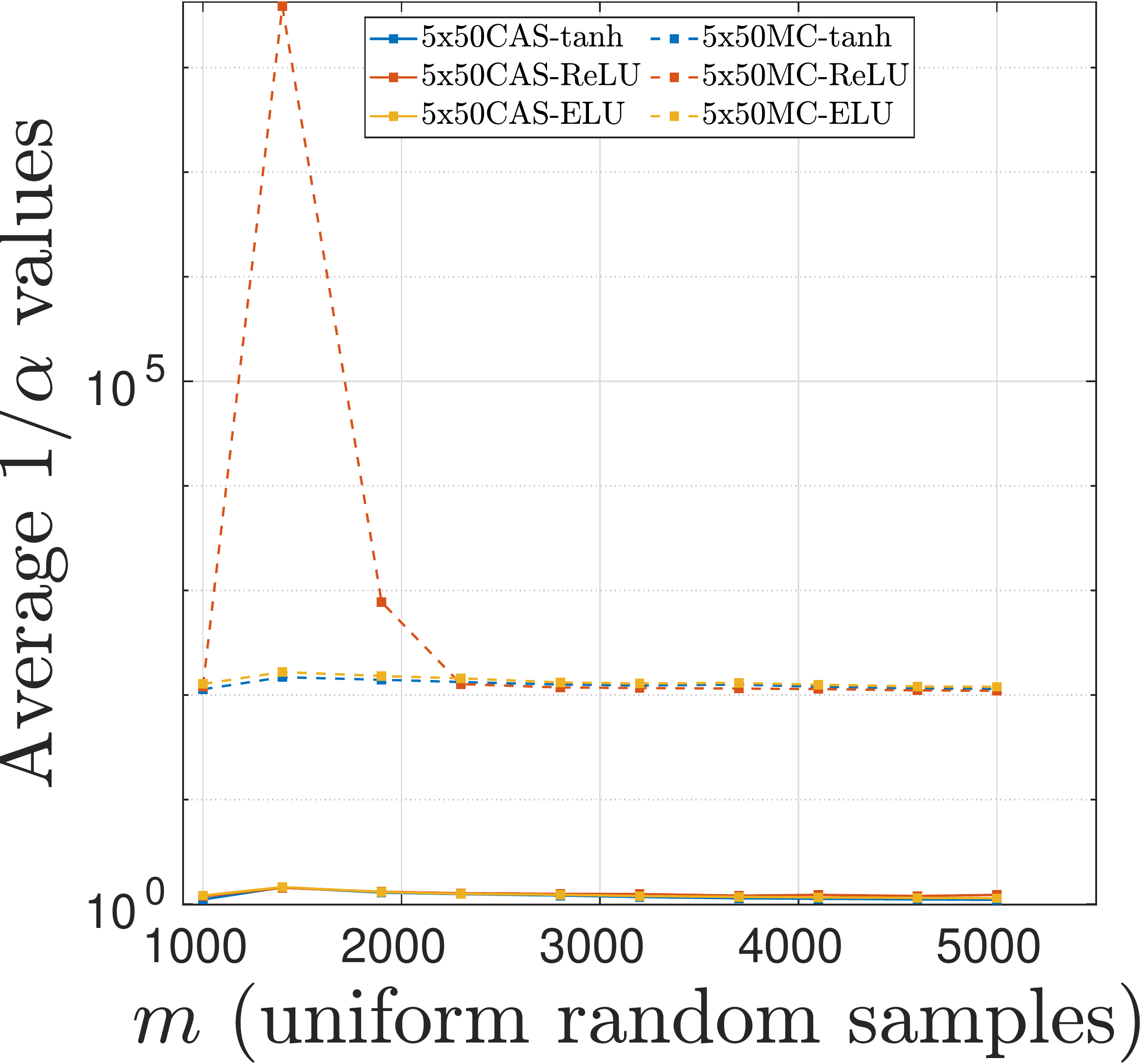} & 
\includegraphics[scale=0.11]{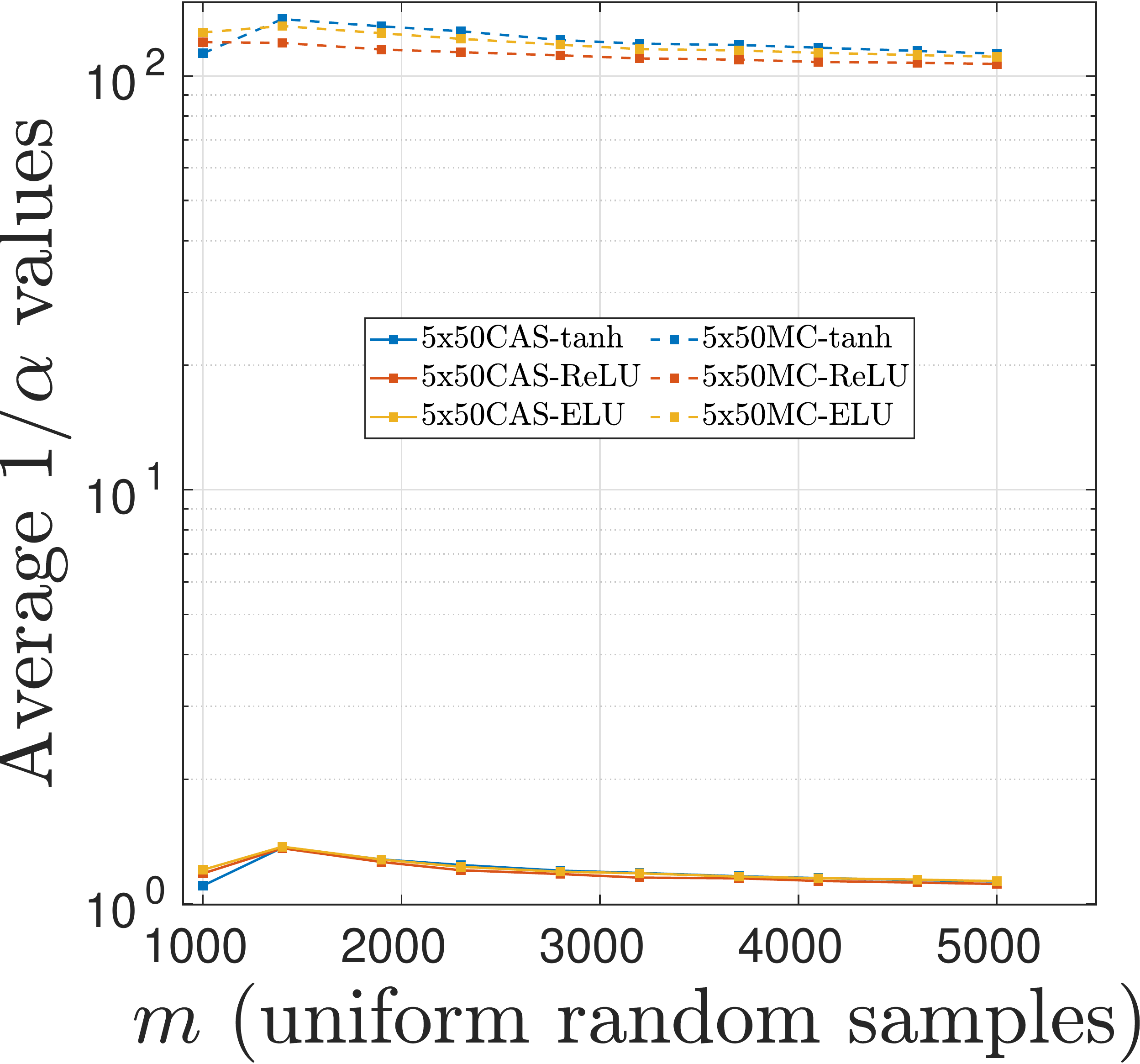} & 
\includegraphics[scale=0.11]{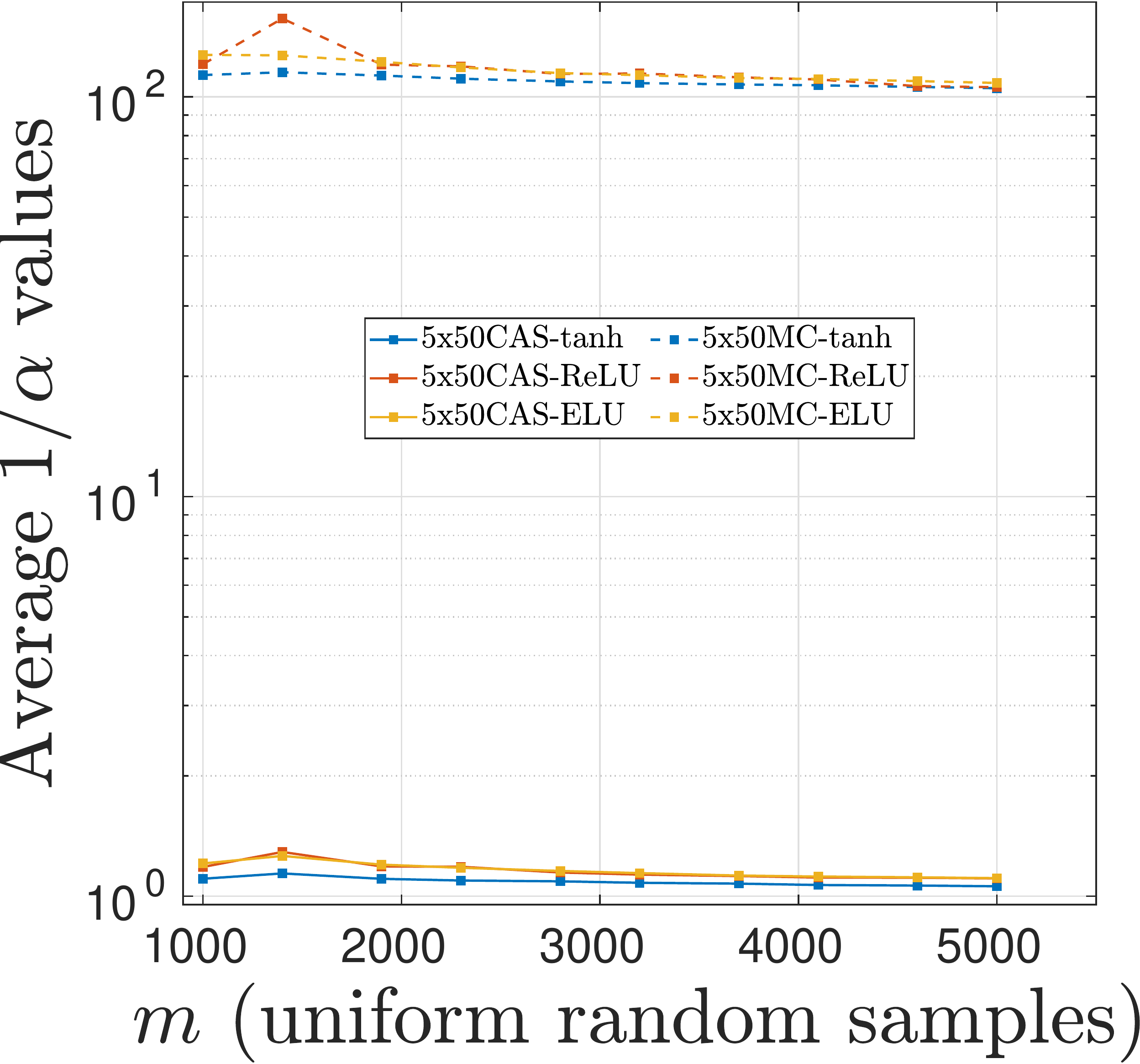} & 
\includegraphics[scale=0.11]{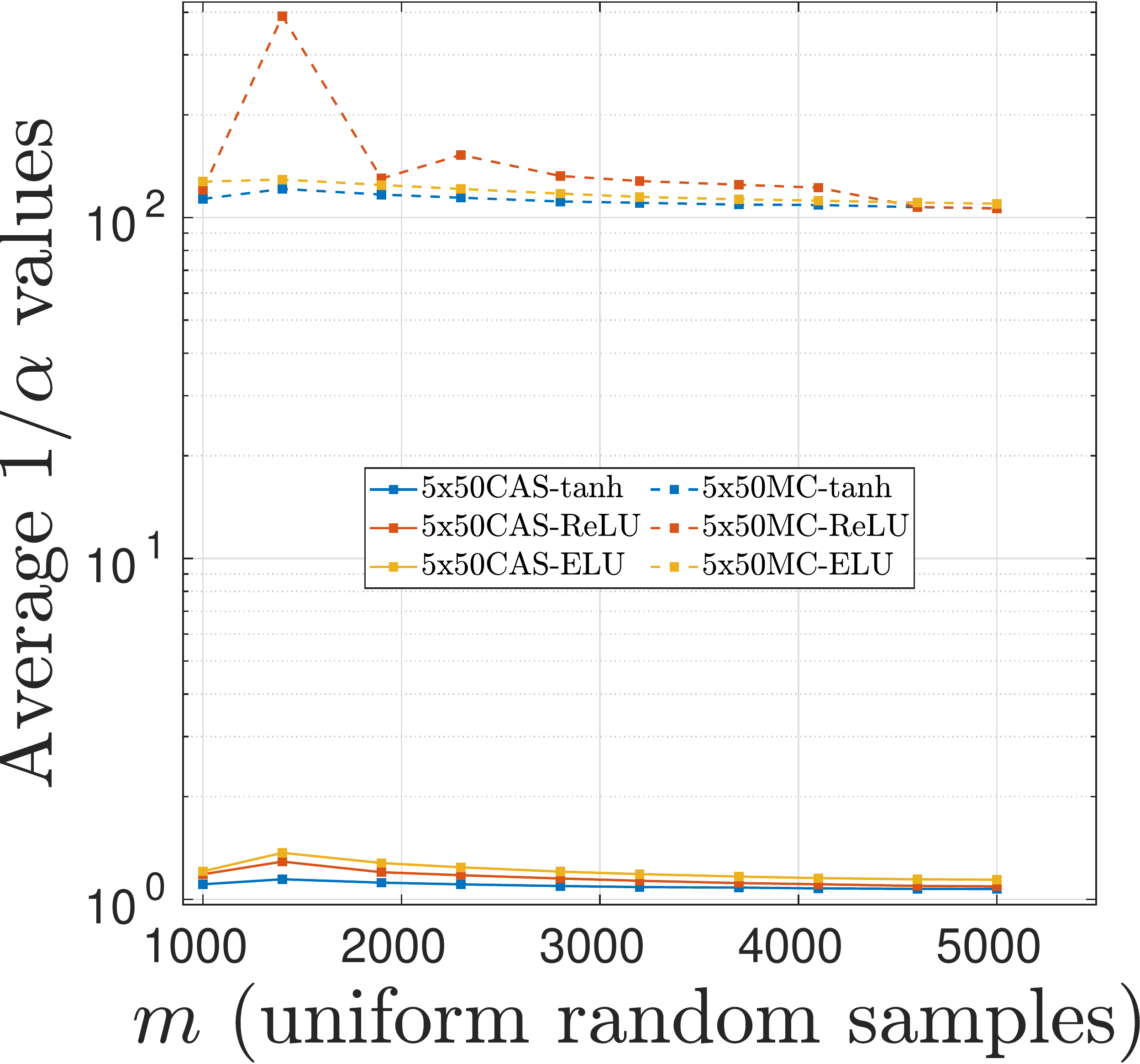} \\
$(f,d)=(f_1,2)$ & $(f,d)=(f_2,2)$ & $(f,d)=(f_3,2)$ & $(f,d)=(f_4,2)$ \\
\hspace{-1cm} 
\includegraphics[scale=0.11]{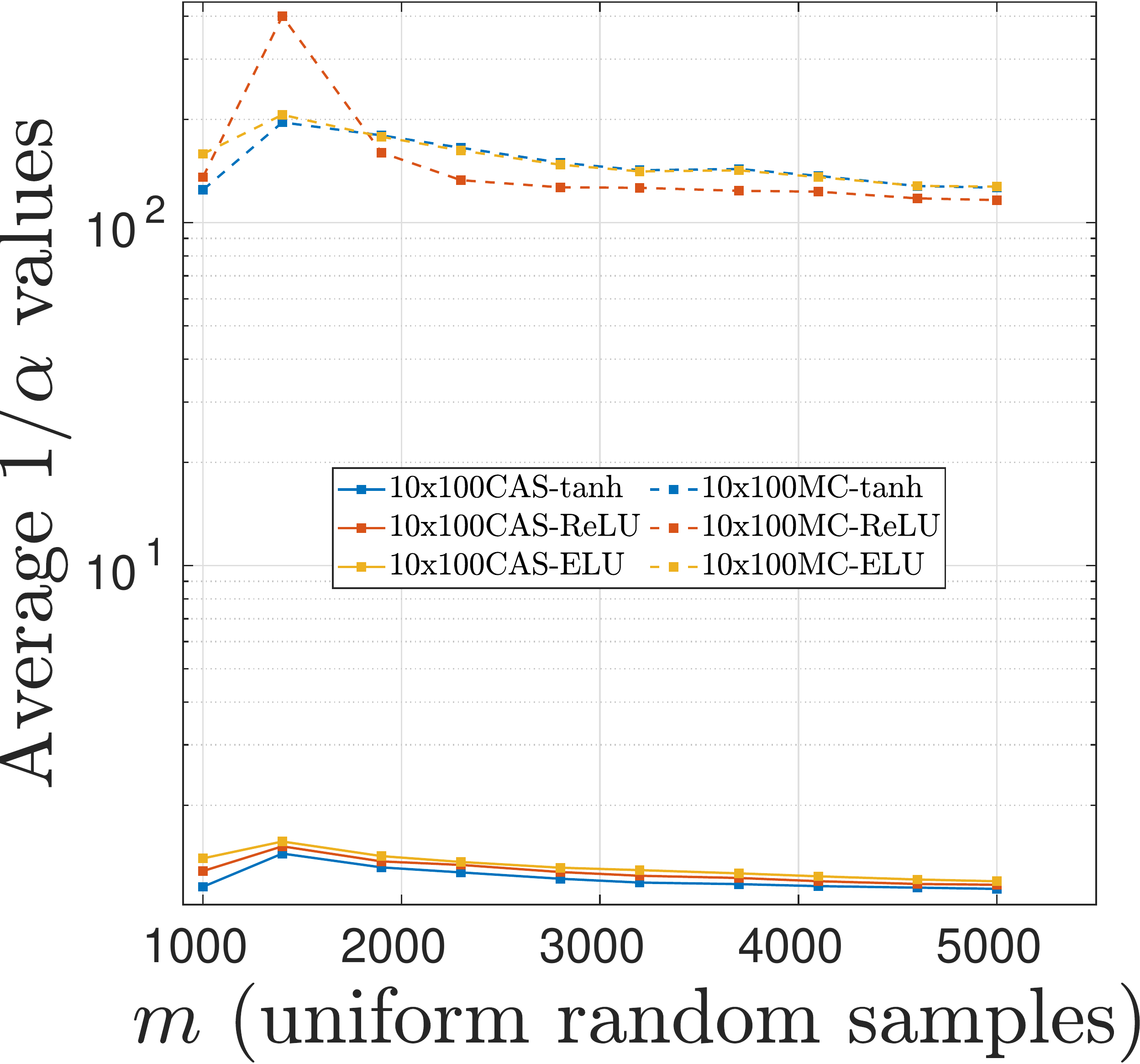} & 
\includegraphics[scale=0.11]{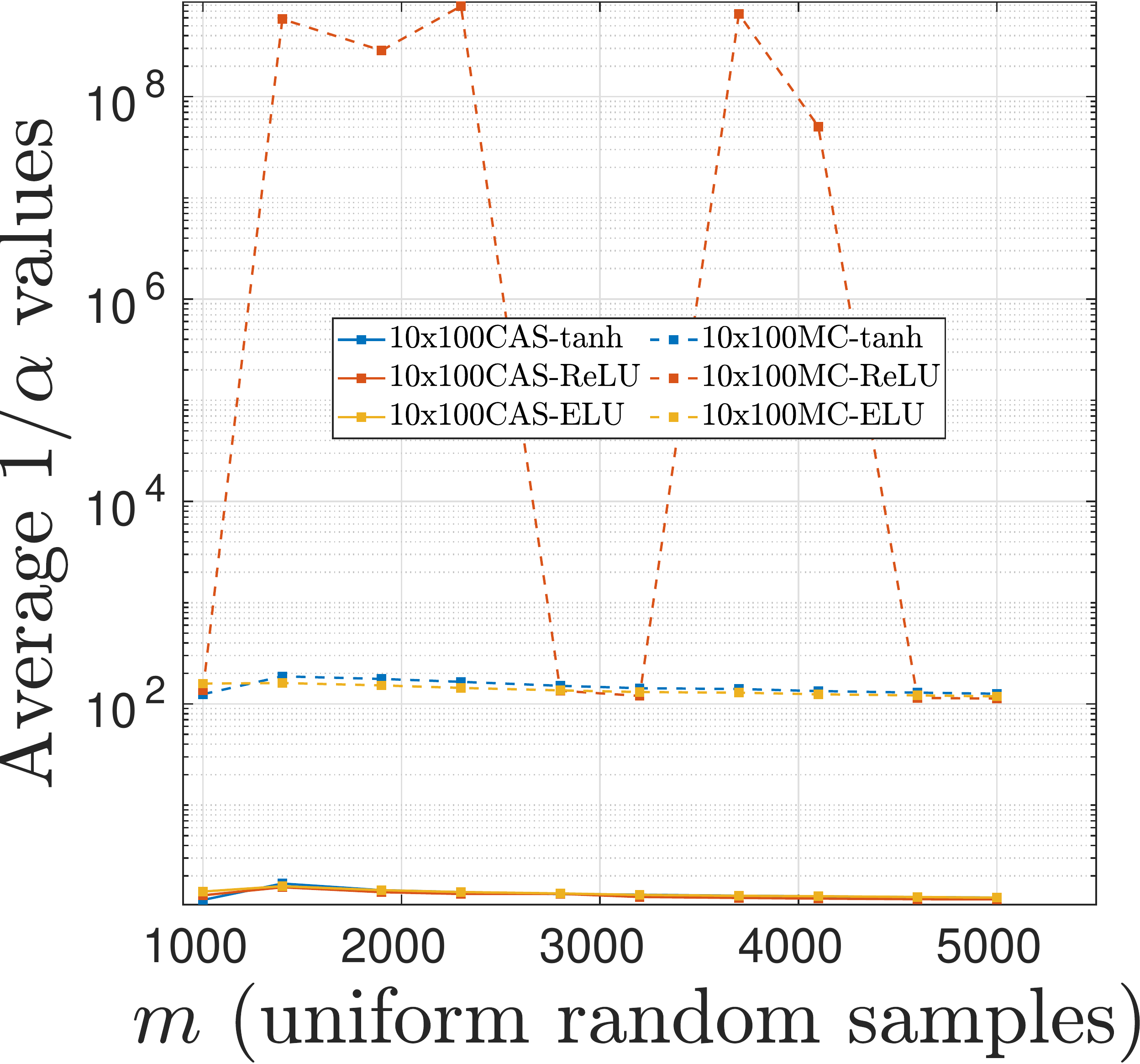} & 
\includegraphics[scale=0.11]{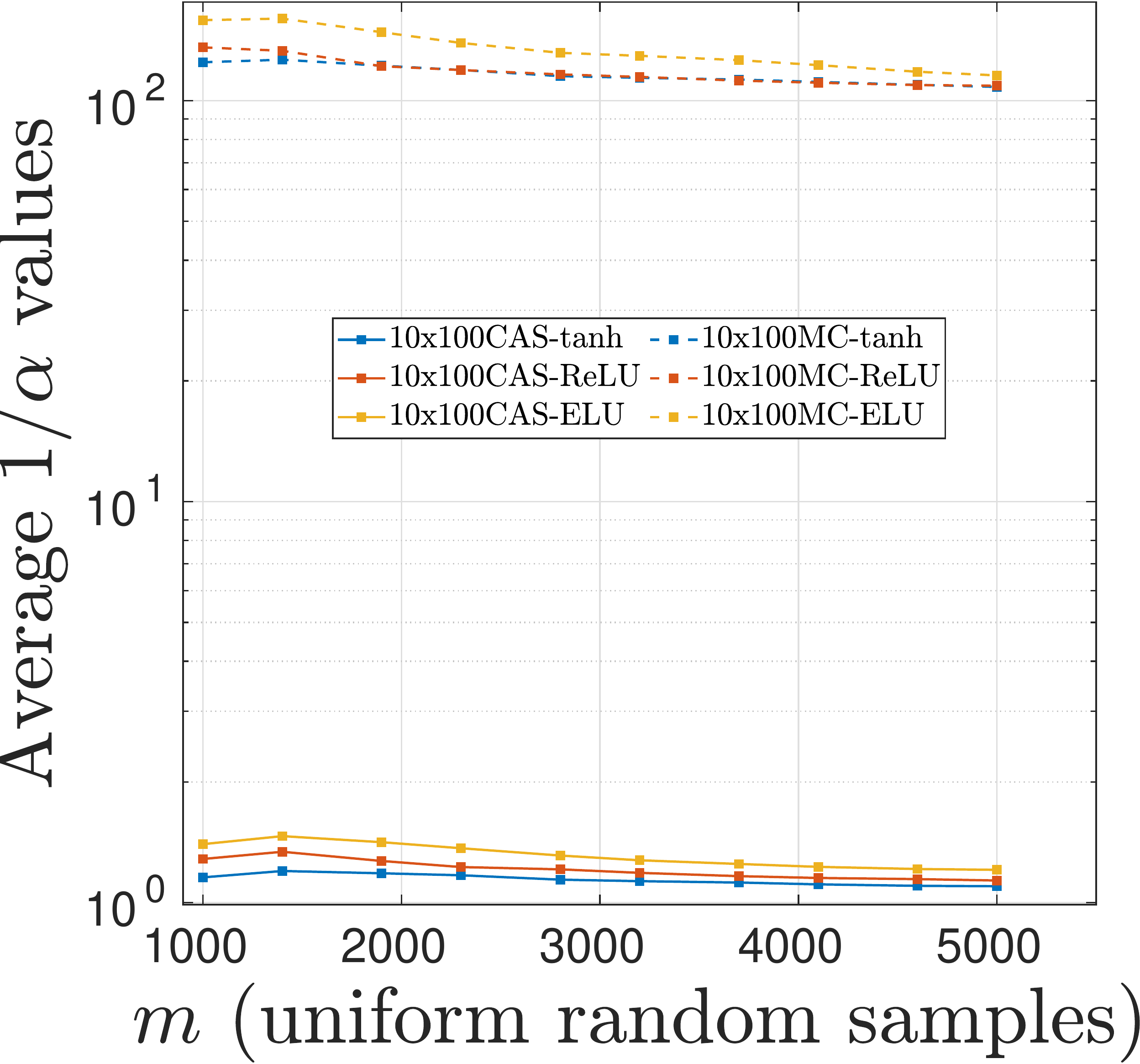} & 
\includegraphics[scale=0.11]{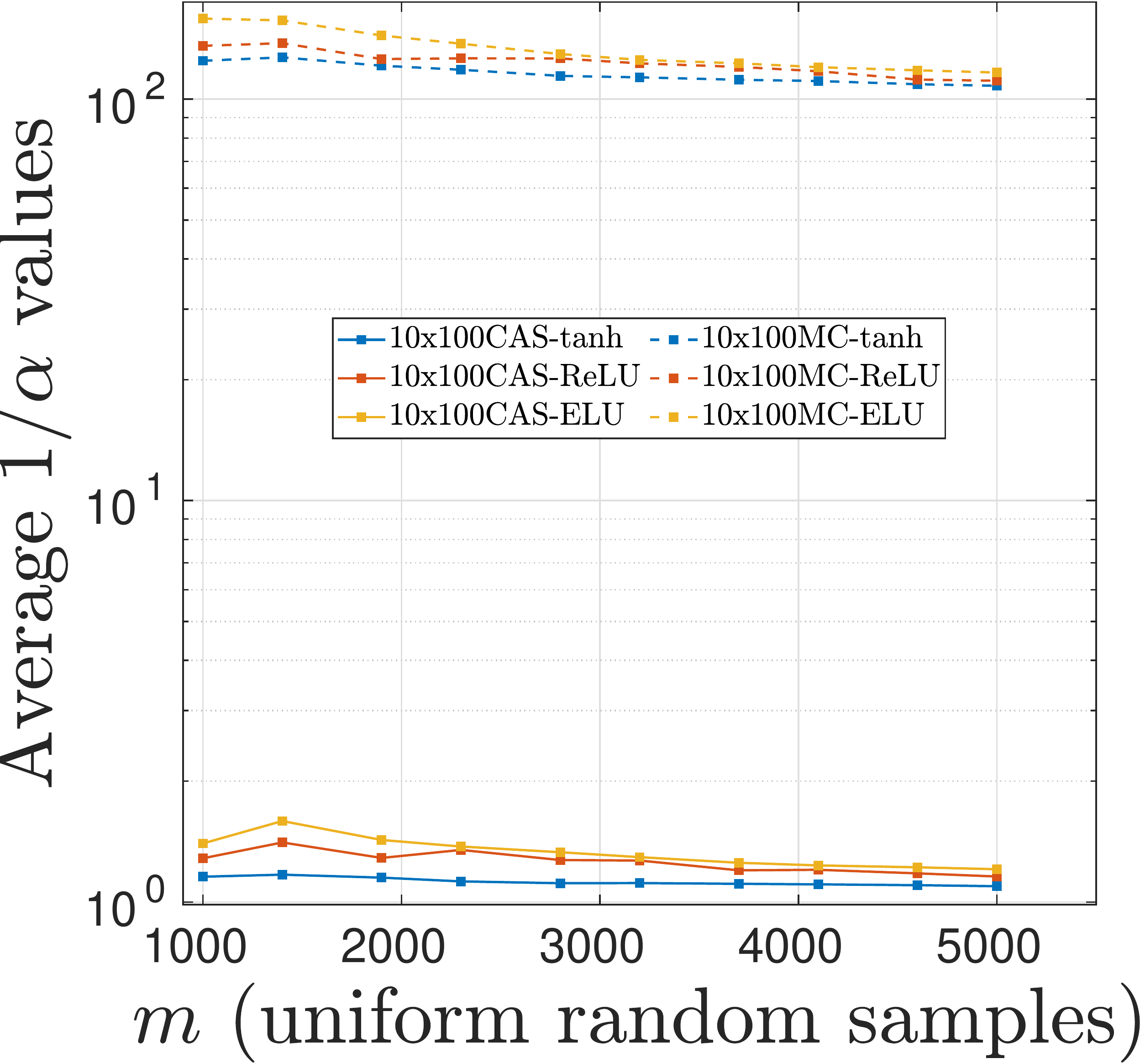} \\
$(f,d)=(f_1,2)$ & $(f,d)=(f_2,2)$ & $(f,d)=(f_3,2)$ & $(f,d)=(f_4,2)$ 
\end{tabular}
}
\caption{Average $1/\alpha(\hat{\Psi})$ values for CAS4DL and MC sampling over 20 trials vs. number of samples $m$ for the functions $f_1$, $f_2$, $f_3$ and $f_4$used in training. \textbf{First row:} $L = 5$ and $N= 50$. \textbf{Second row:} $L=10$ and $N=100$. These DNNs use ReLU, $\tanh$ and ELU activation functions.}
\label{fig:comp_act_C_vals}
\end{figure}

Figure \ref{fig:comp_act_C_vals} shows the values of this constant for DL with MC sampling and CAS4DL. It is clear that CAS4DL is more stable than MC sampling. The constant $1/\alpha(\hat{\Psi})$ is several orders of magnitude smaller for CAS4DL than for MC. The constant for MC is on the order of $10^2$ for all the functions, while the constant for CAS4DL is close to one. Notice that the constant $1/\alpha(\widehat{\bm{\Psi}})$ decreases slightly as the number of measurements increases, but not by a significant amount. This is indicative of the fact that a new subspace $P_{\widehat{\bm{\Psi}}^{(l)}}$ is being learned at each step of the procedure. Indeed, if the subspace (or equivalently, the DNN $\bm{\Psi}$) were fixed, then one generally expects $\alpha(\bm{\Psi})$ to decrease as the number of samples increases. For MC sampling, $\alpha(\bm{\Psi})$ tends to one almost surely for fixed $\bm{\Psi}$ as the number of samples increases (see, e.g., \cite[Rem.\ 5.2]{adcock2021sparse}). The fact that the constants in Figure \ref{fig:comp_act_C_vals} do not appear to be decreasing is indicative of the fact that the subspace is changing at each step.  

\section{Conclusion}
\label{sec:conclusion}

In this work we have introduced a new method, Christoffel Adaptive Sampling for Deep Learning (CAS4DL), for adaptive function approximation via DL. This novel method combines two key ideas. First, ideas of optimal sampling for least-squares approximation in linear subspaces. Second, the adaptive basis viewpoint of DL, where a DNN is viewed as a linear combination of basis functions defined by its penultimate layer. CAS4DL arises from the novel combination of these two ideas, in combination with thresholded SVDs to handle potential numerically redundant learned dictionaries. Our numerical examples show that CAS4DL outperforms MC sampling, both attaining a smaller error from a fixed sampling budget across a range of functions and dimensions and possessing better stability properties.

There are several avenues for further research. First, it is intuitively desirable (and also supported by our experiments) that the numerical dimension $n$ of the learned subspace be as close to full as possible (i.e., $n \approx N$). This raises the question of how one might try to promote this during training -- for instance, by incorporating a regularization term that promotes well-conditioned bases.

This work illuminates an interesting connection between DL and redundant dictionaries. Redundant dictionaries are used throughout signal processing and numerical analysis \cite{christensen2016introduction}, since they are flexible tools that can be deployed in a variety of different settings more easily than classical orthonormal bases \cite{adcock2020frames,adcock2019frames}. One particularly popular approach is to build redundant dictionaries by concatenating together two or more dictionaries.  In this work, we train a full DNN at each step, and therefore learn a new dictionary each time. But the redundant dictionary connection also allows for other possibilities, which can be equally well exploited within CAS4DL. For example, one may combine a learned DNN with unlearned functions such as polynomials. The analysis in Section \ref{sec:accuracy_stability} would then ensure that the learned approximation is at least as accurate as the best approximation in the space spanned by the unlearned functions, thus guaranteeing a certain accuracy.
For this reason, such a hybrid scheme may bring further benefits over using DNNs or polynomials alone. 

Finally, it is notable in this work that the DNN architecture is fixed in each step, while the number of samples is increasing. Therefore, the best achievable accuracy is limited by the best approximation from the DNN space $\cN$. In practice, it may therefore be beneficial to increase the size of the DNN architecture as the number of samples grows. As discussed in, e.g., \cite{AdcockDexter2021}, there is typically a delicate tradeoff between the number of samples and the optimal architecture. Hence, how best to do this in combination with CAS is an open problem.

\bmhead{Acknowledgments}

BA acknowledges support from NSERC through grant R611675. ND acknowledges support from a PIMS postdoctoral fellowship. On behalf of all authors, the corresponding author states that there is no conflict of interest.

\bibliography{sasida}% common bib file
%\input sn-article.bbl

%\input sn-sample-bib.tex

\end{document}